\pdfoutput=1

\newif\ifprint%
\newif\ifhyperlinks%
\newif\ifpublic%
\publictrue%

\documentclass[a4paper, 11pt, twoside]{report}

\usepackage[utf8]{inputenc}
\usepackage[czech, english]{babel}

\usepackage{xcolor}
\usepackage[T1]{fontenc}
\usepackage{XCharter}
\usepackage{libertine}
\usepackage[scaled=0.85]{beramono}
\usepackage{microtype}
\usepackage{everysel}

\usepackage{fancyhdr}
\usepackage[nottoc]{tocbibind}
\usepackage{tocloft}
\usepackage[nobottomtitles]{titlesec}
\usepackage{appendix}
\usepackage{breakcites}
\usepackage{cite}
\usepackage{calc}

\usepackage{graphicx}
\usepackage{pgfplots} \pgfplotsset{compat=newest}
\usepackage{pgf, tikz}
\usepackage{subcaption}
\usepackage[rightcaption]{sidecap}
\usepackage{wrapfig}
\usepackage[font=small,labelfont=bf]{caption}
\usepackage{float}

\usepackage{listings}

\usepackage{amsmath, amsfonts, amssymb}
\usepackage{bm}
\usepackage{bbm}
\usepackage{amsthm}

\usepackage{enumitem}

\usepackage{algorithmicx, algorithm, algpseudocode}
\usepackage{mathtools}

\usepackage{tabularx}
\usepackage{multicol, multirow, makecell}
\usepackage{booktabs} 

\usepackage{pdfpages} 
\usepackage{url} 
\usepackage{tabto}

\definecolor{cvutaccented}{RGB}{56, 56, 56}

\definecolor{cvutblue}{RGB}{13,79,139}
\definecolor{cvutred}{RGB}{159,28,13}
\definecolor{cvutgold}{RGB}{139,91,13}

\definecolor{linksblue}{RGB}{11, 0, 128}

\definecolor{cvutredplot}{RGB}{139,122,13}
\definecolor{cvutgoldplot}{RGB}{139,13,47}

\definecolor{cvut2}{RGB}{7,162,32}
\definecolor{cvut3}{RGB}{152,132,7}
\definecolor{cvut4}{RGB}{152,47,7}
\definecolor{cvut5}{RGB}{124,1,162}

\definecolor{olive}{RGB}{128,128,0}
\definecolor{navy}{RGB}{0,0,128}
\definecolor{cgreen}{RGB}{0, 128, 0}

\pgfplotscreateplotcyclelist{cvutcycle}{
	{cvutgoldplot},
	{cvutredplot},
	{cvutblue},
}

\pgfplotscreateplotcyclelist{cvutcycle5}{
	{cvutgoldplot},
	{cvutredplot},
	{cvutblue},
	{cvut2},
	{cvut5},
}

\makeatletter
\newcommand{\globalcolor}[1]{%
  \color{#1}\global\let\default@color\current@color
}
\makeatother

\AtBeginDocument{\globalcolor{black}}

\ifprint%
    \usepackage[a4paper, margin=1.25cm, top=3cm, bottom=2.5cm, bindingoffset=1.7cm]{geometry}
    \usepackage[unicode, pagebackref=true, hidelinks]{hyperref}
\else
    \usepackage[a4paper, margin=2.1cm, top=3cm, bottom=2.5cm]{geometry}
    \ifhyperlinks%
        \usepackage[unicode, pagebackref=true, breaklinks=true, allbordercolors={cvutgold}]{hyperref}
    \else
        \usepackage[unicode, pagebackref=true, hidelinks]{hyperref}
    \fi
\fi

\usepackage[acronym, nomain, toc]{glossaries}

\renewcommand*{\backref}[1]{}
\renewcommand*{\backrefalt}[4]{%
    \ifcase#1 %
        \relax
        \or%
        Page(s): [#4].%
    \else
        Page(s): [#4].%
    \fi%
}

\let\emph\relax 
\DeclareTextFontCommand{\emph}{\color{black}\em}

\fancypagestyle{beginning} {
    \fancyhf{}
    \fancyhead[LE,RO]{\thepage}
    
}
\fancypagestyle{rest}{
    \fancyhf{}
    \fancyhead[LE, RO]{\color{cvutaccented}\leftmark}
    \fancyhead[RE, LO]{}
    \fancyfoot[C]{\thepage}
    
}

\makeatletter
\patchcmd{\ttlh@hang}{\parindent\z@}{\parindent\z@\leavevmode}{}{}
\patchcmd{\ttlh@hang}{\noindent}{}{}{}
\patchcmd{\f@nch@head}{\rlap}{\color{cvutaccented}\rlap}{}{}
\patchcmd{\headrule}{\hrule}{\color{cvutaccented}\hrule}{}{}
\patchcmd{\f@nch@foot}{\rlap}{\color{cvutaccented}\rlap}{}{}
\patchcmd{\footrule}{\hrule}{\color{cvutaccented}\hrule}{}{}
\makeatother

\LetLtxMacro{\oldalgorithmic}{\algorithmic}
\LetLtxMacro{\endoldalgorithmic}{\endalgorithmic}
\renewenvironment{algorithmic}[1][0]{%
	\hrulefill\par
	\oldalgorithmic[#1]}
{\endoldalgorithmic\par
	\vspace*{-.5\baselineskip}
	\hrulefill\par
}
\algrenewcommand\algorithmicrequire{\textbf{Input:}}
\algrenewcommand\algorithmicensure{\textbf{Output:}}
\algnewcommand\algorithmicinput{\textbf{Input:}}
\algnewcommand\algorithmicoutput{\textbf{Output:}}
\algnewcommand\Input{\item[\algorithmicinput]}%
\algnewcommand\Output{\item[\algorithmicoutput]}%

\newacronym{auprc}{AUPRC}{Area under Precision-Recall curve}
\newacronym{pr}{PR}{\emph{Precision-Recall}}
\newacronym{auroc}{AUROC}{Area under Receiver Operating Characteristic curve}
\newacronym{roc}{ROC}{\emph{Receiver Operating Characteristic}}
\newacronym{ai}{AI}{Artificial intelligence}
\newacronym{ml}{ML}{Machine learning}
\newacronym{url}{URL}{Uniform Resource Locator}
\newacronym{sld}{SLD}{second-level domain}
\newacronym{cc}{C\&C}{Command and control}
\newacronym{iid}{i.i.d.}{independent and identically distributed}
\newacronym{mil}{MIL}{Multi-instance learning}
\newacronym{hmill}{HMill}{Hierarchical Multi-instance Learning Library}
\newacronym{lsass}{LSASS}{Local Security Authority Subsystem Service}
\newacronym{m2m}{M2M}{many-to-many}
\newacronym{m2o}{M2O}{many-to-one}
\newacronym{iot}{IoT}{Internet of Things}
\newacronym{mac}{MAC}{Media Access Control}
\glsunset{mac}
\newacronym{map}{MAP}{maximum a posteriori}
\newacronym{gnn}{GNN}{Graph neural network}
\newacronym{ip}{IP}{Internet Protocol}
\glsunset{ip}
\newacronym{idp}{IDP}{Identity Protection}
\newacronym{json}{JSON}{JavaScript Object Notation}
\newacronym{xml}{XML}{eXtensible Markup Language}
\newacronym{dd}{DD}{Diverse Density}
\newacronym{em}{EM}{Expectation-Maximization}
\newacronym{svm}{SVM}{Support Vector Machine}
\newacronym{is}{IS}{Instance-space}
\newacronym{ddos}{DDoS}{Distributed Denial-of-Service}
\newacronym{bs}{BS}{Bag-space}
\newacronym{es}{ES}{Embedded-space}
\newacronym{ssa}{SSA}{Static Single Assignment}
\newacronym{arp}{ARP}{Address Resolution Protocol}
\glsunset{arp}
\newacronym{dns}{DNS}{multicast Domain Name System}
\glsunset{dns}
\newacronym{ssdp}{SSDP}{Simple Service Discovery Protocol}
\glsunset{ssdp}
\newacronym{mdns}{mDNS}{multicast Domain Name System}
\glsunset{mdns}
\newacronym{dhcp}{DHCP}{Dynamic Host Configuration Protocol}
\glsunset{dhcp}
\newacronym{upnp}{UPnP}{Universal Plug and Play}
\glsunset{upnp}
\newacronym{rbf}{RBF}{Radial Basis Function}
\newacronym{tls}{TLS}{Transport Layer Security}
\glsunset{tls}
\newacronym{http}{HTTP}{Hypertext Transfer Protocol}
\glsunset{http}
\newacronym{https}{HTTPS}{Hypertext Transfer Protocol Secure}
\glsunset{https}
\newacronym{pua}{PUA}{Potentially unwanted application}
\newacronym{lldp}{LLDP}{Link Layer Discovery Protocol}
\glsunset{lldp}
\newacronym{mrf}{MRF}{Markov random field}
\newacronym{ptp}{PTP}{Probabilistic threat propagation}
\newacronym{tcp}{TCP}{Transmission Control Protocol}
 
\makeglossaries%

\newcommand\given[1][]{\:#1\vert\:}

\newcommand*{\blankpage}{%
	\thispagestyle{empty}
	\vspace*{\fill}
	{\centering\small This page is intentionally left blank.\par}
	\vspace*{\fill}\clearpage
}

\let\originalleft\left
\let\originalright\right
\renewcommand{\left}{\mathopen{}\mathclose\bgroup\originalleft}
\renewcommand{\right}{\aftergroup\egroup\originalright}

\DeclareMathOperator*{\argmax}{arg\,max}

\DeclareMathOperator{\mean}{mean}
\DeclareMathOperator{\pnorm}{\mathnormal{p}-norm}
\DeclareMathOperator{\lse}{lse}
\DeclareMathOperator{\relu}{ReLu}
\DeclareMathOperator{\softmax}{softmax}
\DeclareMathOperator{\l1p}{log1p}
\DeclareMathOperator{\pop}{pop}
\DeclareMathOperator{\push}{push}
\DeclareMathOperator{\myroot}{root}

\DeclareMathOperator{\children}{children}
\DeclareMathOperator{\type}{type}

\theoremstyle{definition}
\newtheorem{statement}{Statement}
\theoremstyle{plain}
\newtheorem{theorem}{Theorem}
\newtheorem{definition}{Definition}
\theoremstyle{remark}

\newtheorem*{myproof*}{Proof}

\usetikzlibrary{arrows.meta}

\tikzstyle{genericarrow}=[cvutaccented, -latex]
\tikzstyle{modelarrow}=[cvutaccented, -{Computer Modern Rightarrow[scale width=0.65, scale length=1.5]}]
\tikzstyle{ellipsearrow}=[black, dotted]
\tikzstyle{grapharrow}=[cvutaccented, -{Straight Barb[scale width=0.75]}]
\tikzstyle{grapharrowhuge}=[cvutaccented, thick, -{Straight Barb[scale width=2, scale length=2]}]

\usepackage{amsmath, amssymb}

\DeclareMathOperator{\pdnode}{pn}
\DeclareMathOperator{\bdnode}{bn}
\DeclareMathOperator{\adnode}{an}
\DeclareMathOperator{\pmnode}{pm}
\DeclareMathOperator{\bmnode}{bm}
\DeclareMathOperator{\amnode}{am}
\DeclareMathOperator{\psnode}{ps}
\DeclareMathOperator{\bsnode}{bs}
\DeclareMathOperator{\asnode}{as}

\begin{document}

\pagestyle{beginning}

\ifprint\blankpage\fi%
\ifprint\blankpage\fi%

\begin{titlepage}
    \vspace*{12em}
    \centering {\color{black}\textbf{\Huge{Mapping the Internet: Modelling Entity Interactions in Complex Heterogeneous Networks}}}\\
    \vspace*{4em}
    \centering \LARGE{Master's thesis}\\
    \vspace*{4em}
    \centering \textbf{\LARGE{Bc. Šimon Mandlík}}\\
    \centering \LARGE{Supervisor: doc. Ing. Tomáš Pevný, Ph.D}\\
    \vspace*{3em}
    \centering \LARGE{2020}\\
    \vfill
    \centering \includegraphics[scale=0.5]{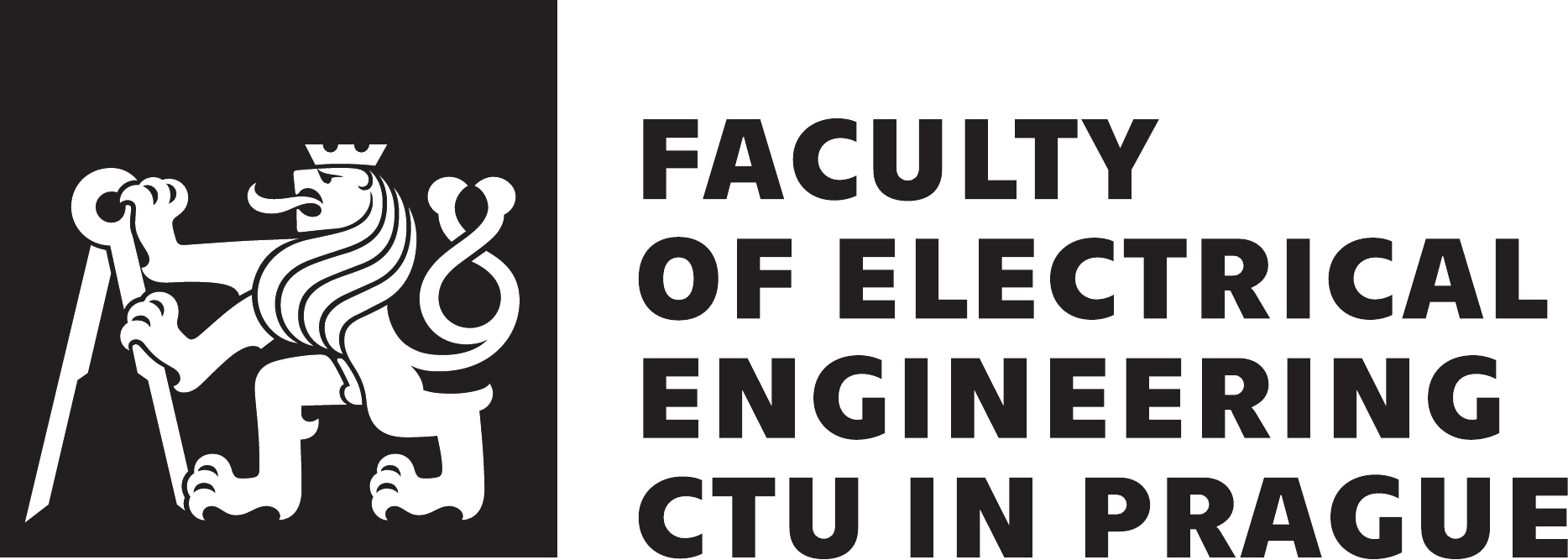}\\
    \vspace*{1em}
    \centering Department of Computer Science
    \thispagestyle{empty}
\end{titlepage}

\ifprint\blankpage\fi%

\pagenumbering{Roman}


\newpage

\ifprint\blankpage\fi%

\vspace*{\fill}

\noindent {\huge \color{black}\textbf{Abstract}} 	
\vspace*{2em}

\noindent 
Even though machine learning algorithms already play a significant role in data science, many current methods pose unrealistic assumptions on input data. The application of such methods is difficult due to incompatible data formats, or heterogeneous, hierarchical or entirely missing data fragments in the dataset. As a solution, we propose a versatile, unified framework called `HMill' (Hierarchical multi-instance learning library) for sample representation, model definition and training, which addresses the discussed problems and meets all requirements for a modern general-purpose instrument. We review in depth the multi-instance paradigm for machine learning that the framework builds on and extends. To theoretically justify the design of key components of HMill, we show an extension of the universal approximation theorem to the set of all functions realized by models implemented in the framework. The text also contains a detailed discussion on technicalities and performance improvements in our implementation, which is published for download under the MIT License. The main asset of the framework is its flexibility, which makes modelling of diverse real-world data sources with the same tool possible. This is done with only minor changes in the pipeline and requires neither further specialization nor performance compromises. Additionally to the standard setting in which a set of attributes is observed for each object individually, we explain how message-passing inference in graphs that represent whole systems of objects can be implemented in the framework. To support our claims, we solve three different problems from the cybersecurity domain using the framework. The first use case concerns IoT device identification from raw network observations. In the second problem, we study how malicious binary files can be classified using a snapshot of the operating system represented as a directed graph. The last provided example is a task of domain blacklist extension through modelling interactions between entities in the network. In all three problems, the solution based on the proposed framework achieves performance comparable to specialized approaches.

\vspace{2em}


\vspace*{\fill}

\newpage

\pagestyle{beginning}
\tocloftpagestyle{beginning}
\tableofcontents
\thispagestyle{beginning}
\newpage
\begingroup
\listoffigures
\listoftables
\renewcommand{\cleardoublepage}{}
\renewcommand{\clearpage}{}
\endgroup

\newpage

\ifprint\blankpage\fi%

\pagestyle{rest}
\pagenumbering{arabic}
\setcounter{page}{1}


\chapter{On challenges posed by modern data sources}%
\label{cha:difficulties}

In recent years, the world has seen a tremendous number of successful applications of \gls{ai} to create value in different sectors. Specifically, in natural language processing, new pre-trained language models~\cite{Devlin2018, Radford2018} significantly improved the accuracy of systems for machine translation, question answering and text summarization. Due to recent advances in speech synthesis~\cite{VanDenOord2016, Amodei2016}, virtual assistants no longer have to rely on soundbanks to produce speech. In computer vision, older black and white movies or photos can now be instantly enhanced through automatic color restoration~\cite{Iizuka2016}. We have seen algorithms capable of plausible reenactment of famous people, synthesizing a completely new video of their talk with precise lip-syncing given only audio as input~\cite{Suwajanakorn2017}. For the inverse task, lipreading models outperforming humans were also developed~\cite{Assael2016}. There are many other fields, that currently employ \gls{ai} techniques to achieve unparalleled results and examples provided here are just a tip of the iceberg.

As impressive as the results oftentimes seem, all of the methods mentioned above are instances of \emph{statistical machine learning} subfield of \gls{ai}, and the whole procedure can be simplified down to a few basic steps. First, a \emph{model} is initialized as an instance of a pre-defined parametrized family of functions that is believed to accurately encompass an unknown distribution of the data. A suitable \emph{loss function} is then defined to bridge the gap between what a human would consider a well-performing system and what can be rigorously expressed in a mathematical language. After that, observations are repeatedly sampled from the dataset, and the model parameters are altered in such a way that the value of the loss function decreases. Various methods  differ in their choice of model architecture, loss function, or training procedure, yet they all have one thing in common---they rely on huge datasets available for their training and large-scale computation that has only recently become attainable. In his essay called \emph{The Bitter Lesson}~\cite{sutton_2019}, Richard S. Sutton reflects, that ``general methods that leverage computation are ultimately the most effective, and by a large margin''. He points out that AI researchers tend to build human knowledge of the domain into their systems, which reaps immediate reward, but eventually plateaus and hinders further progress in the long term. Community is encouraged to focus instead on developing general-purpose meta-methods capable of finding good approximations of the complex world around us by themselves through computation and whose performance further improves with more powerful hardware and larger datasets. 

In history, there are many examples of a clash of these two methodologies. For two-player games such as chess and Go, superhuman performance was achieved by employing massive state space search and self-play~\cite{CAMPBELL200257, Silver1140} after previous approaches based on special structure and features of the game identified by humans faltered. In computer vision, researchers used to define image features based on their belief of how the human mind works, which led, for instance, to SIFT features~\cite{Lowe2001} or predictors based on edge detection technology. However, it turns out that a general low-level convolution operation used in convolutional neural nets coupled with large training dataset and computation leverage was the only thing needed to outperform all prior methods significantly~\cite{AlexNet}.

Last decade saw unprecedented growth in the volume of available data, mainly thanks to an expansion of the Internet and mobile devices as well as recent developments in Web technologies. In 2013, it was estimated that the number of connected devices would increase three times by 2020, reaching more than 30 billion devices worldwide~\cite{ABI}. Facebook generates four petabytes of new data every day, out of which 350 million are photos~\cite{FB}, and Twitter processes 500 million posts per day, leading to around 6 thousand tweets on average every second~\cite{TW}. This trend is nowadays referred to as \emph{big data}~\cite{Gartner2012} and is identified by three main characteristics---\emph{volume}, \emph{velocity} and \emph{veracity}. It is a common belief that with carefully implemented privacy measures, efficiently processed big data will benefit the whole society in many ways, ranging from medical informatics to improved fraud detection and cybersecurity. As the sudden surge of the remarkable \gls{ai} breakthroughs was possible thanks to the increasing amount of data available in the majority of application domains, many people maintain that the humanity is on the verge of another industrial revolution, this time featuring information and knowledge as the primary commodity. Unfortunately, we still lack utterly bulletproof policies for processing of personal data and also sufficiently general methods to deal with the massive amount of data. In this thesis, we address the latter issue and propose a unified framework for learning from diverse real-world data. The following section goes into detail and elaborates further why we believe new approaches are needed.

\section{Pitfalls of real-world datasets}%
\label{sec:pitfalls}

Recently, many works discussed the promises of big data and how its properties present new challenges for current \gls{ml} systems~\cite{Zhou2014, Najafabadi2015, LHeureux2017, Dulac-Arnold2019, Amit2019, Malik2020}. What follows is a compiled list of what we identified as potential problems for current pipelines and focused on in this thesis. The emphasis is put on \emph{deep learning}-based methods, however, most of the reasoning applies also to different approaches to machine learning. We will use words \emph{object}, \emph{individual}, \emph{sample}, or \emph{observation} to refer to all possible real-world entities (executable files, photos, domains) we are dealing with. We provide specific examples mainly from the cybersecurity domain, which this work focuses on and from which all tasks solved in the experiment sections originate. 

\subsection{Independence assumption}%
\label{sub:independence_assumption}

    The classical statistical machine learning paradigm defines the so-called \gls{iid} data assumption stating that there exists an underlying distribution from which all data is independently sampled, and therefore all objects from the dataset are identically distributed~\cite{fukunaga1990, Webb2003, Bishop2006}. This is further utilized when obtaining the \emph{train-test} split of the data, where one part of the dataset is held out from the training phase and evaluated on afterwards in order to estimate the generalizing capabilities of the model. Dynamic real-world systems such as internet networks consist of a large number of interdependent actors, which influence each other on a regular basis, thus reasoning about one individual without including knowledge about others seems like a missed opportunity. Moreover, it is not clear how to disentangle the interdependent system of objects of interest into the training and testing sets so that the true distribution is preserved in both of them.

For example, consider a task of malicious domain classification, in which we observe a computer network and attempt to pronounce each domain that users connect to as malicious or benign. One of the possible approaches for solving this problem is to design a set of features describing a single domain and train one of many off-the-shelf classifier implementations. Examples of such features are $ n $-gram based representation of the domain's \gls{url} or data extracted from the response to a WHOIS query. Nevertheless, a more accurate description of the domain will consider other entities in the network. For instance, it is common that malicious software or malicious servers themselves connect to \gls{cc} servers to receive further instructions. The fact that an object in question connected to such server is a powerful indicator of maliciousness, hence it seems sensible to also include information about outgoing connections to other domains and not only features describing the object itself. Another example, where \gls{iid} assumption does not hold, are polymorphic variations of the same malware, hardware from the same vendor, or hosts in the same network, all of which are very dependent. In all these examples, statistical guarantees provided under a condition that \gls{iid} assumption holds are rendered void.

On the other side of the spectrum are graphical models~\cite{Jordan_graphical}, which aim to represent a joint probability distribution $ p(X) $ of a set of random variables $ X = \left\{ X_1, X_2, \ldots, X_n \right\} $ describing all $ n $ observed objects in the system. The distribution is encoded into a graph, whose vertices represent random variables and edges relationships between them. To represent a real-world system as a graphical model, for every sample we add a variable, which takes values depending on what we want to learn (real numbers for regression tasks, discrete set of values for classification into a finite number of classes, and so forth). Graphical models offer high modelling power at the cost of limited method options for computing marginal probabilities or other inference problems. Exact inference algorithms exist only for special classes of graphs~\cite{Baum1966, Pearl1988}, however, for general graphs the inference is known to be NP-hard. 

Ideally, the modern real-world data processing systems should be able to smoothly interpolate between these two cases by including some knowledge about dependences between samples in order to obtain high expressive power while keeping computational complexity and convergence speed low.

\subsection{Feature engineering and sample representation}%
\label{sub:feature_engineering}

Most machine learning algorithms expect each sample to be represented as a real feature vector $ \bm{x} \in \mathbb{R}^n $ in some real vector space of fixed (finite) dimension $ n $. As $ n $ grows, the maximal possible representation power and the capacity of models increase, nevertheless, the number of model parameters grows as well, which may lead to overparametrization. This vector-of-fixed-size requirement may be too constraining. To further illustrate this, we refer the reader to the discussion about sample interdependence in the previous paragraph. Making a prediction about an object while including knowledge about other objects it depends on means that the amount of information to convey to the model is an increasing function of the number of such related objects. For an observation purely independent of others, this number will be zero, however, in practice, this number may take value from a broad range of values without any obvious upper bound. Therefore, fixing the dimension of the feature vector to some constant seems too limiting.

Additionally, this formulation relies not only on the existence of mapping from samples to feature vectors but also on the fact that we know it or are at least able to sufficiently approximate it. For instance, when performing machine learning on images, it is apparent that the appropriate representations consist of raw pixel values since this is the only input humans also receive and still identify images reliably. Feeding raw observations straight to the model and letting it decide what information to keep and what information to discard is the ideal situation, nevertheless, what is the suitable feature mapping for a web domain or a mobile device? Regrettably, there is no universal answer nor any rule of thumb specifying how to design feature mappings, as empirical evidence suggests that optimal mappings are both data and problem dependent. \emph{Feature engineering} is a process of discovering discriminative features, which often includes a great deal of domain expertise, and is thus deemed as one of the most time-consuming tasks in \gls{ml} pipeline. If the domain knowledge is not sufficient, \emph{feature selection}~\cite{Somol10, feature_selection1} that aims to select the most relevant features comes to rescue. Feature selection improves on a simple trial-and-error selection procedure by using statistical measurements and tests. Even though feature selection may reduce the dimension of the feature vector, it becomes challenging in high dimensions due to spurious correlations. For a long time, the reasonable guiding principle was to define the feature space large enough to properly discriminate, and use feature selection to find a suitable subspace with low redundancy and noise in the data. In recent times methods such as neural networks able to implicitly perform this selection appeared, however, they are constrained in types of input data.

Many real-world sources of data exhibit signs of the \emph{concept drift}, which occurs when the distribution of the data changes in time. As a result, a model can soon become obsolete and degrade in performance, even though it performed well at the time of training. As the world evolves, there will always be a danger of model aging, and because machine learning is designed to learn from past experiences and not predict the future data distribution changes, we have to resort to retraining the whole model. This is possible as long as the appropriate feature set does not change over time, and the only change happens in underlying distribution over the selected features. Once some of the features lose their relevance and on the other hand, other previously discarded features gain in importance, sole model retraining does not help. This leads to an obvious requirement that the whole pipeline should be able to adapt to such changes in a short time without human intervention in an automated fashion. 

A specific example, where we might encounter such problems, is a device identification task from the \gls{iot} domain. In this problem, we aim to recognize a type of an \gls{iot} device using a set of low-level measurements obtainable in the network. Apart from querying \gls{mac} and \gls{ip} addresses of the device, which can be encoded to a fixed size vector, it is recommended to observe the behavior of the device on the network and scan all its opened ports, collect its responses to various queries, and store information broadcasted by the device across the network. As different devices may expose a different number of ports and provide different services, storing this information in a vector of fixed length will likely lead either to overparametrization, when the vector dimension is unnecessarily high, or information loss, if the dimension is not high enough. The other option is not to describe each device as a flat vector, but rather utilize any existing data format capable of encoding structure in observations. As we demonstrate later in this work, the JSON format is very well suited for capturing information like a list of open ports or responses. Nevertheless, how to construct models able to process such input?

Moreover, imagine a (hypothetical) scenario in which we discover that voice assistants of a certain vendor always use the same rather unusual port for communication. In this case, one feature of a classifier or just a rule in a rule-based system could be based on this fact and we would probably achieve good performance. However, what if the vendor changes the implementation, or other devices start using this port? In the worst example, all such specific rules or features lose relevance over time, and need to be rewritten or completely redesigned again by a human. We further discuss the task of device identification in Chapter~\ref{cha:jsons}. 

The main takeaways of this discussion are that modern \gls{ml} systems should cope with non-euclidean data as input and be capable of automatic feature selection from the non-euclidean data in a way neural networks do it for fixed-size vectors. We advocate the resulting model working on raw data with little to no preprocessing or aggregation, as it is the case, for example, in the computer vision domain. Of course, the input data eventually must be aggregated if models are to output only one or several numbers, however, it should be decided by models that can effectively learn how and when to do it in the pipeline as opposed to humans. This is possible with increasing compute and in line with the discussion presented at the beginning of the chapter. In the end, dealing with the concept drift in rapidly changing environments would be solved by fully automated end-to-end retraining of the model on new data. Effectively, resources can be allocated to experimenting with many different data sources, instead of focusing on how to transform them appropriately and/or modify existing algorithms to handle their data formats.

\subsection{Scalability and composability}%
\label{sub:scalability_and_composability}

Another problem of growing importance is the \emph{combinatorial explosion} in the size of datasets, which are needed to describe complexities of real-world phenomena. For instance, by rearranging objects in an image scene or adding occluding objects, current models are often confused~\cite{Wang2017}. A lot of research effort has also been put into discovering adversarial samples and training models robust to them~\cite{yuan2017adversarial}. Even though humans are able to naturally adapt to such changes and extrapolate outside of the training data, this appears not to be the case with the present state-of-the-art models. Consequently, to solve this problem with current tools the training data should contain all possible changes to visual context leading to exponentially large datasets.

At the beginning of this chapter, we mentioned that one of the characteristics of big data is its massive volume, and one may be misled into thinking that enough data to solve these issues is or eventually will be available. However, a lot of deceptive samples are observed with low probability and on the other hand many `common' samples, which the model predicted correctly, are of high prevalence. Therefore, it is difficult to train accurate models with naïve random selection of training samples in reasonable (finite) time. Methods for selecting more representative data have been studied, for example, in~\cite{vapnikteacher}, but they are not frequently used in practice.

\emph{Adversarially robust} training methods have been recently proposed, which train models also on synthetic adversarial samples. This is usually to ensure that small modifications of input, such as adding random noise, do not change the prediction. Unfortunately, this is not a solution to the problem of exponentially large dataset required to capture the intrinsically complex world.

One of the proposed solutions to this is making models \emph{hierarchical} (composable) with the core assumption that real-world complex structures are decomposable into smaller elementary substructures at a finer level of granularity and one can reason about the world by deliberately combining and dismantling blocks of information in a hierarchical way. By introducing this explicit, yet general purpose requirement into the model architecture, it attempts to learn the structure too. It has already been shown that deep learning architectures extract increasingly more complex abstract representations of the data with each layer~\cite{zhou2014object}, however here we require an explicit decomposition into structures and substructures. Thanks to the explicit structure, hierarchical models are easily interpretable, less susceptible to adversarial samples and due to the decomposition, generating new samples is more straightforward. Moreover, (sub)structures tailored for one problem can be further reused in other problems.

Finally, even though hierarchical models allow to better learn real-world phenomena with a smaller number of observations, a new problem of inherent introduction of bias by selecting a specific hierarchical architecture arises. We address this problem in the rest of the thesis and claim that this structure is often naturally encoded in the data itself.

\subsection{Heterogeneous, hierarchical, and missing data}%
\label{sub:heterogeneous_and_hierarchical_data}

Dealing with heterogeneous data involves aggregating data from several different sources. Based on the origin of heterogeneity, we distinguish the following two types~\cite{LHeureux2017}:

\begin{description}
    \item[Syntactic heterogeneity] occurs when incoming data is not encoded in the same way or is of different data types. Current models are not designed to cope with these syntactical differences and perform poorly when asked to simultaneously process inputs of different encodings, such as categorical and numerical data, together.
    \item[Semantic heterogeneity] refers to the case when input data comes from different sources. We will also refer to this as \emph{multi-modal} data coming from diverse \emph{modalities}. Because semantically heterogeneous data has a different interpretation, the underlying distributions may be different, leading to substantial differences in statistical properties of the data across the dataset.
\end{description}

\noindent
The recent adoption of Web technologies, where information is communicated in a hierarchically structured way, led to the origin of various data interchange formats, such as the aforementioned \gls{json} or \gls{xml}. The ubiquity of data formatted in this way also puts pressure on the current methods, which are not designed to process hierarchical data. As opposed to imposing an explicit structure to the model, which was discussed in the last paragraph, this does not introduce any prior on the form of the hypothesis, but rather provides the opportunity to leverage information encoded both in the raw values as well as in structure of the data, leading to more flexible representations of samples. 

Finally, it is often the case that the data is not collected directly at its source. This happens when the dataset is generated by a system, whose properties we want to learn, and is somehow transformed before being observed resulting in information loss and/or addition of noise. Examples of such transformation include irreversible compression, network transportation or data aggregation. A specific case is when parts of the data are missing entirely. This usually occurs when fragments of the data that are missing exist but are not observable, or there is a limited budget of time or resources for obtaining the data. The case when the source of data can be queried at some cost for specific features has been studied~\cite{Janisch2019, Janisch_2020}, however the task is different once the source cannot be manipulated with and hence the observed data is fixed. Present approaches are not developed with this phenomena in mind, and unprincipled, weakly justified ad-hoc solutions are employed.

In the last part of this thesis, we show how a problem of uncovering additional malicious domains in the computer network given a blacklist of existing threats can be solved by simply observing binary relations between domains and other network actors. For example, we can have a binary relation between domains and clients connected to the network, which encodes that a client has connected to a domain. Alternatively, we can define a relation between domains and binary files that indicates that a running instance of a binary has issued a request to a domain. Data of this type is fairly simple to assemble, however, learning from it is not straightforward since binary relations described above are inherently heterogeneous (following completely different distributions), emit weak signals due to their simple definition, and sometimes a relation may not describe a domain at all due to missing information.

To sum up, we add another requirement on modern \gls{ml} approaches to our list, which is to accept multi-modal heterogeneous data from different sources and at different levels of granularity, and still automatically recognize and distill relevant knowledge. Modern algorithms should deal with imprecise observations and missing data in a principled way.

\subsection{Explainability and interpretability}%
\label{sub:explainability}

In many application areas, the strong incentive is to build accurate \gls{ml} models to help human operators either by simplifying rule-based systems previously designed by hand, offloading the most repetitive and time-consuming tasks, or leaving the human entirely out of the loop. Even though offloading work from one's shoulder by systems that have been repeatedly shown to perform better than humans in many tasks seems tempting, there is a major caveat. Modern \gls{ml} algorithms indicate the existence of a tradeoff between interpretability and performance, with deep learning usually being the most performant and at the same time the least interpretable of all. Another example are decision tree-based models, which work with the prior assumption that the decisions can be made in a hierarchical manner by asking simpler questions at each layer of the decision tree and gradually specifying the corresponding output. This hierarchical decomposition is similar to what we justified in the previous paragraph, and the output of a tree-based model is indeed a sequence of decisions similar to how humans reason, however, it was discovered that random forests, which train an ensemble of decision trees together with some randomization involved, outperform a single tree significantly~\cite{rfs}. At the same time, an ensemble of decision trees is much less trivial to explain.

The \emph{Occam's razor} principle implies that methods able to appropriately explain their decisions in simple terms are more likely to generalize well beyond the training data since they seek less complex patterns. Hence, insufficient explainability is currently one of the most criticized limitations of complex models. There are industry sectors, for example, fraud detection systems, in which proper explanation of the decision is crucial to verify system's soundness and gain insights into the origin of potential errors, thus regarding the model simply as a magical `black-box' is not possible. Another exemplar domain is cybersecurity, where a system for filtering out the majority of benign objects in question and discovering threats would be very welcomed by network analysts. The trustworthiness of such discoveries is notably increased if the system also provides an explanation of its decision.

To conclude, modern \gls{ml} systems should strive to achieve explainability in a natural way while keeping the performance high and retaining complexity. Therefore, we advocate approaches that are able to consistently pinpoint and present the main reasons why the decision was made and filtering out negligible nuances.

\section{Thesis structure}%
\label{sec:thesis_structure}

This thesis is dedicated to introducing a versatile and theoretically justified framework called \gls{hmill} for data representation, model definitions and subsequent training. The framework is built upon recent findings in deep learning and multi-instance learning and addresses the aforementioned problems in a principled way. The second chapter revises prior art on multi-instance learning starting from the earliest approaches and ending with the present-day techniques, and provides further motivation for following the multi-instance paradigm. The third chapter is the crux of this thesis, as the framework is formally introduced there and its theoretical properties are investigated. More specific details together with implementation tricks follow in the fourth chapter. In the rest of the thesis, we show how the framework can be used to model diverse and complex data sources with properties discussed in this introductory chapter. Specifically, in the fifth chapter, we deal with \gls{json} documents, in the sixth chapter with graphs with heterogeneous nodes and edges encoding interactions between objects, and in the seventh chapter with binary relations encoded as several bipartite graphs. The thesis is concluded with a chapter containing reflection on the importance of the framework in modern \gls{ml} pipeline and a summary of our future directions.

We have chosen cybersecurity as our `playground' domain since it presents all the challenges we discussed---we deal with typically discrete information at large volumes most of which is irrelevant or uninformative and no canonical and straightforward feature mappings are known. Moreover, we observe whole systems (networks, the Internet, operating systems) of interdependent objects, and the environments change rapidly in the presence of the concept drift. All of these phenomena make the application of machine learning techniques complicated. In the course of the thesis, we introduce three different real-world cybersecurity tasks we solved successfully with the framework. The first problem considers IoT device identification, in which we aim to classify an IoT device into several classes based on information available in the network. In the second problem, we classify binary files in the personal computer based on a snapshot of the operating system represented in a graphical form. The last and perhaps most specific use case, on the basis of which we entitled this thesis, is modelling the computer network based only on interactions between objects in the network encoded in several heterogeneous bipartite graphs. All experiments would not be possible without our cooperation with Avast Software and Cisco Cognitive Intelligence companies, who kindly provided us with data for experimental purposes.

As the main strength of the framework lies in its high modelling flexibility and overall versatility, we assume that the proposed approach to data processing will unlock a new realm of possibilities mainly for data from modern sources, for which we lack suitable technology so far. Taking into account the fact that the framework is easy and straightforward to use, and pipelines for completely different tasks (for instance three tasks solved here) are nearly identical, we are confident that the \gls{hmill} framework will find its use as a universal instrument in other domains.


\ifprint\newpage\blankpage\fi%
\chapter{Multi-Instance Learning}%
\label{cha:mil}

This chapter contains a review of prior art in multi-instance learning, which is essential for the rest of this thesis.
\newline\newline\noindent
\gls{mil} (or Multiple instance learning) is a novel machine learning paradigm first introduced in~\cite{Dietterich1997}. As opposed to standard machine learning, where each sample is represented by a fixed vector $ \bm{x} $ of values, in \gls{mil} we describe each sample with a \emph{set} of vectors. In \gls{mil} nomenclature, this set is called \emph{bag} and vectors it is composed of are called \emph{instances}. Instances live in an \emph{instance space} $ \mathcal{X}$, which can, for example, be an $n$-dimensional Euclidean space. Bags come from \emph{bag space} $ \mathcal{B} = \mathcal{P}_F(\mathcal{X}) $, where $ \mathcal{P}_F(\mathcal{X}) $ denotes all finite subsets of $ \mathcal{X} $. A bag $ b $ can be therefore written down as $ b = \left\{ \bm{x} \in \mathcal{X} \right\}_{\bm{x} \in b}$. Each bag can be arbitrarily large and also empty, therefore the size of bag $ b $, which we will further write as $ \lvert b \rvert  $, can be zero or any natural number. Note that the classical machine learning task is a special instance of the \gls{mil} problem, where for bag representation $ b $ of each sample holds $ \lvert b \rvert  = 1 $. Furthermore, it is also assumed that even though intrinsic labeling of instances may exist, we observe and are interested in predicting only labels at the higher level of bags. This defines one of the main challenges of \gls{mil} problems, which lies in recognition of common patterns at the level of instances and their suitable subsequent aggregation at the bag level.

\begin{figure}[hbtp]
    \begin{subfigure}[t]{.47\textwidth}
        \centering
        \raisebox{0.265cm}{\includegraphics[width=\textwidth]{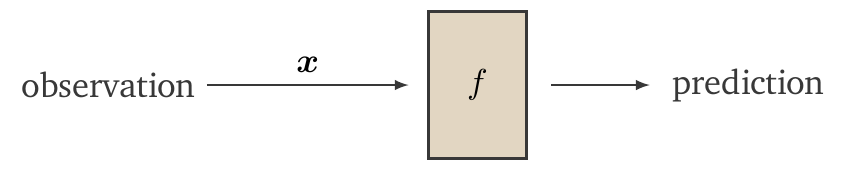}}
        \caption{standard machine learning}%
    \end{subfigure}
    \hfill
    \begin{subfigure}[t]{.47\textwidth}
        \centering
        \includegraphics[width=\textwidth]{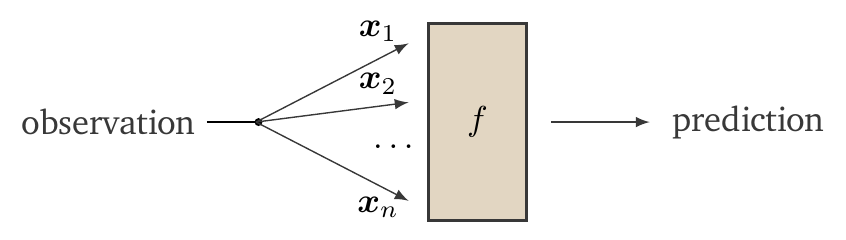}
        \caption{multi-instance learning}%
    \end{subfigure}
    \caption{Difference between standard machine learning and multi-instance learning.}%
    \label{fig:mil}
\end{figure}

In this thesis, we focus classification problems, where bag labels come from a finite set $ \mathcal{C} $, to simplify explanation and notation, however, all models can be modified to perform other tasks such as regression as well. In \gls{mil}, instead of learning a predictor in the form $ f\colon \mathbb{R}^n \to \mathcal{C} $, models are defined as $ f\colon \mathcal{B}\left( \mathcal{X} \right)  \to \mathcal{C}$, in other words, the prediction can be written as $ f(\left\{ \bm{x} \right\}_{\bm{x} \in b} ) $. The difference is visually demonstrated in Figure~\ref{fig:mil}. Additionally, we will consider only \emph{supervised} setting, in which each sample in the dataset is attributed a label, and we will denote the available data by $\mathcal{D} = \bigl\{ \left( b_i, y_i\right) \in \mathcal{B} \times \mathcal{C} \bigm| i \in \left\{1, 2, \ldots, \lvert \mathcal{D} \rvert \right\}   \bigr\} $. We propose to use \gls{mil} formulation as a solution to problems outlined in Chapter~\ref{cha:difficulties}, mainly related to questions of sample representation.

One of the problems that \gls{mil} formulation naturally lends itself to is drug activity prediction in biochemistry. The objective is to decide whether a drug molecule binds strongly to a target site in a much larger protein molecule, which determines the potency of the drug. Unfortunately, drug molecules may take different shapes, each with different binding strengths. Hence, positively tested drug indicates that \emph{one of its} molecule shapes successfully bound with the protein, contrary to the negative test result, which means that \emph{none of} the shapes in a drug sample bound. Moreover, in the positive case, there is no way to determine which exact shape has strong binding potential. A reasonable representation of a molecule is, therefore, a bag with each of its instances defining one of the possible shapes. Another example, where \gls{mil} formulation can be applied, is the cybersecurity domain, which we focus on in this thesis.

In pioneering works~\cite{Dietterich1997, Maron1998}, the \emph{law of inheritance} was adopted, meaning that the whole bag is labeled with a certain label provided it contains an instance with this label. This means, that a bag label can be inferred based on observing only one instance. For binary classification, this translates to a bag being pronounced as positive as long as it contains at least one positive instance and negative otherwise. Nonetheless, this assumption was later relaxed, the requirement on the existence of instance labels was dropped, and it is additionally expected that some labels are inferable from not one but multiple instances only. Still, there may be plenty of instances in the bag that bring little to no insight and can be discarded by the model. All approaches to solving \gls{mil} tasks were categorized in~\cite{Amores2013} into three main paradigms---\emph{instance-space} paradigm, \emph{bag-space paradigm} and \emph{embedded-space paradigm}.

\section{Instance-space paradigm}%
\label{ssub:instance_space_paradigm}

\gls{is} paradigm appeared as the first approach to solving \gls{mil} problems. It is based on the belief that all the information, which is required to make a decision, is specified locally with respect to instances. Hence, no characteristics at the level of bags are needed in the learning algorithm. Assuming that each instance can be attributed one class from a set $ \mathcal{C}_{I} $, an instance-level classifier $ f_I \colon \mathcal{X} \to \mathcal{C}_{I} $ is learned separately and a bag label is constructed from responses of all its instances with a pre-defined aggregating function $ g $:
\begin{equation}
    \label{eq:is_paradigm}
    f(b) = g\left( \left\{ f_I\left( \bm{x} \right) \right\}_{\bm{x} \in b}  \right)
\end{equation}
Binary classification, where $ \mathcal{C}_{I} = \left\{ -1, 1 \right\}  $ making arithmetics on the label values possible, is considered in most of the early works on \gls{mil} problems. Note that contrary to known bag labels, instance labels may not be directly observable and therefore further assumption of how instance and bag labels are related must be made in order to infer $ f_I $. According to this relationship and the definition of aggregating function $ g $, we distinguish methods based on \emph{Standard assumption} and \emph{Collective assumption}.

\subsection{Standard assumption}%
\label{sub:standard_assumption}
Standard assumption defines $ g $ as either $ \max $ function $ g\left( \left\{ f_I\left( \bm{x} \right) \right\}_{\bm{x} \in b}  \right) = \max_{\bm{x} \in b} f_I({\bm{x}}) $ or its approximation. This implies that a bag label is oftentimes inferable merely from a strong response of instance classifier on one of its instances. For binary classification, this also includes the aforementioned assumption that positive bags contain at least one positive instance and negative bags none. This asymmetric condition is projected into methods via different handling of positive and negative samples. In~\cite{Dietterich1997}, authors demonstrate how the drug activity prediction problem described above can be solved with instance-space \gls{mil}. Hypothesis $ f_I $ takes form of an axis-parallel rectangle, which is either shrunk or grown appropriately to include positive instances and exclude negative instances, and $ g $ is defined as the standard $ \max $ function.

A \gls{dd}, introduced in~\cite{Maron1998}, is defined in instance feature space $ \mathcal{X} $ as a measure of how many positive bags contain instances that are near the given point and how distant are all instances from negative bags. Assuming that the underlying concept is a single point $ \bm{t} \in \mathcal{X} $ in this space, one can find a hypothesis by maximizing \gls{dd}, e.g.~by gradient ascent. This procedure yields a point, where many positive bags have at least one instance close by and at the same time instances of negative bags are far away. Authors define a distance metric measuring closeness as the \emph{weighted Euclidean distance}:
\begin{equation}
    \label{eq:dd}
    d(\bm{t}, \bm{x}) = \sqrt{\sum_{i=1}^{D}s_i(t_i - x_i)} 
\end{equation}
where $ \bm{s} $ is a vector of learned parameters. In~\cite{Dooly2001}, authors extend \gls{dd} together with other methods to \gls{mil} on real-valued data. This is further being built on in~\cite{Zhang2002}, where authors combine this approach with \gls{em} algorithm. The key idea of EM-DD is to consider the instance responsible for a positive label of a bag a latent variable, which is used in the \gls{em} algorithm. Given an initial hypothesis $ \bm{t} $ in a feature space, EM-DD first selects the instance most likely responsible for the bag label in E-step, and then uses it in M-step to change hypothesis to maximize \gls{dd}.

A different approach based on modifying the standard soft-margin \gls{svm}~\cite{vapnik_svm} formulation is presented in~\cite{Andrews2003}. Denoting the binary label of a bag $ b $ as $ Y_b = \pm 1$, and the label of its instance $ \bm{x} $ as $ y_{\bm{x}} = \pm 1$, aggregating rule in the form of $ Y_b = \max_{\bm{x} \in b} y_{\bm{x}} $ can be reformulated as a set of linear constraints:
\begin{equation}
    \label{eq:svm}
    \begin{aligned}
        \sum_{\bm{x} \in b} \frac{1 + y_{\bm{x}}}{2} &\geq 1&\forall b\colon Y_b &= 1\\
        y_{\bm{x}} &= -1&\forall b\colon Y_b &= -1, \forall \bm{x} \in b
\end{aligned}
\end{equation}
Instance labels $ y_{\bm{x}} $ are treated as unobserved integer variables subjected to the constraint~\eqref{eq:svm}. The overall goal is to maximize the standard soft-margin on instances, leading to the following formulation in the primal form:
\begin{equation}
    \label{eq:svm_2}
\begin{aligned}
&\underset{(\bm{w}, k, \bm{\xi}, \bm{y})}{\mbox{minimize}} &\frac{1}{2}\|\bm{w}\|^{2}+C \sum_{i=1}^{N} \xi_{i} &&\\
&\mbox{subject to} &y_{\bm{x}_{i}}\left(\langle\bm{w} , \bm{x}_{i}\rangle+k\right) &\geq 1-\xi_{i}, &\forall i=1, \ldots, N\\
&&\xi_i &\ge 0&\forall i = 1, \ldots, N\\
&&\sum_{\bm{x}_{i} \in b} \frac{1 + y_{\bm{x}_{i}}}{2} &\geq 1&\forall b\colon Y_b = 1\\
&&y_{\bm{x}_{i}} &= -1&\forall b \colon Y_b = -1, \forall\bm{x}_{i} \in b
\end{aligned}
\end{equation}
where $ N $ denotes the number of all observed instances (over all bags). A byproduct is either a linear or non-linear instance classifier depending on which kernel is used. Additionally, authors propose another formulation, which maximizes the margin at the level of bags (see~\cite{Andrews2003} for details):
\begin{equation}
    \label{eq:svm_3}
\begin{aligned}
&\underset{(\bm{w}, k, \bm{\xi})}{\mbox{minimize}} &\frac{1}{2}\|\bm{w}\|^{2}+C \sum_{b}\xi_{b} &&\\
&\mbox{subject to} &Y_{b}\max_{\bm{x} \in b}\left(\langle\bm{w} , \bm{x}\rangle+k\right) &\geq 1-\xi_{b}, &\text{ for each bag } b\\
&&\xi_b &\ge 0&\text{ for each bag } b\\
\end{aligned}
\end{equation}
This formulation is useful when we are interested in classifying unseen bags rather than single instances.

A method for solving tasks with sparse bags, in which positive bags contain only few positive instances, was devised in~\cite{Bunescu2007}. Attempts to implement the predictor $ f_I $ as a neural networks were made as well, for instance in~\cite{Zhou2002}. This approach defines the aggregation $ g $ as the $ \max $ function, which is non-differentiable. Alternative method was proposed in~\cite{Ramon2000}, where $ f_I $ is implemented as a feedforward neural network (weights are shared across instances), $ g $ is a smooth maximum, and $ M $ is a real hyperparameter:
\begin{equation}
    \label{eq:ramon}
    g\left( \left\{ f_I\left( \bm{x} \right) \right\}_{\bm{x} \in b}  \right) = \frac{1}{M} \log \left( \sum_{\bm{x} \in b} \exp \left( Mf_I\left( \bm{x} \right)  \right)  \right)
\end{equation}
Other established machine learning methods besides neural networks and \gls{svm}s were adapted to deal with \gls{mil} formulation as well, ranging from boosting~\cite{Auer2004} to decision trees~\cite{Chevaleyre2001, Blockeel2005}.

\subsection{Collective assumption}%
\label{sub:collective_assumption}
Methods in this subcategory expect all instances to inherit the label of the bag they belong to, which better reflects characteristics of some real-world datasets. Hence, a training set for instance-level classifier $ f_I $ is built from instances together with labels of their corresponding bags. The most prominent aggregation function $ g $ is the $\mean$ aggregation $g\left( \left\{ f_I\left( \bm{x} \right) \right\}_{\bm{x} \in b}  \right) = \frac{1}{\lvert b \rvert } \sum_{\bm{x} \in b} f_I({\bm{x}})$, used for example in~\cite{Frank2003}, where probabilities of classes returned by classifier $ f_I $ are averaged across all instances in the bag. There is also a probabilistic perspective of the collective assumption, where a bag is understood as a probability distribution $ p(\bm{x} | b) $ over individual instances and we observe samples of various sizes from this distribution. Under the collective assumption, a bag label $ c \in \mathcal{C} $ is equal to the expected class label of its instances:
\begin{equation}
    \label{eq:collective}
    p(c | b) = \mathbb{E}_{\bm{x} \sim p(\bm{x}|b)} p(c | \bm{x}) = \int_{\mathcal{X}} p(\bm{x} | b) p (c | \bm{x}) \mathrm{d}\bm{x}
\end{equation}
To make a prediction on an unknown bag, this integral is approximated by finite number of instances it contains, which leads to $ p(c | b) \approx \frac{1}{\lvert b \rvert } \sum_{\bm{x} \in b} p(c | \bm{x}) $, which corresponds to $\mean$ aggregation function $ g $ as defined above, as long as instance predictor $ f_I $ provides class probabilities. Collective assumption is further generalized to include weights $ w_{\bm{x}} \in \mathbb{R} $ on individual instances specifying how much an instance impacts the final bag label~\cite{Foulds2010, Mangasarian2008}. This is motivated by an observation that instances in \gls{is} paradigm influence the bag label not only independently but also unequally. Weighted collective assumption bridges the gap between the standard and the collective assumptions, as if we set a constant value to $ w_{\bm{x}} $, we get the collective assumption, and if we manage to all negative instances value of zero and at least one positive instance positive value, we arrive at the standard assumption.

\begin{figure}[hbtp]
    \begin{subfigure}[t]{.33\textwidth}
        \centering
        \includegraphics[width=\linewidth]{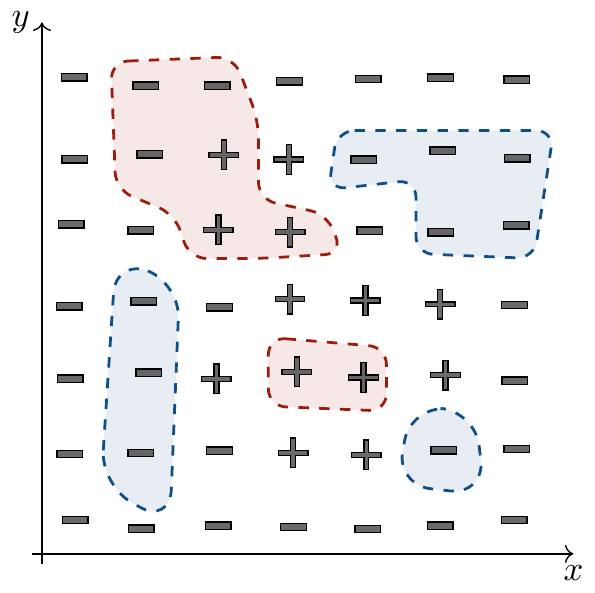}
        \caption{\label{sf:mt1}}%
    \end{subfigure}
    \begin{subfigure}[t]{.33\textwidth}
        \centering
        \includegraphics[width=\linewidth]{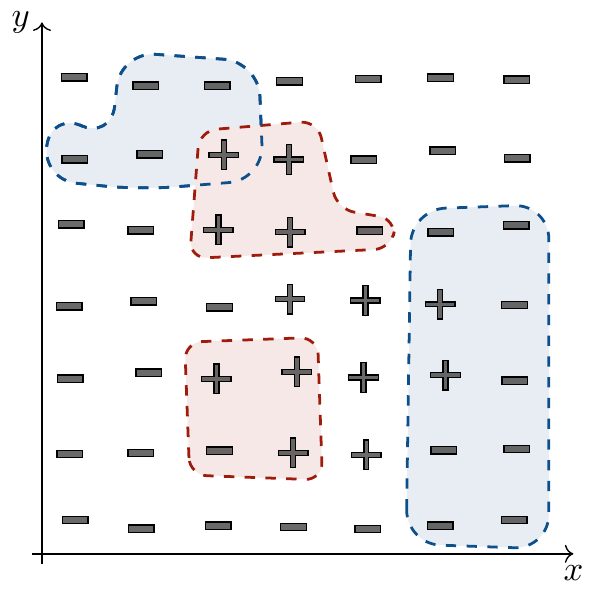}
        \caption{\label{sf:mt2}}%
    \end{subfigure}
    \begin{subfigure}[t]{.33\textwidth}
        \centering
        \includegraphics[width=\linewidth]{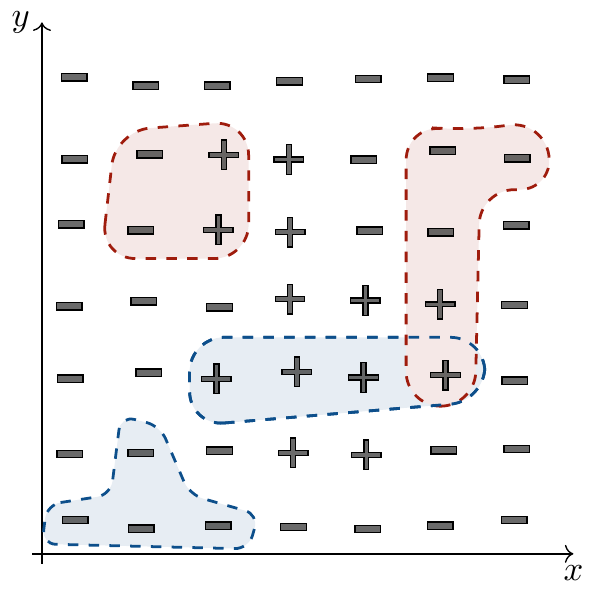}
        \caption{\label{sf:mt3}}%
    \end{subfigure}
    \caption[Examples of problems conforming to different assumptions in multi-instance learning.]{Examples of problems conforming to different \gls{mil} assumptions, all working with instance space $ \mathcal{X} = \mathbb{R}^2 $. Bags with positive labels are drawn in red and bags with negative labels in blue: $\color{cvutred} f(b) = 1$, $\color{cvutblue}f(b) = -1$. In~(\subref{sf:mt1}), the standard assumption is appropriate as all positive bags contain at least one positive instance. In~(\subref{sf:mt2}), the collective assumption holds as the majority of instances in bags retains the bag label, therefore the $ \mean $ aggregation is adequate. Finally, in~(\subref{sf:mt3}), where negative bags contain instances of only one class, none of the assumptions for \gls{is} paradigm holds and it is necessary to consider other paradigms that look on the whole bag first. Note that an instance may be present in more than one bag at once.}%
    \label{fig:mil_tasks}
\end{figure}

Methods following collective assumption are more successful when knowledge is distributed across the bag and therefore it is no longer possible to infer a bag label from one instance only. However, both the standard and collective assumptions still provide limited possibilities, as is illustrated in Figure~\ref{fig:mil_tasks}.

\section{Bag-space paradigm}%
\label{ssub:bag_space_paradigm}

In comparison to the \gls{is} paradigm, where a decision is made by aggregating responses from instance-level predictor $ f_I $ by some pre-defined aggregation rule, methods from the \gls{bs} paradigm first extract global information at the level of a bag using all its instances and then make a decision. Hence, we drop the assumption of the existence of instance labels and infer only bag-level classifier $ f\colon \mathcal{B}\to \mathcal{C} $. Since the bag space $ \mathcal{B} $ is not an Euclidean space anymore, as is usually the case with instance space $ \mathcal{X} $, the existing \gls{ml} methods are hard to extend to process elements of $ \mathcal{B} $.

Most of the existing approaches following the \gls{bs} paradigm therefore involve defining a \emph{distance function} $ d \colon \mathcal{B}\times \mathcal{B}\to \mathbb{R}^+_0$ returning a `distance' between two bags. Once $ d $ is defined, off-the-shelf methods based on distances are used, such as $k$-nearest neighbors~\cite{Dasarathy1991}. Moreover, it is assumed that instance space $ \mathcal{X} $ is equiped with a metric and therefore it is possible to measure a distance between individual instances. To simplify notation in the following paragraphs, we narrow this into an even stricter assumption on the existence of a norm $ \lVert  \cdot \rVert\colon \mathcal{X} \to \mathbb{R}^+_0 $. First example of such distance is \emph{Earth Mover's Distance}~\cite{Rubner2000}:
\begin{equation}
    \label{eq:emd}
    d(b_1, b_2)=\frac{\sum_{\bm{x}_1 \in b_1}  \sum_{\bm{x}_2 \in b_2} l_{\bm{x}_1, \bm{x}_2}\left\|\bm{x}_1-\bm{x}_2\right\|}{\sum_{\bm{x}_1 \in b_1}  \sum_{\bm{x}_2 \in b_2}  l_{\bm{x}_1, \bm{x}_2}}
\end{equation}
This distance is motivated by a well-known transportation problem solvable by linear programming. Intuitively, this distance gives an amount of work needed to transform one bag into another by `moving' instances in instance space $ \mathcal{X} $. Coefficients $ \bm{l} $ are obtained by solving the linear program corresponding to this task.

Another two distances are proposed in~\cite{Wang2000} and are based on \emph{Hausdorff distance} originally designed to measure distance between different subsets of the same metric space. The \emph{minimal Hausdorff distance} between two bags $ b_1 $ and $ b_2 $ is defined as a minimal distance between an instance from $ b_1 $ and an instance from $ b_2 $:
\begin{equation}
    \label{eq:haussdorf}
    d(b_1, b_2) = \min_{\bm{x}_1 \in b_1, \bm{x}_2 \in b_2} \lVert  \bm{x}_1 - \bm{x}_2 \rVert
\end{equation}
\emph{Maximal Hausdorff distance} on the other hand takes maximal such distance, which is equivalent to the minimal number $ D $ such that each instance from one bag falls to a hyperball with radius $ D $ centered at any instance from the other bag:
\begin{align}
    \label{eq:haussdorf_2}
    d(b_1, b_2) &= \max_{\bm{x}_1 \in b_1}\min_{\bm{x}_2 \in b_2} \lVert  \bm{x}_1 - \bm{x}_2 \rVert \\ &= \inf \bigl\{\,D \in \mathbb{R}^+_0 \bigm| \forall \bm{x}_1 \in b_1 \colon \bm{x}_1 \in \bigcup_{\bm{x}_2 \in b_2}\left\{ \bm{y} \in \mathcal{X} \mid \lVert  \bm{x}_2 - \bm{y} \rVert \le D \right\} \,\bigr\}
\end{align}
If the distance function $ d $ is also semi-positive definite, kernel-based methods like \gls{svm}s become available as well. Given any kernel function $ k\colon \mathcal{X} \times \mathcal{X} \to \mathbb{R} $ measuring the similarity between two instances, a general kernel for \gls{mil} was suggested in~\cite{Gartner2002}: 
\begin{equation}
    \label{eq:kernel}
    K(b_1, b_2) = \sum_{\bm{x}_1 \in b_1, \bm{x}_2 \in b_2} k^p(\bm{x}_1, \bm{x}_2)
\end{equation}
where $ p \in \mathbb{N} $ is a constant. Function $ K\colon \mathcal{B} \times \mathcal{B} \to \mathbb{R} $ measures the distance between two bags in the induced Hilbert space and meets all requirements for a kernel. Under the standard assumption, bag kernel $ K $ also separates all bags in the kernel-induced space provided that instances are separable with respect to kernel $ k $. In another work, authors suggested mapping instances to an undirected graph and using graph kernels afterwards~\cite{Zhou2008}.

Last but not least, motivated by the aforementioned probabilistic interpretation of \gls{mil}, where a bag is viewed as a random variable and the set of instances as a finite number of its realizations, in~\cite{Muandet2012} an \gls{svm}-based method using probabilistic kernel~\cite{Christmann2010} was proposed.

\section{Embedded-space paradigm}%
\label{ssub:embedded_space_paradigm}
Methods following the \gls{es} paradigm directly define a vector space for bag representation and specify a mapping from each bag $ b \in \mathcal{B} $ to this space. The most relevant global information at the level of bags is therefore distilled explicitly to a vector representation, as opposed to the \gls{bs} paradigm, where it is extracted implicitly through the computation of distance function. Assuming that the target vector space is $  \mathbb{R}^m $, $ m $ partial mappings $ \phi_i\colon \mathcal{B} \to \mathbb{R}\quad(i = 1, \ldots, m)$ are defined and the overall embedding $ \bm{\phi}\colon \mathcal{B} \to  \mathbb{R}^m $ can be written as:
\begin{align}
    \label{eq:es}
    \bm{\phi}(b) &= \left( \phi_1(b), \phi_2(b), \ldots, \phi_m(b) \right) \\ &= \left( \phi_1(\left\{ \bm{x} \right\}_{\bm{x} \in b}), \ldots, \phi_m(\left\{ \bm{x} \right\}_{\bm{x} \in b}) \right) 
\end{align}
In another perspective, the summarization of relevant bag-level information is a point in a vector space created as a \emph{Cartesian product} of target spaces of all partial mappings $ \phi_i $. The purpose of mappings $ \phi_i $ is to extract and suitably aggregate information from all instances, therefore they are defined as:
\begin{equation}
    \label{eq:partial_mappings}
    \phi_i(b) = g\left( \left\{ k\left(\bm{x}\right) \right\}_{\bm{x} \in b} \right) 
\end{equation}
where $ k\colon \mathcal{X} \to \mathbb{R}^m $ is some instance transformation and $ g\colon \mathcal{P}_F(\mathbb{R}^m) \to \mathbb{R}^m $ is aggregation function. Any standard \gls{ml} algorithm is then applied on the resulting embedded representation of bag samples to infer bag-level classifier $ f_B\colon   \mathbb{R}^m \to \mathcal{C} $ using dataset $ \mathcal{D}_{es} = \bigl\{ \left( \bm{\phi}(b_i), y_i\right) \in \mathbb{R}^m \times \mathcal{C} \bigm| i \in \left\{1, 2, \ldots, \lvert \mathcal{D} \rvert \right\}   \bigr\} $. Using the above definitions, the whole computational model for solving \gls{mil} problems in the \gls{es} paradigm has the concise form $ f(b) = f_B(\bm{\phi}(b))$.

Different definitions of $ \bm{\phi} $ lead to an emphasis on a different type of information. In~\cite{Dong2006, Bunescu2007}, an average of all instances from the bag $ \bm{\phi}(b) = \frac{1}{\lvert b \rvert }\sum_{\bm{x}\in b} \bm{x} $ was used and in~\cite{Gartner2002} a $\max$-$\min$ mapping was proposed:
\begin{equation}
    \label{eq:maxmin}
    \bm{\phi}(b) = \left( \min_{\bm{x} \in b} x_1, \min_{\bm{x} \in b} x_2 \ldots, \min_{\bm{x} \in b} x_d, \max_{\bm{x} \in b} x_1, \max_{\bm{x} \in b} x_2, \ldots, \max_{\bm{x} \in b} x_d \right) 
\end{equation}
Here, $ d $ denotes the dimension of instance space.

Unfortunately, these simple, hand-designed embeddings did not perform well as for example two bags may have similar average of all their instances in spite of these instances having a completely different structure. Therefore, newer methods use more complex rules that attempt to preprocess instances first, identify important patterns in them in an unsupervised way and then obtain the embedding taking these patterns into account. The embedding mapping then written in the form of $ \bm{\phi}(b, \mathcal{V})$, where $ \mathcal{V} = \left\{\theta_1, \ldots, \theta_v \right\}, \theta_i \in \Theta$ denotes a collection of abstract patterns identified in the training set, usually referred to as a \emph{vocabulary}. The mapping is constructed with respect to how well instances in a specific bag follow their patterns. Vocabulary entries can be taken into account in both $ k $ and $ g $, for instance, function $ k $ is in many methods implemented as some kind of a distance measure between an instance and the pattern. According to the definition of $ \theta_i $, how $ \bm{\phi} $ works with the vocabulary, and how $ \phi_i$ defines the aggregation $ g $, \emph{distance-based} and \emph{histogram-based} methods were categorized in~\cite{Amores2013}. All of the following methods derive vocabulary from all instances in the training set regardless of bags they come from, with the only exception being~\cite{Zhang2008}.

\subsection{Distance-based methods}%
\label{par:distance_based_methods}

Distance-based methods identify a pattern $ \theta_i $ as either a collection of instances or one representative from $ \mathcal{X} $ for all instances matching this pattern. The prominent way to obtain a vocabulary is by means of unsupervised clustering. Partial mapping  is defined in distance-based methods as $ \phi_i(b) = \min_{\bm{x} \in b} d(\bm{x}, \theta_i)$, where $ d\colon \mathcal{X}\times \Theta \to \mathbb{R} $ is a distance between instances and patterns. If the pattern is defined with a single instance, $ d $ may be a metric on $ \mathcal{X} $. Hence, the embedding dimension $ m $ is in this case equal to the number of patterns $ v $. Clustering can be performed in a `hard' way with K-means, including a special case, where each instance observed in a training set forms one cluster, or in a `soft' way with Gaussian Mixture Models. Examples of such methods include~\cite{Chen2006, Opelt2006}.

\subsection{Histogram-based methods}%
\label{par:histogram_based_methods}

Histogram-based methods measure how much a bag $ b $ matches pattern $ \theta_i $ not by taking a distance of its `closest' instance, but by quantifying how likely are instances from $ b $ following the pattern  $ \theta_i
$, in other words, $ \phi_i(b) = \frac{1}{Z} \sum_{\bm{x} \in b} l(\bm{x}, \theta_i) $, where $ l\colon \mathcal{X}\times \Theta \to [0, 1]$ is a likelihood function and $ Z $ is a normalizing constant assuring that $ \sum_{i=1}^{v} \phi_i(b) = 1$. For instance, provided each pattern in a vocabulary is represented by one instance $  \bm{x}_\theta$, $l(\bm{x}, \theta)$ can be defined as the likelihood that $ \bm{x} $ falls into a bell curve centered at $ \bm{x}_\theta $ with parameter $\sigma^2$:
\begin{equation}
    \label{eq:hb}
    l(\bm{x}, \theta) = \frac{1}{\sqrt{2\pi\sigma^2} }e^{-\frac{\lVert  \bm{x} - \bm{x}_\theta \rVert}{2\sigma^2}}
\end{equation}
Methods categorizable into the histogram-based approach for solving \gls{mil} problems are for example~\cite{Sivic2003, Foulds2007, VanGemert2010}.
\newline\newline\noindent
All approaches for \gls{mil} problems we discussed above require three components---a function operating at the level of instances $ f_I $ (be it a class predictor, an embedding or other mappings), a form of aggregation or pooling $ g $ (which is either a part of embedding projection $ \phi_i $ or a standalone building block), and a bag-level classifier $ f_B $. One of the crucial ideas used in this thesis, published independently in~\cite{Pevny2016, Edwards2016}, is to jointly optimize the whole pipeline. This is achieved by a gradient-based optimization in conjunction with techniques of differentiable programming. The only restriction this approach poses is that all three components---$ f_I, g $ and $ f_B $ ---need to be (at least piecewise) differentiable with respect to their parameters.

\begin{figure}[hbtp]
    \centering
    \includegraphics[width=0.8\linewidth]{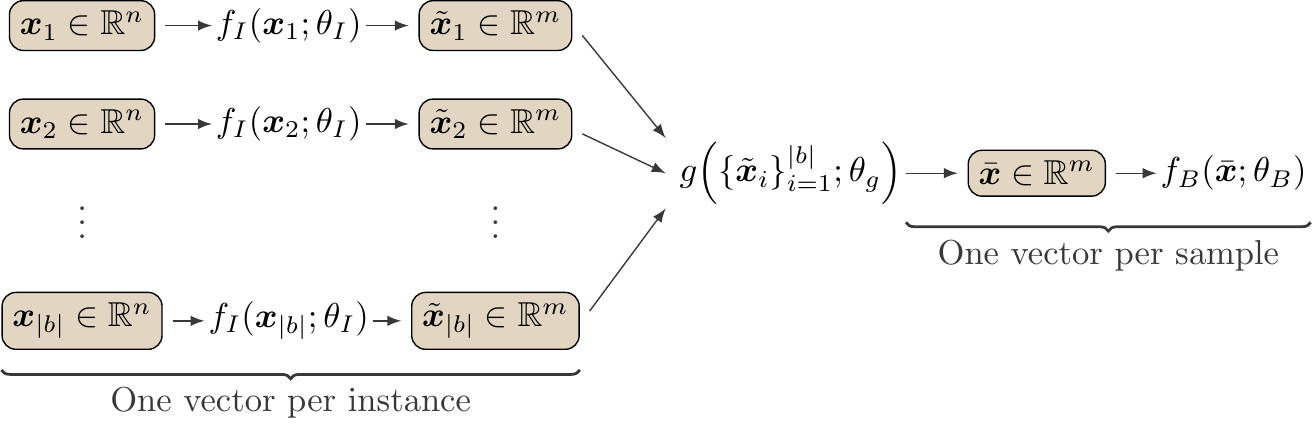}
    \caption[Neural network architecture for solving multi-instance learning tasks.]{A neural network architecture for solving \gls{mil} tasks proposed in~\cite{Pevny2016}.}%
    \label{fig:mil_pevnak}
\end{figure}

End-to-end training mitigates the need to consider how synergic all components are together when designed separately. Instead, it implicitly identifies parts of the instance-space $ \mathcal{X} $ where the difference between probability distributions of instances of bags of different classes is largest and then uses this information to discriminate. Vocabulary $ \mathcal{V} $ is no longer manually designed and precomputed explicitly in an independent process, but is rather encoded in parameters of $ f_I $ and/or $ g $ and optimized together with all other parameters in the model. All of this is done using bag labels only, in an unsupervised way with respect to instances, simply put, no instance labeling is required. Example of such model is sketched in Figure~\ref{fig:mil_pevnak}. In this case, instances are mapped through neural network $ f_I(\cdot; \theta_I) $ with parameters $ \theta_I $, aggregated with an element-wise aggregation $ g$, such as $\mean$ or $\max$, and finally processed by another network $ f_B(\cdot; \theta_B) $ to obtain prediction. Note that network $ f_I $ is the same for each instance, which means that weights $ \theta_I $ are shared between computations involving different instances and updated after the gradient is backpropagated through all instances.

\noindent
This novel model design led to improved performance on both classical and real-world datasets, as demonstrated for example in~\cite{Pevny2016, Pevny2017a, Pevny2020}. The model architecture from~\cite{Pevny2016} is illustrated in Figure~\ref{fig:mil_pevnak}.

Recently, a notion similar to \gls{mil} model formulation solved by jointly optimized embedding-space approach also appeared under a different name \emph{Deep Sets}~\cite{Zaheer2017}.
\newline\newline\noindent
To sum up, Multi-Instance learning has over the years seen many diffferent formulations, approaches and diverse applications, ranging from biochemistry to computer security. We end our long journey through the prior art with one notable conclusion: All best-performing models are composed from three major components, an instance-level function, an aggregation, and a bag-level function, which we will denote from now on by $ f_I $, $ g $ and $ f_B $, regardless of possible different use of the symbols in this chapter. Vital method for the rest of this thesis is the one introduced in~\cite{Pevny2016}.


\chapter{HMill framework}%
\label{cha:hmill_framework}

In this chapter, we formally introduce the \gls{hmill} framework and investigate its theoretical approximation capabilities. First of all, we provide further motivation for the development of the tool.

\section{Motivation}%
\begin{figure}[hbtp]
    \centering
    \includegraphics[width=\linewidth]{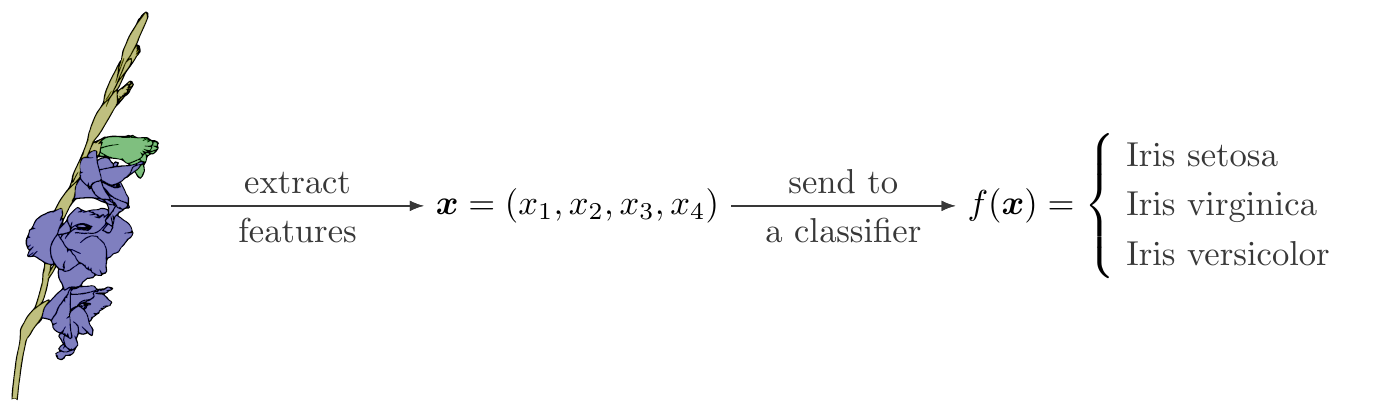}
    \caption[Standard task associated with the \emph{Iris} dataset.]{A depiction of the standard task associated with the \emph{Iris} dataset. Four features $ x_1, x_2, x_3$ and $ x_4 $ are extracted by measuring length and width of sepals and petals of the specimen. These measurements are interpreted as a vector from $ \mathbb{R}^4 $ and a classifier operating on that space is learned.}%
    \label{fig:iris}
\end{figure}

\label{sec:motivation} The most relevant reasons why the \gls{hmill} framework was devised and prototyped are already outlined in Chapter~\ref{cha:difficulties}. Nevertheless, in this section we provide one more argument from the philosophical point of view, this time using the well-known \emph{Iris} dataset~\cite{Fisher1936}. The goal of this nowadays rather `toy' problem is to predict the species of a flowering plant of the \emph{Iris} genus from the measurements of its sepals and petals. For each specimen, four measurements provided are usually interpreted as a point in $ \mathbb{R}^4 $, so that they can be input into various \gls{ml} models operating on Euclidean spaces. This is illustrated in Figure~\ref{fig:iris}\footnote{Due to unavailable assets, we used a different species in our illustrations for the demonstration purposes. The specimen depicted here is \emph{Gladiolus}, another member of the \emph{Iridicae} family. We apologize to careful readers with advanced botanical knowledge for potential confusion.}.

\noindent
Even though sepals and petals of three species of \emph{Iris} genus from the dataset do tend to have different geometrical properties and even higher-bias models such as linear \gls{svm} achieve satisfactory results, there are several fundamental flaws in this procedure. Firstly, if the specimen has more than one flower\footnote{To avoid potential confusion, here we use the term \emph{flowering plant} for one specimen, which consists of a stem, some leaves, and some flowers. In this regard, the term \emph{flower} does not imply the specimen as a whole. We will refer to flowers which have not blossomed as \emph{buds} and to the rest as \emph{blooms}.}, how should we proceed? Should we take the average of measurements over all flowers or their maximum? Or minimum? Secondly, this way of species modelling does not reflect how humans learned to recognise plants. Most of the time, a skilled botanist is able to identify a specimen not by making use of any measuring device, but by visual or tactile inspection of its stem, leaves and blooms. For different species, different parts of the plant may need to be examined for indicators. At the same time, many species may have nearly identical-looking leaves or blooms, therefore, one needs to step back, consider the whole picture, and appropriately combine lower-level observations into high-level conclusions about the given specimen. Intuitively, if we want to develop artificial intelligence capable of doing this task like humans do, we should provide it with the same input.

\begin{figure}[htbp]
    \centering
    \includegraphics[width=\linewidth]{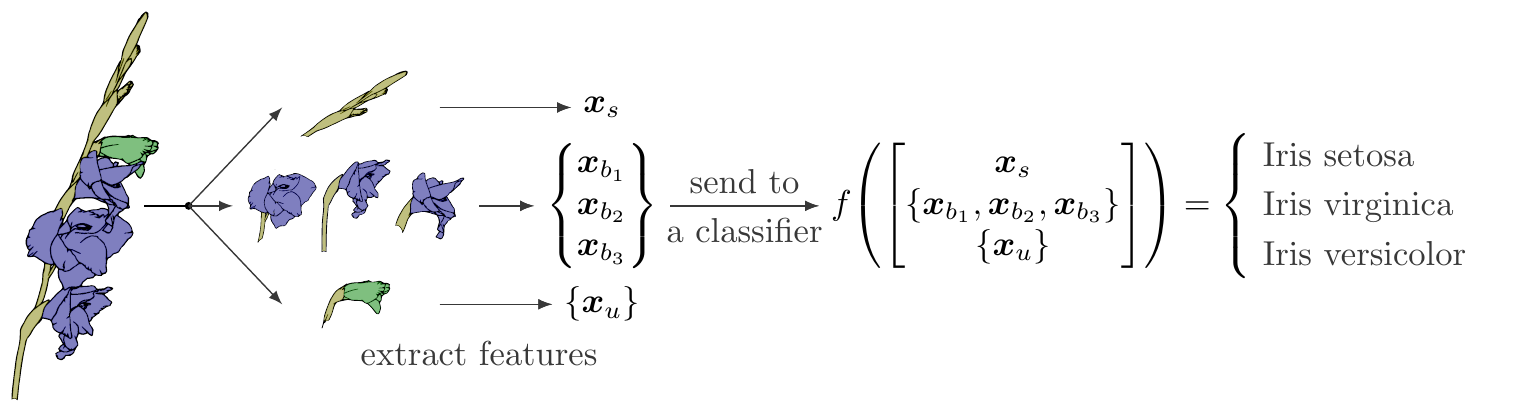}
    \caption[More general representation of a specimen.]{A more general representation of a specimen. In this case, we describe all components of the specimen in a general way, dealing with different amounts of blooms and leaves. In this picture, the specimen consists of a stem ({\color{olive}olive}), three blooms ({\color{navy}navy}) and one bud ({\color{cgreen}green}), which we describe in a hierarchical manner (see text).}%
    \label{fig:iris2}
\end{figure}

For instance, consider the depiction in Figure~\ref{fig:iris2}. We first logically split the plant into a stem, three blossomed blooms and a bud. The stem is represented by vector $ \bm{x}_s $ encoding its distinctive properties such as shape, color, structure or texture. Next, we inspect all blooms. Each of the blooms may have distinctive discriminative signs, therefore, we describe all three in vectors $ \bm{x}_{b_1}, \bm{x}_{b_2}, \bm{x}_{b_3} $, one vector for each bloom. Finally, $ \bm{x}_u $ represents the only flower which has not blossomed. Likewise, we could describe all leaves of the specimen if any were present. Here we assume that each specimen of the considered species has only one stem, but may have multiple flowers or leaves. Hence, all blooms and buds are represented as unordered sets of vectors as opposed to stem representation, which consists of only one vector. Moreover, for simplicity, we decided to stop at the level of flowers and do not decompose them further. Namely, we could as well describe a bloom as a set of vectors representing all of its petals and one vector representing the only sepal. Last but not least, due to various reasons, some collected specimens may have different amounts of flowers, or the flowers may be missing altogether. Still, human botanists are able to identify the species in many cases. This implies that our system should be able to tackle this kind of input as well.

\noindent
The representation we have presented encodes structure together with simple measurements in a hierarchical manner, which is much more flexible than the representation from Figure~\ref{fig:iris2}. However, this also introduces new challenges into the design of the classifier $ f $. The example from this section, together with discussion in Chapter~\ref{cha:difficulties} and benefits and drawbacks stemming from \gls{mil} formulation introduced in Chapter~\ref{cha:mil}, motivated the design of the \gls{hmill} framework.

\section{HMill framework definition}%
\label{sec:hmill_framework_definition}

The \gls{hmill} framework describes how to build powerful hierarchical models with all the aforementioned desirable properties. It defines both a rigorous, yet flexible high-level definition of the structure of data operable by the framework and prescribes how to build models transforming such data. We will call data structures and model topologies conforming to the following definitions \emph{\gls{hmill} samples} and \emph{\gls{hmill} models}, respectively. \gls{hmill} sample is an analogical term to \emph{sample} or \emph{observation} in the standard setting. Both samples and models consist of nodes of multiple types, which are arranged to a rooted directed tree that encodes the structure of the problem. Since in the framework, we think of individual samples as trees, we will use terms `sample', `sample tree', or `tree representing a sample' interchangeably. Furthermore, we call nodes comprising \gls{hmill} sample trees \emph{data nodes}, and nodes from \gls{hmill} model trees \emph{model nodes}. When it is clear from the context, we will sometimes omit the `data' or `model' adjectives. To provide a precise mathematical definition of \gls{hmill} models, we first need to specify how \gls{hmill} sample trees look like. 

\subsection{HMill sample}%
\label{sub:hmill_sample}

Each leaf of a sample tree stores raw low-level information, whereas inner data nodes combine information from their subtrees at a higher level of abstraction. What follows is the description of three abstract data node types with which the framework works and which constitute sample trees. After that, we give a rigorous recursive definition of \gls{hmill} sample tree, since only some of all possible trees composed of \gls{hmill} data nodes qualify as valid samples.

\subsubsection{Array Node}%
\label{sub:array_node}
\begin{wrapfigure}{R}{0.38\textwidth}
    \centering
    \includegraphics[width=\linewidth]{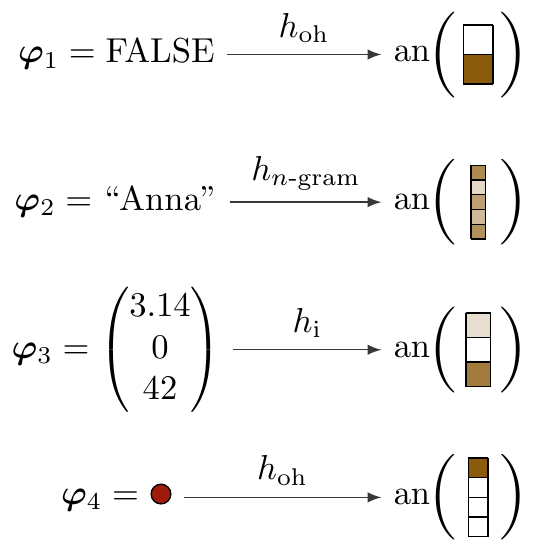}
    \caption[Examples of information stored in array data nodes.]{Examples of information fragments from several different fragment spaces $ \mathcal{F} $, together with results of mapping $ h $ in Euclidean vector space.}%
    \label{fig:mapping_h}
\end{wrapfigure}

Array nodes are responsible for storing all raw low-level observations present in the sample. Because `observation' is a term usually used to denote a whole sample, we will refer to observations made at the lowest levels of abstraction as information \emph{fragments} to avoid confusion. Following our botanical example, fragments are vectors $ \bm{x}_s, \bm{x}_{b_1},\bm{x}_{b_2},\bm{x}_{b_3}$ and $ \bm{x}_u $ representing components of the specimen at the lowest level of decomposition. In \gls{hmill} samples, array node is the only data node type present in leaves of the sample tree.

Let $ \mathcal{F} $ denote a space where all information fragments $ \bm{\varphi} $ carried by array nodes live. As indicated earlier, we aim to process data of diverse formats, therefore $ \mathcal{F} $ is rather loosely defined and may vary a lot depending on the application and also in which part of the tree the node is located. For example, $ \mathcal{F} $ may be a Euclidean space $ \mathbb{R}^m $, linear space over field $ \mathbb{Z}_2 $ consisting of all binary vectors of fixed length, or $ \Sigma^* $, the set of all strings over finite alphabet $ \Sigma $. The only requirement the framework imposes on $ \mathcal{F} $ is the existence of a reasonable representative mapping $ h $ from $ \mathcal{F} $ to a Euclidean space $ \mathbb{R}^n $ of arbitrary dimension $ n $.

We provide several details in Figure~\ref{fig:mapping_h}, although mapping $ h $ may be designed with respect to a specific problem, nature of the data, and domain. First fragment $ \bm{\varphi}_1 $ is a simple binary value ($ \mathcal{F}_1 = \left\{ \text{TRUE}, \text{FALSE} \right\} $), thus one-hot encoding $ h_{\text{oh}}\colon \left\{ \text{TRUE},  \text{FALSE} \right\} \to \mathbb{R}^2$ is used. In the second example, $ \bm{\varphi}_2 $ is a string ($ \mathcal{F}_2 = \Sigma^* $) and $ h_{n\text{-gram}} $ is implemented as a histogram of $ n $-grams. Because $ \bm{\varphi}_3 $ is a three-dimensional vector ($ \mathcal{F}_3 = \mathbb{R}^3 $), Euclidean mapping could be defined as identity $ h_i $. Finally, in the case of $ \bm{\varphi}_4 $, which is a categorical variable with one of four colours ($ \mathcal{F}_4 = \{$ \tikz{\draw[black, fill=cvutred] (0,0) circle (0.1);}, \tikz\draw[black, fill=cvut3] (0,0) circle (0.1);, \tikz\draw[black, fill=cvutblue] (0,0) circle (0.1);, \tikz\draw[black, fill=cvut2] (0,0) circle (0.1); $\}  $), we use one-hot encoding $ h_{\text{oh}} $ again. The examples of $ h $ presented here are one of the most trivial and of course more complex mappings could be used as well, for instance, word2vec~\cite{word2vec} or GloVe~\cite{glove} word embeddings in the case of strings.

The main purpose of mapping $ h $ is to appropriately represent observed low-level fragments from $ \mathcal{F} $ in Euclidean domain which \gls{hmill} models operate within. Therefore, $ h $ can be regarded as the first preprocessing step. We will denote array nodes with data fragment $ \bm{\varphi} \in \mathcal{F} $ by $ \adnode(\bm{\varphi}; \mathcal{F}, h) $. To avoid notational clutter, we will sometimes use notation $ \adnode(\bm{x}) $, where $ \bm{x} = h(\bm{\varphi}) $ is already transformed fragment, in cases where it is assumed that the transformation has already been made and its details are irrelevant or clear from the context.

\subsubsection{Bag Node}%
\label{sub:bag_node}

Bag node type is the pivotal data node of the framework and represents an analogy to a bag from multi-instance learning. Therefore, bag nodes bring the great benefit of increased flexibility in modelling arbitrarily large cardinalities in the data. Every bag node $ \bdnode(b)$ consists of any set of elements (instances) $ b = \left\{ t_1, \ldots, t_{\lvert b \rvert } \right\}  $, which may also be empty. From the graphical point of view, $ t_1, \ldots, t_{\lvert b \rvert } $ are another trees attached to the bag node as its children.\footnote{This slightly abuses terminology from graph theory. In \gls{hmill} parlance, `children' are equivalent to `subtrees'. For instance, bag node $ \bdnode(\left\{ t_1, t_2 \right\} ) $ has two children $ t_1 $ and  $ t_2 $.} If all trees $t_i$ are array nodes $\adnode(x_i)$, where $ x_i $ comes from instance space $\mathcal{X} $, the formulation is identical to the one in Chapter~\ref{cha:mil} and the sample tree corresponding to $ \bdnode(b) $ is equivalent to a bag sample in standard multi-instance learning. Therefore, a bag from \gls{mil} is in \gls{hmill} point of view considered a tree of depth $1$ consisting of one bag node with zero, one, or more array nodes (leaves) attached to it.  Even though a lot of inspiration for the framework came from the \gls{mil} domain, the \gls{hmill} framework takes this notion even further and allows elements $ t_i $ to be further nested, as long as the structure of every element $ t_i $ is the same.\footnote{What it means for two trees to have the same structure will be formalized later.}

\begin{figure}[hbtp]
    \begin{subfigure}[t]{.15\textwidth}
        \centering
        \raisebox{.95cm}{\includegraphics[width=0.8\textwidth]{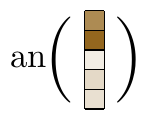}}
        \caption{\label{sf:bn1}Standard ML}%
    \end{subfigure}
    \begin{subfigure}[t]{.4\textwidth}
        \centering
        \raisebox{.225cm}{\includegraphics[width=0.9\textwidth]{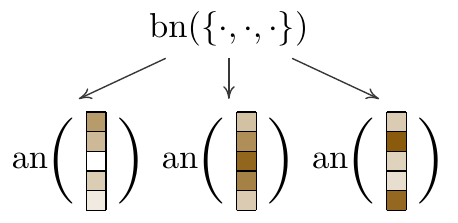}}
        \caption{\label{sf:bn2}Multi-Instance ML}%
    \end{subfigure}
    \begin{subfigure}[t]{.44\textwidth}
        \centering
        \includegraphics[width=0.8\textwidth]{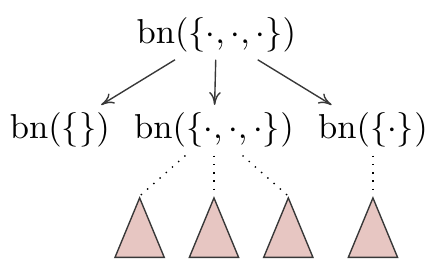}
        \caption{\label{sf:bn3}HMill sample}%
    \end{subfigure}
    \caption[Samples from distinct machine learning paradigms expressed as HMill sample trees.]{Examples of sample trees corresponding to what is considered an observation in different \gls{ml} paradigms. Single array node in~(\subref{sf:bn1}) is equivalent to machine learning in Euclidean vector domains. The second picture~(\subref{sf:bn2}) illustrates how an observation in the form of bag from \gls{mil} paradigm is translated into the framework. The central dot $ \cdot $ suggests that an entire subtree takes this place and is depicted below rather than noted down. The last illustration~(\subref{sf:bn3}) sketches an \gls{hmill} sample tree, where bags are nested. In the whole text, we use arrows $ \tikz[baseline]{\draw[modelarrow] (0,0.1) -> (0.5, 0.1);} $ to signify children relationships when children are drawn (like in~(\subref{sf:bn2})) and dotted lines $ \tikz[baseline]{\draw[ellipsearrow] (0,0.1) -> (0.5, 0.1);} $ in cases when the subtree is compressed (like in~(\subref{sf:bn3})). Arrows of the form $ \tikz[baseline]{\draw[genericarrow] (0,0.1) -> (0.5, 0.1);} $ are used in the rest of the text for all remaining purposes.}%
    \label{fig:bag_node}
\end{figure}

For example, each element $ t_i $ could be another bag consisting of lower-level instances of a different type, these instances themselves may be array nodes or may be further nested, and so forth. In Figure~\ref{fig:bag_node} on the right, there is an illustration of one possible sample tree containing bag nodes. Recall, that we require all instance subtrees of a bag node to have the same structure. Thus, every instance of the highest-level bag representing the sample is itself considered a different type of bag, which is same for all three children. Recursively, grandchildren must also have the same structure. This is depicted by red isosceles triangles, which represent subtrees of possibly different fragments in leaves, but the same structure overall. Note that only two levels of the tree are drawn in detail, however red triangles may further correspond to arbitrarily deep subtrees.

\subsubsection{Product node}%
\label{sub:product_node}
\begin{wrapfigure}{R}{0.2\textwidth}
    \centering
    \includegraphics[width=\linewidth]{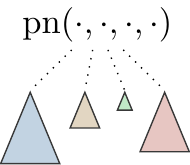}
    \caption[Drawing of a product node with four children.]{A drawing of a product node with four children.}%
    \label{fig:product_node}
\end{wrapfigure}
Product node is the last data node type in the \gls{hmill} framework. Its name is inspired by \emph{Cartesian product}, a non-commutative and non-associative mathematical operation on two or more sets. Like Cartesian products combines, but not irreversibly merges two sets, product nodes are meant to join data coming from different, possibly heterogeneous sources. Let $ t_1, \ldots, t_l $, $ l \geq 1 $ be different \gls{hmill} tree representations of the same sample. We denote product node by $ \pdnode(t_1, \ldots, t_l) $. Contrary to a bag node, whose instances form an unordered set, product nodes impose arbitrary but fixed ordering on their children. Also, unlike bag nodes, product nodes aggregate information from different sources, which may be modelled separately by (sub)trees of different structure and sizes. This is suggested by different colors and areas of triangles. See Figure~\ref{fig:product_node} for illustration. For example, in product data node $ \pdnode(t_1, t_2, t_3, t_4) $, $ t_1$ and $ t_4 $ could be array nodes, $ t_2 $ another product node, and $ t_3 $ a bag node with nested bags. 

\begin{figure}[hbtp]
    \centering
    \includegraphics[width=\linewidth]{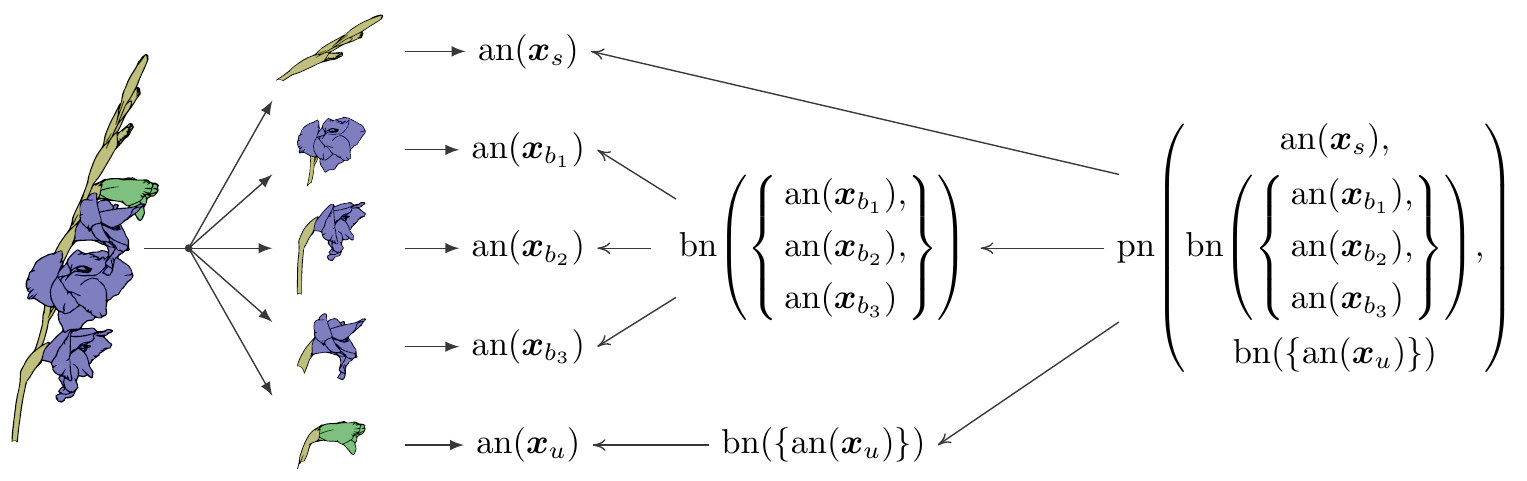}
    \caption[Representation of the plant specimen in HMill.]{Representation of the plant specimen from the beginning of the chapter in the \gls{hmill} framework. See text for details.}%
    \label{fig:iris3}
\end{figure}

\noindent
Equipped with all three data node types, we may now return to the example presented at the beginning of this chapter. See Figure~\ref{fig:iris3}. Recall, that we advocated representing a plant by means of decomposition into several logical parts. In \gls{hmill} terms, we encode all observations about the stem, the only bud and all three blooms into array nodes. After that, we relate all blooms using a bag node. The same is done for all buds, even though in this particular case, we work with the specimen with only one bud. Hence, the corresponding bag node representing all buds contains only one instance. Note that this representation would be applicable as well for cases where specimens have a different number of buds, including zero. The structure of the sample tree would not change, only cardinalities in the corresponding bag. The same applies to blooms. Having done the above, we obtain a unified representation of the stem and all buds and blooms, and finally, we merge this heterogeneous data using a product node. Because product nodes allow their children to have a different structure, array nodes modelling the stem, buds, and blooms may be defined differently (in terms of fragment space $ \mathcal{F} $ and mapping $ h $). Nevertheless, due to restriction imposed by bag nodes, we require all blooms to be modelled by array nodes with same $ \mathcal{F} $ and $ h $, and the same applies to array nodes modelling buds.

\subsubsection{Rigorous sample definition}%
\label{ssub:rigorous_definition}
For writing a definition for general \gls{hmill} sample we lack one more useful notion of \emph{HMill schemata}. Schemata are nowadays standard in many data formats, including \gls{xml}, \gls{json}, BigQuery, or Protobuf. They define how every valid file following the schema looks like in terms of its structure and data types. In the \gls{hmill} framework, we use schemata to define what it means for two (sub)trees to have the same structure. While introducing a bag node, we mentioned that all instances in the bag cannot be an arbitrary composition of data nodes, instead, they are required to have an identical structure. This is formalized by using a schema.
Since a schema is defined very similarly to \gls{hmill} samples, we omit technicalities here and resort to a vague description. Schemas in the framework are trees mirroring the structure of sample trees. For each array data node the schema defines a specific fragment space instance $ \mathcal{F}' $ where the data lives together with a mapping $ h' $ to Euclidean domain in an \emph{array schema node} $ \asnode(\mathcal{F}', h') $. For each bag node the schema defines a \emph{bag schema node} $ \bsnode(s') $ with a subschema (subtree) $ s' $ specifying a structure of instances. Finally, for each product node the schema defines a \emph{product schema node} $ \psnode(s_1, \ldots, s_l) $ with one or more (possibly different) subschemata $ s_1, \ldots, s_l $, one for each subtree of the product node. Hence, in schemata, every bag schema node has one child, every product schema node has multiple children, and every array schema node is a leaf. For the same reasons as in the case of array node, we will sometimes omit details for an array schema node and denote it by $ \asnode(\mathbb{R}^n) $, where $ \mathbb{R}^n $ is the target space of mapping $ h $.

\begin{definition}[Schema matching]\label{def:schema_matching} We say that a sample tree $ t $ follows a schema $ s $, or matches a schema $ s $, denoted by $ t \equiv s$, provided the following conditions hold:
    \vspace{-0.1cm}
    \begin{description}[labelindent=0.5cm]
    \item[(array)] If $ t = \adnode(\bm{\varphi}; \mathcal{F}, h)$ for any $ \bm{\varphi} \in \mathcal{F} $, then $ t \equiv s $ if and only if $ s = \asnode(\mathcal{F}, h)$. That is, $ t $ is an array node using same fragment space $ \mathcal{F} $ and mapping to Euclidean space $ h $ as $ s $ defines.
    \item[(bag)]  If $ t = \bdnode\left(\left\{ t_1, \ldots, t_{k} \right\}\right)$, then $ t \equiv s $ if and only if $ s = \bsnode(s') $ and  $\forall i \in \left\{ 1, \ldots, k\right\} \colon t_i \equiv s'$. In other words, all instances $ t_i $ of the bag node follow the same (sub)schema  $ s' $.
    \item[(product)] If $ t = \pdnode(t_1, \ldots, t_l) $, then $ t \equiv s $ if and only if  $ s = \psnode(s_1, \ldots, s _{l'}) $, $ l = l' $, and  $\forall i \in \left\{ 1, \ldots, l \right\}\colon t_i \equiv s_i $. That is, a matching (sub)schema $ s_i $ is defined for each child of  $ t $.
\end{description}
\end{definition}

\begin{figure}[hbtp]
    \begin{subfigure}[t]{.3\textwidth}
        \centering
        \includegraphics[width=\linewidth]{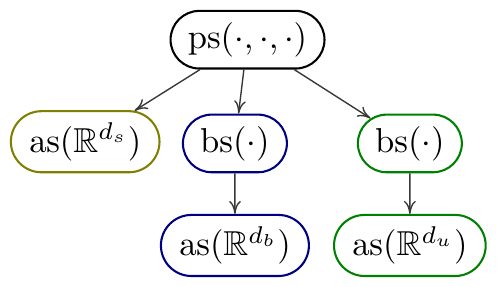}
        \caption{\label{sf:ms1}}%
    \end{subfigure}
    \begin{subfigure}[t]{.39\textwidth}
        \centering
        \includegraphics[width=\linewidth]{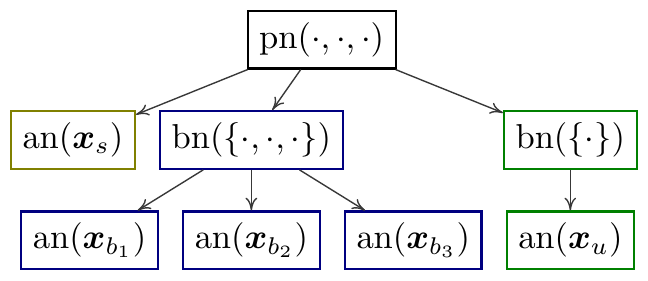}
        \caption{\label{sf:ms2}}%
    \end{subfigure}
    \begin{subfigure}[t]{.3\textwidth}
        \centering
        \includegraphics[width=\linewidth]{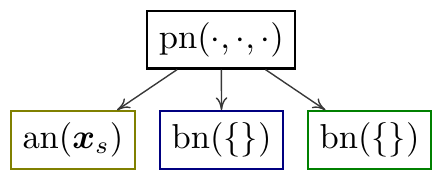}
        \caption{\label{sf:ms3}}%
    \end{subfigure}
    \caption[HMill schema corresponding to the botanical example.]{A schema corresponding to the botanical example, together with two examples of sample trees following it. The {\color{olive}olive}, {\color{navy}navy} and {\color{cgreen}green} subtrees of the schema~(\subref{sf:ms1}) encode representation for the stem, blooms and buds, respectively. Array schema nodes in leaves specify in this example only target spaces $ \mathbb{R}^{d_s} $, $ \mathbb{R}^{d_b} $, $ \mathbb{R}^{d_u} $ of mappings $ h $, which may be different for each of array schema nodes considered. In~(\subref{sf:ms2}) there is the same sample as in Figure~\ref{fig:iris3} and in~(\subref{sf:ms3}) there is a sample with all bags empty, which still follows the schema. Note that according to definition, we require $\bm{x}_s  \in \mathbb{R}^{d_s}$, $  \bm{x}_{b_1}, \bm{x}_{b_2},\bm{x}_{b_3} \in \mathbb{R}^{d_b} $, $ \bm{x}_{u}$, and $ \in \mathbb{R}^{d_u}$.}%
    \label{fig:mil_schema}
\end{figure}

\noindent
To phrase it differently, a sample follows a schema, if their tree representations are `isomorphic'.\footnote{Not isomorphic in a strict mathematical sense---we also need to consider node types and order of children of product nodes and schema nodes. Moreover, because of an arbitrary number of instances in bag nodes, sample trees may have much more nodes than the schemata they follow. In this case, bag nodes are matched with bag schema nodes $ \bsnode(s') $ by `copying' entire subschema subtree $ s' $ and trying to align it with every instance from the bag.}  Note that according to the definition above, empty bag nodes match every schema $ s $ that has bag node as root. As a result, sample trees matching a schema follow the same schema even if we remove all instances from some of its bag nodes. An empty bag can be regarded as containing zero instances of an arbitrary type, and therefore we can match any schema in its instances. For example, if we obtained a plant specimen with all blooms and buds fallen off, we represent it as $ \pdnode\left( \adnode\left( \bm{x}_s \right), \bdnode\left( \left\{  \right\}  \right) , \bdnode\left( \left\{  \right\}  \right)  \right) $ and it is still a valid sample following the same schema. A common schema for all possible specimens observed in our botanical example, together with the two specific samples following it, is sketched in Figure~\ref{fig:mil_schema}.

Not every sample tree consisting of array, bag and product data nodes is a valid sample from the \gls{hmill} point of view. For instance, a tree with a bag data node in its root and two subtrees following different schemata is not a valid sample tree since we require all instances in bag data nodes to follow the same schema. Schemata and valid samples relate to each other in a chicken-and-egg way---given a schema, we can generate a lot of valid artificial samples matching it and on the other hand, given a sufficient number of observed valid samples, we can deduce their schema (provided it exists). The same applies to their definition, as one can be defined using the other. To avoid cyclic definition, we have opted to first introduce the reader to sample trees in general, then define a schema and using a schema, define a set of all valid sample trees. We will denote a set of all valid sample trees by $ \mathcal{T} $. We consider a sample tree valid, if and only if there exists schema, that the tree matches. This can be rewritten in the following recursive definition:

\begin{definition}[Valid \gls{hmill} sample]\label{def:valid_sample} Set of all valid \gls{hmill} samples $ \mathcal{T} $ is defined recursively as follows:
    \vspace{-0.1cm}
    \begin{description}[labelindent=0.5cm]
        \item[(array node)] Let $ \mathcal{F} $ be an arbitrary fragment space, $ \bm{\varphi} \in \mathcal{F}$ its element, and $ h\colon \mathcal{F} \to \mathbb{R}^m $ any mapping. Then, for every array node $ t=\adnode(\bm{\varphi}; \mathcal{F}, h) $ it holds $ t \in \mathcal{T} $.
        \item[(empty bag node)] For every empty bag node $t= \bdnode(\left\{ \right\} ) $ it holds $ t \in \mathcal{T} $.
        \item[(bag node)] Let $t_1 \in \mathcal{T}, t_2 \in \mathcal{T}, \ldots, t_k \in \mathcal{T}$, where $k \geq 1$, be valid \gls{hmill} samples and $ s $ a schema, such that $\forall i\in \left\{ 1, 2, \ldots, k \right\} \colon  t_i \equiv s $. Then, for every bag node $ t= \bdnode\left( \left\{ t_i\right\}_{i=1}^k  \right) $ it holds $ t \in \mathcal{T} $.
        \item[(product node)] Let $t_1 \in \mathcal{T}, t_2 \in \mathcal{T}, \ldots, t_l \in \mathcal{T}$, $l \geq 1$, be valid \gls{hmill} samples. Then, for every product node $t= \pdnode\left( t_1, \ldots, t_l \right) $ it holds $ t \in \mathcal{T} $.
    \end{description}
\end{definition}

\begin{wrapfigure}[15]{L}{0.375\textwidth}
    \centering
    \includegraphics[width=\linewidth]{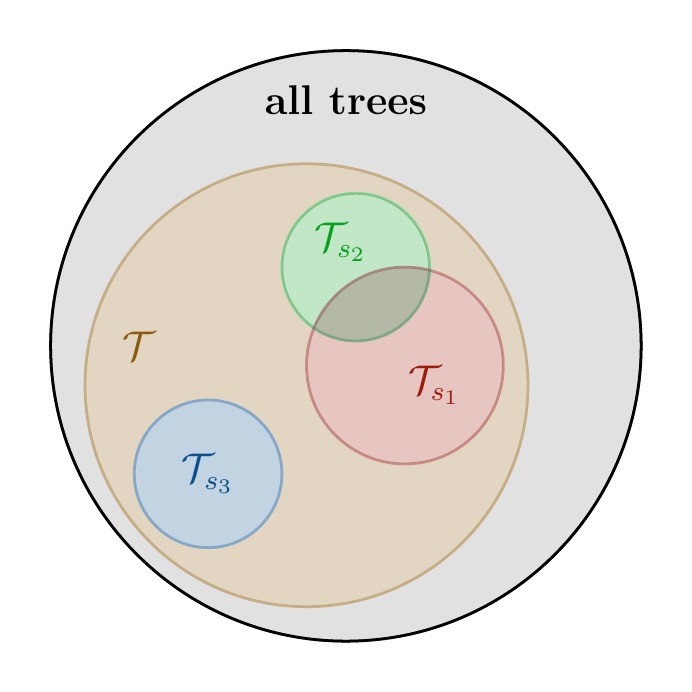}
    \caption[Relationship between $ \mathcal{T} $ and $ \mathcal{T}_s $.]{A relationship between $ \mathcal{T} $ and $ \mathcal{T}_s $.}%
    \label{fig:venn}
\end{wrapfigure}

\noindent
We have defined samples applicable in the \gls{hmill} framework to be arbitrary tree structures following certain rules that can be expressed in schemata. Although the definition above renders many trees invalid, there is still a major portion of samples left in $ \mathcal{T} $ for modelling. Given a specific schema $ s $, we will denote $ \mathcal{T}_s \subseteq \mathcal{T}$ a subset of all valid samples matching the schema. This is illustrated in Figure~\ref{fig:venn}. Valid sample trees $ \mathcal{T} $ form a proper subset of all trees formable from array, bag and product data nodes. Given schemata $ s_1 $, $ s_2 $ and $ s_3 $,  all trees matching them $ \mathcal{T}_{s_1} $, $ \mathcal{T}_{s_2} $, and $ \mathcal{T}_{s_3} $, respectively, are subsets of $ \mathcal{T} $. If schema $s$ contains at least one bag schema node, there is still an infinite number of sample trees in $ \mathcal{T}_s $ following it. Thus, modelling power remains high. The last missing puzzle piece is to describe how \gls{hmill} models transforming such data are built.

\subsection{HMill model}%
\label{sub:hmill_model}
\gls{hmill} model trees are constructed in a way that reflects the structure of corresponding samples, or more specifically, their schema $ s $. Hence, one \gls{hmill} model built by following a particular schema $ s $ accepts any sample matching the schema from $ \mathcal{T}_s $. Models are made of model nodes, which represent functions of different properties. Loosely speaking, given a sample tree, a model tree is first `aligned' with it and all nodes are evaluated. Similarly to \gls{hmill} samples, leaves of model trees are responsible for processing specific, less abstract input, contrary to inner model nodes that function with more abstract concepts derived from lower tree levels. The evaluation of the model is performed in a post-order-like fashion, in other words, parents wait for input provided by their children before computation. The root of the tree serves as the last building block that provides the model's output.

From the mathematical point of view, \gls{hmill} model nodes are hierarchically nested functions forming a tree-structured computational graph, each of which outputs one vector given one sample. However, different model nodes are allowed to produce vectors of a different dimension, even if they are of the same model node type. All nodes are piecewise differentiable with respect to their parameters, and all inner nodes in the tree are differentiable with respect to their input on top of that. Identically to the case of samples or schemata, we also distinguish three main model node types with different characteristics, which we will briefly introduce before formulating the rigorous definition.

\subsubsection{Array Model}%
\label{ssub:array_model}

\begin{wrapfigure}[10]{R}{0.4\textwidth}
    \centering
    \includegraphics[width=0.7\linewidth]{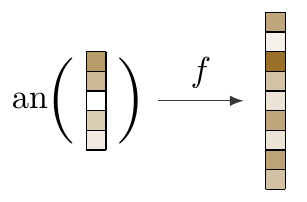}
    \caption[Array model node application.]{Application of $ \amnode(f)$, where $ f\colon \mathbb{R}^5 \to \mathbb{R}^9 $ to an array data node.}%
    \label{fig:array_model}
\end{wrapfigure}
Array model, denoted by $ \amnode(f) $, is the most basic type of model node and performs the first transformation of raw data in leaves of sample trees. We assume that mapping $ h $ on data $\bm{x} \in \mathcal{F}$ has already been used to obtain the Euclidean representation of the data and stored in a corresponding array node. Array model then applies a function $ f\colon \mathbb{R}^m \to \mathbb{R}^n $, mapping input elements to some other Euclidean space of dimension $ n $. Even though any piecewise differentiable mapping is applicable for $ f $, in this thesis, we use solely dense, feedforward neural networks for their high approximation capabilities.

\subsubsection{Bag Model}%
\label{ssub:bag_model}

In the same way as bag data nodes are inspired by data format in multi-instance learning, bag models are motivated by solutions proposed to \gls{mil} problems, discussed in the previous chapter. Therefore, each bag model can be regarded as a solver to specific \gls{mil} (sub)problem in the corresponding bag node. The function implemented by a bag model node $ \bmnode(f_I, g, f_B) $ is a composition of three functions we will call \emph{instance model}, \emph{aggregation} and \emph{bag mapping}. Given a bag data node $ \bdnode(\left\{ t_i\right\}_{i=1}^{k})  $, a bag model is evaluated by first applying instance model $ f_I $ on every subtree (instance) $ t_i $ obtaining one vector for each $ t_i $. This is followed by one or more element-wise aggregations $ g $ into a single vector and finally transformation of this vector via $ f_B $. Instance model $ f_I $ can be any valid \gls{hmill} model with an arbitrarily complex structure, accepting a valid \gls{hmill} tree and producing a vector from $ \mathbb{R}^m $.

In the case when $ f_I $ implements a transformation (or embedding) of instances living in a Euclidean space, the bag model is identical to one proposed in~\cite{Pevny2016}.\footnote{In \gls{hmill} parlance, $ f_I $ in this case corresponds to a tree with one array model $ \amnode(f_I(\cdot; \theta_I)) $, where $ f_I(\cdot; \theta_I) $ has the same meaning as in  Figure~\ref{fig:mil_pevnak}.} However, the power of \gls{hmill} models lies in the hierarchical structure achieved by nesting models. Aggregation $ g $ takes all $k$ vectors produced by  $ f_I $ and applies element-wise aggregation, such as $\mean$ or $ \max $. In this thesis, we will use a compound $ g $ that consists of one or more base element-wise aggregations concatenated together. If there are $ q $ such components, the dimensionality of output of $ g $ is $ qm$. As the last step, bag-level function $ f_B $ is employed, which is a mapping from $ \mathbb{R}^{qm} $ to $ \mathbb{R}^d $, where $ d $ is the output dimension of the whole bag model node. If a bag model node is in the root of the tree and is the last node evaluated, we set $ d $ to be the dimension of the desired output. In the case of classification, we would set  $ d = \lvert \mathcal{C} \rvert  $ and interpret outputs as logits of probabilities of classes. Otherwise, if a bag model is an inner node in the model and therefore models only a subtree of the sample, we set $ d $ appropriately, so that all distilled knowledge can be expressed in a vector from $\mathbb{R}^d$. Again, we require piecewise differentiability of all functions involved. In this work, we implemented bag mappings $ f_B $ as one or more layers of a feedforward neural network and $ g $ as a combination of four base aggregations, which we will elaborate on later.

\begin{figure}[hbtp]
    \centering
    \includegraphics[width=0.8\linewidth]{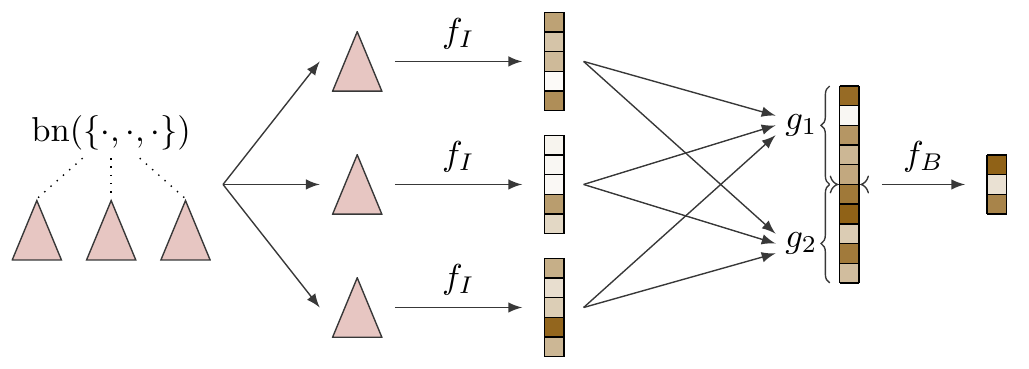}
    \caption[Bag model node application.]{Application of $ \bmnode(f_I, [g_1|g_2], f_B) $ on a bag data node sketched on the left. All instances are first transformed to a vector from $ \mathbb{R}^5 $  using instance model $ f_I $. Then, two elementwise aggregations  $ g_1 $ and  $ g_2 $ are performed, and the results are concatenated into $ 2\cdot 5 $-dimensional vector. The last step is to apply  bag mapping $ f_B $, which in this case maps the result of aggregation from  $ \mathbb{R}^{10} $ to $\mathbb{R}^3 $. If this bag model node is in the root of the tree, the output dimension corresponds to the whole model's output. For instance, this model could be used to solve a classification problem where $ \lvert \mathcal{C} \rvert =3 $.}%
    \label{fig:bag_model}
\end{figure}

\subsubsection{Product Model}%
\label{ssub:product_model}

Product model node $ \pmnode(f_1, \ldots, f_l, f) $ comprises $ l $ submodels $ f_1, f_2, \ldots, f_l $, each of which is another \gls{hmill} model tree that processes one child of a corresponding product data node, and one more mapping $ f $. Contrary to a bag model node, which uses the same mapping $ f_I $ for each of the children, here  $ f_1, \ldots, f_l $ can be mutually different submodels. By applying all of them, we obtain $ l $ vectors of possibly different lengths.  Product model concatenates them together, that is to say, it obtains an element in the Cartesian product of target spaces of $ f_1, \ldots, f_l $. This vector is transformed one more time with $ f $. The same reasoning about the output length as in the case of bag model nodes applies---we set the dimension of the target space of $ f $ according to the position of the node in the tree and provided it is in the root, according to the problem we intend to solve.

\begin{figure}[hbtp]
    \centering
    \includegraphics[width=0.8\linewidth]{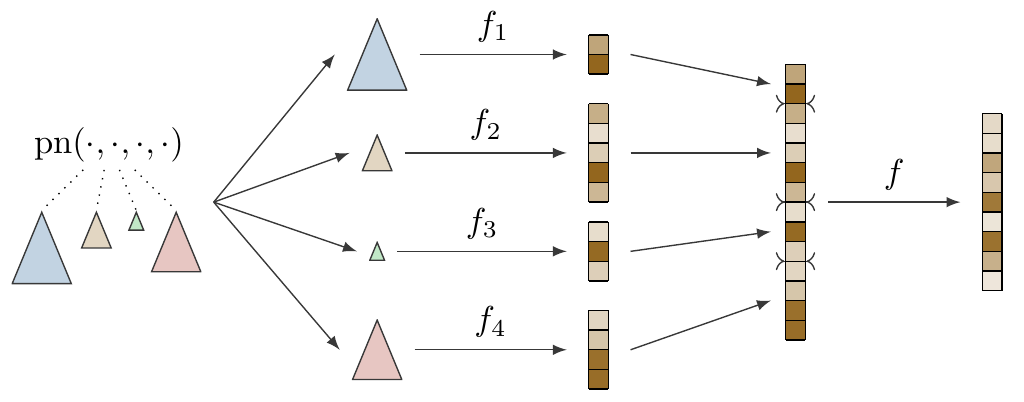}
    \caption[Product model node application.]{A depiction of how product model node $ \pmnode(f_1, f_2, f_3, f_4, f)$ is applied to a corresponding product data node picture on the left. First, mappings $ f_1, \ldots, f_4 $ are applied to children. Since product data node may merge subtrees of different structure (following different subschemata), submodels $ f_1, \ldots, f_4 $ can be defined differently and output vectors of a different dimension. In this case, we obtain vectors of dimension $ 2 $,  $ 5 $,  $ 3 $ and  $ 4 $, that are subsequently concatenated into a vector of size $ 14 $. Finally, mapping $ f $ is applied returning a vector from  $ \mathbb{R}^9 $, which may serve as input to the parent of the model node or the output of the whole model.}%
    \label{fig:product_model}
\end{figure}

\subsubsection{Rigorous model definition}%
\label{ssub:rigorous_model_definition}

Before we provide a complete definition of \gls{hmill} models, let us emphasize that even though we used same letters $ n $,  $ m $, or  $ d $ to generally denote dimensionalities of target spaces in play, this does not imply that each node of the same type maps its input to a space of same dimension (or uses spaces of the same dimensionality for intermediate representations obtained by  $ f_I $ in the case of bag model nodes or concatenation of $ f_1, \ldots, f_l $ in the case of product model nodes). To finish this section, we provide a recapitulation of how \gls{hmill} models are structured and how they are applied with a higher degree of mathematical rigor. We will use $ \mathcal{M} $ to denote a space of all models valid from the \gls{hmill} perspective.

\begin{definition}[Model]\label{def:model} A set of all valid \gls{hmill} models $ \mathcal{M} $ is defined as: 
    \vspace{-0.1cm}
    \begin{description}[labelindent=0.5cm]
        \item[(array model)] For any function $f \colon \mathbb{R}^n \to \mathbb{R}^m$, where $ n, m \geq 1 $, $ \amnode(f) \in \mathcal{M}$.
        \item[(bag model)] Let $f_I\colon \mathcal{T}_s \to \mathbb{R}^m$, where $ s $ is any schema, be an \gls{hmill} model from $\mathcal{M}$, let $g = \left[g_1 | g_2 | \ldots g_q\right]$ be a concatenation of $q$ element-wise aggregation functions, where $g_i \colon \cup_{k=1}^\infty {\left( \mathbb{R}^m \right)}^k \to \mathbb{R}^{m}$~\footnote{${\left( \mathbb{R}^m \right)}^k = \underbrace{\mathbb{R}^m \times \ldots \times \mathbb{R}^m}_k$}, and let $f_B \colon \mathbb{R}^{mq} \to \mathbb{R}^d$ be an arbitrary function. Then, $ \bmnode(f_I, g, f_B) \in \mathcal{M} $.
        \item[(product model)] Let $f_1 \colon \mathcal{T}_{s_1} \to \mathbb{R}^{m_1}, f_2 \colon \mathcal{T}_{s_2} \to \mathbb{R}^{m_2}, f_l \colon \mathcal{T}_{s_l} \to \mathbb{R}^{m_l}$ be all models from $ \mathcal{M} $, where $ s_1, \ldots, s_l $ are arbitrary schemata, and let $ f\colon \mathbb{R}^M\to \mathbb{R}^d $, where $ M = \sum_{i=1}^l m_i $, be any function. Then, $ \pmnode(f_1, f_2, \ldots, f_l, f) \in \mathcal{M} $
    \end{description}
\end{definition}

\noindent
Consider a set of samples $ \mathcal{T}_s $ following a schema $ s $. A model  $ F \in \mathcal{M} $ defines for each array schema node one array model node, for each bag schema node one bag model node and for each product schema node one product model node. The next definition recapitulates, how \gls{hmill} models are evaluated.

\newpage
\begin{definition}[Model evaluation]\label{def:model_evaluation} Given a schema $ s $, a model $ F \in \mathcal{M} $, and a sample $ t \in \mathcal{T}_s $, we evaluate $ F(t) $ as follows:
    \vspace{-0.1cm}
    \begin{description}[labelindent=0.5cm]
        \item[(arrays)] For $ F = \amnode(f) $ and $t = \adnode(\bm{x}) = \adnode(h(\bm{\varphi}))$, $ F(t) = f(\bm{x})  =f(h(\bm{\varphi}))$.
        \item[(bags)] For $ F = \bmnode(f_I, \left[ g_1|g_2|\ldots|g_q \right] , f_B)$ and $ t = \bdnode\left(\left\{ t_1, \ldots, t_{k} \right\}\right)$, $ F(t) = f_B\big(\left[ \gamma_1| \gamma_2|\ldots|\gamma_q \right] \big)$, where $ \gamma_i = g_i(\left\{ f_I(t_1), \ldots, f_I(t_k) \right\}\ \forall i \in \left\{ 1, \ldots, q \right\} $.
        \item[(products)] For $ F = \pmnode(f_1, \ldots, f_l, f)$ and $ t = \pdnode\left(\left\{ t_1, \ldots, t_{l} \right\}\right)$, $ F(t) = f\big( \left[ f_1(t_1)| \ldots| f_l(t_l) \right]  \big)  $.
    \end{description}

\end{definition} 

\begin{figure}[H]
    \centering
    \includegraphics[width=0.6\linewidth]{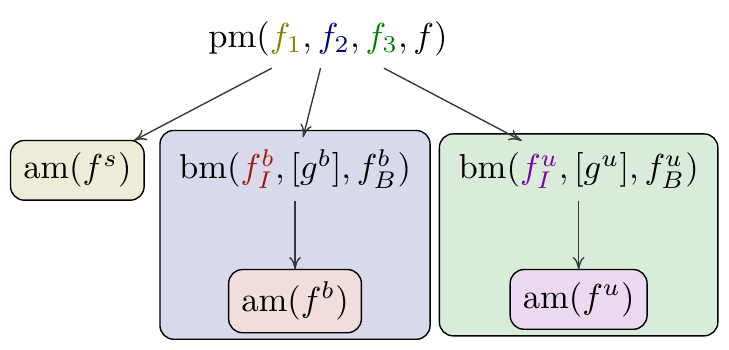}
    \caption[HMill model for the botanical example.]{An \gls{hmill} model for processing samples following schema in Figure~\ref{fig:mil_schema}. Note how schema and model trees are of the same structure. A product model node merges information from three different sources---the stem ({\color{olive}olive}), blooms ({\color{navy}navy}), and buds ({\color{cgreen}green}). Since all specimens have one stem and we consider stems further non-decomposable, simple array model is enough. On the other hand, we employ bag model nodes for processing of blooms and buds, each of which having one array model for processing of instances.}%
    \label{fig:iris_model}
\end{figure}

\begin{figure}[htbp]
    \centering
    \includegraphics[width=\linewidth]{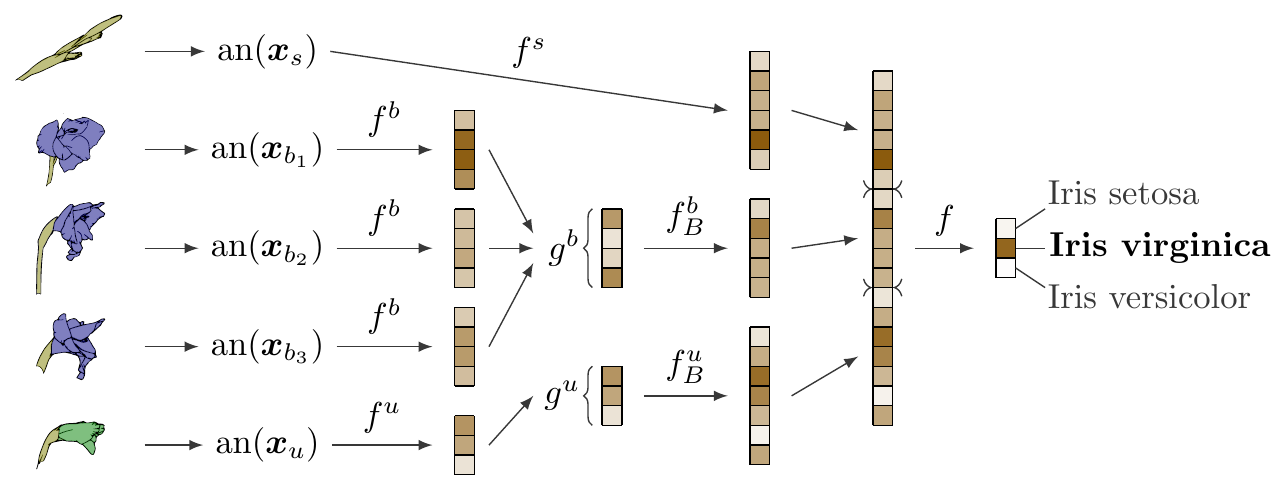}
    \caption[Application of the HMill model for the botanical example.]{Example from Figure~\ref{fig:iris3} processed by a model depicted in~\ref{fig:iris_model}. The stem is processed with array model node $ \amnode(f^s) $ into its representation in $ \mathbb{R}^6 $. All blooms are processed with bag mapping $ \bmnode(f_I^b, [g^b], f_B^b) $, where instance model $ f_I^b $ in this case consists of only one array model node $ \amnode(f^b) $. Hence, we can write down the array model alternatively as $ \bmnode(\amnode(f^b), [g^b], f_B^b) $. All array node representations of blooms are transformed with  $ f_I^b $ first, that is, we apply  $ f^b $ on their vector representation. Then, aggregation consisting this time of only one function  $ g^b $ is performed, and the resulting vector of size $ 4 $ is mapped to  $ \mathbb{R}^5 $ by $ f_B^b $. We apply analogically bag mapping $ \bmnode(f_I^u, [g^u], f_B^u) = \bmnode(\amnode(f^u), [g^u], f_B^u)$ to all buds of the specimen, obtaining a vector of dimension $ 7 $. Finally, we concatenate all partial results together and the resulting vector from $ \mathbb{R}^{18} $ is mapped with $f $ to  $ \mathbb{R}^3 $. This output is interpreted as the logarithm of odds of three classes from the problem and probabilities can be obtained by applying $ \softmax $ function.}%
    \label{fig:iris_eval}
\end{figure}

\noindent
We return one last time to the example from the beginning of this chapter. In Figure~\ref{fig:iris_model}, there is a model tree for processing all samples following the schema from Figure~\ref{fig:mil_schema}. All \gls{hmill} models corresponding to the schema from Figure~\ref{fig:mil_schema} will be structured identically, however, output dimensions of inner mappings can be chosen arbitrarily. Figure~\ref{fig:iris_eval} illustrates, how model evaluation is done given one sample and one specific instance of a model.

\noindent
In this section, we have described the key components of the \gls{hmill} framework and how they are used. We have defined elementary building blocks for observed samples, their schemata and models transforming them in the form of array, bag and product data/schema/model nodes. The short and very general pipeline suggest itself---collect the data, infer a schema from the data or specify it manually, define a model with the same structure and schema, and finally train it. Due to the versatility of an \gls{hmill} sample, we are able to process data of complex hierarchical structure from diverse real-world sources with little to no further preprocessing, which also means disposing of a procedure for transforming structured data to fixed-size vectors, required in AutoML paradigms.  

\clearpage

\section{Universal approximation theorem}%
\label{sec:theoretical_properties}

The \gls{hmill} models as defined previously in this chapter represent a new family of functions from $ \mathcal{T}_s $ to $ \mathbb{R}^d $ for some schema $ s  $ and $ d \in \mathbb{N} $. In the last section of this chapter we attempt to shed light on approximation power of the family and thus justify its practical use. We first formulate the \emph{universal approximation theorem} for standard feedforward neural networks and then demonstrate how it can be extended from real-valued functions on Euclidean spaces to real-valued functions on compact sets of Borel probability measures as well as their Cartesian products, which corresponds to input spaces \gls{hmill} models are defined on. The following text is mainly based on results from~\hbox{\cite{Hornik1991ApproximationCO, Leshno1991MultilayerFN, Pevny2019}}. We will also focus on approximating functions with values from real numbers $ \mathbb{R}  $  and not multi-dimensional mappings, as the results can be trivially extended to them.

\subsection{Universal approximation theorem for dense neural networks}%
\label{sub:universal_approximation_theorem_for_dense_neural_networks}

From their inception, standard neural networks have been regarded as `black-box' function approximators, which performed remarkably well, but for a long time there was little to none theoretical evidence why. Two principal questions suggest themselves:
\begin{enumerate}
    \item Given a function $ f\colon \mathbb{R}^n \to \mathbb{R} $ and the space of all possible different network topologies and parametrizations, how to find the best architecture and its parameters?
    \item Assuming that we know the answer to this, how accurate the approximation can be and how many functions can be approximated this way?
\end{enumerate}
Universal approximation theorem gives an answer to the latter question and states that with certain assumptions any continuous function on compact subset of $ \mathbb{R}^n $ can be approximated to an arbitrary precision. There are several different formulation of the theorem---one of the earliest results in that matter is from~\cite{Cybenko1989ApproximationBS}, where authors prove the theorem for a class of single-layer networks with finite number of neurons and sigmoid activation function. Later, the same statement was proven to hold for any non-polynomial activation function~\cite{Leshno1991MultilayerFN}. The third different formulation comes from~\cite{Hornik1991ApproximationCO} and can be rephrased as follows:
\begin{theorem}%
    \label{th:uat1}
    Let $ C(\mathbb{R}^n)$ be a space of all real-valued continuous functions on $ \mathbb{R}^n , n \in \mathbb{N} $. Then, for any $ f \in C(\mathbb{R}^n) $ and any $ \varepsilon \in \mathbb{R}^+ $, there exists a function $ \widehat{f} \colon \mathbb{R}^n \to \mathbb{R} $ of the form:
\begin{equation}%
    \label{eq:uat_1}
    \widehat{f}(\bm{x}) = \sum_{i=1}^{d} \alpha_i \sigma(b_i +\sum_{j=1}^{n}w_{i,j}x_j)
\end{equation}
where $\alpha_i, b_i, w_{i,j} \in \mathbb{R}$ and $ \sigma \colon \mathbb{R} \to \mathbb{R} $ is a non-polynomial continuous activation function, such that 
\begin{equation}
    \sup_{\bm{x} \in \mathbb{R}^n} \left\lvert f(\bm{x}) - \widehat{f}(\bm{x}) \right\rvert < \varepsilon
\end{equation}
\end{theorem}
\noindent
In other words, the set of all functions of the form~\eqref{eq:uat_1} is dense in $ C(\mathbb{R}^n) $. The same holds for any compact subset of $ \mathbb{R}^n $ as well. This version of the theorem implies that a multi-layer architecture is important for approximation of any continuous function, even though only one non-linearity is present. Because solution to a lot of problems can be expressed as a function, the key insight of the universal approximation theorem is that a neural network can in theory be used to solve any of these problems.

\subsection{Universal approximation theorem for HMill models}%
\label{sub:universal_approximation_theorem_for_hmils}

Now we are ready to formulate analogy of the universal approximation theorem for models in the \gls{hmill} framework. We begin by formulating a general analogy of Theorem~\ref{th:uat1} for the embedded-space paradigm (Section~\ref{ssub:embedded_space_paradigm}) approach to solving \gls{mil} problems. In \gls{hmill}, this corresponds to sample trees of depth one, as in Figure~\ref{sf:bn2}, which are processed with bag model nodes. Recall that a bag with instances from space $ \mathcal{X} $ is considered a random variable distributed according to some probability measure $ p $ on $ \mathcal{X} $, which is observed through a finite number of its realizations (instances). As a result, we attempt to model function $ f $ not from domain $ \mathcal{X}$ (e.g. $ \mathbb{R}^n $ as in the previous paragraph), but rather from probability measures on $\mathcal{X} $, which involve integration over $ \mathcal{X} $ (see~\eqref{eq:collective} for one possible specific example of such $ f $). The following statements specify how neural networks for sufficient approximation may look like:
\begin{statement}\label{st:mil_bag_models}
    Let $ \mathcal{X} $ be a compact subset of $ \mathbb{R}^n $, $ \mathcal{P}_{X} $ a compact set of Borel probability measures on  $ \mathcal{X} $ and $ C(\mathcal{P}_{\mathcal{X}})$ a space of all real-valued continuous functions on $ \mathcal{P}_{\mathcal{X}} $. Any function $ f(p) \in C(\mathcal{P}_{\mathcal{X}}) $ can be approximated to an arbitrary precision by a bag model node $\widehat{f}(p) = \bmnode(f_I, g(p), f_B) $, where $ f_I \colon \mathcal{X} \to \mathbb{R}^m $ is a non-linear neural network layer, $ f_B \colon \mathbb{R}^m \to \mathbb{R} $ is a stack of one non-linear and one linear layer, and  $ g(p) $ is an aggregation function of the form
    \begin{equation}\label{eq:g_int}
        g(p) = \int_{\mathcal{X}} f_I(x) \mathrm{d}p(x)
    \end{equation}
where integral operator $ \int $ is applied element-wise. That is, for any $ \varepsilon \in \mathbb{R}^+ $, there exists a `bag-model' function $ \widehat{f} $, such that
    \begin{equation}
        \sup_{p \in \mathcal{P}_{\mathcal{X}}} \left\lvert f(p) - \widehat{f}(p) \right\rvert < \varepsilon
    \end{equation}
\end{statement}
\newpage\noindent
Therefore, embedding-space paradigm for solving \gls{mil} problems is theoretically justified. This result is a corollary of the more general theorem for any metric space $ \mathcal{X} $ proven in~\cite{Pevny2019}:
\begin{theorem}\label{th:bag_models}
    Let $ \mathcal{X} $ be a metric space and $ \mathcal{P}_{\mathcal{X}} $ be a compact set of Borel probability measures on $ \mathcal{X} $ and $ C(\mathcal{X})$ and $ C(\mathcal{P}_{\mathcal{X}}) $ spaces of all continuous real-valued functions on $ \mathcal{X} $ and $ \mathcal{P}_{\mathcal{X}} $, respectively. Finally, let $ \mathcal{M} $ be a dense subset of $ C(\mathcal{X}) $. Then, a set of functions $ f \colon \mathcal{P}_{\mathcal{X}} \to \mathbb{R} $ of the form
    \begin{equation}\label{eq:bag_models_form}
        f(p) = \sum_{i=1}^d \alpha_i \sigma\left( b_i + \sum_{j=1}^{m_i} w_{i,j} \int_{\mathcal{X}} f_{i,j}(x)\mathrm{d}p(x) \right) 
    \end{equation}
    where $ d, m_i \in \mathbb{N} $, $\alpha_i, b_i, w_{i,j} \in \mathbb{R}$, $ f_{i,j} \in \mathcal{M}$, and $ \sigma $ is a measurable non-polynomial function, is dense in $ C(\mathcal{P}_{\mathcal{X}}) $.
\end{theorem}
\noindent
From Theorem~\ref{th:uat1} we know that if we set $ f_{i,j} $ to be an output of one neuron of a network of the same form as~\eqref{eq:uat_1}, therefore mapping input from $ \mathbb{R}^n $ to $ \mathbb{R} $, the set of all such functions is dense in $ C(\mathbb{R}^n) $. Furthermore, if any common continuous activation function, such as sigmoidal function $ \sigma(x) = 1 / (1 + \exp(-x)) $ or  $ \tanh $, is used in  $ f_I $ and  $ f_B $, and both networks have the corresponding topology, the Statement~\ref{st:mil_bag_models} immediately follows. Note that in this thesis we focus on cases where $ \mathcal{X} $ is a subset of $ \mathbb{R}^n $, however Theorem~\ref{th:bag_models} is more general and applies for any space $ \mathcal{X} $ endowed with a metric and any set of probability measures $ \mathcal{P}_{\mathcal{X}} $ as long as they form a set compact on $ \mathcal{X} $.

In practice, the integral in~\eqref{eq:g_int} cannot be evaluated as only a finite number of observations in bag $ \left\{ x_i \right\}_{i=1}^k $ is available. However, the aggregation can be estimated as
\begin{equation}\label{eq:g_est}
    \frac{1}{k}\sum_{i=1}^k f_I(x_i)
\end{equation}
Because each sample $ \left\{ x_i \right\}_{i=1}^k  $ can be regarded as a mixture of Dirac measures, the theorem is applicable. Furthermore, as all considered functions are continuous with compact support and therefore bounded, the estimation error between aggregation~\eqref{eq:g_int} and its estimate~\eqref{eq:g_est} can be upper-bounded in each dimension by Hoeffding's inequality~\cite{hoeff}.

To conclude, bag models as defined in \gls{hmill} can approximate any function on a compact set of measures on instance space $ \mathbb{R}^n $ using integration (mean) as aggregation. We further elaborate on the choice of aggregation function in the next chapter. 
\newline\newline\noindent
We have extended the universal approximation theorem for spaces of (Borel) probability measures, the next step is to extend the theorem even further to functions on (finite) Cartesian products of compact metric spaces. This justifies the definition of product model nodes in~\eqref{def:model} and~\eqref{def:model_evaluation} as an appropriate choice for modelling product data nodes representing the Cartesian products of multiple input spaces. Recall that the input spaces do not have to be equal, for instance, one product node can model the Cartesian product of a Euclidean space (array data node) and a space of Borel probability measures defined on another Euclidean space (bag data node with array data node children).
\begin{statement}\label{st:mil_product_models}
    Let $ \mathcal{T} $ be a set containing all compact subsets of $ \mathbb{R}^d $ for any $ d \in \mathbb{N} $, which is closed under finite Cartesian products and for each $ \mathcal{X} \in \mathcal{T} $ it holds $ \mathcal{P}(\mathcal{X}) \in \mathcal{T} $, where $ \mathcal{P}(\mathcal{X}) $ is a space of all Borel probability measures on $ \mathcal{X} $. Then, for every $ \mathcal{X} \in \mathcal{T} $, a set of functions represented by \gls{hmill} models $ \mathcal{M} $ is dense in space $C(\mathcal{X})$.
\end{statement}
\newpage\noindent
The statement is a direct consequence of recursive application of Theorems~\ref{th:uat1} and~\ref{th:bag_models} as well as the following theorem from~\cite{Pevny2019}:
\begin{theorem}\label{th:product_models}
    Let $ \mathcal{X}_1, \ldots, \mathcal{X}_l $ be compact metric spaces and $ \mathcal{M}_1, \ldots, \mathcal{M}_l $ be sets of real-valued continuous functions, such that $ \mathcal{M}_i $ is dense in $ C(\mathcal{X}_i) $ for each $ i = 1 ,\ldots,l $. Then, a set of functions $ f \colon \mathcal{X}_1 \times \ldots \times  \mathcal{X}_l \to \mathbb{R} $ of the form
    \begin{equation}\label{eq:product_models_form}
        f(x_1, \ldots, x_l) = \sum_{i=1}^d \alpha_i \sigma\left( b_i + \sum_{j=1}^{l} w_{i,j} f_{i,j}(x_i) \right) 
    \end{equation}
    where $ d \in \mathbb{N} $, $\alpha_i, b_i, w_{i,j} \in \mathbb{R}$, $ f_{i,j} \in \mathcal{M}_i $ and $ \sigma $ is a measurable function, which is not an algebraic polynomial, is dense in $ C(\mathcal{X}_1 \times \ldots \times  \mathcal{X}_l) $.
\end{theorem}
\noindent
Hence, under the assumption that bags represent probability distributions, the universal approximation theorem holds for hierarchically defined input spaces (such as $ \mathcal{T} $ from \gls{hmill}) provided the suitable activation functions and topology of neural network is used. The bound for error introduced by using finite number of instances in bags can be again obtained using Hoeffding's inequality.
\newline\newline\noindent
All results presented here imply, that the design of the key components in the \gls{hmill} framework is well-founded. The rest of this thesis is dedicated to demonstrating our claims using three completely different tasks and further discussion of the properties of \gls{hmill} models.


\ifprint\newpage\blankpage\fi%
\chapter{HMill framework technicalities}%
\label{cha:hmill_extensions}

In the previous chapter, we provided general definitions and theoretical properties of the \gls{hmill} framework, whereas this section describes specific details. We will discuss rather technical issues concerning the implementation of data representation, evaluation and gradient computation. To simplify experiments, we implemented\footnote{The public implementation is a joint effort of several people. At the time of writing this thesis, the author was responsible for approximately 65\% of all changes in the git repository.} the proposed functionality in \emph{Julia} programming language~\cite{Bezanson2014} using the highly efficient automatic differentiation over a \gls{ssa} representation of programs~\cite{Innes2018, Innes2018a, Innes2018b}. The framework was published as an open-source project \texttt{Mill.jl}~\cite{Mill2018} under MIT license and we encourage everyone to try examples presented there and apply \gls{hmill} to their own data. Version \texttt{1.2.0} of the framework is also attached to this thesis.

\section{Weighted bag nodes}%
\label{sub:weighted_bag_nodes}
The purpose of bag data nodes is to capture higher-level knowledge scattered in their instances. Motivated by the weighted collective assumption described in~\ref{sub:collective_assumption}, we introduce the concept of \emph{weighted bag data nodes}, an extension of standard bag data nodes. Weighted bag data nodes allow instances in the bag to contribute to output to a different extent. This can be specified by assigning a real weight $ w_i \in \mathbb{R}_{0}^{+} $ to each instance, which is taken into account during aggregation. We denote weighted bag nodes by $ \bdnode_w(b_w) $, where  $ b_w = \left\{ (t_1, w_1), \ldots, (t_{\lvert b \rvert }, w_{\lvert b \rvert }) \right\}  $. Since this extension only adds one data element to each bag node and modifies how aggregation $ g $ in bag model nodes is computed, we could straightforwardly extend Definitions~\ref{def:schema_matching},~\ref{def:valid_sample},~\ref{def:model} and~\ref{def:model_evaluation} to fit weighted bag node into the framework. For brevity, we omit this and instead of rewriting all four definitions again, we elaborate on aggregation functions in the next section, where we also illustrate how to take weights into account. As we show later, a need for weights sometimes emerges naturally in many diverse problems.

One typical example is domains in which we observe bags with large cardinality. Recall that instance $t$ (in \gls{hmill} world a subtree) can be interpreted as a realization of the random variable represented by bag $ b $. To compute statistics over the whole bag (for instance its class probabilities, see~\eqref{eq:collective}), we resort to the following estimate:\footnote{Here we expect that all assumptions from theorems in Section~\ref{sub:universal_approximation_theorem_for_hmils} hold and we have all the tools needed to integrate over $ \mathcal{T} $. }
\begin{equation}
    \label{eq:est_bag}
    \mathbb{E}_{t \sim p(t|b)} f(t) = \int_{\mathcal{T}} p(t | b) f(t) \mathrm{d}t \approx \frac{1}{k}\sum_{i=1}^k f(t_i)
\end{equation}
where $ k $ is a size of observed bag $ b $.  If the unknown probability distribution  $ p(t| b) $ of the random variable has large support, a lot of samples of $ t $ are needed for accurate computation. If instances require a lot of resources to obtain or process, the evaluation of sum~\eqref{eq:est_bag} using all instances in the bag is still too expensive. In this case, \emph{importance sampling} can be utilized. Importance sampling approximates the expected value by sampling instances using different distribution $ r(t | b) $ and reweights their contribution. Equation~\eqref{eq:collective} can be expanded into:
\begin{equation}
    \label{eq:importance}
    \mathbb{E}_{t \sim p(t|b)} f(t) = \int_{\mathcal{T}} p(t | b) f(t) \mathrm{d}t = \int_{\mathcal{T}} r(t | b) \frac{p(t | b)}{r(t | b)} f(t) \mathrm{d}t = \mathbb{E}_{t \sim  r(t|b)} \left[\frac{p(t | b)}{r(t | b)} f(t) \right] \approx \frac{1}{n}\sum_{i=1}^{n}  \frac{p(t_i | b)}{r(t_i | b)} f(t_i)
\end{equation}
We can define $ r(t | b) $ ourselves and estimate $ p(t | b) $ from the data. Then, it is natural to represent this bag as a weighted bag node with weights $ w_i = p(t_i | b)/r(t_i | b) $ that are taken into account in bag model nodes during computation. One of the examples, when importance sampling helps is when bags contain two types of instances, one of which tends to be much more represented than the other one. Let us for simplicity call these prevalent instances negative and rare instances positive. To accurately represent the bag, we need to sample its instances in a way which follows the true distribution $ p(t|b) $ skewed in favor of negative instances, and at the same time, positive instances are not underrepresented. In this case, naïve uniform sampling leads to only a small number of positive instances in the sampled set. With importance sampling, we can design $ r(t | b) $ to be higher for positive instances. Therefore, the sampled set contains examples of both categories, and larger weights $ w_i $ for negative sampled instances correct the biased sampling procedure. More specific examples are provided later.

\section{Aggregation functions}%
\label{sub:aggregation_functions}
\gls{hmill} gives the user a relatively free hand in defining model components. Recall that we have implemented all transformations $ f $ in array and product model nodes as one or more feedforward neural network layers. In the case of bag model nodes, bag mappings $ f_B $ were implemented by dense neural networks as well, however, the form of aggregation $ g $ is yet to be specified. As shown in the last chapter, simple $ \mean $ aggregation is sufficient in theory. Nevertheless, in our experience, models using more types of aggregations than only $ \mean $ perform better in practice. Aggregation in \gls{hmill} bag model node takes form of a concatenation $g = \left[g_1 | g_2 | \ldots g_q\right]$ of $q$ base aggregation functions $g_i \colon \cup_{k=1}^\infty {\left( \mathbb{R}^m \right)}^k \to \mathbb{R}^{m}$. Different choices of base aggregation $ g_i $, or their combinations, are suitable for different problems. Nevertheless, because the input is interpreted as an unordered bag of instances, every aggregation function should be invariant to permutation of its parameters and also should not scale with increasing size of the bag.

In this section, several options for $ g_i $ are proposed. All base aggregations proposed in this thesis are element-wise operations---each element in the resulting vector is computed using only entries at the same position in input. Given a set of vectors $\{\bm{x}_1, \ldots, \bm{x}_k\}$, $ g_i $ would first select from all vectors their element at the first index and compute one number from the resulting set of real numbers $ \left\{ x_1, \ldots, x_k \right\}  $, then all elements at the second index, and so forth. This way, we obtain $ m $ numbers, or in other words, a vector from $ \mathbb{R}^m $, a space of the same dimension as is the size of vectors on input. For simplicity, all the following formulas below represent computation of only one output dimension of each base aggregation. We repeat the procedure $ q $ times for all base aggregations, which leads to the final result in  $ \mathbb{R}^{qm} $.

\subsection{Non-parametric aggregations}%
\label{ssub:nonparam_aggregation}
The most simple choices for base aggregation $g_i$ are $ \max $:
\begin{equation}
    \label{eq:max}
    g_{\max}(\{x_1, \ldots, x_k\}) = \max_{i = 1, \ldots, k} x_i
\end{equation}
and $ \mean $:
\begin{equation}
    \label{eq:mean}
    g_{\mean}(\{x_1, \ldots, x_k\}) = \frac{1}{k} \sum_{i = 1}^{k} x_i
\end{equation}
The motivation for either of this form of base aggregation function comes from the standard assumption and the collective assumption introduced in Sections~\ref{sub:standard_assumption} and~\ref{sub:collective_assumption}, respectively. However, let us emphasize that in \gls{hmill} instances may be represented not only by vectors in real vector spaces, but by entire sample (sub)trees.

The $ \max $ aggregation is suitable for cases when one instance in the bag may give evidence strong enough to predict the label. This is oftentimes the case in the computer security domain, where signals indicating the maliciousness are rather rare, yet usually strongly incriminating. On the other side of the spectrum lies the $\mean$ aggregation function, which detects well trends identifiable globally over the whole bag.

\subsection{Parametric aggregations}%
\label{ssub:parametric_aggregations}

Whereas non-parametric aggregations do not use any parameter, parametric aggregations represent an entire class of functions parametrized by a real vector of parameters, which is learned during training. In this thesis, we use two parametric aggregations---$\lse$ (log-sum-exp)~\cite{Kraus2015} and $\pnorm$~\cite{Gulcehre2014}.

Log-sum-exp aggregation is parametrized by a vector of positive numbers $ \bm{r} \in {(\mathbb{R}^+)}^m $ that specifies one real parameter for computation in each output dimension and is computed as follows:
\begin{equation}
    \label{eq:lse}
    g_{\lse}(\{x_1, \ldots, x_k\}; r) = \frac{1}{r}\log \left(\frac{1}{k} \sum_{i = 1}^{k} \exp({r\cdot x_i})\right)
\end{equation}
Here, $ r $ denotes an element of  $ \bm{r}$, corresponding to the dimension, where the result is being computed. With different values of $ r $, log-sum-exp behaves differently and in fact both $ \max $~\eqref{eq:max} and  $ \mean $~\eqref{eq:mean} are limiting cases for $ \lse $:
\begin{align}
    \lim_{r \to 0} g_{\lse}(\left\{ x_i \right\}_{i=1}^k; r) &= g_{\mean}  (\left\{ x_i \right\}_{i=1}^k)
      \label{eq:r0} \\
      \lim_{r \to \infty} g_{\lse}(\left\{ x_i \right\}_{i=1}^k; r) &= g_{\max}  (\left\{ x_i \right\}_{i=1}^k)\label{eq:rinf}
\end{align}
If $ r $ is very small, the output of the function approaches simple average~\eqref{eq:r0}, on the other hand, if $ r $ is a large number, log-sum-exp becomes a smooth approximation of the $ \max $ function~\eqref{eq:rinf}. Since computing the value of $ g_{\lse} $ according to~\eqref{eq:lse} may lead to numerical instabilities, the following equality can be derived:
\begin{equation}
    \label{eq:lse_numstability}
    g_{\lse}(\{x_i\}_{i=1}^k; r) = \alpha + g_{\lse}(\{x_i - \alpha\}_{i=1}^k; r)
\end{equation}
which holds for any $ r > 0 $ and  $ \alpha \in \mathbb{R} $.\footnote{Proofs of all three claims can be found in Appendix~\ref{ap:proofs} (Statements~\ref{st:statement1}, \ref{st:statement2} and \ref{st:statement3}).} Therefore, to avoid overflow in summation of exponents, we use $ \alpha = \max_i x_i $. Finally, to satisfy the constraint that parameters $ \bm{r} $ are strictly positive, we represent them as a vector of (unconstrained) real values $ \bm{\rho} \in \mathbb{R}^m $. To evaluate $ g_{\lse} $, we re-parametrize with $ \rho \to \log(1 + \exp(\rho)) $ (softplus function\footnote{Specifically, its stable implementation $ \rho \to \max(\rho, 0) + \l1p(\exp(-\vert\rho \vert)) $.}) or any other  transformation that maps from $ \mathbb{R} $ to $ (0, \infty)$.

\begin{figure}[htp]
    \begin{subfigure}[t]{.245\textwidth}
        \centering
        \includegraphics[width=\textwidth]{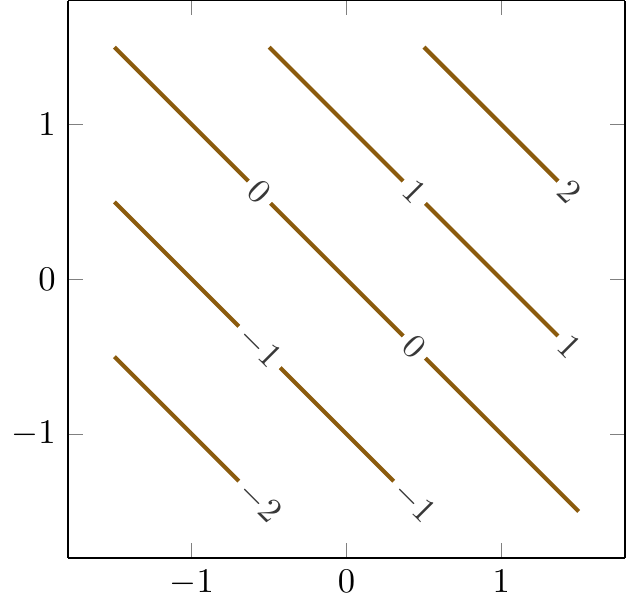}
        \caption{\label{sf:r0}$ r \to 0 $}%
    \end{subfigure}
    \begin{subfigure}[t]{.245\textwidth}
        \centering
        \includegraphics[width=\textwidth]{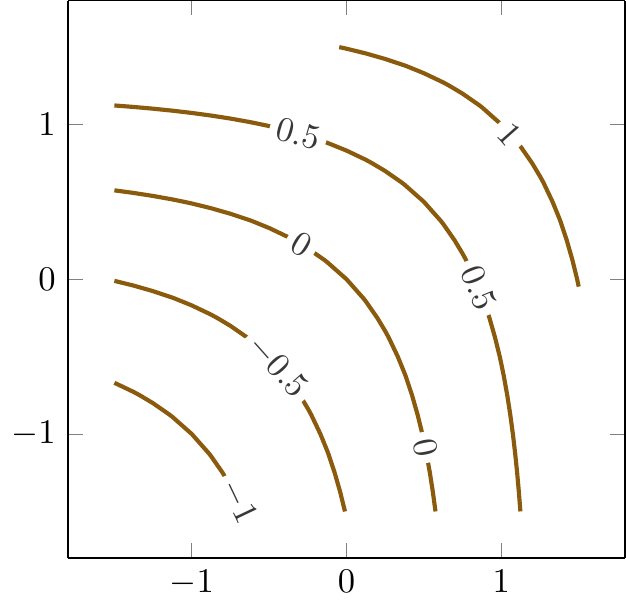}
        \caption{\label{sf:r1}$ r = 1 $}%
    \end{subfigure}
    \begin{subfigure}[t]{.245\textwidth}
        \centering
        \includegraphics[width=\textwidth]{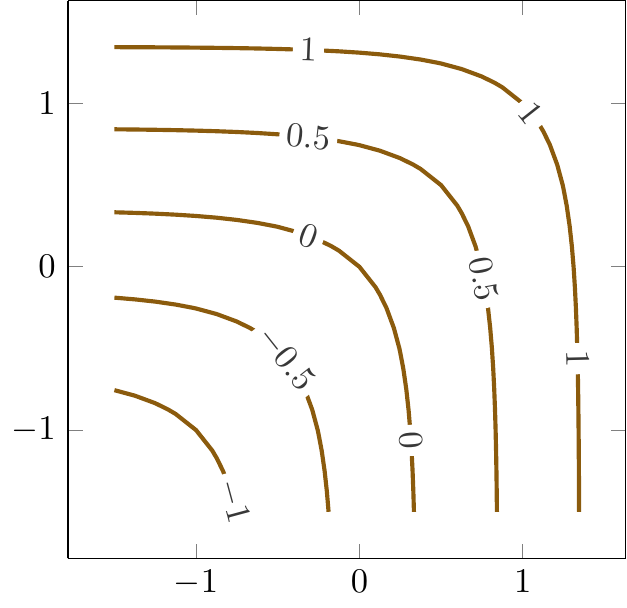}
        \caption{\label{sf:r2}$ r=2 $}%
    \end{subfigure}
    \begin{subfigure}[t]{.245\textwidth}
        \centering
        \includegraphics[width=\textwidth]{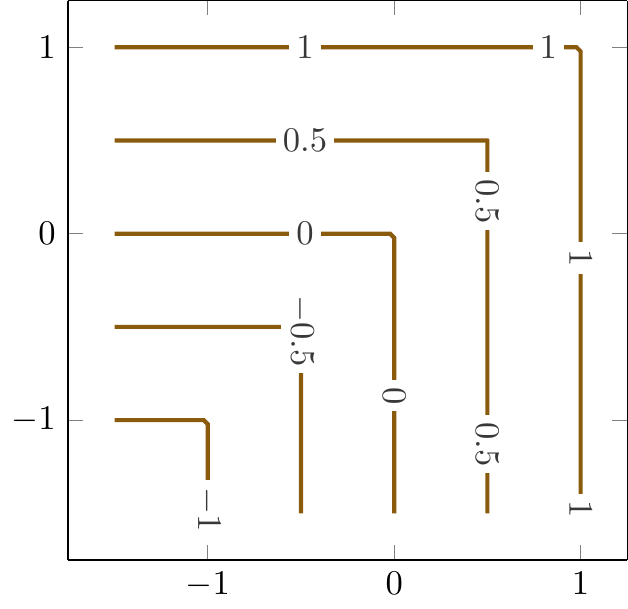}
        \caption{\label{sf:rinf}$ r \to \infty $}%
    \end{subfigure}
    \caption[Contours of $ \lse $ aggregation function.]{Contours of $ g_{\lse}(x_1, x_2; r) $ base aggregation defined in~\eqref{eq:lse} for the case when $ k=2 $ and  $ r $ takes different values. In case~(\subref{sf:r0}), where $ r \to 0 $, $ \lse $ aggregation is equivalent to the $ \mean $ aggregation and in case~(\subref{sf:rinf}), $ \lse $ is equivalent to $ \max $. Here, $ x $ axis represents values of  $ x_1 $ and  $ y $ axis values of  $ x_2 $, however, as all aggregations are invariant to permutations, contours would be the same even with swapped axes. }%
\end{figure}

\begin{wrapfigure}[16]{R}{0.35\textwidth}
    \centering
    \includegraphics[width=\linewidth]{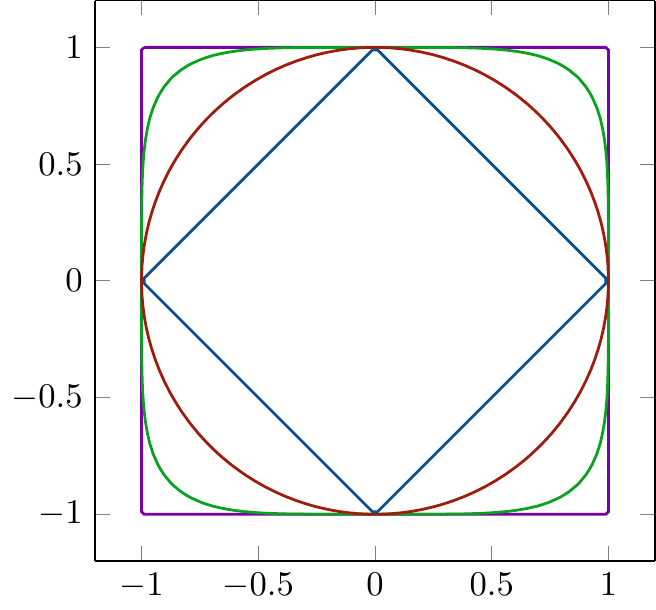}
    \caption[Visual representation of $ \pnorm $ aggregation function.]{A visual representation of $ g_{\pnorm }(x_1, x_2; p, c)$ when $ k = 2$, and colors represent different values of $ p $. Here $ \color{cvutblue}p=1 $, $ \color{cvutred}p=2 $, $ \color{cvut2}p=5 $ and $ \color{cvut5} p\to \infty $ . For clarity, $ c $ is set to $ 0 $ and also normalization  $ 1 / k $ is not used. The plot represents a set of points  $ (x_1, x_2) $ for which $ g_{\pnorm}(x_1, x_2; p, 0) = 1 $. }%
    \label{fig:pnorm}
\end{wrapfigure}

\noindent
Last aggregation is (normalized) $ \pnorm $, which is parametrized by a vector of real numbers $ \bm{p}$, where $ \forall i \in \left\{ 1, \ldots, m \right\}\colon p_i \geq 1 $, and another vector $ \bm{c} \in \mathbb{R} $. It is  computed with formula:
\begin{equation}
    \label{eq:pnorm}
    g_{\pnorm}(\{x_1, \ldots, x_k\}; p, c) = \left(\frac{1}{k} \sum_{i = 1}^{k} \vert x_i - c \vert ^ {p} \right)^{\frac{1}{p}}
\end{equation}
Due to the parametric interpretation of both $ p $ and  $ c $,  the $ \pnorm $ aggregation~\eqref{eq:pnorm} also generalizes on several well-known cases (for simplicity we assume here that $ c = 0 $):
\begin{equation}
    \label{eq:pnorm_cases}
    g_{\pnorm}(\left\{ x_i \right\}_{i=1}^k; p, 0) = \begin{cases}
     \frac{1}{k}\sum_{i=1}^k\lvert x_i \rvert, & \text{for } p = 1\\
        \sqrt{\frac{1}{k} \sum_{i=1}^k(x_i)^2 } , & \text{for } p = 2\\
        \max_{i=1, \ldots, k}\lvert x_i \rvert, & \text{for } p \to \infty\\
        \end{cases}
\end{equation}
Hence, for different values of $p$, $ \pnorm $ ~\eqref{eq:pnorm} interpolates between the (normalized) $\ell_1$ norm and the $ \ell_\infty $ norm.\footnote{For a proof for the case where $ p \to \infty$, see Statement~\ref{st:statement4} in Appendix~\ref{ap:proofs}.} Parameter $ c $ allows to shift the origin for each coordinate. Figure~\ref{fig:pnorm} represents the behavior of the $ \pnorm $ aggregation. Similarly to the case of $ \lse $, for stable implementation we make use of equality:\footnote{A proof can be found in Appendix~\ref{ap:proofs}, Statement~\ref{st:statement5}.}
\begin{equation}
    \label{eq:pnorm_numstability}
    g_{\pnorm}(\{x_i\}_{i=1}^k; p, c) = \beta \cdot g_{\pnorm}(\{x_i/\beta\}_{i=1}^k; p, c/\beta)
\end{equation}
which is valid for any $ \beta > 0, p \geq 1, c \in \mathbb{R} $. Therefore, we set $ \beta = \max\{1, \max_i \lvert x_i - c \rvert\}  $. Finally, to satisfy the constraint that all elements of vector $ p $ are greater or equal to 1, we employ a similar trick as in the case of the $ \lse $ aggregation, this time using transformation $ \rho \to 1 + \log(1 + \exp(\rho)) $.

\subsection{Weighted aggregations}%
\label{ssub:weighted_aggregations}

Following the motivation in~\ref{sub:weighted_bag_nodes}, we define the weighted versions of $ \mean $ and $ \pnorm $ aggregations:
\begin{align}
    g_{\mean}(\{(x_i, w_i)\}_{i=1}^k) &= \frac{1}{\sum_{i=1}^k w_i} \sum_{i = 1}^{k} w_i \cdot x_i \\
    g_{\pnorm}(\{x_i, w_i\}_{i=1}^k; p, c) &= \left(\frac{1}{\sum_{i=1}^k w_i} \sum_{i = 1}^{k} w_i\cdot\vert x_i - c \vert ^ {p} \right)^{\frac{1}{p}}
\end{align}
Note that since one weight is defined for each instance in the bag, weights influence result identically across output dimensions of $ g $, whereas aggregation parameters $ \bm{r} $, $ \bm{p} $ and $ \bm{c} $ define one parameter for each dimension and therefore may differ. The former case corresponds to altering a contribution of the given instance to the aggregation output. The latter case on the other hand allows to aggregate representations obtained from instance models differently according to dimension.

For instance, some elements of the instance representation are best aggregated with the $\mean$ function and others with the $\max$ function. By considering $ \bm{r} $ a learnable vector of parameters, a suitable $ \bm{r} $ can be inferred for each output dimension during training. To finish, since $ \max $ and its smooth approximation $ \lse $ identify extreme phenomena across the bag, it is reasonable not to use weights in their computation. Thus, weights are ignored:
\begin{align}
    g_{\max}(\{(x_i, w_i)\}_{i=1}^k) &= g_{\max}(\{x_i\}_{i=1}^k) \\
    g_{\lse}(\{(x_i, w_i)\}_{i=1}^k; r) &= g_{\lse}(\{x_i\}_{i=1}^k; r)
\end{align}

\section{Missing data}%
\label{sub:missing_data}

One detail that was deliberately left out in the description of the \gls{hmill} framework is how it handles incomplete or missing data. This phenomenon is nowadays ubiquitous in many data sources and occurs due to various reasons---a high price of obtaining an observation, information being unreachable due to privacy reasons, a gradual change in the definition of data being gathered, or a faulty collection process, to name some examples. Regardless of the origin of missing data, it is wasteful to throw away the incomplete observations altogether. Possibly missing labels gave rise to fields such as \emph{supervised}, \emph{semi-supervised} and \emph{unsupervised} learning.

The \gls{hmill} framework helps to leverage knowledge hidden in incomplete observations, not labels. Thanks to the hierarchical structure of both samples and models, we can still represent samples with missing information fragments at various levels of abstraction. Problems of this type can be categorized into three not necessarily separate types:

\begin{wrapfigure}{R}{0.3\textwidth}
    \centering
    \includegraphics[width=\linewidth]{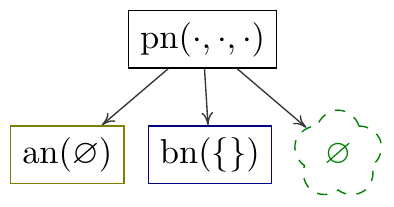}
    \caption[Missing data example.]{Example of a sample tree following schema in Figure~\ref{fig:mil_schema} with missing data. In this particular case all three possible types of missing data appear.}%
    \label{fig:missing_data}
\end{wrapfigure}
\begin{enumerate}
    \item Missing information fragments in array nodes 
    \item Empty bag nodes
    \item Whole data source missing in product nodes 
\end{enumerate}

\noindent
One example of a sample tree with missing data is in Figure~\ref{fig:missing_data}. In this particular example, information fragment in an array node responsible for modelling stems ({\color{olive}olive}) is not observed, which is denoted by the $ \varnothing $ sign. The subtree modelling blooms ({\color{navy}navy}) is a bag node with no instances observed. The third source of data---buds ({\color{cgreen}green})---is missing altogether. According to our experience, the latter two cases are more common than the first one. To incorporate missing data into the framework, we provide model nodes with additional functionality to handle these situations and also declare sample trees with missing data to match the same schema as their counterparts without incomplete information so that they can be processed with same structured models. Although this requires a slight modification of Definitions~\ref{def:schema_matching},~\ref{def:valid_sample},~\ref{def:model} and~\ref{def:model_evaluation}, we will not delve into details and will instead explain, how models can be extended to deal with missing data.

Recall that any \gls{hmill} model node can be interpreted as a function from a set of sample trees following a certain schema $ \mathcal{T}_s $ to $ \mathbb{R}^m $, where $ m \in \mathbb{N} $. This means that during evaluation, each model node always receives vectors of the same size from all of its children. To deal with missing data, we introduce a vector of real parameters $ \bm{\psi} $ into each model node. For an array model node implementing a mapping $ f\colon \mathbb{R}^n \to \mathbb{R}^m $, $ \bm{\psi} $ has $ m $ elements. During both training and evaluation of the model, whenever the array model node encounters a missing information fragment  $ \varnothing $, instead of attempting to evaluate $ f $, it returns vector $ \bm{\psi} $.

For bag model nodes, $ \bm{\psi} $ has the same dimensionality as the output of aggregation $ g $. If a bag is empty, we feed to bag model $ f_B $ the vector of default values $ \bm{\psi} $ instead of aggregated instances.

In the case of product model node, we keep one vector $ \bm{\psi}_i $ for each of its children. Whenever is the $ i$-th child missing, we provide the corresponding  $ \bm{\psi}_i $ as input to concatenation and further processing. Since all these parameters are learned, this technique enables effortless processing of both samples containing all information and samples with missing data, all at the price of a tiny increase in the overall number of parameters.  Note that even though the three distinguished cases of missing data may seem equivalent, in fact, each of them represents different situations and results in a different treatment.

For example, consider the sample from Figure~\ref{fig:missing_data}, but at this time missing the middle branch altogether. This would result in filling in default values from $ \bm{\psi}_2 $ at the level of the root. However, in the current situation, we use $ \bm{\psi} $ from the bag model node, which means that different default values are used and also in this case they are futher processed by bag model $ f_B $. Therefore, a small change in the sample results in an entirely different computation.

Parameters $ \bm{\psi} $ can be either fixed or even learned during training. This requires a procedure for computing gradients of the loss function for all learnable parameters in the model. This is the main topic of the following section.

\section{Gradient computation}%
\label{sub:gradient_computation}

As all functions composing models are defined to be (piecewise) differentiable, and the computational graph is a tree, gradients with respect to all parameters can be computed by adaptation of the original backpropagation algorithm~\cite{Rumelhart1986}, which we describe in Algorithm~\ref{alg:backprop}. It proceeds in a preorder fashion and for each model node in the tree computes a derivative of loss with respect to the parameters of the node and also to its input. The derivative with respect to the input is used subsequently by the children of the node. In step 8, we use a combination sum and chain rule for aggregation functions and the standard backpropagation algorithm for neural network layers to backpropagate through a given model node.

Note that this pseudocode is supposed to give a rough idea of how gradients are computed and some details are omitted. For instance, because all instances in a bag data node are processed with the same instance model in a corresponding bag model node, the parameters are shared. As a result, gradients of all submodels that instance model consists of have to be appropriately aggregated over all instances in the bag, for example, averaged in the case of $ \mean$ aggregation. Also, if default vectors $ \bm{\psi} $ are used, Algorithm~\ref{alg:backprop} has to be modified to change the control flow according to whether the default values have been filled in or not. Delving further into details is beyond the scope of this thesis and we refer the reader to the literature about differentiable programming and our implementation.

\begin{algorithm}[htb]
	\begin{algorithmic}[1]		
		\Procedure{Backprop}{$ s $, $F$, $\mathcal{L}$, $t$}
		\Input a schema $s$,  a model $F\colon \mathcal{T}_s \to \mathbb{R}^n$ constructed from $ s $, a loss function $\mathcal{L}\colon \mathbb{R}^n \to \mathbb{R}^+_0$, a sample $t \in \mathcal{T}_s$ 
		\Output $ \frac{\partial\mathcal{L}}{\partial \theta}   $ for all model parameters $ \theta $
        \State evaluate $ F(t) $ and save inputs $ i_f(t) $ and outputs $o_f(t)$ and  of every model node $ f $ in  $ F $
        \State compute $ \frac{\partial\mathcal{L}}{\partial F(t)}  $
        \State $ q = \{\myroot(F)\} $
		\Comment{push the root model node to the queue}
		\While{$\lvert q \rvert > 0$}
        \State~$f \gets \pop(q)$
		\Comment{pop one model node from the queue}
		\State $\Theta_f \gets$ all parameters of $ f $
        \State compute $ \frac{\partial\mathcal{L}}{\partial i_f(t)}   $ and $ \frac{\partial\mathcal{L}}{\partial \theta} $ for all $\theta \in \Theta_f $  using $ \frac{\partial\mathcal{L}}{\partial o_f\left( t \right) } $ computed in the parent 
        \State $\forall c \in \children(t)\colon \push(q, c)$
		\EndWhile
		\EndProcedure
	\end{algorithmic}
    \caption[Backpropagation algorithm adaptation for HMill models.]{Adaptation of original backpropagation algorithm~\cite{Rumelhart1986} to a tree-structured computational graph.}%
    \label{alg:backprop}
\end{algorithm}

\noindent

\section{Batching procedure}%
\label{sub:batching}

Nowadays, full \emph{Batch gradient descent}, which computes gradients of model parameters using the whole dataset in each iteration, is no longer employed for many problems. The reasons are twofold. Firstly, iterating over all training samples requires a lot of time and is also considered inefficient given the fact that many samples are very similar. Secondly, it is known that stochasticity in training procedure improves performance. Instead of full Batch gradient descent, its stochastic variants are used---\emph{Stochastic gradient descent}, which selects\footnote{The selection is performed either by sampling or some deterministic procedure, such as iteration over the dataset.} one observation for every update, or \emph{Mini-batch gradient descent}, which interpolates between these two, by selecting a predefined number of observations from the dataset. We have already defined how a single \gls{hmill} sample looks like. In this section, we describe how to group samples into (mini)batches and efficiently compute gradients with respect to the whole (mini)batch.

Two prominent guiding principles in modern scientific computing are \emph{array programming} and a \emph{static computational graph}. Array programming refers to writing computer programs in a way that facilitates the application of operations to multiple values at once, so that specialized vector instructions in CPUs can be utilized. Regardless of whether the data are processed on GPU or on vector processors, the focus is to group (possibly independent) data together and perform operations on it simultaneously. For instance, computation of an output of a neural network can be expressed as a composition of matrix-vector multiplications and elementwise application of activation functions. If matrix-matrix multiplications are performed instead, a whole batch of samples can be run through the network at once. This is much faster than feeding samples to the network one by one.

The second principle is to keep the computational graph as static as possible so that both high-level optimizations at the level of computational graph and low-level optimizations in the compiler can be utilized. A naïve approach to compute gradients with respect to each sample tree from the batch first and summing the results violates both of these principles. Nevertheless, \gls{hmill} offers a way to group sample trees following the same schema into a single tree representing the whole batch in a manner respecting both principles. In other words, as in the standard neural network case, all matrix-vector operations can be transformed into matrix-matrix operations, and the computational graph is identical.

The high-level overview of the whole procedure is that all sample trees from the batch are first horizontally overlapped and the data nodes in the same positions in the tree are merged into one. This is possible due to the fact that all samples match the same schema and requires small extensions of all data node types so that they can carry more than one sample. All model nodes will instead of one vector per sample return a matrix per multiple samples, in which outputs for samples are stored in columns.

\subsection{Array data nodes}%
\label{sub:array_data_nodes}

\begin{wrapfigure}[14]{R}{0.4\textwidth}
    \centering
    \includegraphics[width=1\linewidth]{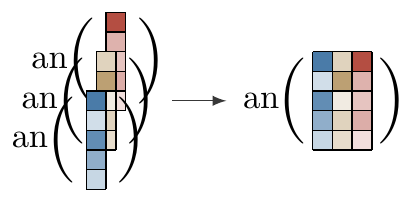}
    \caption[Array data node merging.]{Illustration of merging three array data nodes representing three different samples (drawn by different colors) into a single array node.}%
    \label{fig:batch_array}
\end{wrapfigure}

In the case of an array data node, the modification is done in the same way as in the aforementioned example, that is, instead of one vector representing the information fragment of the sample in the given position in the tree, it carries multiple vectors concatenated into a matrix for multiple samples. This is illustrated in Figure~\ref{fig:batch_array}, where three array nodes representing three different samples are merged together into one. Given an array data node and its matching model node, evaluation is performed in the same way as in Figure~\ref{fig:array_model}, however, this time we input a matrix into function $ f $. As a result, we obtain not one vector, but a whole matrix, where each column corresponds to one sample.

\subsection{Bag data nodes}%
\label{sub:bag_data_nodes}

Merging several bag data nodes together is possible thanks to the following observation. Because all samples in the batch follow the same schema, all instances from all samples in the batch are represented by trees that also follow the same (sub)schema. This implies that all trees representing instances can be collapsed into a single tree as well recursively by following the same algorithm, and this observation holds for any position of the bag node in the schema tree. Hence, given a batch of samples, to merge a corresponding bag data nodes together, we collect all instances from all samples in the batch and apply the same algorithm. Put differently, the bag node now has only one child, which is a result of merging all instances of all samples it stores. Moreover, we remember the indices of instances in order to assign them to bags during evaluation.

\begin{figure}[htp]
    \centering
    \includegraphics[width=0.8\linewidth]{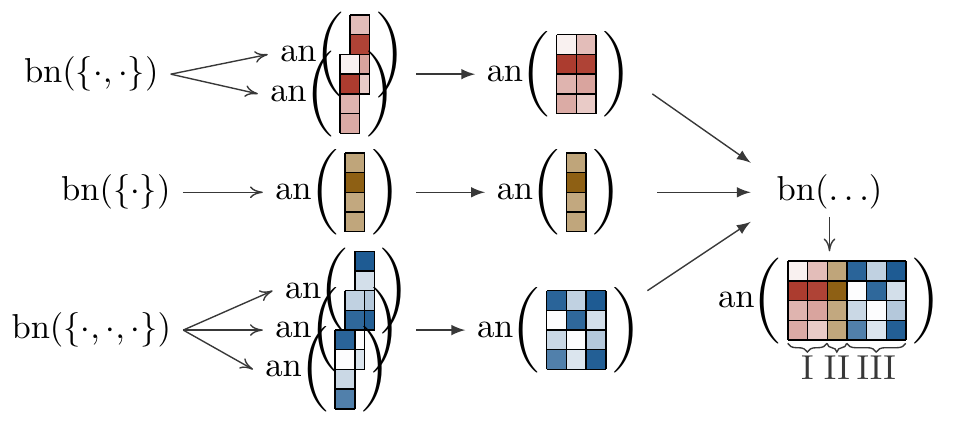}
    \caption[Bag data node merging.]{Merging three bag nodes corresponding to different samples (drawn by different colors). See text for details.}%
    \label{fig:batch_bag}
\end{figure}

For a specific example, refer to Figure~\ref{fig:batch_bag}, where a process of merging three bag nodes with two, one and three instances in their bags is illustrated. As all instance trees (in this example trees consisting of only one array node) follow the same schema, they can be collapsed into one tree. In this case, this means that instead of six different array data nodes representing six different instances in three bags, we have one array node, which becomes the only child and stores all information. Note, that this child can also be another arbitrarily complex subtree---bag data node or product data node, depending on the schema, however, thanks to the observation we made, it is always possible to merge all instances of all samples into a single subtree. Therefore, the resulting bag data node always has only one child after merge.

The evaluation of a merged node with a given bag model node is performed analogically to the single sample case (Figure~\ref{fig:bag_model}). We first use the instance model $ f_I $ on the only child obtaining a matrix, where each column stores a result of $ f_I $ on the corresponding instance. In example in Figure~\ref{fig:batch_bag}, we would get a matrix with the number of rows equal to the dimensionality of target space of $ f_I $ and six columns. Then, aggregation  $ g $ is performed using the indices stored in the merged bag node, so that instances belonging to the same bags are correctly aggregated together. We store the results into a narrower matrix having as many columns as there are samples (in this case three) and apply bag model  $ f_B $ on it, which may change the number of rows (the dimensionality of target space of  $ f_B $), but the number of columns stays the same.

\subsection{Product data nodes}%
\begin{figure}[htp]
    \centering
    \includegraphics[width=0.7\linewidth]{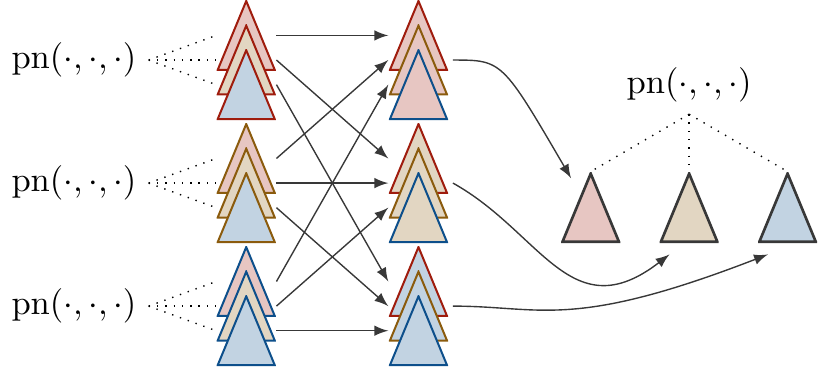}
    \caption[Product data node merging.]{Procedure of merging three different product nodes matching same schema in one. Here, the fill colors indicate different (sub)schema that children follow (see Figure~\ref{fig:product_node}) and the same border colors represent subtrees coming from the same sample.}%
    \label{fig:batch_product}
\end{figure}

\label{sub:product_data_nodes}
Finally, if all samples in the batch follow the same schema, this also means that all product data nodes at the same position in different sample trees have children that can be merged into one subtree for all samples. In other words, even though children of a single product data node may match different (sub)schemata and therefore they may not be combinable as is the case with bag nodes, we can merge children across multiple sample trees provided that they follow the same (sub)schema. This is done by merging all first children into a single tree, then all second children into a single tree, and so forth. See Figure~\ref{fig:batch_product}, where this procedure is applied to product data nodes from three different samples with three children each. Children are grouped according to the (sub)schema they follow\footnote{or equivalently by their order in product data node, since we have defined product nodes to impose ordering on their children} and all groups are recursively collapsed into one subtree. The evaluation of a product model node is done in a similar way as in Figure~\ref{fig:product_model}. We apply submodels $ f_1, \ldots, f_l $ to children and obtain a matrix with as many columns as there are samples for each child. In our specific example, we would obtain three matrices with three columns and possibly different number of rows. Then, all matrices are vertically concatenated and mapping $ f $ is applied, which leads to one column for each sample on the output.

\newpage\noindent
Applying this recursive procedure on any number of samples leads to a single tree, where all intermediate representations are stored in one matrix. Static schema across the batch guarantees static computational graph, and therefore the computation of gradients on the resulting tree is effective. To simplify the explanation, we have assumed that there are no missing data present in any sample. In practice, this is not true, and slightly more complex functionality is required. For instance, if any of the bags is empty when merging bag data nodes together, we use the methodology described in Section~\ref{sub:missing_data}. Hence, the resulting bag data node remembers that this particular bag has zero instances and when performing aggregation, the resulting column is filled with the default values. 
\newline\newline\noindent
Having read this and the previous chapter, the reader should now be familiar with sample representation and model definition in \gls{hmill}, as well as the technical details of its implementation. Should the reader wish to learn even more, we recommend referring to our implementation~\cite{Mill2018}. The source code of version \texttt{1.2.0} of the framework is attached to the submission. In the remaining chapters, we delve into the practical adoption of \gls{hmill}. 


\chapter{Towards automated processing of structured data}%
\label{cha:jsons}

In this chapter, we demonstrate how the \gls{hmill} framework lends itself to the processing of structured hierarchical data. We mentioned in Chapter~\ref{cha:difficulties} that a majority of \gls{ml} algorithms is applicable only to fixed-size input, even though real-world data is hardly ever found in a form in which each observation is described by a vector of fixed length. On the contrary, plenty of data appears naturally structured as graphs or trees, with information being specified not only in nodes but also in the structure itself. This brings higher flexibility when opting for formats in which the data is stored, however brings new challenges to methods processing such input. We focus on such hierarchical formats in this part, and in the remaining Chapters~\ref{cha:hmill_on_graphical_data} and~\ref{cha:cisco} on the processing of graphs.

\begin{figure}[hbtp]
    \centering
    \includegraphics[width=11cm]{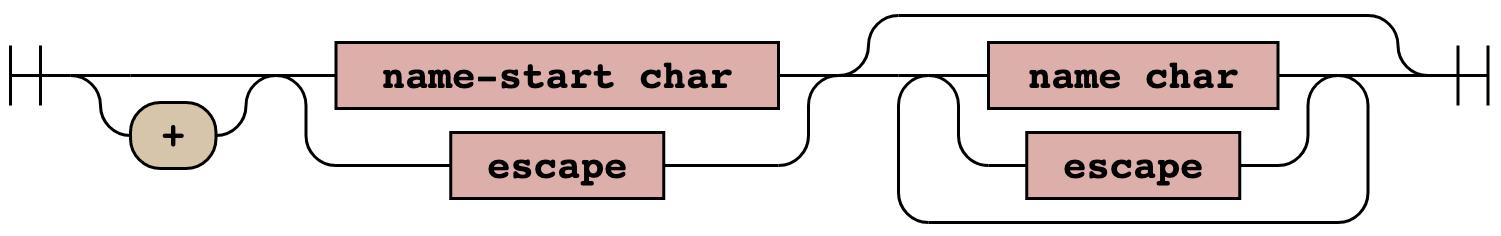}
    \caption[Syntax diagram of a JSON value.]{Syntax diagram of a \gls{json} value.}%
\end{figure}

\noindent
Specifically, we consider hierarchically structured data formats ubiquitous in today's Internet communication, such as \gls{json} and \gls{xml}. We will demonstrate how these formats seamlessly integrate into the framework and therefore how data stored in this way can be learned from. We will illustrate this on \gls{json} format in particular, however, everything applies to \gls{xml}s as well.

\gls{json} is a lightweight, human-readable format designed for data exchange independently of the programming language used. The standard\footnote{\url{https://json.org}} defines three main structures (data types):

\begin{description}
    \item[Values] are the most general type of \gls{json} structure and can be either primitive data types, such as strings, numbers, booleans or \texttt{null}s, or composite \emph{object} or \emph{array} types.
    \item[Objects] are unordered collections of zero, one, or more key-value pairs, which abstract on \emph{dictionaries}, \emph{hash tables}, or \emph{associative arrays} in a modern programming language. Keys are defined as arbitrary strings and values are \gls{json} values.
    \item[Arrays] are, contrary to objects, \emph{ordered} lists of values, corresponding to standard \emph{arrays} in programming languages. Elements of \gls{json} arrays can be any \gls{json} values.
\end{description}

\begin{figure}[hbtp]
    \begin{subfigure}[t]{.623\textwidth}
        \centering
        \includegraphics[width=10.27cm]{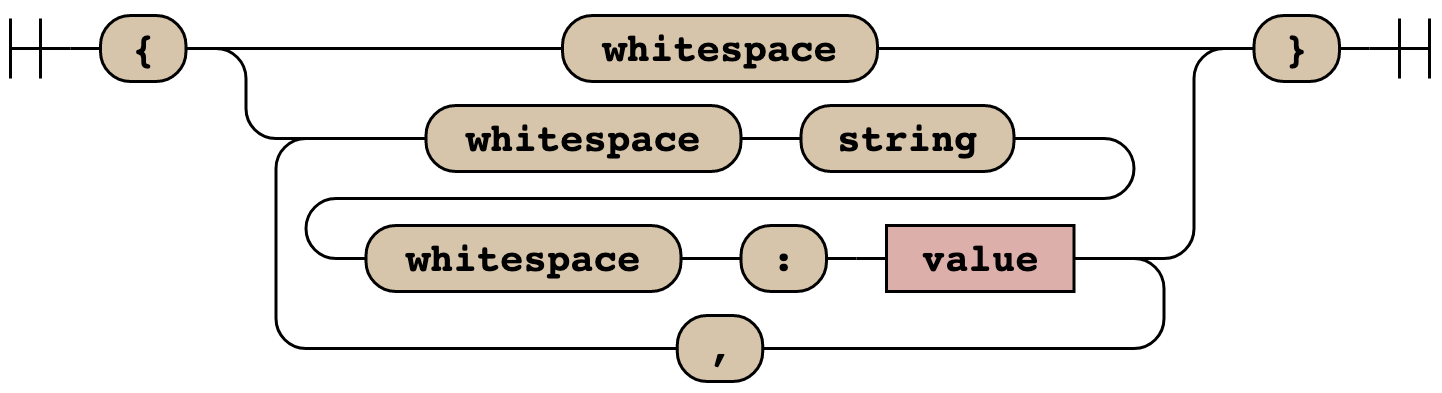}
    \caption{\label{sf:jsonobject}}%
    \end{subfigure}
    \begin{subfigure}[t]{.376\textwidth}
        \centering
        \includegraphics[width=6.19cm]{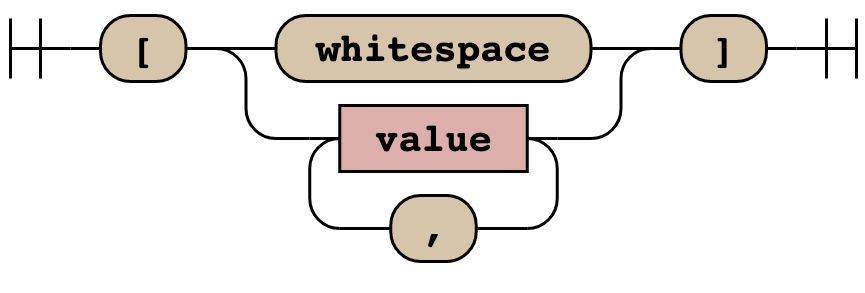}
        \caption{\label{sf:jsonarray}}%
    \end{subfigure}
    \caption[Syntax diagram of a JSON object and a JSON array.]{Syntax diagram of a \gls{json} object~(\subref{sf:jsonobject}) and a \gls{json} array~(\subref{sf:jsonarray}).}%
\end{figure}

\begin{wrapfigure}{L}{0.4\textwidth}
    \includegraphics[width=0.4\textwidth]{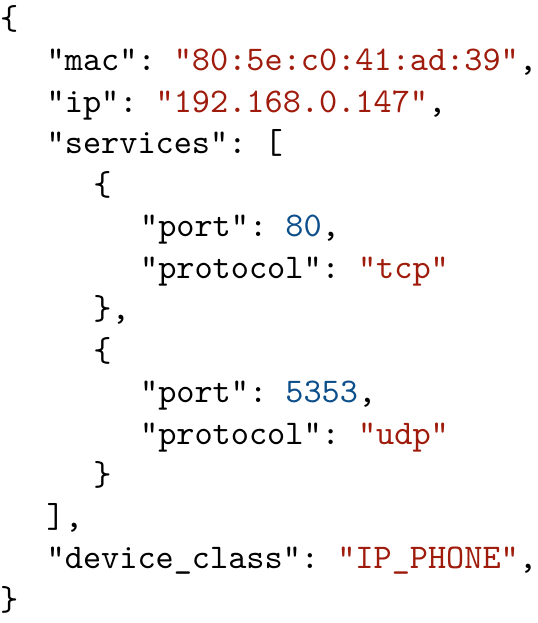}
    \caption{Example of a JSON document.}%
    \label{fig:json}
\end{wrapfigure}

\noindent
Apparently, \gls{json} format provides higher expressive power for objects of interest, as the single-vector representation can be encoded as \gls{json} array with real numbers as its elements and constant size across observations. Whereas \gls{json} arrays are most of the time used to store elements with the same interpretation\footnote{\gls{json} standard though allows elements to be any \gls{json} value.}, \gls{json} object is a data type intended to accommodate multiple possibly heterogeneous attributes with an arbitrary representation. Needless to say, thanks to the recursive definitions of \gls{json} data types, both arrays and objects can be further nested, creating complex composite structures capable of capturing many real-world phenomena.

An example of a valid \gls{json} file  storing information about an \gls{iot} device is in Figure~\ref{fig:json} and it is based on the task we attempt to solve in the experiment section of this chapter. The whole top-level object consists of data entries specifying the \gls{mac} address and the \gls{ip} address of the device, services it provides and the class of the device, which in this case serves as a label. As one device can expose multiple ports and use different protocols for communication, we model this as an array of objects, each specifying one service. Every \gls{json} value can be regarded as a node in a tree representation of the document, with arrays and objects having several children and primitive values, such as numbers or strings, being leaves of the tree. Schemata representing the graphical structure can be defined in a similar way as in \gls{hmill},  and therefore, \gls{json} documents can be validated against schemata.

In this thesis, we focus mainly on the classification task with valid \gls{json} documents or similar data formats as input. Traditional approaches include \emph{rule-based} systems or \emph{flattening}. Rule-based systems combine manually designed simple rules based on domain knowledge with transformations of suitable subsets of data. Referring to Figure~\ref{fig:json}, one example of such rule could be that the observed device is a phone as long as it uses the UDP protocol on port $ 5353 $. Hence, rules, such as logical \texttt{AND} or \texttt{OR} help with dealing with \gls{json} arrays of arbitrary length. To increase the space of such rule-based hypotheses a transformation on the appropriate subsets of the document can be defined or even learned.

For instance, in the previous example, we could collect all port-protocol pairs in documents describing phones and infer a detector returning a score specifying how likely this pair appears in the descriptions of phone devices. Then, a device would be classified as a phone if all of its elements in \texttt{"services"} array reach score above a certain threshold. The rule-based approach to handling tree-structured data in many ways resembles the instance-space paradigm for solving \gls{mil} problems from Section~\ref{ssub:instance_space_paradigm} with all of its benefits and drawbacks. The main advantage of the rule-based approach is that rules are easily interpretable, however, they require deep domain knowledge and can be very difficult to design properly. Moreover, in the presence of the concept drift, rules may become quickly obsolete, and sometimes, the language of simple rules may not even be expressive enough to reasonably discriminate samples, especially for more complex structures.

\noindent
The flattening approach deals with hierarchical data by defining a procedure that yields a vectorized representation of any given tree. This is convenient since a lot of well-researched \gls{ml} algorithms can be used with the resulting vector representation. One major disadvantage of this approach is that it does not deal with complex variable structures in a principled way---as different tree-structured observations may contain different amount of information (for instance by storing \gls{json} arrays of different length), transformation to a fixed-size vector can be either lossy if the size is too small, or wasteful if it is too high.

The hierarchical nature of tree-structured data presents an opportunity to increase the capacity of models by leveraging structural information from the data while keeping the number of parameters sufficiently low.  The prior art on models learning from both raw data in leaves and topologies of such structures is scarce and existing methods use limiting assumptions, such as all leaves being present in the same depth~\cite{jsons1, jsons2}, or the tree having a very specific structure, such as \emph{syntactic parse trees} in natural language processing~\cite{jsons4, jsons5, jsons3}. To the best of our knowledge, the only fully general approach to end-to-end learning on hierarchical data is proposed in~\cite{jsons6}.

\section{Modelling JSONs with HMill}%
\label{sec:sec_hmill_jsons}

To use \gls{hmill} for learning from \gls{json} documents, we need to make one mild assumption that all documents in the dataset must be structured in the same way (in \gls{hmill} parlance, they must follow the same schema). Given a set of \gls{json} documents $D$, this assumption, loosely speaking, translates to the following conditions:

\begin{itemize}
    \item Keys of every \gls{json} object located at a certain position in the tree representation of any document $d \in D$ must come from a finite `super' set. In other words, given a fixed position in the tree, we know which keys can be encountered in this particular position before training procedure starts. 
    \item Values corresponding to the same key of \gls{json} objects located at the same position must have the same structure for every document $d \in D$.
    \item \gls{json} Arrays must be either empty or contain elements with the same structure. For a certain position in the tree, elements of arrays of different documents from $ D $ must also have the same structure.
\end{itemize}

\noindent
The whole situation is analogical to the notion of \gls{hmill} schema, which is defined and discussed in Section~\ref{ssub:rigorous_definition}. To demonstrate this condition on the example from Figure~\ref{fig:json}, if we wanted to train our model on a set of multiple similar \gls{json} documents $D$, we would require each one of them to contain at the top level only key-value pairs with keys from set $\{$ \texttt{"mac"}, \texttt{"ip"}, \texttt{"services"}, \texttt{"device\_class"} $\}$, with every value corresponding to any key apart from \texttt{services} being represented as a string of characters. Note that this does not imply that all documents from $ D $ must have the same key sets---some keys might be missing (see Section~\ref{sub:missing_data}). All documents in $D$ would further have either no \texttt{services} key present, or its value would be an array of another type of \gls{json} objects with keys from set $\{$ \texttt{"port"}, \texttt{"protocol"} $\}$ with values represented as an integer for ports and string for the protocol. Even though the assumption of the common schema for all documents might seem limiting, according to our experience, this is rarely the case. Despite the \gls{json} standard being rather forgiving in how the documents can be structured, when we observe documents from one specific data source or a collection of data sources, they usually follow the same schema directly, or only a trivial transformation is required. 

\begin{figure}[hbtp]
    \centering
    \includegraphics[width=\linewidth]{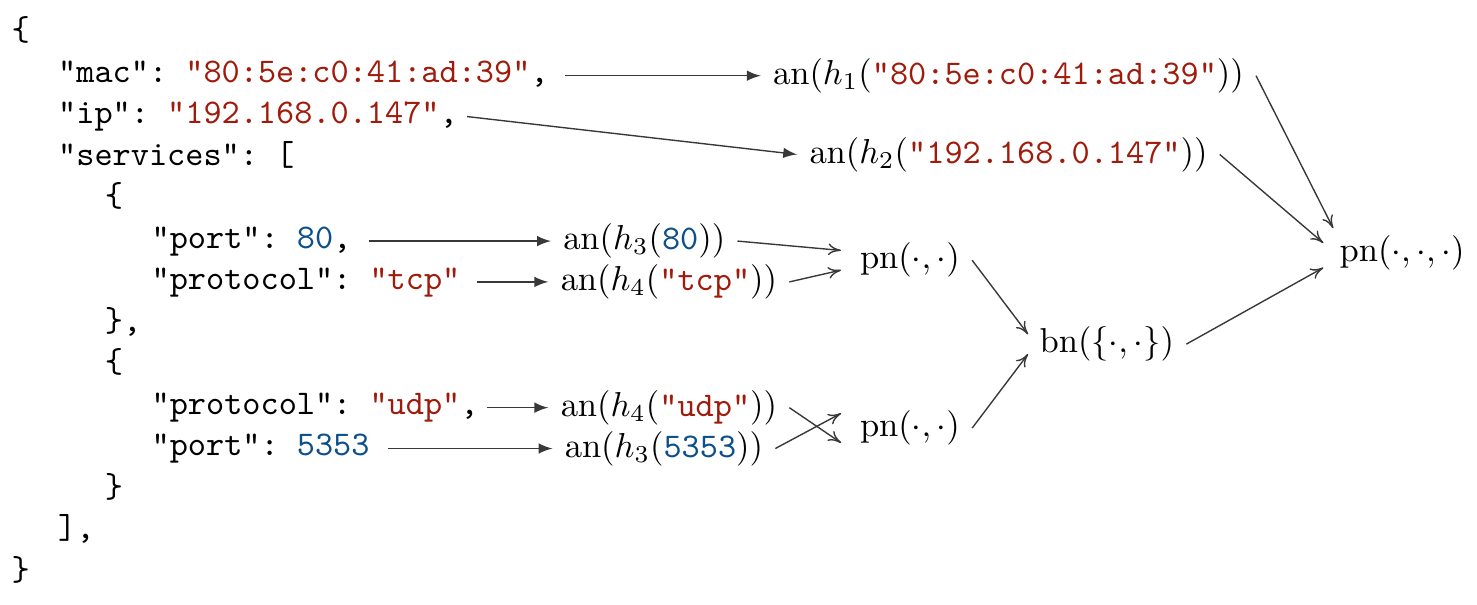}
    \caption[Transformation of JSON document to a valid HMill sample.]{A transformation of \gls{json} document from Figure~\ref{fig:json} into a valid \gls{hmill} sample. See text for details.}%
    \label{fig:json_sample}
\end{figure}

Translating a \gls{json} document to an \gls{hmill} sample is straightforward. Each leaf of the tree with primitive data type entries such as numbers, booleans, or strings is transformed into an array data node $ \adnode(h(\ldots)) $, where $ h $ is a suitable mapping to some real vector space (see Section~\ref{sub:array_node}). Therefore, the entry is converted to a vector of real numbers. Note that mapping  $ h $ can be defined differently for leaves at different positions in the tree, however, it is the same for leaves at the same position in the tree across all documents. For instance, mapping $ h $ in an array node modelling \gls{ip} addresses is the same for all documents but may differ from a mapping used for \gls{mac} addresses.

To model \gls{json} objects, we use product data nodes. Recall that given a particular position in the tree, a finite set $K$ of all keys used by \gls{json} objects at this position is known. First, an arbitrary ordering $ \prec $ is imposed on all elements $ k \in K $, for instance, trivial lexicographical order of key strings. This way, we can assign a unique number from $ \{1, \ldots , |K|\} $ to each key, according to its rank with respect to $ \prec $. This number is regarded as the index of the child (subtree)  corresponding to the \gls{json} value stored with this key in the product node. This is merely a technical detail, stemming from the fact that children of product data nodes can be semantically heterogeneous, however, key-value pairs from \gls{json} objects are originally unordered. Therefore, we need to define this arbitrary, yet precise mapping from keys to children indices. As children of product data nodes do not have to match the same schema, we can model heterogeneous values stored under different keys with an arbitrary substructure. For instance, each of the \texttt{mac}, \texttt{ip} and \texttt{services} entries could be modelled by a different subtree and therefore processed by submodels with variable topology.

Finally, \gls{json} arrays are modelled as \gls{hmill} bag data nodes by interpreting each homogeneous entry in the array as an instance in the corresponding bag. Note that \gls{hmill} handles empty arrays as well as arrays of any length. The whole transformation is depicted in Figure~\ref{fig:json_sample}. The root of the resulting sample tree is a product data node, with the first two children being subtrees for modelling \gls{ip} and \gls{mac} addresses. The last child is a bag data node representing an array of elements from the \texttt{services} field, each of which is another \gls{json} object and is therefore modelled as another product data node. Four distinct mappings $ h_1 $,  $ h_2 $,  $ h_3 $ and  $ h_4 $ are defined for modelling different information fragments (\gls{ip} and \gls{mac} addresses, port and protocol). For specific examples of various mappings $ h $, refer to Figure~\ref{fig:mapping_h}. Also, as \gls{json} object is unordered, the order of keys in the second element of \texttt{services} arrays is changed so that each product node modelling entries in the array has submodel for ports as the first child, and submodel for protocols as the second child. 

The resulting pipeline for learning from \gls{json} documents is not much different from the pipeline in the current machine learning. We first collect the data, in this case, a collection of documents $ D $ and deduce the schema. This can be done in an automated way, for instance with  \texttt{JsonGrinder.jl} library~\cite{Grinder2019}. Then, a suitable model is defined according to the schema and trained in an end-to-end fashion on both primitive types in leaves of the documents and the whole topology. As models learn from the structure as well, a need for the tedious, trial-and-error design of a flattening procedure to encode the structure in a fixed-size vector is alleviated. Also, learning \gls{hmill} models from the data is more general than rule-based approaches, and a more complex hypothesis can be learned, as aggregation functions in bag model nodes can easily substitute hand-written rules. We believe that \gls{hmill} will enable out-of-the-box learning from tree-structured data expressed as \gls{json}s, \gls{xml}s and other similar formats in tasks across variable domains and will help other researches with processing data so far considered incompatible with current methods.

\section{IoT device identification (use case)}
The first of three use cases we will present throughout the thesis concerns \gls{iot} domain. \gls{iot} is a term referring to a network of billions of existing devices in the Internet that communicate with each other and collect and share data through special sensors. Thanks to advances in multiple disciplines and cheaper manufacturing, the \gls{iot} network saw immense growth in the last decade and, for instance, nearly half of the North American households have an Internet-connected television or a streaming device~\cite{Kumar2019AllTC}. It is believed that \gls{iot} will benefit society across sectors, ranging from health services and home automation for the elderly and disabled, to energy savings in `smart homes'. However, The progress is hindered by security and privacy concerns and rightfully so---some of the examples include up to $600\,000$ infected devices being used in massive \gls{ddos} attacks by the infamous \emph{Mirai} botnet~\cite{Antonakakis2017UnderstandingTM}, or smart home devices being compromised~\cite{Hernandez2014SmartNT, Kumar2018SkillSA}. Thus, it is of utmost interest of security companies and device vendors to turn their attention to the security of the \gls{iot} network and develop suitable solutions.

In~\cite{Kumar2019AllTC}, researchers from University of Illinois Urbana-Champaign, Stanford University and Avast Software cooperated to conduct the first large-scale investigation and empirical analysis of \gls{iot} devices in homes that are otherwise invisible in Internet scanning. Data was collected by local subnet scanning manually initiated by users. One of the problems that emerge when the network is scanned is an identification of the device type (TV, wearable, voice assistant, \ldots) using only information obtained from the scan. In the following sections, we demonstrate how \gls{hmill} helps with solving this task.

\subsection{Data description}
The scan is performed by a tool called \emph{WiFi Inspector} built into the antivirus client.\footnote{All data about devices is collected only for research purposes, sensitive information is anonymized or aggregated and users are informed and can opt out anytime. See~\cite{Kumar2019AllTC} for further details about privacy policies.} First, the list of candidates is generated using the \gls{arp} table and \gls{arp}, \gls{ssdp} and \gls{mdns} scans of the network. Then, additional knowledge about each candidate is gathered from both the network and the application layer. Specifically, ports used for listening and protocols for communication are further scanned, as well as \gls{dhcp}, \gls{mdns}, \gls{ssdp} and \gls{upnp} broadcasts made by the device. Two specific examples are shown in Apendix~\ref{ap:deviceid} (Figures~\ref{fig:deviceid_sample1} and~\ref{fig:deviceid_sample2}). 

\begin{wrapfigure}{L}{0.5\textwidth}
    \includegraphics[width=0.5\textwidth]{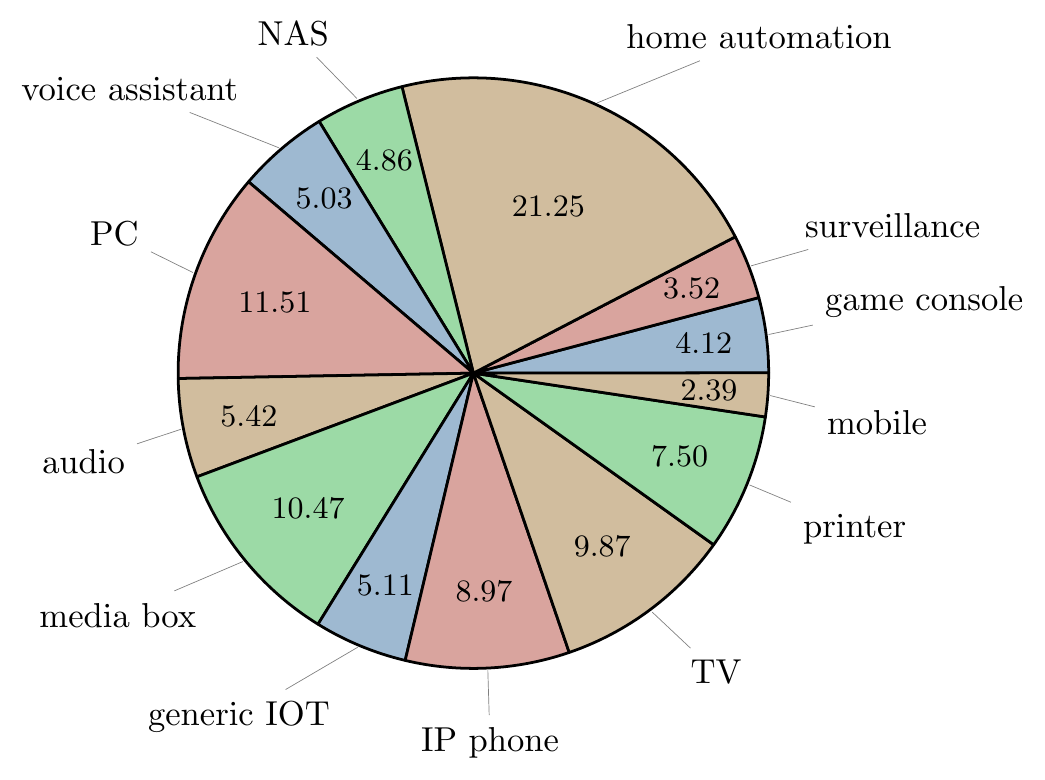}
    \caption[Class label distribution in the \emph{IoT device identification} task.]{Distribution of class labels in the training set of the dataset obtained from Kaggle. All ratios are stated in percents.}%
    \label{fig:class_distr_deviceid}
\end{wrapfigure}

We worked with two specific datasets. The first dataset is publicly accessible and comes from the Kaggle challenge\footnote{\url{https://www.kaggle.com/c/cybersecprague2019-challenge/overview}}. It contains $57\,906$ documents for training and $77\,777$ documents for testing grouped into $ \lvert \mathcal{C} \rvert = 13$  different classes. The schema of the training data as obtained from \texttt{JsonGrinder.jl} can be found in Figure~\ref{fig:deviceid_schema} in Appendix~\ref{ap:deviceid}. All fields except of \texttt{device\_class}, which contains the label, and \texttt{device\_id}, which is a unique identifier of the sample, can be used as input to the model. Describing the data in full detail is beyond the scope of this thesis, and we refer interested readers to the Kaggle competition description or the cited paper. A pie chart of the class distribution in the training data is in Figure~\ref{fig:class_distr_deviceid}. The second dataset containing $8\,816$ documents for training, $99 719$ documents for testing, and $ 13 $ classes was kindly provided by Avast. There is more information in its documents, such as \gls{tls} cipher suites list or \gls{http} user agents, therefore the schema of this dataset is slightly different.

Similarly to other problems tackled in the computer security domain, identifying a device type from \gls{json} documents of this schema is trivial for humans most of the time, however, designing a system for automatic recognition is a challenging task. Device labels can often be simply inferred from information in a single string field in a \gls{json} document, for instance, manufacturers often inform about the model name and device specifications in responses to the aforementioned probes. Nevertheless, this is not always the case and for some devices, their type can be only deduced by inspection of their overall communication interface and behavior. Moreover, the domains of information fragments are typically discrete (strings, categorical variables, or natural numbers) and many fields may contain an empty string or a key-value pair may not be specified at all. This is due to protocols being unsupported on the device side, timeouts or failures in the network or other reasons. Finally, most of the information from \gls{json} document is usually redundant and not required to make the correct decision.

\subsection{Experimental results}%
\label{sub:device_id_experiment}

Learning to classify \gls{iot} device type from \gls{json}s is rather straightforward using the process described in Section~\ref{sec:sec_hmill_jsons}. We inferred a schema of the available documents and defined a hierarchical \gls{hmill} model. Mapping $ h $ in each array data node, where fragment space  $ \mathcal{F} $ contained strings of characters, was defined to compute histograms of tri-grams, in cases, where there was only a handful of unique string values (for instance ports, protocols, or manufacturers), we interpreted the field as a categorical variable and used one-hot encoding, and in the remaining cases (integers, reals), $ h $ was set to identity. To promote effortless processing of sample trees following an arbitrary schema, \texttt{Mill.jl} provides an auxiliary procedure that helps with defining complex composite models. As arguments, it accepts a schema and a prescription specifying topology of neural network components in \gls{hmill} models (mapping $ f $ in array model nodes, $ f_B $ in bag model nodes, and $ f $ in product model nodes) and types of aggregation functions $ g $. For simplicity, we opted for all neural networks components to consist of one feedforward layer with output dimension $ 50 $, dropout regularization~\cite{dropout} with probability $ 0.5 $ and the $ \tanh $ activation function. In every bag model node, we used a concatenation of all four functions discussed in~\ref{sub:aggregation_functions}. Note that specifying output dimensions for mappings $ h $ in array data nodes as well as for neural network components is enough to specify the whole model tree, since the input dimension of every component can be deduced. The resulting model therefore mapped every \gls{json} document to a vector with dimension $ 50 $, which was followed by one final layer to obtain a vector of size $ \lvert \mathcal{C} \rvert  $ and $\softmax$ transformation to obtain class probabilities conditioned on the input.

\newpage\noindent
In each experiment, the model was trained for $ 30 $ epochs (iterations over the whole training set) using minibatches of size $ 100 $ and  \emph{Adam}\footnote{with the default parameters $ \alpha = 0.001 $,  $ \beta_1 = 0.9 $,  $ \beta_2 = 0.999 $ and  $ \epsilon = 10^{-8} $} optimizer~\cite{kingma2014adam}. For each neural network layer, the \emph{Glorot} normal initialization~\cite{glorot} was employed for weights, and all biases were initialized to zero. For a loss function, we used the standard multiclass cross entropy (also known as multinomial logistic loss) on the conditional probabilities output by the model:
\begin{equation}
    \label{eq:crs_entropy}
    \mathcal{L}(\theta) = - \frac{1}{n} \sum_{i=1}^{n} \sum_{c \in \mathcal{C}} \mathbbm{1}\left\{ c = c_i\right\} \log p(c | d_i; \theta)
\end{equation}
where $n$ is the number of samples (documents) in the batch, $ d_i $ a representation of  $ i $-th document as an \gls{hmill} sample,  $ c_i $ its class label, and  $ p(c | d_i ; \theta) $ the conditional probability of  $ i $-th document having label  $ c $ according to the model parametrized by $ \theta $.

After the training was finished, we measured the accuracy on the training set as well as averaged F1 score over all classes. Measured over $ 10 $ runs with a different random seed, we reached accuracy $ 0.9709 \pm 0.003$\ \footnote{$ \mu \pm \sigma $ notation specifies the average and the standard deviation over all runs.} and F1 score $ 0.9582 \pm 0.007$ for the Kaggle dataset, which is on par with the winning entry. Metrics measured on the second dataset are in Table~\ref{tab:device_id}. We compared the results to the system described in~\cite{Kumar2019AllTC}, which is deployed in Avast production. It consists of more than a thousand rules written by experts and a machine learning-based predictor, each of which returns a class prediction together with a confidence score. The overall prediction is then derived by combining predictions of individual components. The \gls{ml}-based part is implemented as an ensemble of classifiers as described in the paper. This ensemble was recently replaced by a single model---one-vs-rest multiclass adaptation of soft-margin \gls{svm} classifier~\cite{vapnik_svm}, with \gls{rbf} kernel ($ \gamma = 1/(2\sigma^2) = 10^{-4} $) and regularization constant $ C = 10^5 $. It uses a manually designed set of features extracted from \gls{json} document, mainly from \gls{upnp} and \gls{mdns}. In the table, we present the accuracy of the whole system, that is, expert rules together with an \gls{ml} classifier, for both ensemble and \gls{svm}-based implementation.

\begin{table}[htb]
    \caption[Results from the \emph{IoT device identification} experiment.]{Metrics measured on the testing set from the dataset obtained from Avast.}%
    \label{tab:device_id}
    \centering
    {\renewcommand\arraystretch{1.25}
        \begin{tabular}{ccc}
            \toprule
            method & accuracy & F1 score \\
            \midrule
            ensemble & $ 0.878 $ & - \\
            \gls{svm} & $ 0.925 $ & - \\
            \gls{hmill} & $ \mathbf{0.969 \pm 0.004} $ & $ \mathbf{0.924 \pm 0.006} $ \\
            \bottomrule
        \end{tabular}
    }
\end{table}

\noindent
It needs to be emphasized that due to the fact that datasets represent merely a fraction of devices in the world, the results presented here are likely slightly skewed in comparison to the performance measured in the production environment. Furthermore, accuracy may not be the most suitable metric if there are different costs of misclassification for labels, or if we are most interested in a subset of high-confidence predictions and do not want to measure accuracy on `hard' samples.
The fact that simple baseline \gls{hmill} models trained on raw data without any fine-tuning achieve competitive performance on both datasets shows great promise. We assume that by further experimenting with model components (for example adding layers to submodels in model nodes), even better results are attainable. This, however, is not the goal of this thesis in which we aim to demonstrate the versatility of the framework on different problems.


\ifprint\newpage\blankpage\fi%
\chapter{HMill for graph inference}%
\label{cha:hmill_on_graphical_data}

The previous chapter showed how the \gls{hmill} framework can be used to learn from tree-structured data together with one particular use case concerning solving the device identification problem. Yet, one of the main requirements outlined in Chapter~\ref{cha:difficulties}---dealing with the violated \gls{iid} assumption---remains unaddressed. As it was discussed before, independence between individual samples is in some tasks rather unrealistic assumption, especially in the computer security domain. Two remaining use cases of the framework in this thesis are prime examples of this. The first one concerns malicious binary file classification at the end-user's machine from a `snapshot' of the state of the operating system. In the last use case, we explore possibilities for second-level domain classification based on its behavior in the network. Even though both tasks are essentially different problems with different data inputs, they have one property in common---in the system, there are many objects of interest of the same type (binaries in the former case and domains in the latter), all of which contribute to the developments in the surrounding system.

To give a specific example, in the operating systems many binary files are being run at once, some of them multiple times in different processes. Since processes may further install other binary files and spawn or terminate other processes, the whole operating system is interconnected in a complex way. To infer the maliciousness of a binary, we should take into account the maliciousness of other binaries or processes in the system it interacted with. Nevertheless, the maliciousness of other objects is unknown as well and deducing it corresponds to essentially the same task. Thus, the \gls{iid} assumption does not hold. In this chapter, we first briefly revise statistical frameworks and methodologies for representing such complex real-world systems and then demonstrate how \gls{hmill} is used to perform inference in these systems. The final part of this chapter presents experiment results from the task of binary file classification based on inference in a graph representation of the operating system.

\section{Prior art}%
\label{sec:prior_art_on_graphical_models}

In the following text, we denote $ X = \left\{X_1, X_2, \ldots, X_n \right\} $ a system of $ n $ objects we want to model, where each $ X_i $ is regarded as a random variable taking values from some domain $ \mathcal{X} $, which depends on the problem. For instance, if we want to classify objects into several classes, we define all variables to be binary $ X_i \in \left\{0,1 \right\}  $ in the case of binary classification, discrete-valued $ X_i \in \left\{ 1, \ldots, \lvert \mathcal{C} \rvert  \right\} $, where $ \mathcal{C} $ is a set of classes, for multi-class classification, and real-valued $ X_i \in \mathbb{R} $ for regression tasks. Under the independence assumption, it is enough to consider only marginals $ p_i(X_i) $ since the joint probability can be eventually computed using the product rule as $ p(X) = \prod_{i=1}^n p_i(X_i) $. When this assumption does not hold, we have no choice but to model the whole joint probability $ p(X) $, so that all dependences between variables can be captured. It is often not the case that all random variables in $X$ are mutually dependent. This happens when $X$ consists of several independent subsets of variables, or when there are pairs of variables that are conditionally independent\footnote{Random variables $ X $ and $ Y $ are said to be conditionally independent given a set of variables $Z$ if $ p(X | Y, Z) = p(X | Z) $, or equivalently, $ p(X, Y | Z) = p(X | Z)p(y|Z)$.} of each other given some (smaller) subset of variables in the model. Consequently, the joint probability distribution $ p(X) $ can be factorized into a product of several terms (usually called \emph{factors}), each computed from a subset of all variables:
\begin{equation}
    \label{eq:factorization}
    p(X) = \prod_{S \subseteq X} f(S)
\end{equation}
This is a generalization of the case when \gls{iid} assumption holds, as we can have one factor for each of the variables equal to its marginal probability $ p_i(X_i) $.

\subsection{Graphical models}%
\label{sub:markov_random_fields}

\emph{Graphical models}, or \emph{Structured probabilistic models}, is an umbrella term for all models, where the possible factorization is encoded into a graph, whose set of vertices are random variables and an edge between two random variables indicates dependence. Therefore, the more (conditionally) independent random variables are, the more efficient their representation as a graphical model is, compared to a full description of probabilities in a table. According to the exact definition of factors $ f $ and the type of the graph, graphical models can be further split into directed graphical models, referred to as \emph{Bayesian networks}~\cite{pearl2013probabilistic}, and undirected models, known as \emph{Markov random fields}~\cite{MRF, Jordan_graphical}.

\begin{figure}[hbtp]
    \begin{subfigure}[t]{.5\textwidth}
        \centering
        \includegraphics[width=0.7\textwidth]{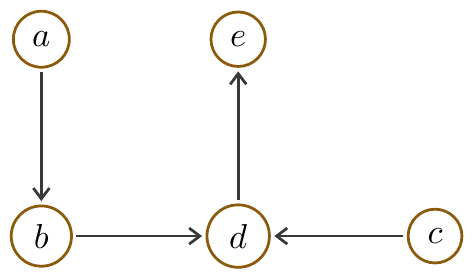}
    \caption{\label{sf:bn}}%
    \end{subfigure}
    \begin{subfigure}[t]{.5\textwidth}
        \centering
        \includegraphics[width=0.7\textwidth]{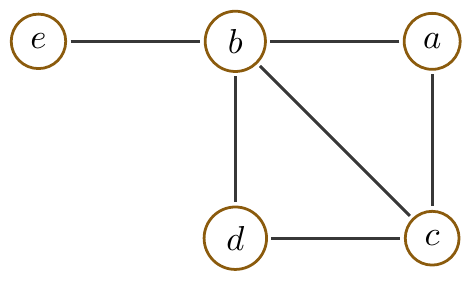}
        \caption{\label{sf:mrf}}%
    \end{subfigure}
    \caption[Examples of a Bayesian network and a Markov random field.]{Example of a Bayesian network and a Markov random field representing a system of five variables. In (\ref{sf:bn}), there is a Bayesian network representing a probability distribution factorizable as $p(a,b,c,d,e) = p(a)p(b\given a)p(c)p(d\given b, c)p(e \given d)$. In Bayesian networks, edges signify direct conditioning, that is, $b$ is dependent on $\left\{ a \right\} $, $d$ is dependent on $\left\{ b, c \right\} $, and $e$ is dependent on $\left\{ d \right\} $. Additionally, $e$ is conditionally independent of all other vertices given its parent $d$, and $d$ is conditionally independent of $a$ given $b$ and $c$. A representation of a (different) probability distribution of five variables as a Markov random field is in (\ref{sf:mrf}). This distribution can be factorized as $p(a,b,c,d,e) = \frac{1}{Z} \psi(a, b, c) \psi(b,c,d) \psi(b,e)$, where $\{a, b, c\}$, $\{b, c, d\}$ and $\{b, e\}$ are maximal cliques in the graph. For example, $e$ is conditionally independent of every other node given $b$, and $a$ and $d$ are conditionally independent given their separating subset $\{b, c\}$. }%
    \label{fig:bn_mrf}
\end{figure}

As opposed to Bayesian networks, where factors $ f $ are given by conditional distributions and reflected in edges of a directed graph as depicted in Figure~\ref{fig:bn_mrf}, an analogical procedure for undirected graphs is not that straightforward. Intuitively, two random variables not connected by an edge should be conditionally independent of each other given any separating subset.\footnote{The separating subset between two vertices $x$ and $y$ in the graph is a minimal set of other vertices such that $x$ becomes unreachable from $y$ after the removal of the separating subset from the graph.} The corollary is that a random variable is conditionally independent of any other variable in the graph, given all variables in its neighborhood. This idea is theoretically justified in the Hammersley-Clifford~\cite{MRFproof} theorem, which states that every strictly positive probability distribution can be factorized into a product of \emph{potential functions} defined over maximal cliques\footnote{Maximal clique is a maximal complete subgraph (adding another node from graph breaks the completeness property).} in the graph and vice versa, every function represented as this product describes a valid probability distribution. Therefore, a factorization of the probability distribution in a Markov random field is proportional to the product of non-negative potential functions $\psi_{C}$ defined on maximal cliques of the graph:
\begin{equation}
    p(X) \propto \prod_{C} \psi_{C}(X_{C})
    \label{nonnormalized factorization}
\end{equation}
where $X_{C}$ is the set of variables in clique $C$. Potential functions do not have to integrate to one and therefore are not proper probability distributions themselves, however, their normalized product corresponds to a valid joint distribution:
\begin{equation}
    p(X) = \frac{1}{Z}\prod_{C} \psi_{C}(X_{C})
    \label{normalized factorization}
\end{equation}
where $ Z = \int_{\mathcal{X}} \prod_{C} \psi_{C}(X_{C}) \mathrm{d}X $ is a normalizing constant, also referred to as a \emph{partition function}.

Assume we are given a joint probability distribution $ p(X) $ represented as a graphical model and a set of observed variables with known values $X_{O} = \left\{ x_1, \ldots, x_l \right\}  \subseteq X $. Denoting the remaining variables to be inferred $ Y = X \setminus X_{O} = \left\{ Y_1, \ldots, Y_k \right\}  $, one of the relevant tasks is to find the most probable state $ Y^* = \argmax_{Y} p(Y | X_O) = \argmax_{Y} p(Y, X_O)$, that is, to perform \gls{map} inference. Another problem is to compute marginals $ p(Y_i | X) $ for each $ i = 1 ,\ldots, k $. For a specific loss function $ \mathcal{L}\colon \mathcal{X} \times \mathcal{X} \to \mathbb{R}$, marginals can be then used to obtain the optimal Bayes classifier for unobserved variables $ Y $. Note that $ X_{O} $ can be empty. For graphs without loops, the canonical \emph{Belief propagation}~\cite{Pearl1988} algorithm exists. Unfortunately, exact computation of either of these probabilities and even evaluation of constant $ Z $ is a computationally intractable problem in the general case when graphs may contain loops. In this case, approximation methods such as \emph{graph cuts}~\cite{graph-cuts} or \emph{Monte Carlo Markov chains}~\cite{Wang2000a} can be utilized. A prominent family of approximation algorithms for graphical models is the family of so-called \emph{message-passing} algorithms containing the standard \emph{Loopy belief propagation}~\cite{LBP, Sum-product} algorithm  also known as the \emph{sum-product} algorithm. In message-passing inference, messages representing local statistical information are iteratively exchanged between vertices, which update their beliefs accordingly, until some convergence criterion is met. The key idea is that even though messages are computed from a small neighborhood of each vertex, after a sufficient number of iterations, information is spread across the entire graph and with appropriate message definitions, marginals can be computed in each vertex separately.

There are several drawbacks in practical modelling and inference with graphical models. Namely, one needs to suitably construct the graph and define potential functions, and algorithms for approximate inference usually lack theoretical guarantees. In~\cite{MPILearning1, MPILearning2, MPILearning3} authors attempted to learn the definition of potential functions, however, this required an introduction of another assumption, for instance, fixed number of neighbors of each vertex. Finally, it has been shown in~\cite{Learning-Message-Passing-Inference-Machines-for-Structured-Prediction} that Loopy belief propagation and its variants can be regarded as special cases of a general iterative scheme in which a sequence of interdependent local classifications is performed. This suggests that not only potential functions but also messages themselves, which are in Loopy belief propagation computed using explicit formulas, can be learned. The notion of message-passing inference, where information encoded in the graph is repeatedly collected from neighbors, aggregated and further propagated, together with the key insight that we can learn message predictors, are main motivations for performing inference with \gls{hmill} as explained later.

\subsection{Graph neural networks}%
\label{sub:graph_neural_networks}

Since graphs are a natural representation of many real-world phenomena occurring in biology, sociology, computer security, or finance, research efforts have been turned to development of methods for effective learning from such structures~\cite{Bronstein2016}. \gls{gnn}s, first proposed in~\cite{gnnsfirst}, are an extension of classical feedforward and recurrent architectures to graph input. The goal is to learn an embedding vector $ \bm{h}_{\mathcal{G}} $ for the whole graph or a set of embeddings $ \left\{ \bm{h}_{v_1}, \bm{h}_{v_2}, \ldots, \bm{h}_{v_n} \right\} $ for each vertex $ v_i $ of the graph. Depending on the application, this embedding is further processed. Graph Convolutional Networks (GCNs)~\cite{Kipf2016} reflect the structural dependences together with features from vertices by generalizing a convolution operation to the graph domain, leading to the following update rule:
\begin{equation}
    \label{eq:gcn}
    \bm{H}^{(l+1)} = \sigma\left(\bm{\hat{D}}^{-\frac{1}{2}} \bm{\hat{A}} \bm{\hat{D}}^{-\frac{1}{2}} \bm{H}^{(l)} \bm{W}^{(l)}\right)
\end{equation}
where activations $ \bm{H}^{(0)} $ in the first iteration are initialized with a matrix of features $ \bm{X} $, $ \bm{\hat{A}} = \bm{A} + \bm{I}$ is the adjacency matrix of the graph with added loops, $ \bm{\hat{D}} $ is a diagonal vertex degree matrix, $ \bm{W}^{(l)} $  is a matrix of parameters in $ l $-th layer, and $ \sigma $ is a non-linear activation function. By stacking multiple layers of~\eqref{eq:gcn}, information is aggregated from the graph, and hidden states of vertices are gradually refined. GCNs belong to the family of \emph{spectral} approaches, which utilize spectral representation of the graph as proposed in~\cite{Bruna2013}. The disadvantage of spectral methods is that the parameters are learned using a specific graph structure, and the model cannot be applied once the input graph is even slightly perturbed.

\emph{Spatial} approaches leverage spatiality in the graph and define graph convolution based on vertex relations. GraphSAGE~\cite{Hamilton2017} is one of the first representatives of spatial methods, which defines graph convolution as follows:
\begin{equation}
    \label{eq:graph_sage}
    \mathbf{h}_{v}^{t}=\sigma\left(\mathbf{W}^{t} \cdot a\left(\mathbf{h}_{v}^{t-1},\left\{\mathbf{h}_{u}^{t-1}, \forall u \in \mathcal{N}(v)\right\}\right)\right)
\end{equation}
where $ a $ is a permutation-invariant aggregation function, and $ \mathcal{N}(v) $ denotes the neighborhood of vertex $ v $. Note that in spatial approaches the central idea is to iteratively update hidden state $ \mathbf{h}_{v} $ at each node $ v $ based on its neighborhood, which corresponds to the message-passing scheme. Indeed, in~\cite{Gilmer2017}, authors reformulated a handful of existing \gls{gnn} architectures into a common general message-passing-based framework. In~\cite{Yoon2019}, it has been shown how \gls{gnn}s can be utilized to learn the messages from the original Loopy Belief Propagation algorithm and eventually outperform it.

Lately, research efforts have been put into modelling a \emph{heterogeneous network}, a graph with multiple vertex and edge types. As it is the case with vanilla \gls{gnn}s, to deal with this intrinsic property, many different methods were proposed~\cite{Chang2015, Gui2016, Fu2017, Shi2017, Dong2017, Zhang2018, Sun2018, Fan2019, Wang2019}.

Prior art on \gls{gnn}s is vast and specialized approaches have been proposed for a particular task being solved (vertex classification, graph embedding, et cetera), different graph types and sizes, and whether the model should generalize to changes in the graph structure. We merely scratched the surface of the family of \gls{gnn}s and heterogeneous network modelling to provide motivation for the \gls{hmill}-based approach to inference described in the next section. We refer the reader to~\cite{Bronstein2016} for a review of methods operating on non-Euclidean data and to~\cite{Zhang2018a, Zhou2018a, Zhou2018a} for comprehensive reviews of \gls{gnn}s.

\section{Modelling heterogeneous networks with HMill}%
\label{sec:modelling_heterogeneous_networks_with_hmill}

In this section, we demonstrate how \gls{hmill} models can be utilized to perform graph inference. In our examples, we focus mainly on the task of vertex classification, however, the general idea can be easily extended to other tasks. We show that models implemented in bag and product nodes are appropriate for simulating a message-passing procedure, effectively adapting the framework for modelling of not only tree-structured data but also complex interactions between samples encoded in a graph. Ultimately, both vertices and edges can be described not by a fixed vector of features, which is assumed in \gls{gnn}s, but by any of the data types processable by \gls{hmill}. Again, this is consistent with the fact that the main contribution of the \gls{hmill} framework is in its flexibility, which enables modelling of a wide range of data sources.

In the following text, we assume a graph $ \mathcal{G} = \left( \mathcal{V}, \mathcal{E} \right) $ to be a tuple consisting of a finite set of vertices (nodes) $ \mathcal{V} $, and a finite set of edges $ \mathcal{E} \subseteq \binom{\mathcal{V}}{2}$ or $ \mathcal{E} \subseteq \mathcal{V} \times \mathcal{V} $ for undirected and directed graphs, respectively. For a vertex $ v \in \mathcal{V} $, we will denote its neighboring vertices by $ \mathcal{N}(v) = \left\{u \mid \left\{ u, v \right\} \in \mathcal{E} \right\} $ in an undirected graph, and in a directed graph, we will distinguish two types of neighborhoods based on the orientation of edges: $ \mathcal{N}^+(v) = \left\{u \mid (u, v)  \in \mathcal{E} \right\} $ and $ \mathcal{N}^-(v) = \left\{u \mid (v, u ) \in \mathcal{E} \right\} $. Moreover, we will call `$ k $-step neighborhood' the set of all vertices whose (unoriented) distance from $ v $ is less than or equal to $ k $. Furthermore, we can have multiple types of vertices and edges specified in finite sets $ T_{\mathcal{V}} $ and $ T_{\mathcal{E}} $, respectively. Mapping $ \type(v) \in T_{\mathcal{V}}  $ attributes the type to every vertex $ v \in \mathcal{V} $ and, analogically, mapping $ \type(e) \in T_{\mathcal{E}} $ returns the type of each edge $ e \in \mathcal{E} $. Given a vertex $ v $ and the graph it comes from, the task is to classify the vertex into one of the classes from set $ \mathcal{C} $. This task is the most general one---for instance, we may be interested in classifying only vertices of a certain type $ t_v \in T_{\mathcal{V}} $, or given labels of a subset of vertices in the graph inferring the labels of the rest, or a combination of both. Examples are provided later.

As opposed to Euclidean domains, graph domains pose new challenges in data transformation. For example, generalizing a convolution operation for graphs is not straightforward. In an image, the neighborhood of each pixel contains a constant number of surrounding pixels, which can be ordered by their relative position to the given pixel, whereas in graphs, number of neighboring vertices of a vertex can be an arbitrary number, and neighbors are not ordered. Therefore, formulas for message computation in a sound inference procedure should accept any number of arguments and should be invariant to the permutation of the arguments.

The key idea of \gls{hmill}-based message computation is that neighborhood $ \mathcal{N}(v)$ of a node $ v \in \mathcal{V}$ can be considered a bag and each of vertices neighboring $ v $ can be regarded as an instance of this bag. A message predictor can then be implemented using \gls{mil} paradigm, which guarantees that all requirements for a message computation are fulfilled since a bag in a \gls{mil} problem is defined to be an unordered, arbitrarily large set of instances. Thus, all existing knowledge about \gls{mil} models can be applied. Formally, for a node $ v \in \mathcal{V} $, we construct an \gls{hmill} bag data node  $ \bdnode(\mathcal{N}(v)) $ and define a message predictor using the framework. As we shall demonstrate in the next paragraphs, using the \gls{hmill} framework for graph inference should be regarded not as a specific method, but rather a modelling toolkit for diverse types of input graphs, which due to its flexibility opens a new realm of modelling choices.

\subsection{Unrolling the procedure}%
\label{sub:unrolling_the_procedure}

The standard message-passing procedure in \gls{gnn}s goes as follows:

\begin{enumerate}
    \item Hidden states $\mathbf{h}_{v}^{0}$ are initialized for each vertex $ v $ in the graph
    \item In each iteration $ t $, hidden states $\mathbf{h}_{v}^{t}$ of each vertex $ v $ are updated using a function taking hidden states of vertices in $ \mathcal{N}(v)$ from iteration $ t-1 $ as input.
    \item After performing $ k $ iterations, hidden states $\mathbf{h}_{v}^{k}$ are processed in a `readout' phase. This usually involves classification of vertices based on their states, or somehow aggregating hidden states of all vertices in the graph to produce a representation of the whole graph.
\end{enumerate}

\noindent
Each iteration is computed synchronously in the sense that hidden states of all vertices are updated at once. The main advantage of this approach is that intermediate results can be reused and, therefore, no redundant computation is performed. Rather than this graph-centric point of view, we advocate a different, node-centric perspective in this work. We assume that a single vertex is one sample (observation) in machine learning terms, and the graph is only queried for information characterizing the vertex. Let us emphasize that `unrolling' a computation on a single vertex in this way does not contradict the dependence claim from the beginning of this chapter. We will still be able to infer knowledge (for instance a class label) about all vertices in the graph taking into account their interactions with other vertices. The aforementioned perspective only demonstrates that inference on different vertices in the graph can be considered a completely independent procedure and computed as such at the price of limited reuse of intermediate results.

\begin{figure}[hbtp]
    \centering
    \includegraphics[width=\linewidth]{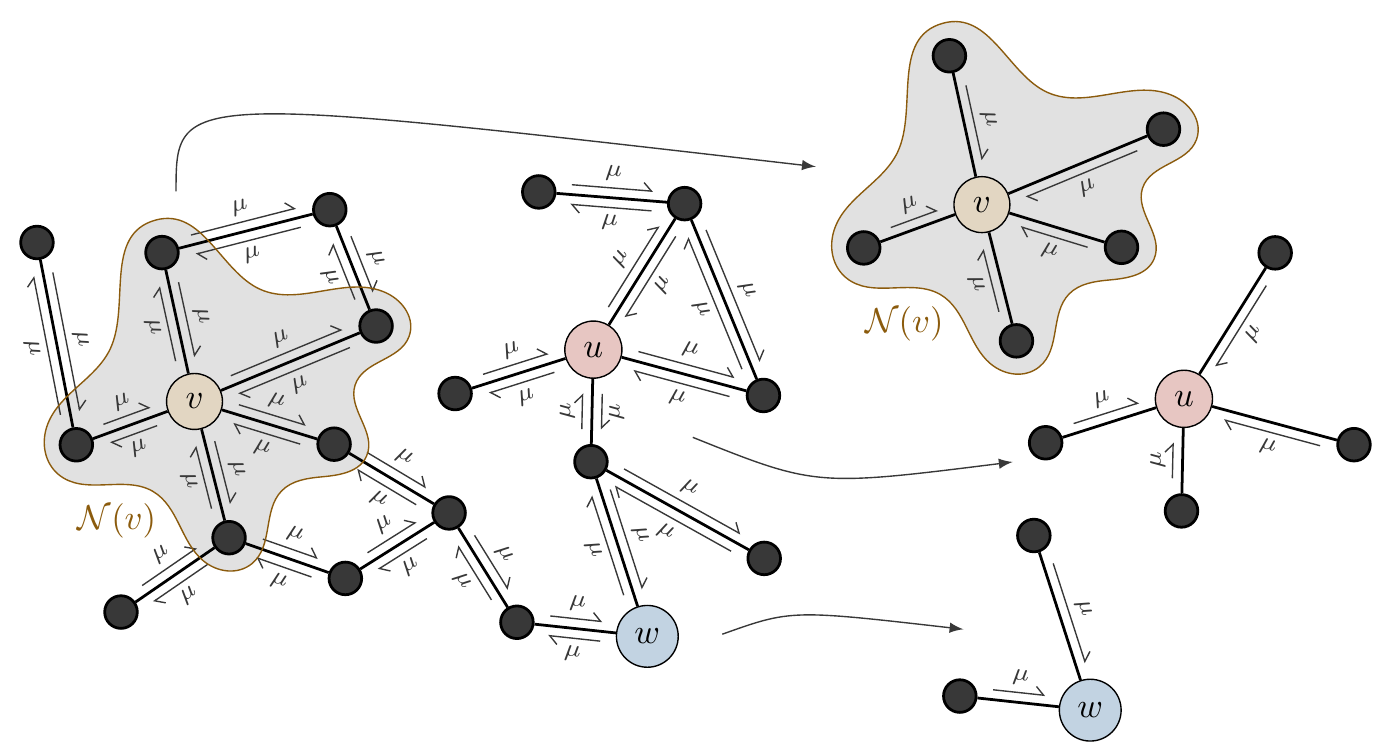}
    \caption[Unrolling message-passing inference.]{A depiction of how a message-passing-based inference can be unrolled. Arrows $\tikz{\draw[cvutaccented] (0,0.1) edge[-{Straight Barb[left]}] (0.5, 0.1); \node () at (0.25, 0.25) {\color{cvutaccented}\scriptsize$\mu$};}$ signify the flow of information (messages). Three random vertices $ u $,  $ v $ and  $ w $ from the graph on the left are highlighted, and the $ 1 $-step neighborhood of vertex $ v $ is shown as well. The equivalent and independent procedures for the hidden state computation are illustrated on the right.  See text for details.}%
    \label{fig:unrolling}
\end{figure}

See Figure~\ref{fig:unrolling} for a specific example, where for simplicity, we show only one step of message-passing inference. On the left, there is an illustration of how one step of message-passing inference is carried out in an undirected graph. Messages, which we universally denote by $ \mu $, are sent in both directions across each edge. This way, we can compute a new hidden state in each vertex in the graph by aggregating information from $ 1 $-step neighborhood. Equivalently, the new hidden states can be computed for each vertex `independently' from smaller subgraphs of the original graph, which is illustrated for three specific vertices on the right.

Assume a vertex $ v $, which we will also refer to as the `current vertex', for which we want to run the inference procedure using the graph as a source of information. This information may include features defined in the vertex itself or in its incident edges, or similar knowledge derived from other vertices in the graph, which are locally connected to the node, directly with an edge or longer path. In this manner, we can consider a `hidden state' $ \mathbf{h}_u^t  $ of a node $ u $ located close to  $ v $ in the graph, as a `feature'. The fact that these types of features derived from the vertex or its neighborhood are inherently heterogeneous, differently structured, and some of them recursively computed, directly motivates the use of \gls{hmill} for their processing. In the following text, we explain how this can be done, and two more specific use cases are presented afterwards. 

First, assume that $ v  $ and all vertices in its neighborhood  $ \mathcal{N}(v) $ are described as fixed size vectors $ \mathbf{h} $. We construct a bag data node $\bdnode(\left\{\adnode(\mathbf{h}_u)  \mid u \in \mathcal{N}(v)\right\} )$ containing as many instances as is the degree of $ v $. Due to the properties of bags in \gls{mil} paradigm, we can obtain such bag data node for every node in the graph and all of them will follow the same schema. Thus, we can define a single model for processing all of them. In the \gls{hmill} framework, bag model nodes consist of an instance model $ f_I $, an aggregation function $ g $ and a bag model $ f_B $. In this case, $ f_I $ would be responsible for the initial transformation of vector representation of neighboring vertices, $ g $ would output a single vector, and  $ f_B $ would perform the final transformation.

The first modelling choice is how to include representation of vertex $ v $ itself to the computation. We elaborate on this later. The described \gls{hmill}-based inference procedure using $ v $ and its $ 1 $-step neighborhood  $ \mathcal{N}(v) $ is equivalent to performing one step in the message-passing scheme. Recall that an instance model $ f_I $ can be any valid \gls{hmill} model. By a recursive argument, if we define $ f_I $ to be itself a bag model node with the same architecture as in the $ 1 $-step case, the overall model performs two-step message-passing procedure. Since every \gls{hmill} model returns a vector, $ f_I $ would also return a vector for each neighbor (instance) and the rest of the computation in $ g $ and  $ f_B $ remains the same. Therefore, with \gls{hmill} we can implement a message-passing procedure with an arbitrary number of steps. The depth of the model and resulting tree sample representations will be $ k+1 $, where $ k $ is the number of steps.

\subsection{Modelling vertices}%
\label{sub:modelling_vertices}

Vertex $ v $ itself and other nodes in the graph may be described by a vector of features or raw data encoded in different data types, or a combination of both. One of the main advantages of using \gls{hmill}-based inference is that a node description can be provided as an arbitrarily complex structure, for instance, a \gls{json} document. The inference procedure with a complex description of vertices can be split broadly into two abstract parts---in the first one, important knowledge is distilled from each node, and in the second part, it is distributed across the graph and combined with data in other vertices. We have already demonstrated how hierarchical structures are modelled with \gls{hmill} in Chapter~\ref{cha:jsons}. Thanks to the fact that message-passing inference itself can be implemented with \gls{hmill} models as well, the whole computation seamlessly integrates into the framework. With the reasonable assumption that node representations all match the same schema, the whole inference procedure for one node can be implemented as a single \gls{hmill} model, which can be trained in an end-to-end manner. Encoding of the $ \type(v) $ into the node's representation is also possible either trivially as a one-hot encoded vector, or as some more complex encoding.

We have shown that information from the neighborhood of vertex $ v $ can be aggregated with a bag model node to compute a message, however, how is a representation of the vertex itself added to computation is yet to be specified. This is done with an \gls{hmill} product node, whose children are one bag node describing the neighborhood and one other data node for vertex representation, whose type depends on the data type of $ v $'s description---array node if $ v $ is described by only a vector of features, or more complex data nodes in other cases. In other perspective, knowledge stored in the neighborhood of $ v $ can be regarded as just another deep subtree in product node representation of $ v $. In \gls{json} parlance, this is equivalent to adding another key-value pair into the \gls{json} object that describes $ v $.

\begin{figure}[hbtp]
    \centering
    \begin{subfigure}[b]{0.49\textwidth}
        \centering
        \includegraphics[width=\textwidth]{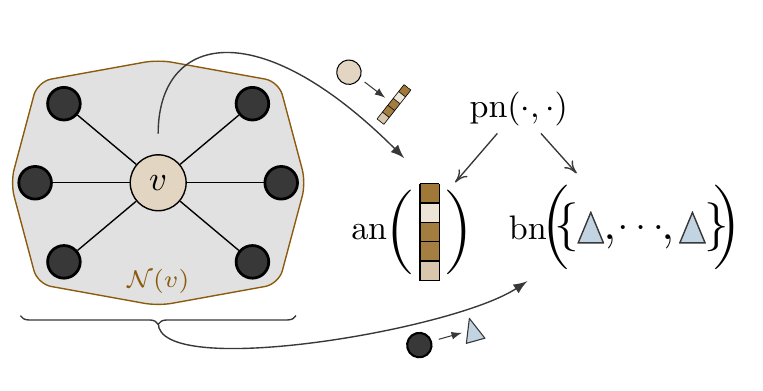}
        \caption{\label{sf:ssf1}feature vector}%
    \end{subfigure}
    \hfill
    \begin{subfigure}[b]{0.49\textwidth}  
        \centering
        \includegraphics[width=0.925\textwidth]{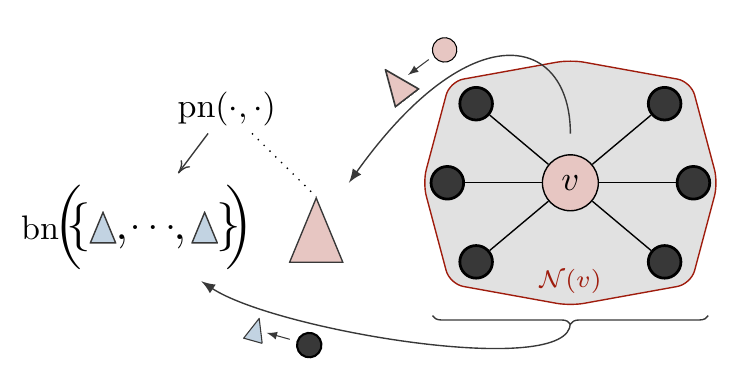}
        \caption{\label{sf:ssf2}\gls{hmill} subtree}%
    \end{subfigure}
    \caption[Modelling nodes in graph inference.]{An illustration of how information about current node $ v $, which receives the message, is added to its representation using a product data node. The information (features) describing $ v $ can be in the form of a feature vector (\subref{sf:ssf1}) or a more complex \gls{hmill} representation (\subref{sf:ssf2}). The fact that a vertex in the graph is described by an \gls{hmill} compatible structure is illustrated with an arrow \tikz[baseline]{\draw[genericarrow] (0,0.1) -> (0.3, 0.1);}. The types of arrowheads used for different purposes is consistent with the rest of the thesis---for graph edges, we use \tikz[baseline]{\draw[grapharrow] (0,0.1) -> (0.3, 0.1);}, for illustrating parent-children relation in sample and model trees \tikz[baseline]{\draw[modelarrow] (0,0.1) -> (0.3, 0.1);} or \tikz[baseline]{\draw[ellipsearrow] (0,0.1) -- (0.3, 0.1);}, and \tikz[baseline]{\draw[genericarrow] (0,0.1) -> (0.3, 0.1);} are used as generic pointers. See text for details.}%
    \label{fig:node_repr}
\end{figure}

Refer to Figure~\ref{fig:node_repr} for a depiction of this process. Node $ v $, together with its $ 1 $-step neighborhood of six vertices, is drawn on the far left and right sides. In both cases, the neighborhood of $ v $ is modelled by a bag node with six instances in this particular case, each of which is described further by an arbitrarily complex \gls{hmill} sample tree, depicted by a blue subtree. This subtree can describe not only the neighboring node itself (for instance its features) but its neighborhood as well. In the left picture, $ v $ is described by a feature vector, whereas in the right one, we used a red subtree signifying that in fact any valid \gls{hmill} (sub)tree can be used to describe $ v $.

Finally, \emph{direct feedback} is a phenomenon that occurs when an output of a vertex is influenced by the vertex itself due to information being propagated in several iterations. For instance, consider an edge $ \left\{ v, u \right\} \in \mathcal{E} $. If we want to infer a label for $ v $, according to the inference procedure we first evaluate results of $ f_I $ on all vertices in  $ \mathcal{N}(v) $ and then aggregate them. However, if more steps are performed and therefore an intermediate result on vertex $ u $ is computed from its neighborhood as well, it holds that $ v \in \mathcal{N}(u) $, and output from vertex $ v $ will be influenced twice by its representation. Depending on the domain, this may be desirable or not. In \gls{hmill}-based inference, this can be easily remedied by redefining a bag representing $ u $ to contain vertices $ \mathcal{N}(u) \setminus v$.

\subsection{Modelling edges}%
\label{sub:modelling_edges}

In some applications, most useful information is stored not in vertices themselves, but in relationships between them signified by edges. Analogically to vertices, each edge can be represented by a more or less complex \gls{hmill} structure. Instead of regarding each neighbor $ u \in \mathcal{N}(v)$ as an instance, we can consider the relationship $ \left\{ u, v \right\} $ an instance and include information about the relationship $ \left\{ u, v \right\} $ instead of about $ u $ only. If both $ u $ and  $ \left\{ u, v \right\}  $ are important, bag instances can be defined as a tuple $ \left( u, \left\{ u, v \right\}  \right) $. The most straightforward way to represent such instance is by a product data node with two subtrees, one for $ u $  and one for $ \left\{ u, v \right\}  $.

Graph edges of different types, or directed edges, which are just a special case of the former, or both, can be modelled in different ways. The more naïve option is to use one-hot encoding on the type of an edge and add this to the representation of $ \left\{ u, v \right\}  $. Another way is to model each type of edges separately. This corresponds to constructing $ \lvert T_{\mathcal{E}} \rvert $ bags ($2 \lvert T_{\mathcal{E}} \rvert $ if graph edges are both typed and directed), one for each type of edges. For each edge type $ \tau \in T_{\mathcal{E}} $, we will include only neighbors connected with the edge of type $ \tau $ into the bag. In other words, for each $ \tau \in T_{\mathcal{E}} $, we define one bag $ b_{\tau} = \left\{u \mid u \in \mathcal{N}(v), \type(\left\{ u, v \right\} ) = \tau \right\} $. Which way is more suitable again depends on application and specific examples are provided later.

\begin{figure}[hbtp]
    \centering
    \includegraphics[width=0.7\linewidth]{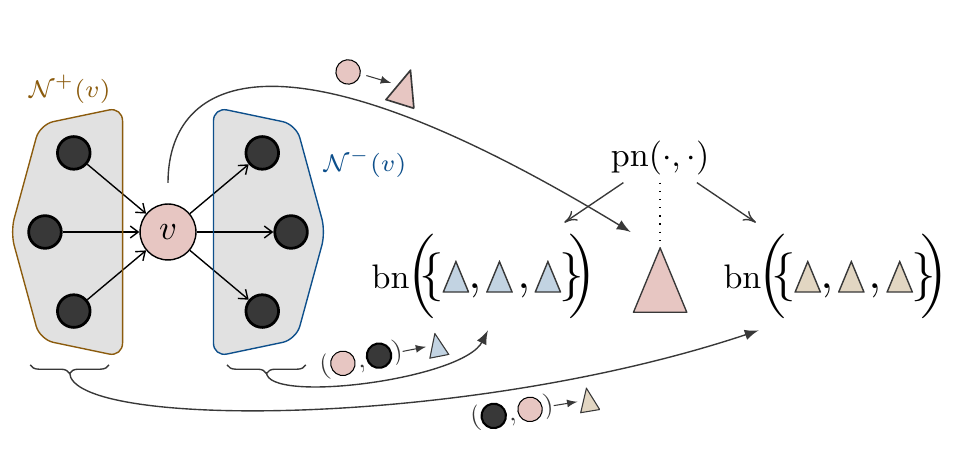}
    \caption[Modelling edges in graph inference.]{One possible way of dealing with multiple edge types. In this particular case, edge types correspond to incoming and outcoming edges with respect to current node $ v $. See text for details.}%
    \label{fig:edge_repr}
\end{figure}

In Figure~\ref{fig:edge_repr}, we sketch the latter option. The situation is similar to one depicted in Figures~\ref{sf:ssf1} and~\ref{sf:ssf2}, however, in this case the graph being modelled is oriented. This means that we deal with two types of edges with a possibly different interpretation. Thus, we use two different bag nodes, one for incoming and one for outcoming edges. Recall, this implies that each incoming edge will be modelled with the same (sub)model tree and the same applies to outcoming edges. However, the parameters will not be shared between incoming and outcoming edges. In the picture, we use golden and blue subtrees to illustrate a representation of incoming and outcoming relations in a general way, nevertheless, both representations can be defined as equal. Note, that these subtrees can be used to describe not only the neighboring node $ u $ but also the relationship $ \left( v, u \right)  $ or $ \left( u, v \right)  $, which is depicted schematically in the figure.

\subsection{Training the model}%
\label{sub:training_the_model}

\begin{figure}[hbtp]
    \centering
    \begin{subfigure}[b]{0.25\textwidth}
        \centering
        \raisebox{1cm}{\includegraphics[width=\textwidth]{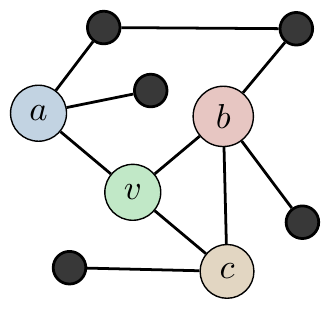}}
        \caption{\label{sf:os}Original subgraph}%
    \end{subfigure}
    \begin{subfigure}[b]{0.3\textwidth}  
        \centering
        \raisebox{.15cm}{\includegraphics[width=\textwidth]{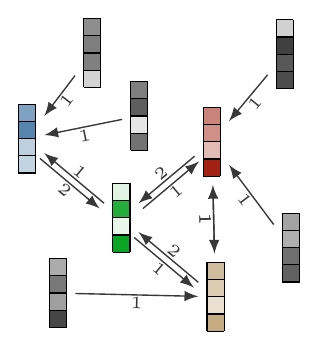}}
        \caption{\label{sf:nr}Node representations}%
    \end{subfigure}
    \begin{subfigure}[b]{0.43\textwidth}   
        \centering
        \includegraphics[width=\textwidth]{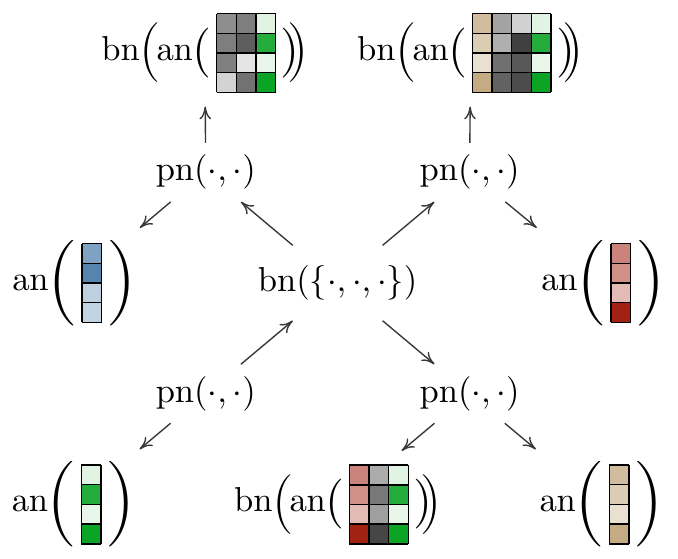}
        \caption{\label{sf:rs}Resulting sample}%
    \end{subfigure}
    \caption[Example of $ 2 $-step graph inference.]{\gls{hmill} sample tree extracted from the neighborhood of vertex $ v $ for performing $ 2 $-step message-passing inference in a graph, where vertices are described by vectors of features. The original subgraph induced by vertices in the $ 2 $-step neighborhood of $ v $ is depicted in (\subref{sf:os}). The picture (\subref{sf:nr}) in the middle  shows  the same subgraph, where feature vectors of dimension $ 4 $ describing vertices are drawn, together with arrows illustrating, how information is transferred in each of two steps. The uppermost horizontal edge is not present in (\subref{sf:nr}), since more than $ 2 $ inferential steps would be required to reach it. The final sample tree, drawn aligned with the original subgraph to facilitate comparison, is in the third figure (\subref{sf:rs}). For bag node representations of neighborhoods of vertices $ a $,  $ b $, and  $ c $, we used the batch version to save space (see Section~\ref{sub:batching}). See text for details.}%
    \label{fig:inference_steps}
\end{figure}

Refer to Figure~\ref{fig:inference_steps} for a specific example of how message-passing inference can be translated into \gls{hmill}. For simplicity, we assume that no information about edges themselves is known, however, each node is described by a vector of features. The figure shows how \gls{hmill} sample for implementing two steps of message-passing inference looks like. First, each of the vertices $ a $,  $ b $, and  $c$ is described by a product node merging feature vector representing the vertex and information obtained from its neighborhood, which is stored in a bag node. Then, we store all three product data nodes into another bag data node, which is merged with feature vector of $ v $. Note that all bag node definitions are valid, as the product data node representation of vertices in $ \mathcal{N}(v) $ follow the same \gls{hmill} schema.

As the whole model is a single \gls{hmill} model, it can be trained in an end-to-end fashion. Thus, processing of the data in vertices and edges is optimized jointly with a message-passing part. If the node-centric perspective on message-passing inference is used, the standard training procedure can be employed. One or more graphs are collected, and in each iteration of the training loop, several vertices are sampled to form a minibatch, the unknown quantity (for instance a label) is inferred, and errors are backpropagated through the model according to Algorithm~\ref{alg:backprop}. Due to the fact that \gls{hmill} offers a principled way of dealing with missing data at different levels of abstraction (Section~\ref{sub:missing_data}), the only requirement on the data is that \gls{hmill} subtrees describing vertices (or edges) must follow the same schema. The framework elegantly deals with other possible discrepancies in the data, such as missing information on vertices or edges or empty neighborhoods.

One concern about working with any method for learning from graphs is limited scalability. The number of vertices in a $k$-step neighborhood of a vertex grows exponentially, and with increasing $ k $ gradient computation quickly becomes infeasible. This is still an open problem~\cite{Wu2019}, however, it can be partially remedied by sampling. One possible solution is to employ a variant of importance sampling, which in \gls{hmill} is easily done by replacing bag data nodes representing neighborhoods with their weighted variants (Section~\ref{sub:weighted_bag_nodes}). For instance, in problems from the computer security domain, positive signs of malicious activity can be sometimes observed sparsely, however, they convey the most useful information. In this case, the sampling procedure can be designed to be biased towards malicious observations. The specific examples are provided later in this chapter and in the next one. In our experience, the performance gain achieved by increasing the number of steps $ k $ in the inference procedure quickly saturates. A pleasant side-effect of sampling from bags (neighborhoods) is an introduction of stochasticity into learning, which is known to be beneficial for optimization with the gradient descent class of algorithms.

\subsection{Differences to GNNs}%
\label{sub:graph_neural_networks_differences}

We have shown how the \gls{hmill} framework can be extended to perform message-passing inference in graphs, using bag nodes to model neighborhoods and product nodes to merge heterogeneous information. The resulting method retains all favorable properties of the message-passing computation scheme while offering several additional benefits. Due to the fact that universal function approximators are used instead of analytical closed-form expressions for message computation, the space of all inferable hypotheses becomes larger. At the same time, the method does not require specifying a family of functions for message computation, avoiding potential approximation error stemming from selecting a wrong family. The \gls{hmill}-based inference, as well as \gls{gnn}s, assumes that a subgraph induced by $ k $-step neighborhood contains all relevant information and that the further a node or edge is located in the graph, the less important it is. The proposed inference method starts at the most distant vertices and proceeds towards $ v $, in each step using bag and product model nodes to aggregate and merge distilled knowledge relevant for inferring the class label or other unknown quantity. Upon closed inspection, it turns out that in comparison to how \gls{gnn}s operate, our method uses a slightly different computational model, parameters are shared at different locations, and different overall philosophy to the graph inference is adopted. Whereas \gls{gnn}s drew most inspiration from recurrent networks with each layer estimating graph convolution operation, which leads to powerful, but hard-to-scale models, the \gls{hmill}-based approach regards the graph merely as a knowledge database for querying information about vertices and their relationships. Hence, both approaches are suitable for different circumstances.

As the main goal of the framework is to unlock new possibilities in the processing of raw data, the scalability and modelling flexibility are two of the main concerns. Specifically, in this thesis, we provide examples of use cases from the computer security domain, where graphs are huge and node degrees come from a wide range of values. In our opinion, bag nodes are the perfect candidate for modelling neighborhoods, which in most graphs observed in real-world scenarios follow a power law distribution, and described bag sampling schemes make inference tractable. Moreover, the alternative node-centric view to inference enables straightforward parallelization as inferring unknown values for vertices can be considered independent. Also, because a single value can be inferred at the time, sampling techniques at the level of batches are possible. For instance, if we want to perform binary classification of vertices in graphs, where the vast majority of vertices are positive, it is still possible to construct balanced minibatches for training. Another benefit is that a problem of missing data at different levels is solved in the \gls{hmill} framework in a principled way and any heterogeneity in the graph, such as different vertex and edge types can easily be modelled with product nodes.

Finally, thanks to the modelling flexibility of \gls{hmill} in general, complex descriptions of vertices and edges can be integrated and the resulting model is still one \gls{hmill} model, which means that the inference procedure together with the processing of representations is optimized jointly. We believe that due to little requirements imposed on the input graphs in terms of size, degree distributions and structure, the proposed approach is versatile enough for utilization in many potential applications.

\section{Behavior-based malware classification in graphs (use case)}%
\label{sec:idp_graphs}

An integral component in any antivirus software is a capability to recognize and protect users from malicious programs. Nowadays, harmful software takes many forms.  \emph{Trojans} allow unauthorized access to computers, \emph{worms} spread themselves over many machines, \emph{ransomware} programs encrypt the whole hard drive of the user and demand payment in exchange for decryption secret, \emph{spyware} records sensitive information like passwords, and finally,  \emph{adware} programs generate revenue by displaying an advertisement on the screen. These examples are just a few of many types of existing malware. Current technologies designed for defense can be generally categorized into two categories.

Traditional \emph{signature-based} mechanisms rely on extracting a predefined `signature' from the inspected file and comparing it to a known database of threats. A signature could be based for instance on MD5 or SHA1 hash of the whole executable or its part in the most trivial case. The advantages of the signature-based approach is low false positive rate, and, due to its simplicity, high detection speed, crucial characteristics of any antivirus software. On the other hand, once a new threat appears, anti-malware solution providers have to generate its signature and distribute it to clients. Oftentimes, this leads to blacklists becoming outdated and users being vulnerable to the evolved malware for a period of time. The Cisco 2017 Annual CyberSecurity Report~\cite{Cisco2017} states that 95\% of all analyzed malware files were not even a day old, which indicates that obsolete signature databases are more the rule than the exception. Moreover, a great deal of resources is required to keep blacklist databases updated all the time. More sophisticated malware makes use of encrypted or polymorphic code segments, which makes signature generation very hard, since many executables with the same behavior but different signatures can be generated. 

\newpage\noindent
These issues can be addressed by designing a rule-based system or a machine-learning system that takes into account the structure of the file or the behavior of the program. In \emph{static analysis}, dangerous attributes are sought after within the structure of the executable, its code or its other properties, whereas in \emph{dynamic analysis}, the executable is run and observed in a sandboxed environment. For this reason, this complementary approach to signature-based is usually referred to as \emph{behavior-based}. Once suspicious behavior is detected, the antivirus may remove some important permissions of the program or stop the execution altogether. Different aspects of the program's behavior can be observed, for example, changing register values, installing rootkits, accessing specific files or services, locking the executable from removal, disabling security controls, registering the executable for autostart, attempting to detect and evade sandbox, and so forth. Behavior-based techniques effectively tackle all of the signature-based approach flaws, as malicious behavior is the defining feature of all malware and is therefore difficult to conceal. Two programs with different signatures but identical behavior are treated in the same way as well. Furthermore, this approach detects new malware families, which might be programmed in a completely different way, but exhibit the same behavior as known malware. The downside of behavior-based methods is that they are usually slower.

In some cases, malware programs can resist detection and removal by spawning two processes that restart each other, should one of them get killed. In other cases, malware activities are exhibited not by a single process, but a group of processes that may not all be instances of the same executable. Malware is usually first installed with a \emph{dropper} program, the malicious action is performed by another program, and both can use additional modules. For instance, Windows hooks play a major role in most of keyboard logger programs, and droppers that install this malware have an extra dynamically linked library or dll file~\cite{patent_removal}. As a result, malware activities can be hard to detect when all the components are examined independently. This motivates the use of a graph structure to encode the overall behavior in the system for the detection.

\subsection{Data description}%
\label{sub:graph_based_detection}

For the experiment presented in this chapter, we once again partnered with Avast researchers to try \gls{hmill}-based graph inference on real-world data from clients. Specifically, we focus on the Windows operating system, which is installed on a bulk of clients. As described in patents~\cite{patent_removal, patent_detection}, the Avast client keeps a regularly updated snapshot of the operating system for the detection purposes, which is encoded as a graphical structure called \gls{idp} graph.

The graph consists of two types of vertices representing executable files and processes, therefore $T_{\mathcal{V}} = [\texttt{EXECUTABLE}, \texttt{PROCESS}]$. Executable vertices correspond to executable files in the system that have been run, and are identified by the full file path. Therefore, multiple copies of the same file will result in multiple vertices in the graph. Process vertices represent all processes which either run in the system or spawned another process that is still running. Executable files and their processes are represented separately, because different processes may behave differently, for example when foreign code is injected into a running process.

Multiple types of graph edges encode events that occurred in the system measured from the Windows kernel and other sensors. Edge types (relationships) are summarized in Table~\ref{tab:relationships}. Note that the table contains only the most relevant relationships published in the aforementioned patents, which is a proper subset of all $ 24 $ types of edges.

\begin{table}[htp]
    \caption[Edge types (relationships) used in IDP graphs.]{Edge types (relationships) $ T_{\mathcal{E}} $ used in \gls{idp} graphs that were published in~\cite{patent_removal}.}%
    \label{tab:relationships}
    \centering
    {\renewcommand\arraystretch{1.25}
        \begin{tabular}{ccp{10cm}}
            \toprule
            name & participants & interpretation \\
            \midrule
            \emph{INSTALLS} & $ p $ \tikz[baseline]{\draw[grapharrow] (0,0.1) -> (0.3, 0.1);} $ e $ & Process $ p $ wrote an executable  $ e $ to disk.\\
            \emph{SPAWNS} & $ p_1 $ \tikz[baseline]{\draw[grapharrow] (0,0.1) -> (0.3, 0.1);} $ p_2 $ & Process $ p_1 $ started another process $ p_2 $.\\
            \emph{INSTANCE OF} & $ e $ \tikz[baseline]{\draw[grapharrow] (0,0.1) -> (0.3, 0.1);} $ p $ & Process $ p $ is an instance of executable  $ e $. \\
            \emph{REGISTERS} & $ p $ \tikz[baseline]{\draw[grapharrow] (0,0.1) -> (0.3, 0.1);} $ e $ & Process $ p $ stored some configuration data about executable $ e $, for instance a registry key causing reboot survival. \\
            \emph{CODE INJECT} & $ p_1 $ \tikz[baseline]{\draw[grapharrow] (0,0.1) -> (0.3, 0.1);} $ p_2 $ & Process $ p_1 $ caused process $ p_2 $ to run foreign code.\\
            \emph{KILLS} & $ p_1 $ \tikz[baseline]{\draw[grapharrow] (0,0.1) -> (0.3, 0.1);} $ p_2 $ & Process $ p_1 $ attempted to kill another process  $ p_2 $. \\
            \bottomrule
        \end{tabular}
    }
\end{table}

Moreover, each vertex in the graph is described by an \gls{xml} document storing additional multimodal information. So far, we have described only how \gls{json} documents are processed with the \gls{hmill} framework. Even though \gls{xml} documents are processed analogically, in this case, documents can be directly converted to \gls{json}s, thus one can think of this as a preprocessing step and then apply exactly the same procedure defined previously for \gls{json} documents. An \gls{hmill} schema of the vertex description inferred from the training data can be found in Figure~\ref{fig:idp_schema} in Appendix~\ref{ap:idp}. It includes a unique $ \texttt{"id"} $, a fully qualified $ \texttt{"name"} $, a $ \texttt{"size"} $ of the file in bytes, a $ \texttt{"sha256"} $ hash, information about \texttt{"connections"} made, \texttt{"named\_objects"} used, or registry operations (\texttt{"regops"}) performed for executables, and finally, a set of `characteristics'. Characteristics are binary predicates measured by means of static and dynamic analysis of the given executable or process designed to discriminate between malicious and clean software. For this experiment, we used a set of $ 164 $ such characteristics. Refer to Table~\ref{tab:characteristics} in Appendix~\ref{ap:idp} for specific examples. Some information is specified only for one type of vertices, for instance, it does not make sense to define a size of a process node. The same holds for characteristics, therefore, ternary logic with values $ \left\{ \texttt{TRUE}, \texttt{FALSE}, \texttt{UNKNOWN}  \right\} $ is used for their specification. 

In Appendix~\ref{ap:idp}, we provide Figure~\ref{fig:whole_idp} sketching how the shape of one weakly connected component of an \gls{idp} graph looks like. Also, in Figures~\ref{fig:idp_g2} and~\ref{fig:idp_g1} we show specific instances of subgraphs, the first one coming from an operating system without infection and the other one from an infected system. For even more details, please refer to the patents cited above.

\subsection{Experimental settings}%
\label{sub:experimental_setting}

For experimental purposes, we obtained a labeled, anonymized dataset consisting of around $ 4\cdot 10^5 $ anonymized graphs collected from Avast clients during February and March 2020. The goal was to classify a subset of approximately $ 7,9\cdot 10^6$ executable nodes present in graphs into a malicious or benign class based on only information encoded in the graph. We did not attempt to classify all executable vertices in graphs since some of them are signed by trusted companies and are therefore considered clean and ignored by the antivirus client.

\begin{figure}[hbtp]
    \begin{subfigure}[t]{.15\textwidth}
        \centering
        \raisebox{1.5cm}{\includegraphics[width=0.75\textwidth]{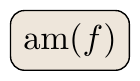}}
        \caption{\label{sf:f1}$ m_1 $}%
        \hfill
    \end{subfigure}
    \begin{subfigure}[t]{.30\textwidth}
        \centering
        \raisebox{1.05cm}{\includegraphics[width=0.87\textwidth]{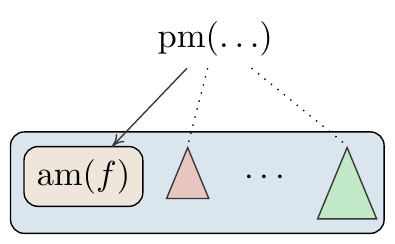}}
    \caption{\label{sf:f2}$ m_2 $}%
\end{subfigure}
\begin{subfigure}[t]{.5\textwidth}
    \centering
    \includegraphics[width=\textwidth]{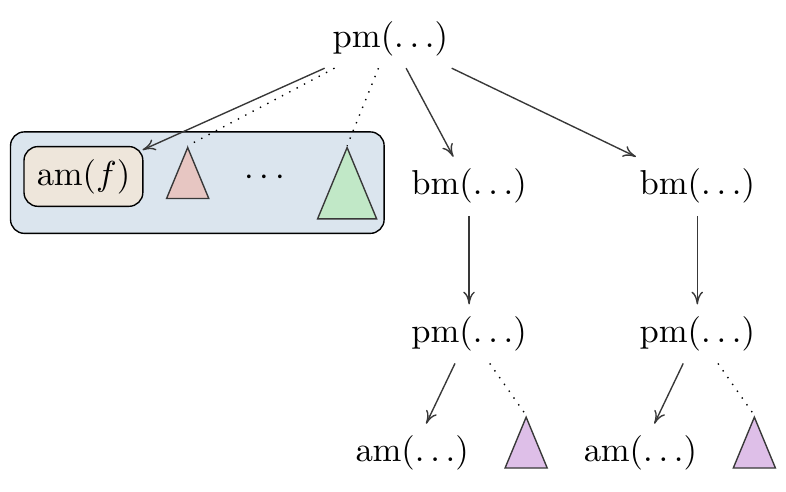}
    \caption{\label{sf:f3}$ m_3 $}%
\end{subfigure}
\caption[Three different models defined for the \emph{IoT device identification} task.]{Three different models used in experiments. Golden and blue rectangles indicate that the same architecture (schema) was used for the model. However, inner neural network components may have different numbers of neurons in the last layer. For example, model $ m_1 $ (\subref{sf:f1}) has $ 2 $ neurons in the last layer of $ f $, one for each class. (Sub)model in the golden subtree in $ m_2 $ does not output $ 2 $ values, as the output is transformed further in the model.}%
\label{fig:idp_models}
\end{figure}

In this experiment, we demonstrate that the main strength of the \gls{hmill} framework lies in its modelling flexibility. We defined three types of increasingly more complex models operating with different amounts of information, all depicted in Figure~\ref{fig:idp_models}. Assume we want to classify an executable vertex $ v $.

The first model (Figure~\ref{sf:f1}) $ m_1 $ processes only characteristics of $ v $ encoded into a vector of length $ 2\cdot 164 $ with one-hot encoding\footnote{We used two `bits' per each characteristic. We set the first bit if the characteristic is \texttt{TRUE}, the second one if it is \texttt{FALSE}, and neither if it is \texttt{UNKNOWN}.}. It is defined as a standard feedforward neural network with two hidden layers of $ 300 $ neurons with $\relu$ activation function and $ 2 $ output neurons, one for each class. A similar model is deployed in Avast pipeline as well. In \gls{hmill} terminology, $ m_1 $ is a single array model node $ \amnode(f) $, where $ f $ is the mapping implemented by the network.

The second model $ m_2 $ (Figure~\ref{sf:f2}), defined as an \gls{hmill} product model node, uses all information available on $ v $, which is summarized in the schema in Figure~\ref{fig:idp_schema}. Thus, it contains a subtree of depth one for modelling characteristics as well, with the same structure as $ m_1 $.\footnote{In fact they differ in a minor detail that $ 300 $ neurons were used in the last layer instead of $ 2 $ as in the case of $ m_1 $, since the output it is used further in the model.} In the figure, this is indicated with a golden frame. Other than that, a product model node $ m_2 $ contains a child for every key of the aforementioned schema, with the exception of \texttt{id}, \texttt{node\_type}\footnote{Modelling node types does not make sense in this context since all modelled vertices are executables.} and \texttt{sha256}. For sizes in the \texttt{size} field, we used mapping $ x \mapsto \log(1+x) $. The procedure for constructing an \gls{hmill} sample tree for $ m_2 $ from an \gls{xml} document is the same as used in the device identification problem from Section~\ref{sub:device_id_experiment}---we used histograms of tri-grams for modelling string values, and one-hot encoding for the representation of categorical variables. An \gls{hmill} model tree was built so that every neural network component in model nodes has the output dimension of $ 50 $ if not specified otherwise, in exactly the same way as in Section~\ref{sub:device_id_experiment}.

Last model $ m_3 $ implements a $ 1 $-step message-passing inference as described in this chapter, therefore, apart from data about $ v $, it also utilizes information stored in $ 1 $-step neighborhood $ \mathcal{N}(v) $ of the vertex in the graph. This was implemented exactly as described in Section~\ref{sub:modelling_edges}. The children (subtrees) of the topmost product model node of $ m_3 $ consist of three sets---one for modelling data stored in the vertex itself, which is the same architecture as in $ m_2 $ (marked with a blue box in Figure~\ref{sf:f3}), and two bag model nodes for modelling incoming and outcoming edges. For each sample, every edge (instance in one of the corresponding bag data nodes) is represented as the one-hot encoded edge type and the description of the neighbor, merged together in a product data node. The neighbor description, in the figure depicted as a purple subtree, was implemented using the same architecture as for the description of $ v $, note however that model parameters are not shared. Consequently, model $ m_3 $ can infer the label not only from the information known about the executable vertex itself but also from how the file (or its process instances) relates to other entities in the whole system. Since both process and executable vertices may be present in $ \mathcal{N}(v) $, we include a one-hot encoded vertex type into all neighbor representations. Again, all neural network component architectures were defined using the simple rule described above for the case of $ m_2 $. We used $ \relu $ non-linearity everywhere, except of last layers of instance models in bag model nodes, where we used the $ \tanh $ activation function since in our experience it performs better with subsequent aggregations. The very last layer in the root of every model tree was followed by a $ \softmax $ layer to obtain class probabilities. 

\newpage\noindent
In some cases, vertex $ v $ has a high number of neighbors. This happens, for example, when a ransomware program spawns a lot of processes to perform data encryption. As most of the time all of these processes have the same attributes, it is enough to inspect a small subset of them. Therefore, to save time, we employed an importance sampling scheme and instance weighting described in Section~\ref{sub:weighted_bag_nodes}. We include into both bag data nodes describing incoming and outcoming edges of vertex $ v $ at maximum $ K = 100 $ neighbors connected with the edge of the same type. This way, all sparsely represented relationships are placed in the bag every time, however neighbors connected to $ v $ with a common edge type are sampled.

A $ 15\% $ proportion of all $ 4\cdot 10^5 $ graphs in the dataset was left out for testing. We counted $ 8\cdot 10^5 $ positive vertices out of $ 6\cdot 10^6 $ vertices in the training graphs and $ 1,5\cdot 10^5 $ positive vertices out of  $ 1.2\cdot 10^6 $ in the graphs held out for testing. All models were trained in the exact same way, performing one epoch (iteration over the whole training set) using minibatches of size $ 100 $ and \emph{Adam} optimizer~\cite{kingma2014adam} with the default parameters. Weights in each neural network layer were initialized with the \emph{Glorot} normal initialization~\cite{glorot}, and all biases were initialized to zero. Loss function was weighted binary cross entropy:
\begin{equation}
    \label{eq:weighted_crs_entropy}
    \mathcal{L}(\theta) = - \frac{1}{n} \sum_{i=1}^{n} w_1y_i \log p(1 | v_i; \theta) + w_0(1-y_i) \log p(0 | v_i; \theta)
\end{equation}
where $n$ is the number of classified vertices in the batch, $ v_i $ \gls{hmill} representation of  $ i $-th vertex in the batch,  $ y_i \in \left\{ 0, 1 \right\} $ its binary label, and  $ p(1 | v_i ; \theta) $ conditional probability of  $ i $-th vertex being malicious according to the model parametrized by $ \theta $. Note, that $ p(0 | v_i ; \theta) = 1 - p(1 | v_i ; \theta) $ for each $ i $. Weights $ w_0 $ and  $ w_1 $ are real numbers, specifically, we used  $ w_1 = 0.1 $ and  $ w_0 = 0.9 $ to express that we prefer false-negative to false-positive errors.

\subsection{Experimental results}%
\label{sub:idp_results}

\begin{wrapfigure}{R}{0.5\textwidth}
    \centering
    \includegraphics[width=\linewidth]{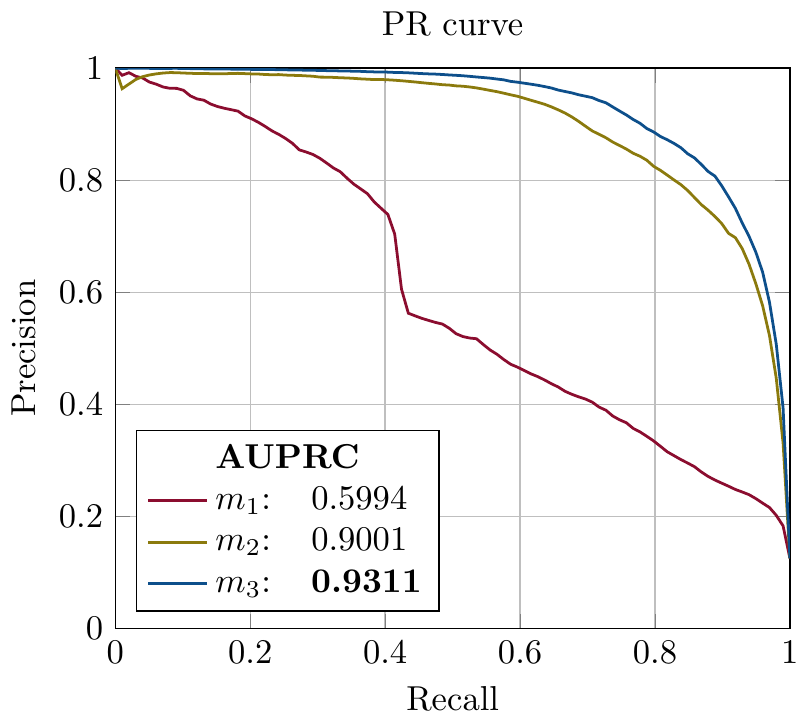}
    \caption[PR curve from the \emph{behavior-based malware classification in graphs} use case.]{Description of the performance of models $ m_1 $,  $ m_2 $ and  $ m_3 $ with a \gls{pr} curve.}%
    \label{fig:idp_pr}
\end{wrapfigure}
A canonical way to compare the performance of several binary classifiers in the presence of imbalanced classes is to plot a \gls{pr} curve or a \gls{roc} curve. The standard scalar metrics computed from a confusion matrix require a specific threshold value for their computation. \gls{pr} and \gls{roc} curves enable us to study values of such scalar metrics for multiple threshold values simultaneously. \gls{pr} curve captures a tradeoff between \emph{recall} and \emph{precision}, two inversely related quantities. \gls{roc} curve describes how \emph{true positive rate} (recall) and \emph{false positive rate} change for different values of the threshold.

To express the performance with a single number taking into account multiple thresholds, we can measure the area under such curves. This gives rise to \gls{auprc} and \gls{auroc} metrics, which both range from zero to one. Even though this is a convenient way to quantify the performance, the values of \gls{auprc} and \gls{auroc} may sometimes be misleading. Depending on the circumstances, different parts of the curves may be of different significance. For instance, in the cybersecurity domain, false positives are much less desirable than false negatives. Usually, high enough threshold values are selected to keep false positive rate sufficiently low, and precision sufficiently high. This corresponds to rather small regions in both curve types, however, \gls{auprc} and \gls{auroc} are evaluated using the whole curve. For the same reason, we include an \gls{roc} curve with a logarithmic $ x $-axis (false positive rate). All curves drawn in this work are made of one hundred points with $ x $-values sampled uniformly from the linear (or logarithmic) scale and $ y $-values obtained with linear interpolation.

\begin{figure}[hbtp]
    \begin{subfigure}[t]{.5\textwidth}
        \centering
        \includegraphics[width=\textwidth]{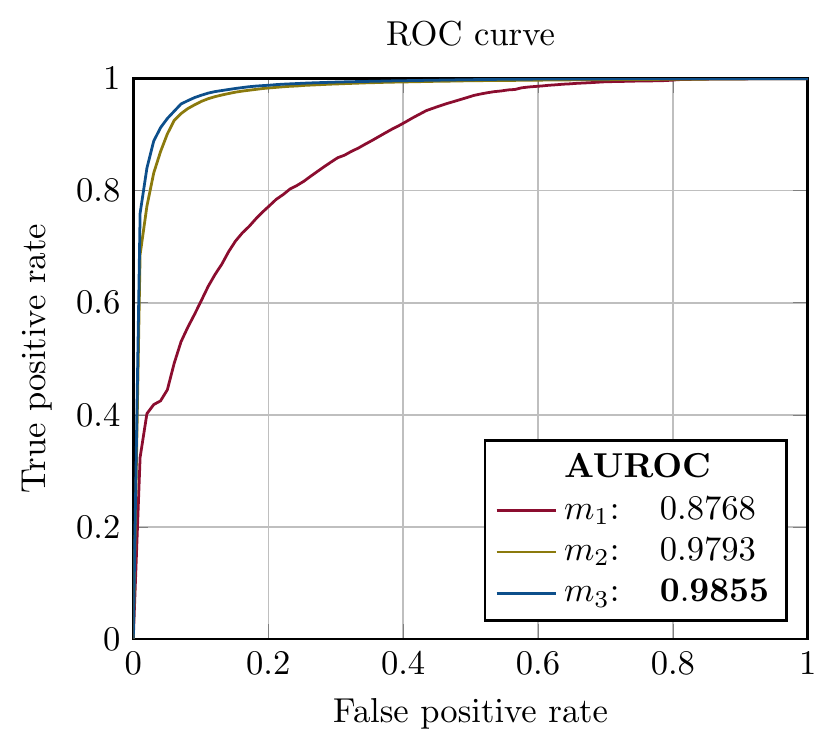}
    \caption{\label{sf:idproc}}%
    \end{subfigure}
    \begin{subfigure}[t]{.5\textwidth}
        \centering
        \includegraphics[width=\textwidth]{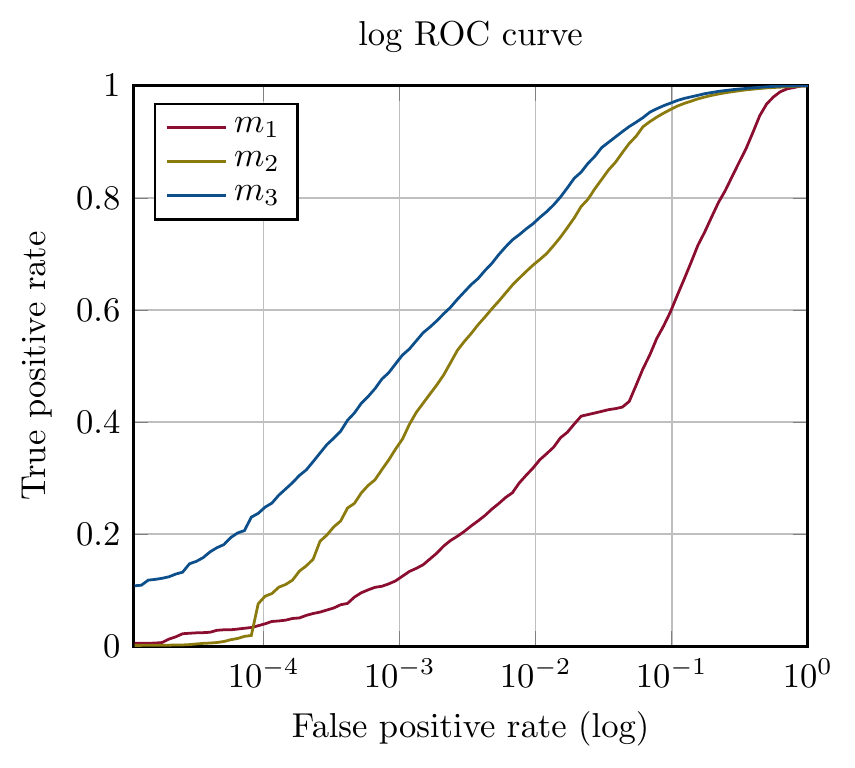}
        \caption{\label{sf:idproclog}}%
    \end{subfigure}
    \caption[ROC curve from the \emph{behavior-based malware classification in graphs} use case.]{ROC curve corresponding to models $ m_1 $,  $ m_2 $ and  $ m_3 $ used in the experiment. We provide both variants with linear (\subref{sf:idproc}) and logarithmic (\subref{sf:idproclog}) scale of $ x $-axis.}%
    \label{fig:idp_rocs}
\end{figure}

We trained all three aforementioned types of model $ m_1 $,  $ m_2 $, and  $ m_3 $, and evaluated their performance. The resulting \gls{pr} curve is drawn in Figure~\ref{fig:idp_pr}, and \gls{roc} curve with linear and logarithmic scale of $ x $-axis is in Figure~\ref{fig:idp_rocs}. Perhaps not surprisingly, the more information the model processes, the better results it achieves. By inspecting a $ 1 $-step neighborhood of the vertex corresponding to an executable implemented in model $ m_3 $, we were able to attain high true positive rate while keeping false positive rate low. We could not achieve this with model $ m_2 $, which processes only an \gls{xml} document description of the vertex. This makes model $ m_3 $ much more useful in practice than $ m_2 $, despite the little difference in the \gls{auroc} metric.

This experiment confirms that simply using more data leads in most cases to better results. All three models $ m_1 $,  $ m_2 $ and  $ m_3 $ were designed as `baseline' and architectures and training procedure can surely be improved. However, unutilized data sources are often the low-hanging fruit. Thanks to \gls{hmill}, modelling of such sources is straightforward.

\chapter{Modelling interactions in heterogeneous networks}%
\label{cha:cisco}

In this next-to-last chapter, we present one last and the most specific use case, where the \gls{hmill} framework comes in handy. In this problem, very little is known about objects of interest themselves. Instead, interactions between objects are provided in several bipartite graphs specifying binary relations between objects. Thus, we move from a purely `feature-based' approach presented in Chapter~\ref{cha:jsons} and also in Chapter~\ref{cha:hmill_on_graphical_data} to learning merely from relations between objects. The two approaches, each with its own advantages and flaws, are to some extent orthogonal and can be combined as we discuss later. To empirically test the proposed method, we investigate the task concerning an extension of a blacklist of malicious computer domains based on their interactions with other entities in the network. In the last part of the chapter, we present the results of the experiments on the data provided by researchers from Cisco Cognitive Intelligence.

In the domain of network security analysis, the focus is shifted from the level of individual client machines to whole networks of clients and their protection by detecting and neutralizing malicious activities happening in the network. Being able to decide for any entity (IP address, domain, email) in the network whether it is safe for communication or not is a crucial requirement for implementing passive security measures in the first line of defense, for instance, simply blacklisting known threats. However, keeping such blacklist updated is a difficult task for the same reasons we listed earlier when discussing malware signature databases---due to the quick growth of the number of existing threats, it is nearly impossible to keep lists up to date manually. Therefore, methods for the automatic extension of blacklists with uncovered threats are researched.

\section{Prior art}%
\label{sec:prior_art}

Prior art on the problem of blacklist extension can be broadly split into two fundamental approaches. In \emph{classifier-based} approach~\cite{zhang2008_blacklist, Antonakakis2010BuildingAD, bilge2011}, blacklists are used to obtain labels for training a classifier for determining the maliciousness of unseen objects in the future. One of the most constraining assumptions to achieve good performance is having high-quality training data. It is known that public blacklists are oftentimes incomplete due to the delay in their updates or the fact that some existing threats have not been discovered yet. If all entities not present in the blacklist are labeled positive in the training phase, the classifier will be inevitably shown samples with an incorrect label.

The second approach is \emph{graph-based}~\cite{coskun2010friends, philips2012detecting, Carter2014a, oprea2015, Rezvani, PlumeWalk} which assumes that malicious activities are localized in a graph, forming communities with sharp boundaries~\cite{collins2007using, Yu2010}. Methods following this approach build a graph representation of relations between objects such that graph analysis algorithms can be utilized to extend knowledge from the blacklist. High malicious score is first initialized in vertices corresponding to objects found in the blacklist, which are referred to as \emph{seeds} or \emph{tips}, and afterwards propagated through the graph using formulas involving the maliciousness of the neighbors and weights defined on edges. After some number of steps, vertices with highest malicious scores that are not present in the blacklist are returned as candidates for blacklist extension. As a result, no data apart from relations themselves are collected. This is useful in situations when data is inaccessible or expensive to obtain. On the other hand, to the best of our knowledge, algorithms for propagating the maliciousness through graphs use fixed formulas for propagation, cannot use complicated representations of relations, and have few to none tunable parameters. For instance, the most popular \gls{ptp} algorithm~\cite{Carter2013, Carter2014a}, which we further elaborate on below, propagates malicious scores in the form of scalar values throughout a graph with scalar weights on edges. Definition of edge weights is the only parameter of the method together with a number of steps performed. The small number of parameters limits the use of methods for different problems, as the algorithms are oftentimes tailored to perform a particular task on a specific dataset and may not be robust to changes in the application domain.

The graph-based approach to blacklist extension is similar to the inference problem in probabilistic graphical models we described in the previous chapter and in fact some methods are built on the same ideas~\cite{Carter2013, Carter2014a, Manadhata2014}. However, the possibility of using \gls{gnn}s for the task has not been studied for two main reasons---their high computational complexity coupled with characteristically large size of graphs observed in the domain, and a lack of public datasets. Therefore, the cybersecurity domain requires different techniques, despite many similarities in the task at hand.

We end this section by reviewing a \gls{ptp} algorithm~\cite{Carter2013, Carter2014a}, which is a canonical graph-based algorithm used in cybersecurity. Given an undirected graph, it estimates the probability $P(v_i)$ of a vertex $v_i$ being malicious from its connection to other vertices and from a set of known malicious vertices, which are assumed to come from the blacklist, using the following equation:
\begin{equation}
\label{eq:PTP1}
P(v_i) = \sum\limits_{v_j \in \mathcal{N}(v_i)} w_{ij} P(v_j | v_i = 0),
\end{equation}
where $w_{ij}$ are positive edge weights assumed to be normalized to one, $\sum_j w_{ij} = 1$. Conditioning the threat to $ v_i = 0 $ avoids unwanted direct feedback of the vertex to itself. Since exact solving of~\eqref{eq:PTP1} is generally intractable for larger graphs, authors propose a message-passing algorithm for finding an approximate solution.  It initializes $P^0(v)$ to one for vertices in the blacklist and zero for vertices outside of it, and updates the solution in each iteration as
\begin{equation}
\label{eq:PTP2}
P^t(v_i) = \sum\limits_{v_j \in \mathcal{N}(v_i)} w_{ij} (P^{t-1}(v_j) - C^{t-1}(v_i, v_j)),
\end{equation}
where $C^{t-1}(v_i, v_j)$ is the portion of $P^{t-1}(v_j)$ propagated from vertex $v_i$ in the previous step. In each iteration, $P(v)$ for the blacklisted vertices is set to $ 1 $ again to reinforce the signal.

In~\cite{Kazato2016}, PTP was successfully used to infer malicious domains from a bipartite DNS graph. To the best of our knowledge, the most scaled and comprehensive use of PTP is in~\cite{Jusko2017}, where it was used for malicious domain discovery and successfully implemented in the pipeline of Cisco Cognitive Intelligence products. Unlike in most prior art, the problem of constructing the graph from binary relations and setting weights to edges is discussed in~\cite{Jusko2017}. We consider methods from~\cite{Jusko2017} state of the art and compare the proposed method to them.

\section{Method description}%
\label{sec:method_description}

Our method for blacklist detection is based on recent progress in both cybersecurity and machine learning. All bipartite graphs representing relations on the input are first transformed into unipartite graphs and message-passing inference with \gls{hmill} models is performed afterwards. Since messages are learned, this leads to more general procedure and unlocks possible applications outside of the domain as well, because the algorithm can be optimized for a particular problem and circumstances. This part of the thesis describes an extension of our work presented in~\cite{thesis}, where we learned messages for inference algorithm using only one bipartite graph on the input. In this thesis, we expand the method to work with several bipartite graphs with heterogeneous vertices and edges. Also, thanks to \gls{hmill}, the introduction of a rather complicated method seems natural. Since the resulting model is implemented as an \gls{hmill} model and raw data (features) can be therefore specified in both vertices and edges, the method effectively bridges the gap between the classifier-based and the graph-based approaches. We elaborate on this matter later after the method is explained.

\subsection{Input}%
\label{sub:input}

Formally, the method expects $ n $ undirected bipartite graphs  $ \left( \mathcal{G}_{1}, \mathcal{G}_{2}, \ldots, \mathcal{G}_{n}\right)  $, where $ \mathcal{G}_{i}$ is a tuple  $ \left( \mathcal{A}_i \cup \mathcal{B}_i , \mathcal{E}_i\right) $. Here, all vertices are split into two disjunct sets $ \mathcal{A}_i$  and $ \mathcal{B}_i $, such that for all $ i = 1, \ldots, n $ and all $ \left\{ u, v \right\}  \in \mathcal{E}_i $, it holds $ u \in \mathcal{A}_i $ and  $ v \in \mathcal{B}_i $. Each vertex in the graph represents a specific object of some type in real-world. For instance, in the cybersecurity domain,  a vertex may correspond to a \gls{sld}, a client, an \gls{ip} address, an email address, and so forth. Moreover, we assume that one vertex set in each of the graphs contains vertices of the same constant type, that is, each vertex set $ \mathcal{A}_i$ is a subset of some superset $ \mathcal{A} $ and $ \type(v) $ is constant for each $ v \in \mathcal{A} $. It may happen that some vertex $ v \in \mathcal{A} $ is present in two different bipartite graphs, in other words, $ \mathcal{A}_i \cap \mathcal{A}_j \neq \varnothing $. For vertices in second partites we only assume that for each $ i $,  $ \type(v) $ is constant across all vertices in  $\mathcal{B}_i $, however, this type may vary with different $ i $. Each edge represents some relationship between two objects corresponding to its incident vertices, and each graph $ \mathcal{G}_i $ may differ in a definition of the relationship in real-world.

\begin{figure}[hbtp]
    \centering
    \begin{subfigure}[b]{0.47\textwidth}
        \centering
        \includegraphics[width=0.95\textwidth]{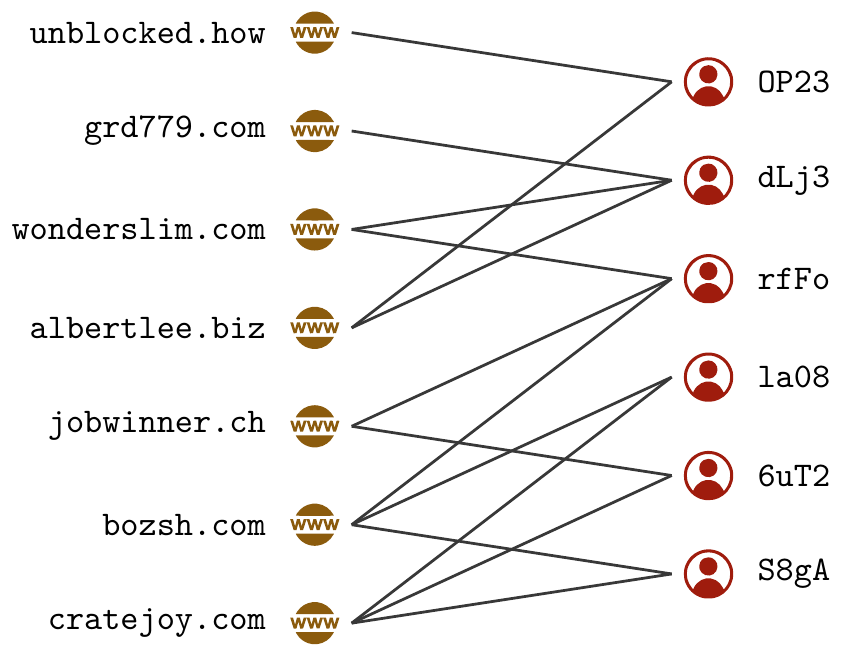}
        \caption{\label{sf:bg1}$ \mathcal{G}_1 $}%
    \end{subfigure}
    \hfill
    \begin{subfigure}[b]{0.52\textwidth}  
        \centering
        \raisebox{0cm}{\includegraphics[width=0.95\textwidth]{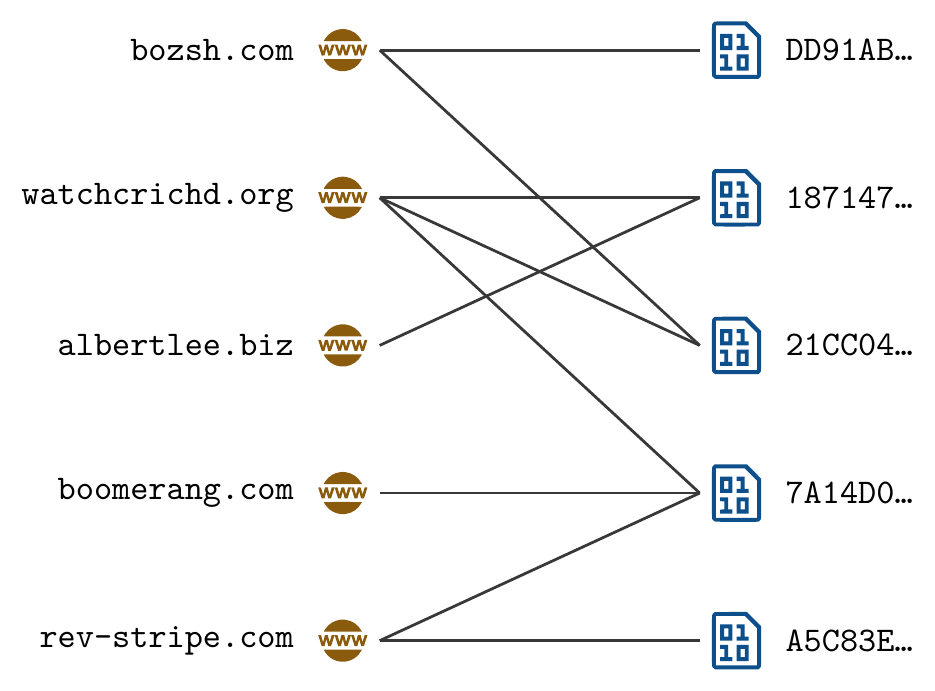}}
        \caption{\label{sf:bg2}$ \mathcal{G}_2 $}%
    \end{subfigure}
    \caption[Two examples of bipartite graphs encoding entity interaction.]{Example of two bipartite graphs, a \emph{domain-client} graph $ \mathcal{G}_1 $ (\subref{sf:bg1}) and a \emph{domain-binary} graph $ \mathcal{G}_2 $ (\subref{sf:bg2}), that can be used to describe the behavior of entities in a network. Clients are identified by a four-character-long identifier and binary files by their \texttt{sha256} hash (shortened in the figure). See text for further details.}%
    \label{fig:graphs_cisco}
\end{figure}

In Figure~\ref{fig:graphs_cisco}, there is an example of two such bipartite graphs that can be used to describe the behavior of entities in a network. In this example, left vertex sets of both graphs contain vertices representing \gls{sld}s appearing in network traffic. The right set in the first graph comprises clients connected to the network (computers, smartphones, printers, and others), and each edge $ \left\{ d, c \right\}  $ encodes the fact, that a client $ c $ connected to a domain  $ d $ during the time when the network was observed. In the second graph, the other set represents individual binary files identified by their hashes, and each edge $ \left\{ d, b \right\}  $  indicates  that  a  process instance of $ b $  has issued a request to a domain $ d $. 

Additionally, we assume that a blacklist of malicious objects $ L \subseteq \mathcal{A}$ is available. In our example, this means that we know some domains in graphs exhibit malicious activity. The task is to discover additional domains that are not present in list $ L $, based on the behavioral information encoded in input bipartite graphs.

\subsection{Graph transformation}%
\label{sub:graph_transformation}

\begin{figure}[hbtp]
    \centering
    \begin{subfigure}[b]{0.45\textwidth}
        \centering
        \includegraphics[height=4.8cm]{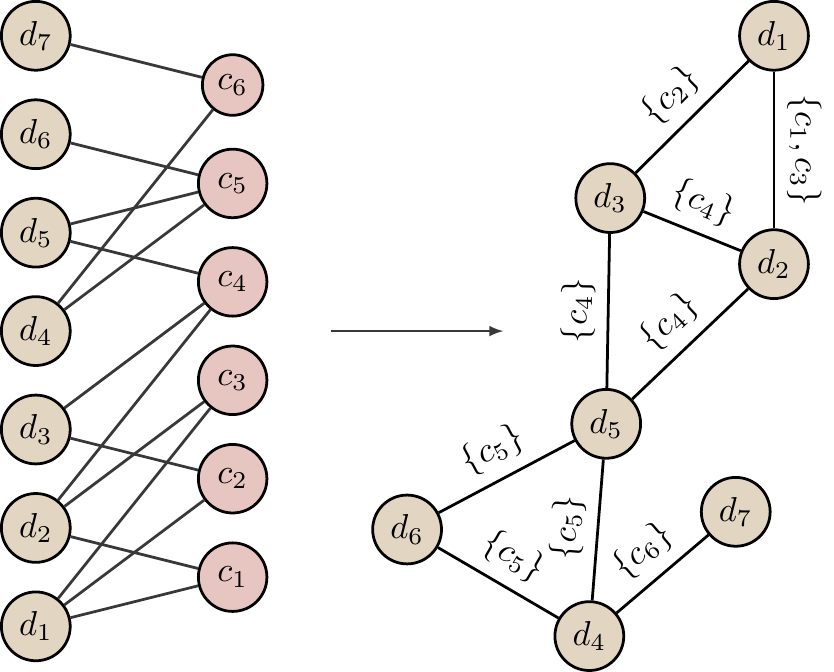}
        \caption{\label{sf:gt1}$  \mathcal{G}_1 \mapsto G_1 $}%
    \end{subfigure}
    \hfill
    \begin{subfigure}[b]{0.45\textwidth}  
        \centering
       \raisebox{0.885cm}{\includegraphics[height=3.384cm]{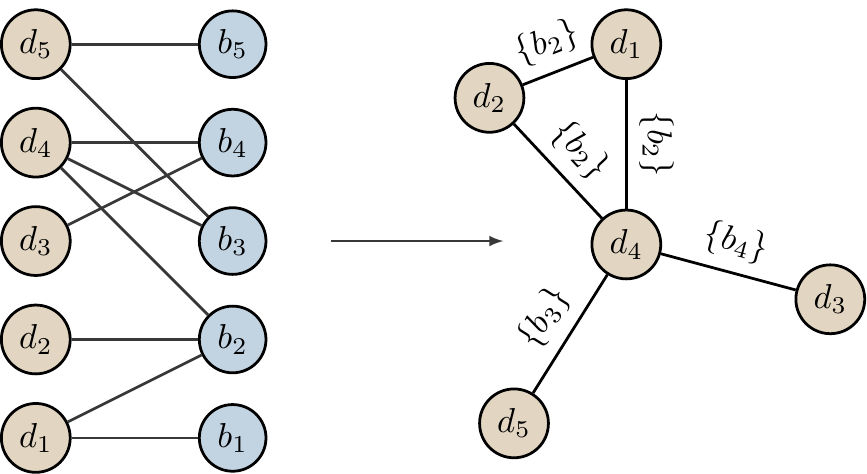}}
       \caption{\label{sf:gt2}$ \mathcal{G}_2 \mapsto G_2 $}%
    \end{subfigure}
    \caption[Two examples of graph transformation.]{Examples of graph transformation performed on both graphs from Figure~\ref{fig:graphs_cisco}. Each client and binary file creates a clique in the transformed graph of the same cardinality as its degree in the original bipartite graph. Consequently, $ b_1 $ and  $ b_5 $ in (\subref{sf:gt2}) do not contribute at all. Each edge in transformed graphs is labelled with all clients/binaries that caused its appearance.}%
    \label{fig:graph_transf}
\end{figure}

In the next step, each graph $\mathcal{G}_i = (\mathcal{A}_i, \mathcal{B}_i)$, is transformed to a unipartite \emph{transformed graph} $G_i = ( \mathcal{A}_i, E_i )$ with vertices equal to vertices in the left vertex set $\mathcal{A}_i$ and edges constructed with the following rule:
\begin{equation}
    \label{eq:transformation}
    \{ u, v\} \in E_i \iff \exists b \in \mathcal{B}_i \colon \{\{u, b\}, \{v, b\}\} \subseteq \mathcal{E}_i ,
\end{equation}
for all $u, v \in \mathcal{A}_i$. Thus, in the \emph{domain-client} graph example, the transformed graph contains an edge between two domains if and only if there exists a client that communicated with both of them. Analogically, in the \emph{domain-binary} graph, each edge represents that at least one binary file issued a request to both domains involved. The semantics of this transformation were already discussed in~\cite{Liu2014, Jusko2017, Manadhata2014}. For example, when a computer is infected over a network, the malicious file is downloaded from some server. Once the malware becomes active, the vital step for attackers is to establish a communication channel to the infected machine. This usually means contacting another server, which is called a command-and-control (C\&C) server. As the infected client connected to both servers, an edge connecting them together appears in the transformed graph. The transformation of the \emph{domain-client} and \emph{domain-binary} graphs from Figure~\ref{fig:graphs_cisco} can be found in Figure~\ref{fig:graph_transf}.

\subsection{Message-passing phase}%
\label{sub:message_passing_phase}

After performing graph transformation on all $ n $ bipartite graphs, we are left with $ n $-tuple  $ \left( G_1, \ldots, G_n \right) $, where each $ G_i $ is a unipartite graph of vertices of a single type, in our case \gls{sld}s. Assume that we want to infer an unknown quantity for a vertex $ v \in \mathcal{A} $, for example, estimate the probability of $ v  $ being malicious. In each of the $ n $ transformed graphs, we perform $ k $-step \gls{hmill}-based message-passing inference centered in the current vertex $ v $ exactly in the same way as described in the previous chapter. Recall that neighborhoods are modelled with (weighted) bag model nodes and knowledge extracted from heterogeneous multi-modal sources is merged together with product model nodes. In this case, however, we are dealing with not one, but $ n $ graphs at once. This is again easily solved with the framework---as inference in each of the transformed graphs is implemented by an \gls{hmill} model, we create a product model node as a new root of the model tree with each of the $ n $ inferential models as its child. Recall that when evaluated, each \gls{hmill} model returns a vector. We can therefore regard evaluation of the resulting product model node as first distilling a vector describing interactions of $ v $ in each of the available relations and merging them afterwards to obtain a result.

\begin{figure}[hbtp]
    \centering
    \includegraphics[width=\textwidth]{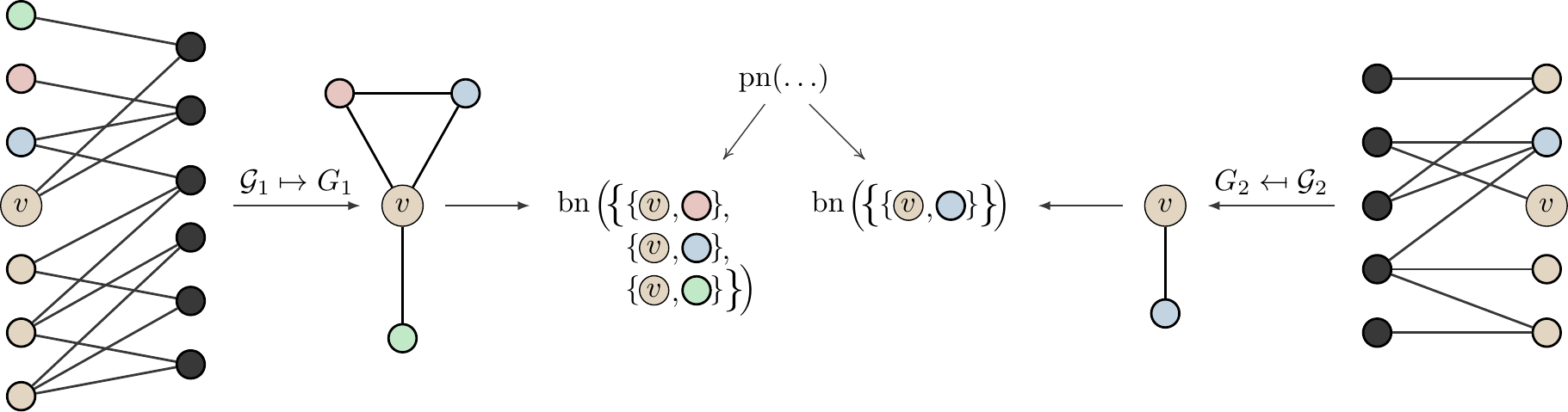}
    \caption[Obtaining an HMill sample tree for a vertex using transformed graphs.]{A process of obtaining an \gls{hmill} sample tree for a specific vertex $ v $. Neighbors of  $ v $ in unipartite graphs are colored. We do not show the whole transformed graph, but only its subgraph induced by vertices in the neighborhood of  $ v $. This example shows the same graphs as examples in Figures~\ref{fig:graphs_cisco} and~\ref{fig:graph_transf}. Vertex $ v $ corresponds to \texttt{albertlee.biz} domain, aliased as $ d_4 $ and $ d_3 $ in Figure~\ref{fig:graph_transf}. See text for details.}%
    \label{fig:cisco_model}
\end{figure}

The process of how a sample tree is obtained for our case where $ n = 2 $ is sketched in Figure~\ref{fig:cisco_model}. Given a vertex $ v $ for prediction, all input bipartite graphs are transformed to unipartite graphs. The neighborhood of $ v $ in each graph is represented in $ n $ bag data nodes, which are merged together with a product data node afterwards. Again, each instance in these bag data nodes represents one edge $ \left\{ u, v \right\}  $ in the transformed graph, which is depicted with different colors in the figure. Since $ u $ is also a part of this adjacency relation, we could describe the edge (instance) in the bag node as a deeper subtree and perform multiple-step inference, as explained in the last chapter. Also, information (features) on both vertices and edges can be included, which we have already discussed as well.

\noindent
In this section, we show how to deal with circumstances when no additional information about vertices and edges is known, and input to the method comprises only $ n $ bipartite graphs representing binary relations between vertices. To describe edge $ \left\{ u, v \right\} $ where $ u, v \in \mathcal{A}_i $ in transformed graph $ G_i $, we  can use any information about $ u $,  $ v $ or the edge itself. Recall, that when performing the graph transformation, all vertices from $ \mathcal{B}_i $ causing an edge to appear can be stored, as illustrated in Figure~\ref{fig:graph_transf}. Nonetheless, in this setting, no external information about  $ u $,  $ v $, or any vertex from  $ \mathcal{B}_i $ is available. For this purpose, we designed a set of graphical features that can be directly computed from bipartite graphs and their transformations. They are summarized in Table~\ref{tab:g_features} and can be either precomputed during the construction of a transformed graph, or computed on demand in linear time with respect to the number of vertices in a bipartite graph at worst. All features apart from label of the neighbor obtained from the blacklist $ v \in L $, which is binary, are in range  $ [1, \infty) $, and are further transformed with mapping  $ x \mapsto \log(x) + 1 $. For the label, we used one-hot encoding. This feature mapping is an example of mapping $ h $ as explained in Section~\ref{sub:array_node}, which in this case maps from edges in a transformed graph to a Euclidean space of dimension $ 9 $. 

\begin{table}[htb]
    \caption[Graphical features for the \emph{Modelling Internet communication} use case.]{Specification of graphical features used in experiments to describe edge $ \left\{ u, v \right\}  $, where $ v $ is the current vertex for inference. $ \mathcal{G} $ and $ G $ denote bipartite and transformed graphs, and $\mathcal{N}_{\mathcal{G}}(v)$ and $\mathcal{N}_{G}(v)$ neighborhoods of $ v $ in $ \mathcal{G} $ and $ G $, respectively.}%
    \label{tab:g_features}
    \centering
    {\renewcommand\arraystretch{1.25}
        \begin{tabular}{cc}
            \toprule
            feature definition & property of \\
            \midrule
            $\left\vert \mathcal{N}_{G}(v) \right\vert$ & $v$ \\
            $\left\vert \mathcal{N}_{\mathcal{G}}(v) \right\vert$ & $v$ \\
            $\left\vert \mathcal{N}_{\mathcal{G}}(u) \right\vert$ & $u$ \\
            $\left\vert \mathcal{N}_{\mathcal{G}}(u) \cap \mathcal{N}_{\mathcal{G}}(v) \right\vert$ & \{$u$, $v$\} \\
            $\left\vert \mathcal{N}_{\mathcal{G}}(u) \cup \mathcal{N}_{\mathcal{G}}(v) \right\vert$ & \{$u$, $v$\} \\
            ${\left\vert \mathcal{N}_{\mathcal{G}}(u) \cap\ \mathcal{N}_{\mathcal{G}}(v) \right\vert} / {\left\vert \mathcal{N}_{\mathcal{G}}(u) \cup\ \mathcal{N}_{\mathcal{G}}(v) \right\vert}$ & \{$u$, $v$\} \\
            $\left\vert \mathcal{N}_{\mathcal{G}}(u)\right\vert \left\vert\mathcal{N}_{\mathcal{G}}(v) \right\vert$ & \{$u$, $v$\} \\
            $ u \in L $ & $u$ \\
            \bottomrule
        \end{tabular}
    }
\end{table}

\subsection{Method merits}%
\label{sub:method_merits}

Using the method, we are able to learn merely from the behavior of the objects of interest in any system (e.g. computer network). This is done by a set of simple and efficiently computable features, which are obtained simply from input graphs representing interactions. We claim that this `relation-based' representation generalizes better as well and adapts well in rapidly changing environments. In the experiments below, we add more relations than only \emph{domain-client} and \emph{domain-binary} relations. Besides others, we also used \emph{domain-TLS authority}, which assigns to each \gls{sld} a certificate authority that issued its \gls{tls} certificate. One possibility to model the fact that an \gls{sld} uses a certificate issued by an authority is to use one-hot encoding. However, this may lead to overparametrization once the set of authorities becomes too large. Moreover, this approach also does not allow for the future, when new authorities may arise---to capture this occurrence, one needs to either resize the one-hot encoded vector, thus having to retrain all models on new data, or unify all entities arising in the future into one class, which leads to information loss. Also, if the adversary changes the authority for their malicious domains, but everything else stays the same, the model will obtain a different input in the feature-based approach. Instead, if relations are modelled, the transformed graph will in this case contain a clique consisting of all \gls{sld}s with a certificate obtained from the same authority. If the adversary changes the authority used, the clique remains intact.

With the proposed approach, one can smoothly transition between two different approaches to learning, the first one based on collecting `hard' data about entities in the form of features (or structured raw data in case of \gls{hmill}), and the second one based on simply describing relationships between entities. Both approaches can be arbitrarily combined using the same tools from the framework, independently on the circumstances and the nature of available data. For example, in the computer security domain, informative features are either hard to define or unobservable, however, as discussed later, binary relations are easily obtainable and require no feature engineering. If on the other hand, information about vertices or edges is available, modelling structured data with \gls{hmill} is simple, as discussed earlier. Also, the issue of missing data, which in this case can happen, when the current vertex $ v $ is not present in some of input bipartite graphs, is solved. All these properties lead to a richer and more accurate description of the application domain and consequently, higher accuracy of learned models.

Finally, due to the fact that a bag may contain an arbitrary number of instances, the method is designed to deal with any degree distribution in the graph. This is a substantial difficulty in the transformed graphs, where the size of the neighborhood of a domain varies greatly for different vertices. For instance, frequently visited \gls{sld}s are neighbors of a majority of vertices in the transformed domain-client graph. Thanks to the relative simplicity of graphical features and their effective computation, together with the possibility to use importance sampling technique, a (weighted) bag data vertex representation of neighborhoods are obtained fast, and we can perform inference on graphs with a large number of vertices and edges.

In our previous work~\cite{thesis}, we consider a less general setting, where only one bipartite graph is on the input ($ n=1 $), whereas here, we extend this approach to work with multiple graphs. On the other hand, the cited work provides more details on a domain-specific graph pruning techniques and evaluation methodology, both applicable here as well.

\section{Modelling Internet communication (use case)}%
\label{sec:cisco_usecase}

In this section, we demonstrate how the proposed method performs on the real-world task of extending a known blacklist of (second-level) domains by modelling Internet communication. The problem specifics are huge input bipartite graphs, heavy-tailed degree distributions in transformed graphs, and a very low ratio of positive domains with respect to the total number.

\subsection{Data description}%
\label{sub:data}

The data was kindly provided by Cisco Cognitive Intelligence and describes relations between (second-level) domains and other network entities in several bipartite graphs. Three following relations, two of which were used for the examples above, were collected from a subset of anonymized web proxy (W3C) logs processed by Cisco Cognitive Intelligence\footnote{\url{https://cognitive.cisco.com}}, and Cisco AMP telemetry\footnote{\url{https://www.cisco.com/c/en/us/products/security/advanced-malware-protection/index.html}}:
\begin{itemize}
    \item \emph{domain-client} --- an edge signifies that a client has communicated with (issued an \gls{http}/\gls{https} request to) the domain.
    \item \emph{domain-binary} --- an edge indicates that a process has communicated with a given domain.
    \item \emph{domain-\gls{ip} address} --- an edge signifies that the domain hostname was resolved to the \gls{ip} address using \gls{dns}.
\end{itemize}
These relations are of the \gls{m2m} cardinality type meaning that one domain may be connected with an edge to multiple clients/binaries/\gls{ip}s and vice versa. All bipartite graphs representing these relations were constructed from interactions observed during one week. Bipartite graphs collected during the same week were linked into the same dataset, and we identify them by a starting date of the time window. Hence, here we call a `dataset' a collection of bipartite graphs, each for one relation. During three months of year 2019, we assembled twelve datasets altogether---\texttt{05-23}, \texttt{06-03}, \texttt{06-10}, \texttt{06-17}, \texttt{06-24}, \texttt{06-26}, \texttt{06-27}, \texttt{07-01}, \texttt{07-08}, \texttt{07-15}, \texttt{07-22} and \texttt{07-29}. The model was first trained on a subset of these dates, and then labels were inferred for domains from the remaining dates and the performance was evaluated. Specifically, we used \texttt{06-10}, \texttt{06-27} and \texttt{07-22} for testing, \texttt{06-24} for validation, and the rest for training.
The datasets were further enriched with following three relations (bipartite graphs) extracted from fields in (the latest) \gls{tls} certificate issued to the domain\footnote{more specifically one of its hostnames}:
\begin{itemize}
    \item \emph{domain-\gls{tls} issuer} --- a relation connecting domains to the issuer of the \gls{tls} certificate. The cardinality type is \gls{m2o} and therefore the transformed graph contains a fully connected component for each of the issuers.
    \item \emph{domain-\gls{tls} hash} --- a \gls{m2o} relation connecting a domain to the hash of the \gls{tls} certificate. In the transformed graph, domains using the same certificate are connected.
    \item \emph{domain-\gls{tls} issue time} --- domain and the time when validity period of the certificate starts, stored as a \emph{Unix timestamp}.
\end{itemize}
Lastly, we also leveraged information publicly available in \emph{WHOIS} registries, and extracted relations
    \emph{domain-WHOIS email}, \emph{domain-WHOIS nameserver}, \emph{domain-WHOIS registrar name}, \emph{domain-WHOIS country}, \emph{domain-WHOIS registrar id}, and \emph{domain-WHOIS timestamp}. All of these are of the \gls{m2o} cardinality type. For further information, please refer to the WHOIS specification.

Thus, we had twelve datasets altogether, each containing eleven bipartite graphs. All relations used in our experiments are summarized with several examples in Table~\ref{tab:relations} in Appendix~\ref{ap:cisco}. To gain an idea about the size of the input graphs, we provide Tables~\ref{tab:vertex_sizes} and~\ref{tab:edge_sizes} in Appendix~\ref{ap:cisco}. On average, the bipartite graphs we are dealing with contain around $7\cdot 10^6$ vertices and $3 \cdot 10^8$ edges. On the first three relations mentioned (\emph{domain-client}, \emph{domain-binary} and \emph{domain-IP}), the pruning preprocessing step was performed as described in~\cite{thesis}.

A blacklist of malicious domains, $ L, $ was also provided by Cisco Cognitive Intelligence. At the time of data collection, the blacklist tracked $ 335 $ malicious campaigns, each representing a different threat type, such as ransomware, trojans or click frauds. For each dataset, we used the current version of blacklist available at the time to reflect the real situation. Around five hundred domains from the blacklist were observed during each of the weeks: this is due to the fact that many domains did not appear in any communication during the week-long time window. All domains outside of the blacklist were considered benign. Precise numbers of malicious domains in each of the datasets together with ratios to all observed domains are listed in Table~\ref{tab:mal_numbers} in Apendix~\ref{ap:cisco}. 

\subsection{Experimental setting}%
\label{sub:hyperparameters}

For this experiment, we have decided to train a model for performing only one-step inference in each of the transformed graphs. Firstly, in cybersecurity transformed graphs are dense with a lot of cliques due to vertices of large degrees in bipartite graphs. As a result, neighborhoods of vertices in the transformed graph are huge, and it is possible to traverse between two arbitrary vertices in several hops. It has also been discussed in~\cite{Yu2010} that malicious entities in computer networks tend to form dense communities. For these reasons, we believe that one-step inference is sufficient for this task and leave the exploration of multi-step inference for future work. Secondly, the model for one-step inference in this task has a less complicated structure compared to models for both previous use cases in this thesis. This enables us to further investigate to which extent the results depend on the architecture of inner neural network components, which we have so far always defined using a trivial procedure.

\begin{figure}[htp]
    \centering
    \includegraphics[width=0.8\textwidth]{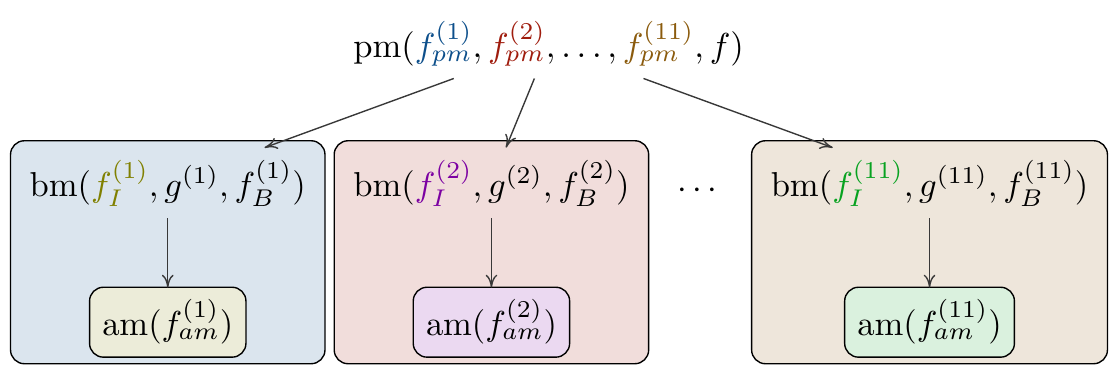}
    \caption[Tree structure of models used in the \emph{Modelling Internet communication} use case.]{A tree structure of models used in this use case. Colors distinguish distinct submodels. See text for details.}%
    \label{fig:cisco_model_specific}
\end{figure}

Every sample tree was constructed in the same way as illustrated in Figure~\ref{fig:cisco_model}. A product data node is used to merge together multi-modal information contained in transformed graphs. Each instance in a bag-data-node representation of neighborhoods is described by an array data node with a vector of real-valued features from Table~\ref{tab:g_features}. As a result, every sample tree and also every corresponding model tree has a depth of three. Let us now turn our attention to the model's architecture. Recall that the topmost product model node in the corresponding model will consist of eleven model (sub)trees, each for processing one of the transformed graphs, and one more mapping we denote here by $ f $ that is applied on the concatenation of results from each of the submodels. This is illustrated in Figure~\ref{fig:cisco_model_specific}. Each of the submodels $ f_{pm}^{(1)}, f_{pm}^{(2)}, \ldots, f_{pm}^{(11)} $ itself is a bag model node, consisting of an instance model $ f_I ^{(i)} $, an aggregation function $ g^{(i)} $ and a bag model $ f_B^{(i)} $. Instances models $ f_I^{(i)} $ are in this case array models for the initial transformation of graphical features with neural-network mappings $ f_{am}^{(i)} $.

\noindent
For simplicity, we have defined all bag model nodes $ f_{pm}^{(i)} $ in the same way. This means that all $ f_{am}^{(i)} $ and $ f_B^{(i)} $ are neural-network components with constant architecture across different indexes $ i $. However, parameters are not shared so that each bag model node can learn different representations. We constructed three distinct models---a `baseline' model $ m_b $, a `wider' model  $ m_w $, and a `deeper' model  $ m_d $. Numbers of neurons in layers of components $ f_{am}^{(i)} $ and  $ f_B^{(i)} $ for each $ i $, as well as $ f $ in the topmost product model node, are in Table~\ref{tab:neuron_numbers}. For activation functions, we used again $ \tanh $ in last layers before aggregations  $ g^{(i)} $, and  $ \relu $ otherwise. For aggregations $ g^{(i)} $, we used a concatenation of all four functions presented in Section~\ref{sub:aggregation_functions}.

\begin{table}[htb]
    \caption[Numbers of neurons in models from the \emph{Modelling Internet communication} experiment.]{Numbers of neurons in models $ m_b $,  $ m_w $ and  $ m_d $. Each number in tuples represent one feedforward layer with the given number of neurons. The length of the tuple specifies a number of layers in the component. Note that all neural-network components $ f_{am}^{(i)} $ contain $ 9 $ input neurons, which corresponds to an input dimension for edge (instance) representation. Also, the first layer of $ f_B^{(i)} $ contains four times as many neurons as the last layer of  $ f_{am}^{(i)}$, which is due to a concatenation of four aggregation functions. The input layer of $ f $ is eleven times wider than the output layer of  $ f_B^{(i)} $, because there are eleven relations altogether.}%
    \label{tab:neuron_numbers}
    \centering
    {\renewcommand\arraystretch{1.25}
        \begin{tabular}{cccc}
            \toprule
            model & $ f_{am}^{(i)} $ & $f_B^{(i)}$ & $ f $ \\
            \midrule
            $ m_b $ & $(9, 30, 30)$ & $ (120, 60, 20) $  & $ (220, 100, 2) $ \\
            $ m_w $ & $(9, 60)$ & $ (240, 120) $  & $ (1320, 2) $ \\
            $ m_d $ & $(9, 15, 15, 15, 15)$ & $ (60, 60, 30, 20) $  & $ (220, 60, 60, 2) $ \\
            \bottomrule
        \end{tabular}
    }
\end{table}

\noindent
For various reasons, it may happen that for a particular domain some relations are not available at all, or the domain has no neighbors in some of the transformed graphs. This leads to empty bags, which we handle as described in Section~\ref{sub:missing_data}. On the other hand, as some of the bags may be very large (millions of instances), we also employed a sampling procedure (Section~\ref{sub:weighted_bag_nodes}). Because positive domains usually emit a stronger signal, and there is also a very low number of malicious domains in input graphs, we include all edges incident to positive vertices to the bag representing the neighborhood of $ v $. From instances corresponding to edges incident to a negative neighbor of $ v $, we sample $ K $ instances without replacement and put them to the bag as well. Of course, all instances are appropriately reweighted again. In all experiments presented here, we set $ K = 100 $.

\newpage\noindent
We used weighted binary cross entropy \eqref{eq:weighted_crs_entropy} for loss function, \emph{Adam}~\cite{kingma2014adam} optimizer with the default parameters, the \emph{Glorot} normal initialization~\cite{glorot} for weights, and zero initialization for biases. During training, we first loaded one of the eight datasets (dates) for training into the memory, and sampled one thousand balanced minibatches each containing $ 256 $ vertices (sampled without repetition). After all minibatches were processed by the model, we loaded another dataset and repeated the whole process, until this was done for each of the training datasets five times.

\subsubsection{Evaluation metrics}%
\label{ssub:evaluation_metrics}

Recall that the primary goal of the proposed method is to find unknown malicious domains not present in the blacklist $ L $. To simulate this goal as realistically as possible, one has to be aware of the interdependence between domains while splitting the blacklist into a part disclosed to the classifier and a part tested on. Moreover, the split should reflect the situation in real-life, when some recently registered or simply hard-to-detect domains are not in the blacklist, but appear on the input. As malicious domains tend to group into dense clusters (communities) of the same family (malware campaign), it is less complicated to merely extend a cluster of known domains than to discover a completely unknown family. Taking all of the above into account, we use the following procedure to evaluate performance. Once the training on the whole blacklist is finished, we first randomly split domains in each blacklist cluster into $ k $ disjunctive parts of approximately the same size. Then, the whole blacklist is divided into $ k $ \emph{folds}, always using  one of the parts for every threat category. This yields $ k $ disjunctive folds and every threat category is represented equally frequently in each of the folds.

The accuracy of the method is estimated from $ k $ separate inference runs in each of which one of the folds is left out and the remaining $ k-1 $ folds are used as tips (seeds). With this procedure, we obtain an estimate of the maliciousness for all domains in the blacklist---the estimate is taken from the only inference run when the domain was not seeded. For domains that are not in the blacklist and therefore treated as benign, $k$ estimates of the maliciousness are acquired, one from each of $k$ runs. These were aggregated with maximum function, which corresponds to the worst case. Eventually, each domain is assigned a single scalar value, which can be subsequently used in the standard evaluation of the performance of a binary classifier. Here, this is realized with \gls{pr} and ($ \log $) \gls{roc} curves. For more elaborate discussion on evaluation in these circumstances and blacklist seeding procedure, please refer to~\cite{thesis}.

\subsection{Experimental results}%

We will now present the results we achieved with our models $ m_b $,  $ m_w $,  $ m_d $ in different circumstances. The experimental setting (training procedure, hyperparameters, initialization and others) is the same in every section unless stated otherwise. Recall, that we used eight datasets (dates) for training and three for testing. For every model and every dataset, we plot a \gls{pr} curve and an \gls{roc} curve with logarithmically scaled $ x $-axis. We do not show \gls{roc} curves with linear $ x $-axis, as in this problem they offer little to none insight. However, both \gls{auprc} and \gls{auroc} are provided. Full results can be found in Appendices~\ref{ap:cisco_ptpcomp},~\ref{ap:cisco_additional_relations},~\ref{ap:cisco_lessgraphs}, and~\ref{ap:cisco_gtest}.

\subsubsection{Comparison to \gls{ptp}}%

In the first experiment, we compare to the \gls{ptp} method. We used the implementation from the Cisco Cognitive Intelligence pipeline, which performs $ 20 $ iterations of the algorithm. Further details are in~\cite{Jusko2017}. The \gls{ptp} method expects only on a single (bipartite) graph and no extensions for multiple input graphs are known. To achieve a fair comparison, we used in this experiment only the \texttt{domain-client} bipartite graph, even though the ability to handle more such graphs is one of the main advantages of the proposed \gls{hmill}-based approach. For this purpose, we define another three models $ m_B $,  $ m_W $, and  $ m_D $, which are built in the same way as  $ m_b $,  $ m_w $ and  $ m_d $, respectively, however, their topmost product model node contains only one child for the \texttt{domain-client} graph processing instead of the original eleven children.

Judging from the results listed in Table~\ref{tab:cisco_results_ptpcomp} and Figures~\ref{fig:ptpcomp_pr} and~\ref{fig:ptpcomp_roc_log}, it is safe to say that the proposed method performs at least as well as the \gls{ptp} algorithm and even surpasses it by a larger margin in some cases. Note, however, that as we explained in Section~\ref{sub:idp_results}, the difference (measured by the \gls{auprc} metric) between the best-performing model $ m_W $  and other models is not that significant, since a lot of area is measured in rather irrelevant parts of the plotted curve. The differences in the \gls{auroc} are minuscule. We further discuss the influence of `baseline', `wide', and `deep' architectures in the next section.

\subsubsection{Additional relations}%

In Table~\ref{tab:cisco_results_all} and Figures~\ref{fig:all_pr} and~\ref{fig:all_roc_log}, we present the results of the \gls{hmill}-based inference performed by models $ m_b $,  $ m_w $ and  $ m_d $, this time using all eleven relations. For evaluation, we used only domains from the \texttt{domain-client} graph, using the remaining ten relations merely to enrich information about these domains. This way, we can compare the results to the instance of the experiment presented in the previous section. Nevertheless, with the proposed method we can obtain estimates of the maliciousness for every domain ever observed (denoted above by $ \mathcal{A} $), since when a domain is not present in the \texttt{domain-client} graph (or any other graph), we treat this information as a missing value and \gls{hmill} copes with it.

With more data, models perform approximately three times better than before in terms of the \gls{auprc} metric. This confirms again, that a quick performance boost can be oftentimes attained merely by feeding more data to the model, which \gls{hmill} excels at. Despite the fact, that the wider architecture $ m_w $ helped in the previous experiment, here we observe that the width of the layers plays little to no role as $ m_b $ and  $ m_w $ perform more or less the same. On the other hand, when a higher number of thinner layers is employed, the performance severely decreases.

\subsubsection{Fewer datasets trained on}%
\label{ssub:fewer_datasets}

To assess the generalizing capabilities of the method and also its resilience to the potential concept drift, in this experiment, we trained the models exactly in the same way, however, with less data. Specifically, we sorted the datasets by date and used only the first $ D $ of them for training, where $ D \in \left\{ 1,3, 5, 8 \right\}  $. Thus, in the first instance we train the model on the data observed during only week (\texttt{05-23}) and in the last instance on all of the data available. Because the testing dates are dispersed over the whole range of considered dates, this enables us to study the decline in the models' performance over time (aging). In each of the examples, the training procedure was prolonged appropriately so that the total number of sampled minibatches remained the same as in the case when we iterate over all training datasets five times. We used only `baseline' model $ m_b $ in this experiment for simplicity. Results are presented in Table~\ref{tab:cisco_results_lessgraphs} and Figures~\ref{fig:lessgraphs_pr} and~\ref{fig:lessgraphs_roc}. We can see that in this case, even a lower number of datasets is sufficient to learn a reasonable model, probably because the network traffic observed during different weeks is highly correlated, and the concept drift present in the data (if any) is not significant enough to alter the performance.

\subsubsection{Grill test}%

In spite of being one of the defining characteristics of the proposed method, the ability to learn messages and the whole inference procedure actually introduces one additional concern. When the same domain appears in both the data for training and the data for testing, the methods that employ learning may come up with a model that simply memorizes the domains in the training set, gaining an unfair advantage. This does not happen with methods that do not contain a training phase, such as \gls{ptp}, since the formulas involved are fixed, regardless of the input data. Even though it is certainly not possible to be able to fully differentiate between all distinct domains, as the mapping to the representation in the form of several bipartite graphs is noisy and thus not injective (also, the same domain will have different neighborhoods in graphs corresponding to different dates), this phenomena may still occur to some extent.

To investigate further whether this occurs in our case, we carried out one more experiment, where we leveraged a technique from~\cite{Grill2016}, which we refer to here as \emph{Grill test}. First, we sampled a subset of malicious domains from the blacklist proportionally to the size of each cluster. During the training phase, we consider these domains benign so that the model cannot memorize them specifically. For evaluation, we use all negative domains as before, however, only positive domains that the model did not see during training are picked (around one tenth of all positive domains in the dataset). This way, we make sure that a specific malicious domain is first seen during the testing phase. Moreover, adding noise to labels by inverting some positive domains in the blacklist during training simulates the situation, when some positive domains are undiscovered and thus not present in the blacklist, but they are observed.

Again, for simplicity, we did this procedure only for the baseline model $ m_b $. We trained it in the Grill test `mode' and compared it on all three testing datasets to the case when $ m_b $ is trained on (and evaluated) on all domains. Unsurprisingly, the performance drops, as can be seen in Table~\ref{tab:cisco_results_gtest} and Figures~\ref{fig:gtest_pr} and~\ref{fig:gtest_roc_log}. Nevertheless, let us emphasize that with Grill test, both training and testing data follow the different distributions and therefore, the task being solved is slightly different.
\newline\newline\noindent
In this chapter, we showed that it is possible to estimate the maliciousness of the domain merely from its behavior in the network. The proposed solution implemented with the \gls{hmill} framework and the popular \gls{ptp} algorithm performed comparably when on only one relation was used. Nonetheless, the \gls{hmill}-based solution achieved approximately three times better results with more input data.


\chapter{Conclusion and future work}%
\label{cha:conclusion_and_future_work}

\section{Tying loose ends together}%
\label{sec:tying_loose_ends_together}

In this thesis, we proposed a novel general-purpose machine learning framework for sample representation and model definition based on the multi-instance learning paradigm.  The main defining characteristic of the \gls{hmill} framework is its versatility. It excels at modelling data sources with attributes typical for real-world data, such as heterogeneity, incompleteness and hierarchical structure, elegantly dealing with all of them. As a result, raw observations can be input into models with little to no preprocessing required. \gls{hmill} models are able to discover relevant and informative features and extract them at different levels of abstraction, which effectively mitigates the need to design a mapping from complex input data types to vectors, with which the majority of current \gls{ml} methods operate. In our opinion, valuable domain expertise should be used to decide \emph{what} discriminative data to collect and feed to the models instead of \emph{how} to do it, and the framework helps with that.

Moreover, we investigated the approximation capabilities of the framework at the theoretical level. We showed that the design of the framework is well-founded by extending the Universal approximation theorem to functions realized by the \gls{hmill} model nodes. Thanks to the explicit structure of \gls{hmill} model trees derived directly from input data in the form of a schema, learned models are natural to interpret and decisions easier to explain. We also discussed how \gls{hmill} models deal with missing data, how effective sampling techniques can be leveraged to make inference scalable, and how effective minibatching is implemented.

We have demonstrated the flexibility of the framework in three different tasks from the computer security domain, which gained notoriety for containing many problems that are not solvable by machine learning methods in a straightforward manner. The prime example are data-serialization formats used for information exchange over the Internet, such as \gls{xml} or \gls{json} documents, which may be structured in a complex, hierarchical way. Specifically, we attempted to solve the \emph{\gls{iot} device identification} problem, in which every \gls{iot} device is described by a \gls{json} document containing data that can be queried in the network.

Furthermore, we have also shown how to learn from graphs containing dependent observations and how to implement \gls{hmill}-based message-passing graph inference using bag nodes for neighborhood representation. This was accompanied by the practical example of how executable files can be classified as malicious or benign from the snapshot of the operating system represented as a graph. In this example, both vertices and edges in the graph stored useful information, which in the vertex case was encoded as an \gls{xml} document. We demonstrated that the processing of features on edges and vertices as well as the message-passing procedure can be jointly optimized and tailored to the corresponding problem as a single \gls{hmill} model.

In the last part of the thesis we discussed the task, in which no data is observable in vertices or edges, and the only available information is in the structure of the graph. We then investigated the real-world task of blacklist extension based on the behavioral patterns in the network. This represents a different approach to the problem, in which we describe interactions between observed entities, rather than attributes of individuals. With the framework, both approaches can be fluently transitioned between.

In all three tasks, `baseline' \gls{hmill} models achieved comparable or even better performance than current specialized algorithms, which suggests that high modelling flexibility is not traded for compromises in performance. The only downside is the relatively high computational complexity of training caused by demanding gradient computation in possibly deep and broad model trees. We note that this issue is of little concern with ever-increasing computing power.

To conclude, we firmly believe that the flexibility in modelling coupled with the straightforward out-of-the-box application of the framework will enable application in problems, where current methods may have struggled.

\section{Future work}%
\label{sec:future_work}

Since feature selection and decision explaining are considered two tasks similar in many regards, and we have shown that \gls{hmill} deals with the former one well, we believe that explanations for \gls{hmill} models are possible. Although explainability is not specifically addressed in this text, we assume that the hierarchical structure of the model and the fact that it has to learn to extract relevant information in each layer of the tree makes precise explanations feasible. Also, as models from the framework can process raw data, the outputs can be explained directly using data from the input-level as well as from higher-level representations. If a model is trained on manually extracted features, this opportunity is lost. Moreover, in some cases, explaining at the lowest level may be much more useful. Consider the IoT device identification task. We could produce an explanation of a device type prediction involving a specific protocol being used at a specific port, which seems reasonable. However, this is not possible if this information is somehow flattened into a feature vector beforehand.

Other than that, we would like to investigate further, how the performance increases once we make more than one step in graph inference, and how the exponential growth in sizes of the neighborhood can be tackled. One possible approach is to use attention mechanisms~\cite{bahdanau2014neural, Vaswani2017, ilse2018attentionbased} in an aggregation function in bag model nodes, which would allow models to first examine many neighbors and pay attention to those which seem most relevant.

In spite of competitive performance achieved in all three tasks, the models presented here are rather basic and do not make use of any recent discoveries in deep learning, for example, residual learning~\cite{he2015deep} or batch normalization layers~\cite{ioffe2015batch}. This presents another possible research direction for the future. 

Last but not least, we aim to apply the \gls{hmill} framework on other datasets not from the cybersecurity domain alone.
 
\newpage
\begin{appendices}

\newpage
\thispagestyle{empty}
\vspace*{\fill}
\begin{center}
{\Large Appendices}
\end{center}
\vspace*{\fill}

\ifprint\newpage\blankpage\fi%

\chapter{Proofs}
\label{ap:proofs}

\begin{statement}\label{st:statement1}
For any $ k \in \mathbb{N} $, $ \alpha \in \mathbb{R} $, any $ r > 0 $ and all $ x_i \in \mathbb{R} $, it holds
\begin{equation*}
    g_{\lse}(\{x_1, \ldots, x_k\}; r) = \alpha + g_{\lse}(\{x_1 - \alpha, \ldots, x_k - \alpha\}; r) 
\end{equation*}
where 
\begin{equation*}
    g_{\lse}(\left\{ x_1, \ldots, x_k \right\}; r) =  \frac{1}{r}\log \left(\frac{1}{k} \sum_{i = 1}^{k} \exp({r\cdot x_i})\right).
\end{equation*}
\end{statement}

\begin{myproof*} 
\begin{align*}
    g_{\lse}(\{x_i\}_{i=1}^k; r) &= \frac{1}{r}\log \left(\frac{1}{k} \sum_{i = 1}^{k} \exp({r\cdot x_i})\right)
    = \frac{1}{r}\log \left(\frac{1}{k} \sum_{i = 1}^{k} \exp({r\cdot \left(   x_i + \alpha - \alpha \right)})\right) \\
                                 &= \frac{1}{r}\log \left(\exp(r\alpha)\frac{1}{k} \sum_{i = 1}^{k} \exp({r\cdot (x_i - \alpha)})\right) = \frac{1}{r}\log \exp(r\alpha) + \frac{1}{r}\log \left(\frac{1}{k} \sum_{i = 1}^{k} \exp({r\cdot (x_i - \alpha)})\right) \\ &= \alpha + \frac{1}{r}\log \left(\frac{1}{k} \sum_{i = 1}^{k} \exp({r\cdot (x_i - \alpha)})\right) = \alpha + g_{\lse}(\{x_i- \alpha\}_{i=1}^k; r)
\end{align*}
\end{myproof*}

\begin{statement}\label{st:statement5}
For any $ \beta > 0$, $p \geq 1$, and $c \in \mathbb{R} $, it holds
\begin{equation*}
g_{\pnorm}(\{x_1, \ldots, x_k\}; p, c) = \beta \cdot g_{\pnorm}(\{x_1/\beta, \ldots, x_k/\beta\}; p, c/\beta)
\end{equation*}
where
\begin{equation*}
g_{\pnorm}(\{x_1, \ldots, x_k\}; p, c) = \left(\frac{1}{k} \sum_{i = 1}^{k} \vert x_i - c \vert ^ {p} \right)^{\frac{1}{p}}.
\end{equation*}
\end{statement}
\begin{myproof*} 
    \begin{align*}
        g_{\pnorm}(\{x_i\}_{i=1}^k; p, c) &= \left(\frac{1}{k} \sum_{i = 1}^{k} \vert x_i - c \vert ^ {p} \right)^{\frac{1}{p}} = \left(\frac{1}{k} \sum_{i = 1}^{k}\frac{\beta^p}{\beta^p} \vert x_i - c \vert ^ {p} \right)^{\frac{1}{p}} \\ &= \beta\cdot \left(\frac{1}{k} \sum_{i = 1}^{k} \left\vert \frac{x_i - c}{\beta} \right\vert ^ {p} \right)^{\frac{1}{p}} = \beta \cdot g_{\pnorm}(\{x_i/\beta\}_{i=1}^k; p, c/\beta)
    \end{align*}
\end{myproof*}

\begin{statement}\label{st:statement2}
For any $ k \in \mathbb{N} $ and all $x_i \in \mathbb{R} $, it holds
\begin{equation*}
    \lim_{r \to 0} g_{\lse}(\left\{ x_1, \ldots, x_k \right\}; r) =  \lim_{r \to 0} \frac{1}{r}\log \left(\frac{1}{k} \sum_{i = 1}^{k} \exp({r\cdot x_i})\right) = g_{\mean}\left( \left\{ x_1, \ldots, x_k \right\}  \right) = \frac{1}{k} \sum_{i=1}^{k} x_i.
\end{equation*}
\end{statement}
\begin{myproof*} 
\begin{align*}
    \lim_{r \to 0} \frac{1}{r}\log \left(\frac{1}{k} \sum_{i = 1}^{k} \exp({r\cdot x_i})\right) &\overset{\shortstack{\tiny L'Hospital's,\\ \tiny Rule}}{=}\lim_{r \to 0} \frac{\sum_{i=1}^k x_i\exp(r \cdot x_i)}{\sum_{i=1}^k \exp(r \cdot x_i)} = \frac{1}{k}\sum_{i=1}^k x_i
\end{align*}
\end{myproof*}

\begin{statement}\label{st:statement3}
For any $ k \in \mathbb{N} $ and all $x_i \in \mathbb{R} $, it holds
\begin{equation*}
    \lim_{r \to \infty} g_{\lse}(\left\{ x_1, \ldots, x_k \right\}; r) =  \lim_{r \to \infty} \frac{1}{r}\log \left(\frac{1}{k} \sum_{i = 1}^{k} \exp({r\cdot x_i})\right) = g_{\max}\left( \left\{ x_1, \ldots, x_k \right\}  \right) = \max_{i=1, \ldots, k} x_i.
\end{equation*}
\end{statement}
\begin{myproof*} Using Statement~\ref{st:statement1}
\begin{align*}
    \lim_{r \to 0} \frac{1}{r}\log \left(\frac{1}{k} \sum_{i = 1}^{k} \exp({r\cdot x_i})\right) &= \max_{i=1, \ldots, k} x_i + \underbrace{\lim_{r \to \infty} \frac{1}{r}\log \frac{1}{k}}_{=0} + \lim_{r \to \infty}  \frac{1}{r}\log \left(\sum_{i = 1}^{k} \exp({r\cdot (\underbrace{x_i - \max_{i=1, \ldots, k} x_i}_{\le 0})})\right) \\ &= \max_{i=1, \ldots, k} x_i + \underbrace{\lim_{r \to \infty}  \frac{1}{r}\log (\exp(1))}_{=0} = \max_{i=1, \ldots, k} x_i 
\end{align*}
\end{myproof*}

\begin{statement}\label{st:statement4}
For any $ k \in \mathbb{N} $, $ c \in \mathbb{R} $ and all $x_i \in \mathbb{R} $, it holds
\begin{equation*}
    \lim_{p \to \infty} g_{\pnorm}(\left\{ x_1, \ldots, x_k \right\}; p, c) = \max_{i=1, \ldots, k}\lvert x_i - c \rvert
\end{equation*}
where
\begin{equation*}
g_{\pnorm}(\{x_1, \ldots, x_k\}; p, c) = \left(\frac{1}{k} \sum_{i = 1}^{k} \vert x_i - c \vert ^ {p} \right)^{\frac{1}{p}}.
\end{equation*}
\end{statement}

\begin{myproof*} Assume that $ \widetilde{x} = \max\limits_{i=1, \ldots, k} \vert x_i - c \vert $. Then for $ p \geq 1 $, 
\begin{align*}
    \left(\frac{1}{k} \sum_{i = 1}^{k} \vert x_i - c \vert ^ {p} \right)^{\frac{1}{p}} \leq \left(\frac{1}{k} \sum_{i=1}^k \widetilde{x} ^ {p} \right)^{\frac{1}{p}} = \left(\frac{1}{k} k \widetilde{x} ^ {p} \right)^{\frac{1}{p}} = \widetilde{x}
\end{align*}

and

\begin{align*}
    \left(\frac{1}{k} \sum_{i = 1}^{k} \vert x_i - c \vert ^ {p} \right)^{\frac{1}{p}} \geq \left(\frac{1}{k} \widetilde{x} ^ {p} \right)^{\frac{1}{p}} = \left(\frac{1}{k}\right)^{\frac{1}{p}} \widetilde{x}
\end{align*}

Using the \emph{squeeze theorem}

\begin{align*}
    \lim_{p \to \infty} \left(\frac{1}{k}\right)^{\frac{1}{p}} \widetilde{x}  = \widetilde{x} \leq \lim_{p \to \infty} g_{\pnorm}(\left\{ x_i\right\}_{i=1}^k; p, c) \leq\lim_{p \to \infty} \widetilde{x} = \widetilde{x}
\end{align*}
\end{myproof*}

\chapter{IoT device identification (use case)}
\label{ap:deviceid}

\begin{figure}[htp]
    \includegraphics[width=\textwidth,height=0.95\textheight,keepaspectratio]{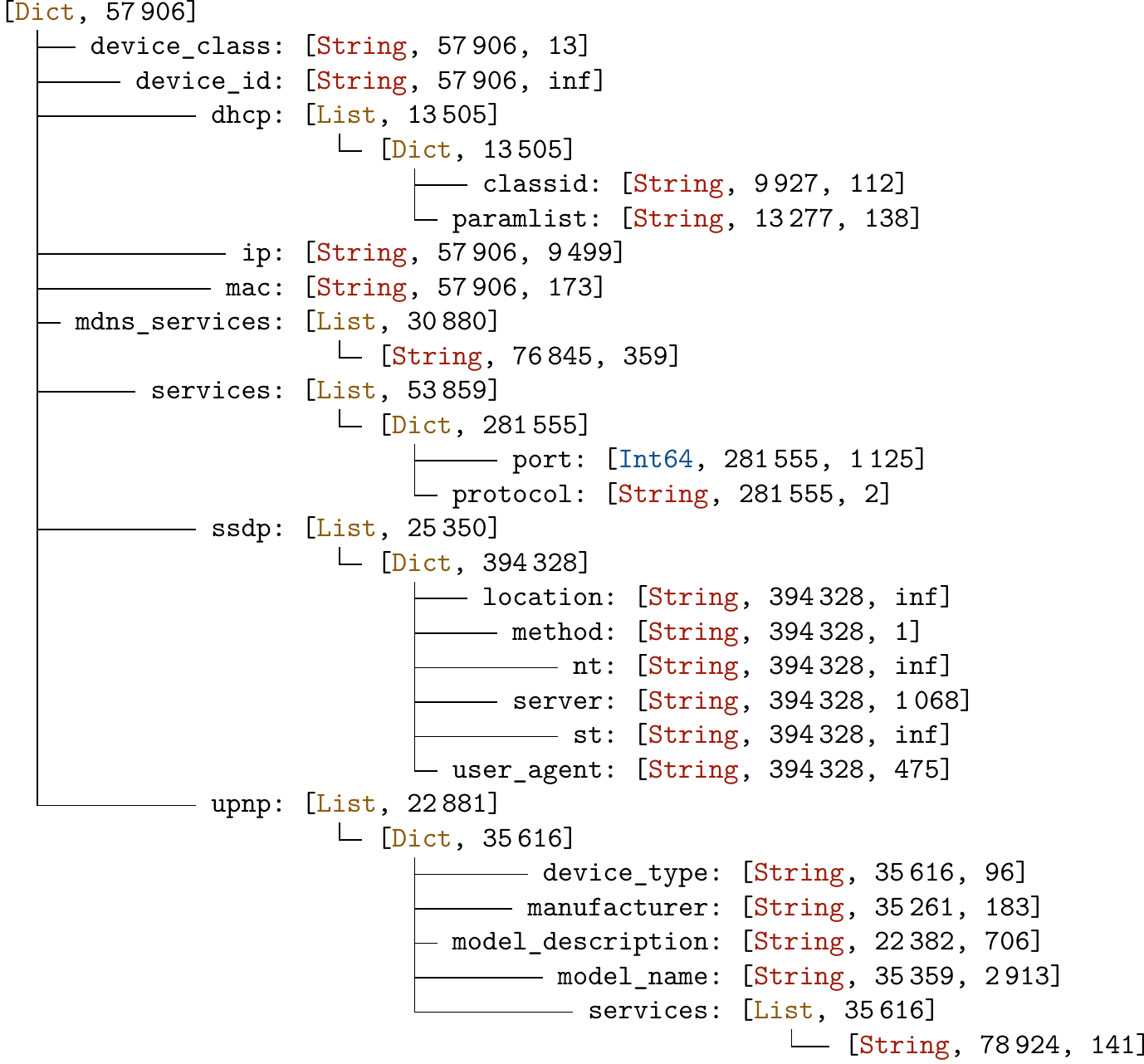}
    \caption[Schema of JSON documents from the \emph{IoT device identification} problem.]{A (shortened) schema deduced from $ 57\,906$ training \gls{json} documents for the device identification task from the dataset downloaded from \emph{Kaggle.} Each entry is accompanied by at most two numbers. The first one specifies the number of times the entry was updated summed over all samples (for a single sample, this can range from zero in case of missing values to any natural number in the presence of arrays) and the second number is the number of unique values of the given field for identification of categorical variables. This is not measured for composite entries, such as dictionaries (\gls{json} objects) or lists (\gls{json} arrays). If there are too many unique values, \texttt{JsonGrinder.jl} stops the count to save memory and represents this with $ \texttt{inf} $. From this schema we can learn, for example, that only \texttt{device\_class}, \texttt{device\_id}, \texttt{ip} and \texttt{mac} fields are specified for every sample, that \texttt{dhcp} entry (if specified) always contains an array of a single element, and all \gls{json} arrays representing \texttt{ssdp} scans consist of \gls{json} objects with no missing keys.}%
    \label{fig:deviceid_schema}
\end{figure}

\begin{figure}[htp]
    \includegraphics[width=\textwidth,height=0.95\textheight,keepaspectratio]{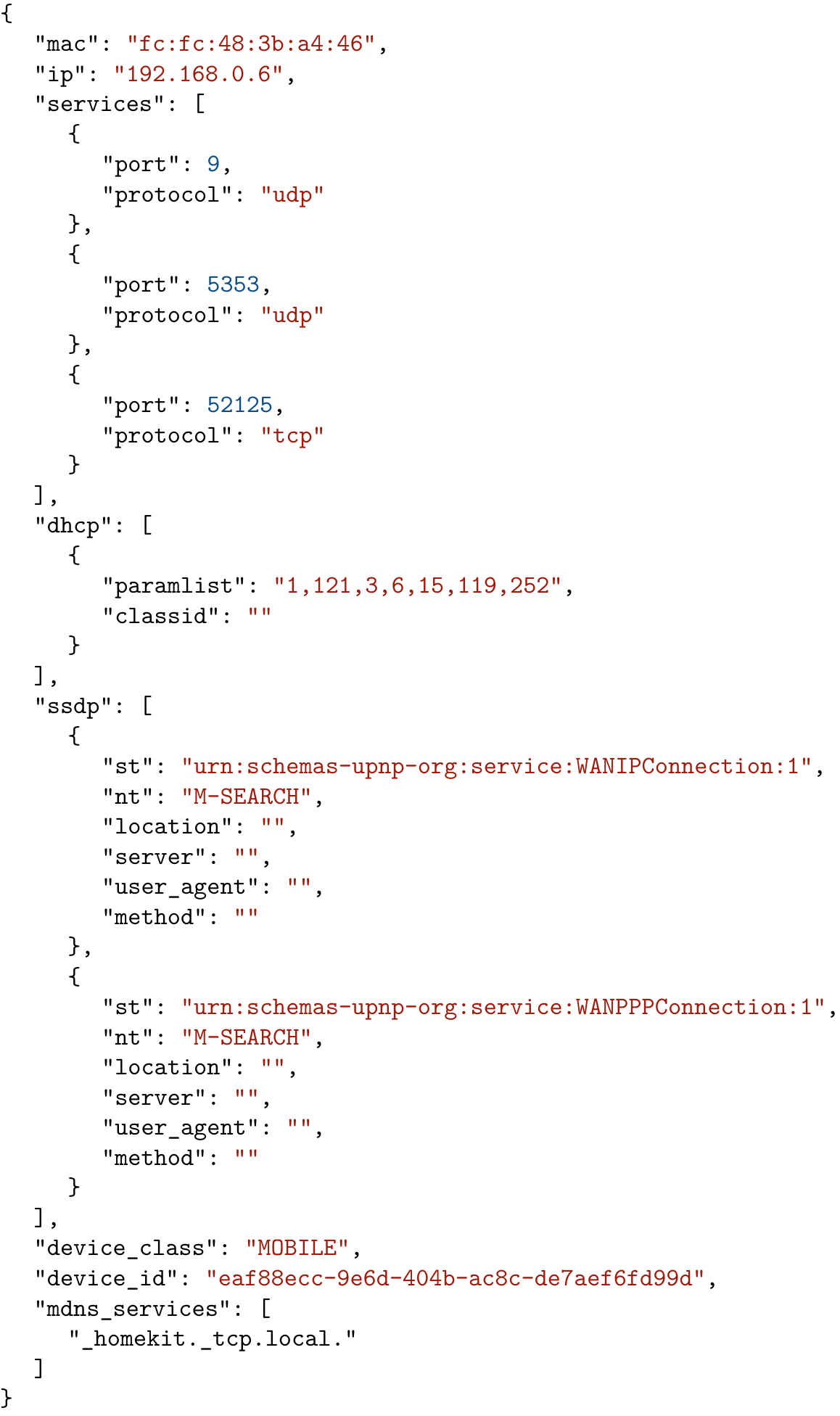}
    \caption[One JSON representation of an IoT device.]{An example of one observation for the device identification task.}%
    \label{fig:deviceid_sample2}
\end{figure}

\begin{figure}[htp]
    \includegraphics[width=\textwidth,height=0.95\textheight,keepaspectratio]{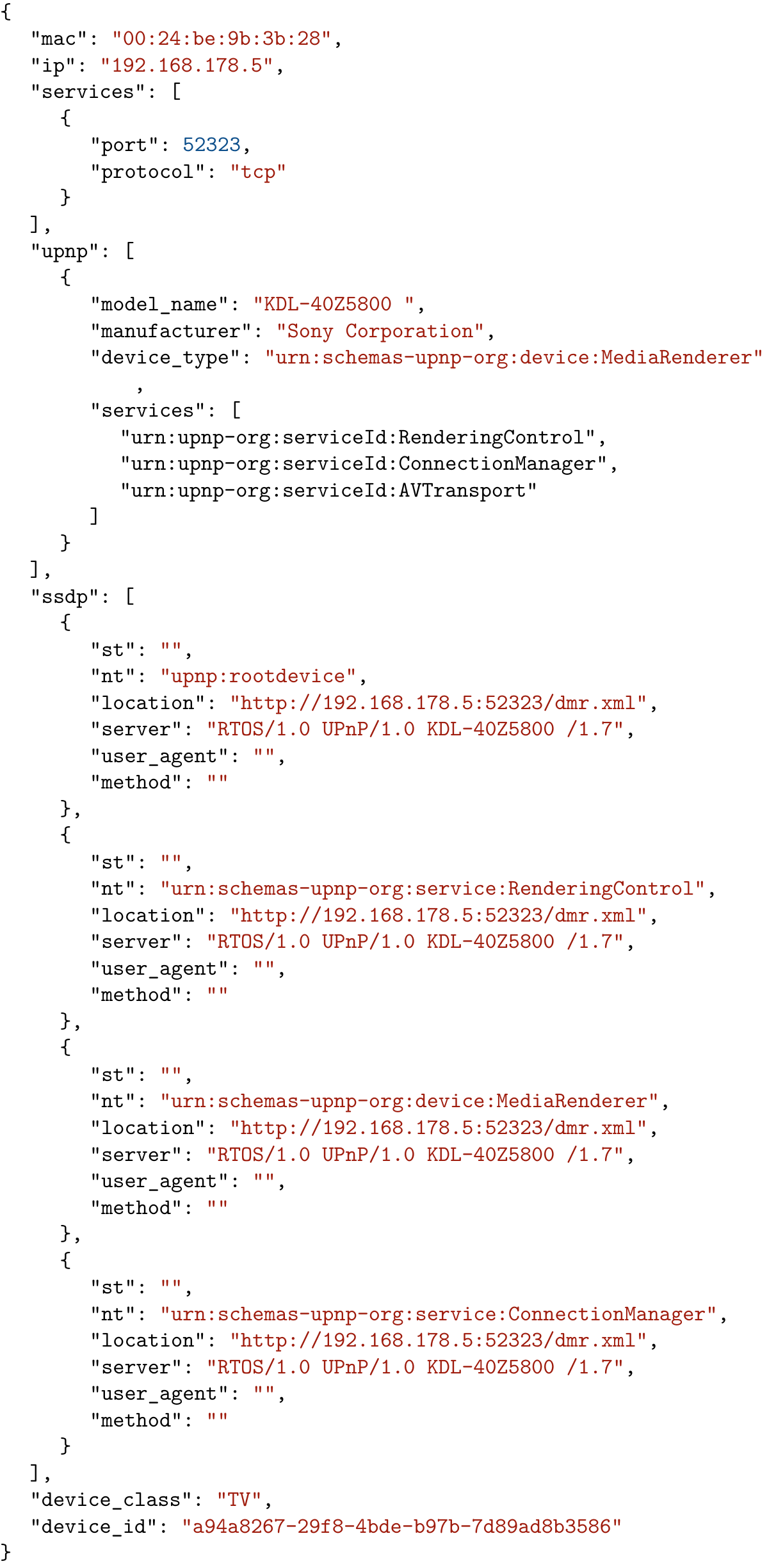}
    \caption[Another JSON representation of an IoT device.]{An example of one observation for the device identification task.}%
    \label{fig:deviceid_sample1}
\end{figure}

\chapter{Behavior-based malware classification in graphs (use case)}
\label{ap:idp}

\vfill

\begin{figure}[h]
    \centering
    \includegraphics[angle=270, width=\textwidth]{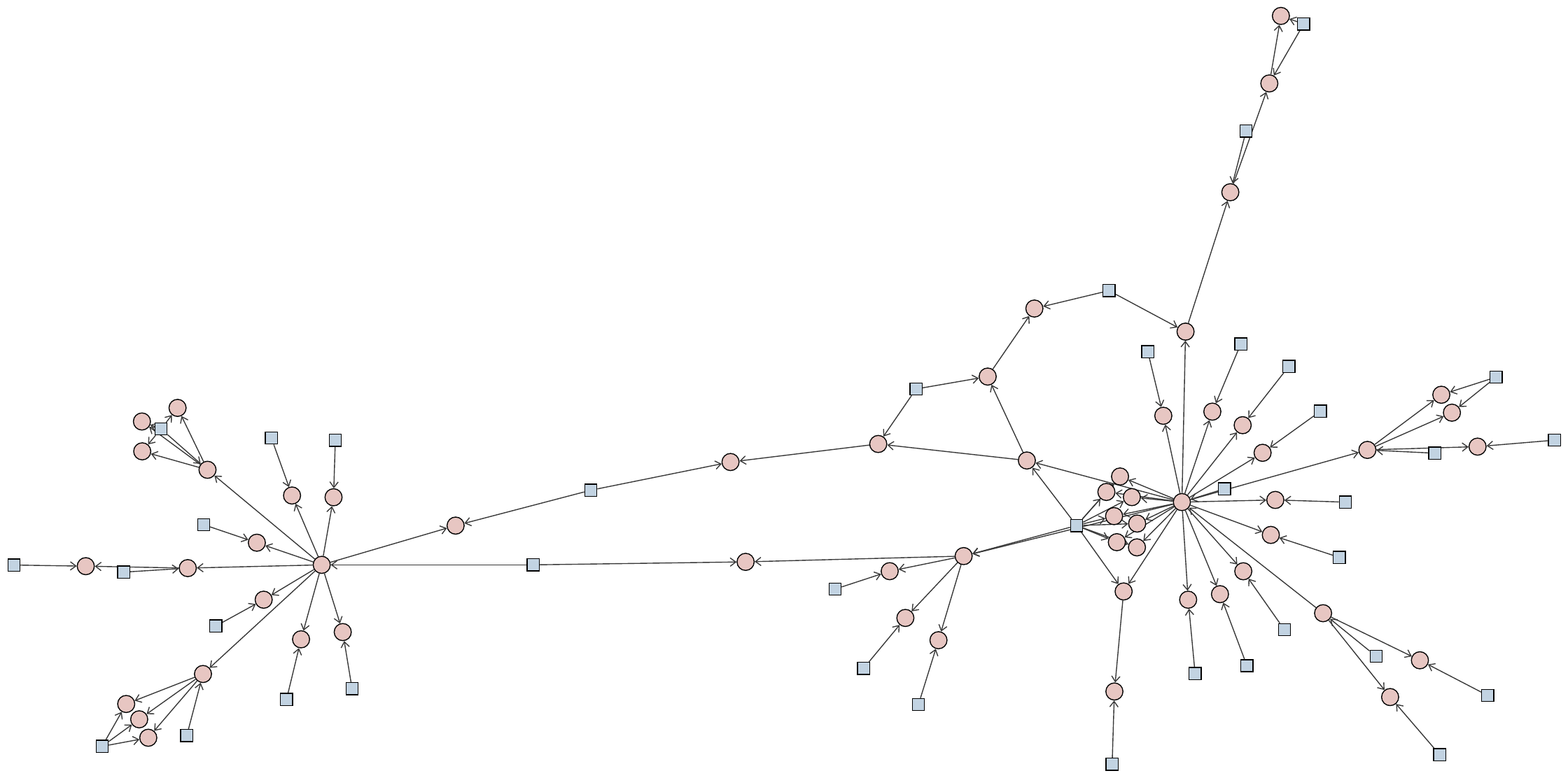}
    \caption[IDP graph example.]{A drawing of the largest (weakly) connected component in one of the \gls{idp} graphs from the training set. For demonstration purposes, we cherry-picked one of the smallest graphs with a smaller number of nodes and smaller degree values. Blue squares represent executable nodes and red circles process nodes. This figure should provide a rough idea of the structure of \gls{idp} graphs. For a more detailed description of smaller subgraphs, refer to Figures~\ref{fig:idp_g2} and~\ref{fig:idp_g1}. Process nodes with large degrees in the middle of most prominent `clusters' in this particular graph correspond to \texttt{explorer.exe} (left) and \texttt{services.exe} (right) processes.}%
    \label{fig:whole_idp}
\end{figure}

\vfill
\newpage
\vfill

\begin{figure}[p]
    \centering
    \includegraphics[width=\textwidth,height=0.95\textheight,keepaspectratio]{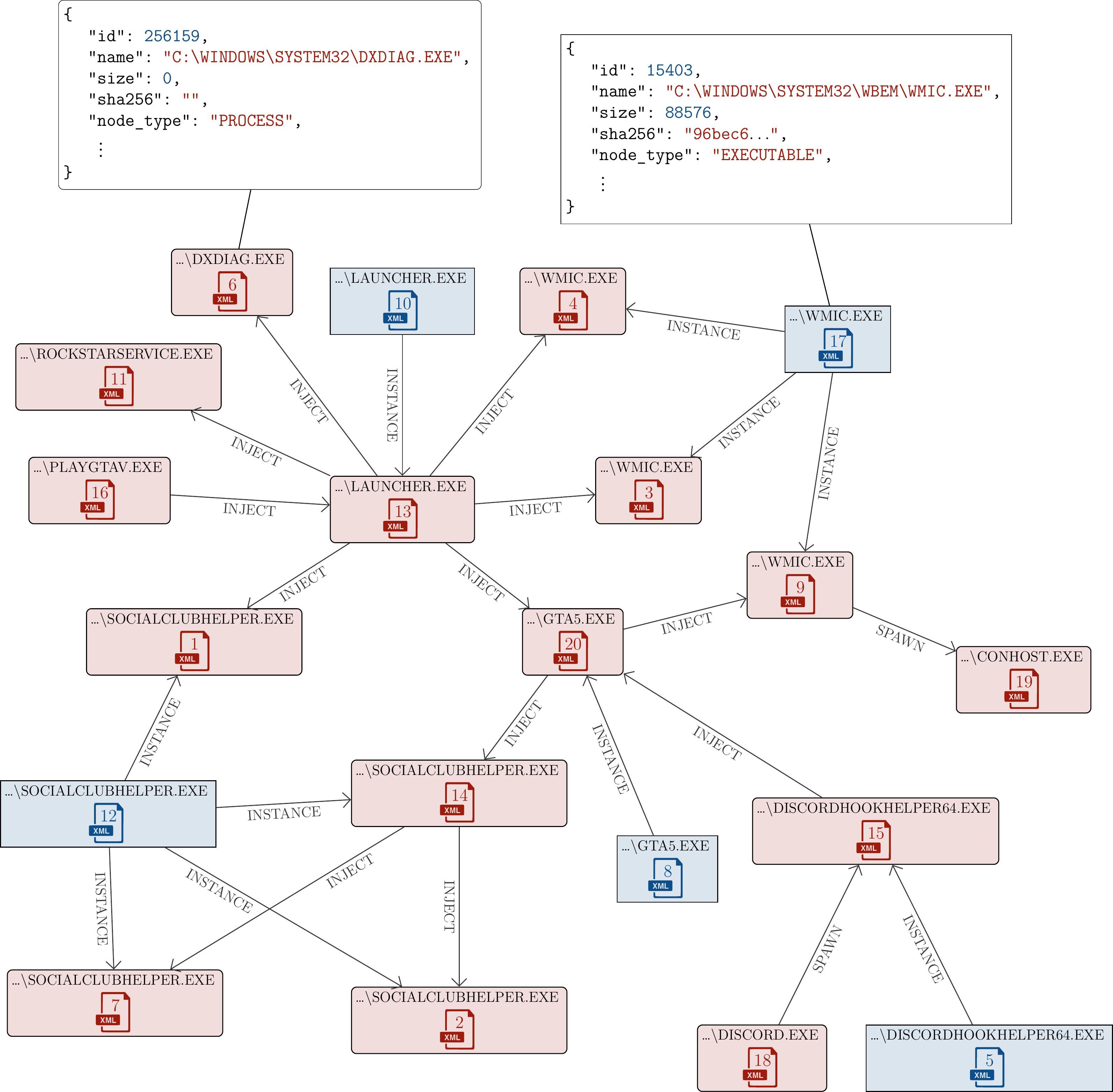}
    \caption[Example of an IDP graph describing an uninfected operating system.]{One of the (weakly) connected components of an \gls{idp} graph representation of the system.  Red rectangles with rounded corners represent processes, and blue rectangles represent executable files. The document icon signifies that each vertex is further described by an \gls{xml} document. For two nodes (number $ 6 $ and  $ 17 $), the documents are shown specifically. To remain consistent with the rest of the thesis, we use the equivalent \gls{json} representation of these documents (see Section~\ref{sub:graph_based_detection}). Fully qualified names are in Figure~\ref{fig:g2l}. This connected component captures a launch of the \texttt{Grand Theft Auto 5} computer game. In this case, there is no ongoing malicious activity---\emph{INJECT} relationships usually signify the presence of an external (anti)cheat tool.}%
    \label{fig:idp_g2}
\end{figure}

\vfill
\newpage
\vfill

\begin{figure}[p]
    \centering
    \includegraphics[width=\textwidth,height=0.95\textheight,keepaspectratio]{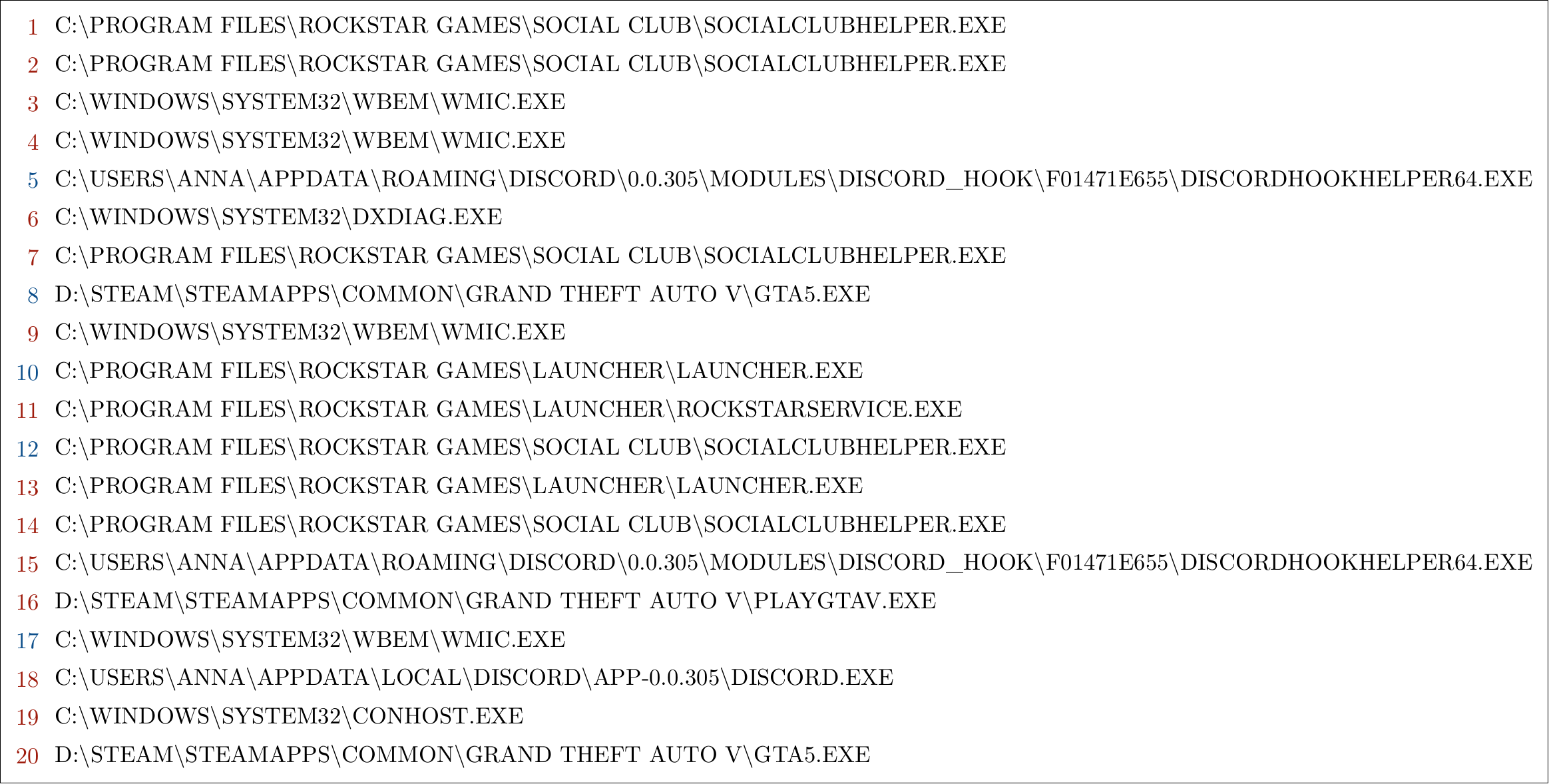}
    \caption[Legend for the second example.]{Fully qualified names of executables and processes in Figure~\ref{fig:idp_g1}. Generic username \texttt{"ANNA"} was used for anonymization of the original user.}%
    \label{fig:g2l}
\end{figure}

\vfill
\newpage
\vfill

\begin{figure}[p]
    \centering
    \includegraphics[width=\textwidth,height=0.9\textheight,keepaspectratio]{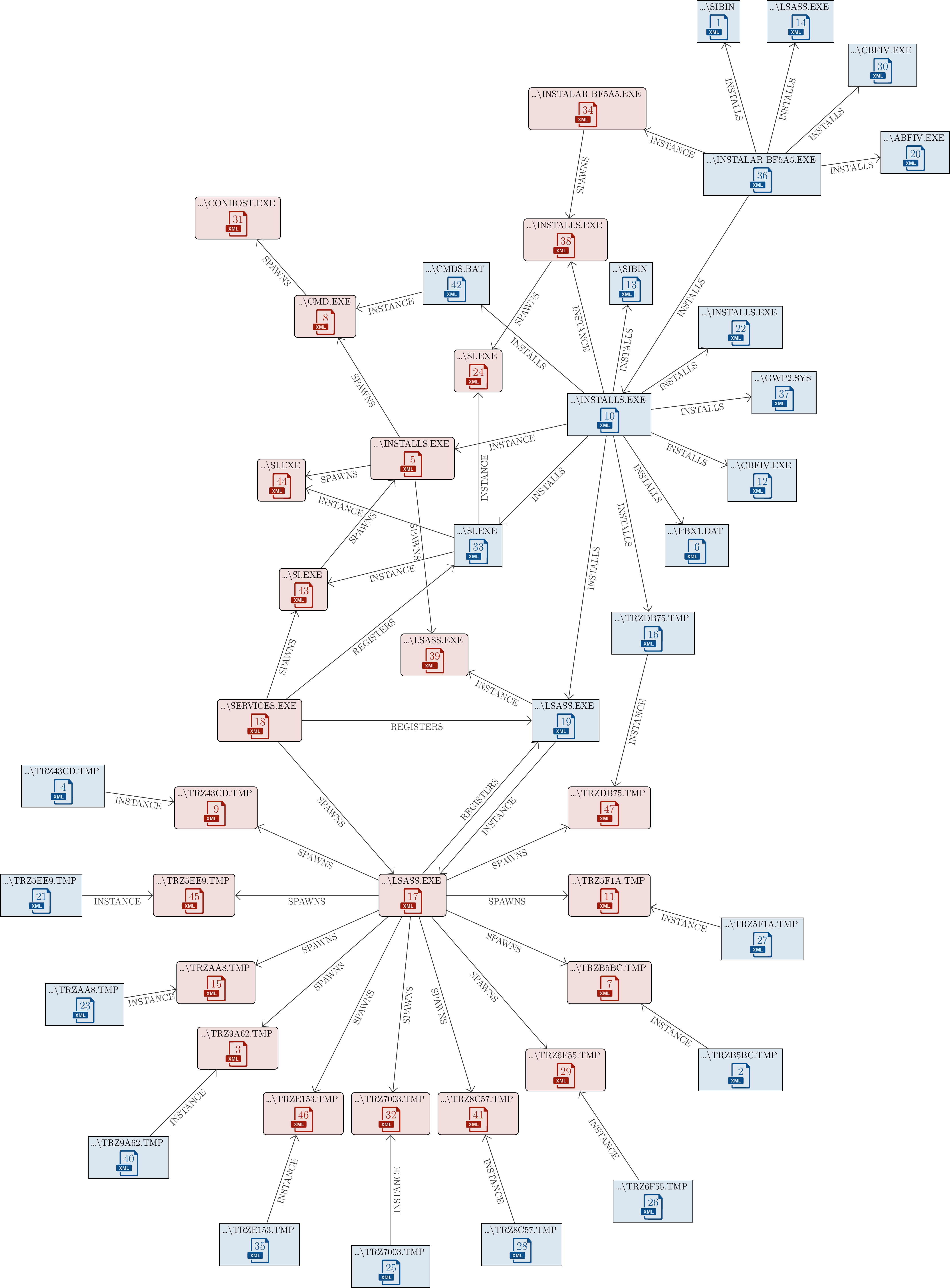}
    \caption[Example of an IDP graph describing an operating infected system.]{One (weakly) connected component of an \gls{idp} graph representation of a system most probably infected through \texttt{EternalBlue} exploit. Red rectangles with rounded corners represent processes, and blue rectangles represent executable files. The document icon signifies that each vertex is further described by an \gls{xml} document. Fully qualified names are in Figure~\ref{fig:g1l}. In this specific example, malicious payload is present in node number {\color{cvutblue}$19$}. The binary mimicks the standard \texttt{\gls{lsass}} system file. However, the malicious binary is stored in a different directory than \texttt{\%WINDIR\%\textbackslash System32}.}%
    \label{fig:idp_g1}
\end{figure}

\vfill
\newpage
\vfill

\begin{figure}[p]
    \centering
    \includegraphics[width=\textwidth,height=0.95\textheight,keepaspectratio]{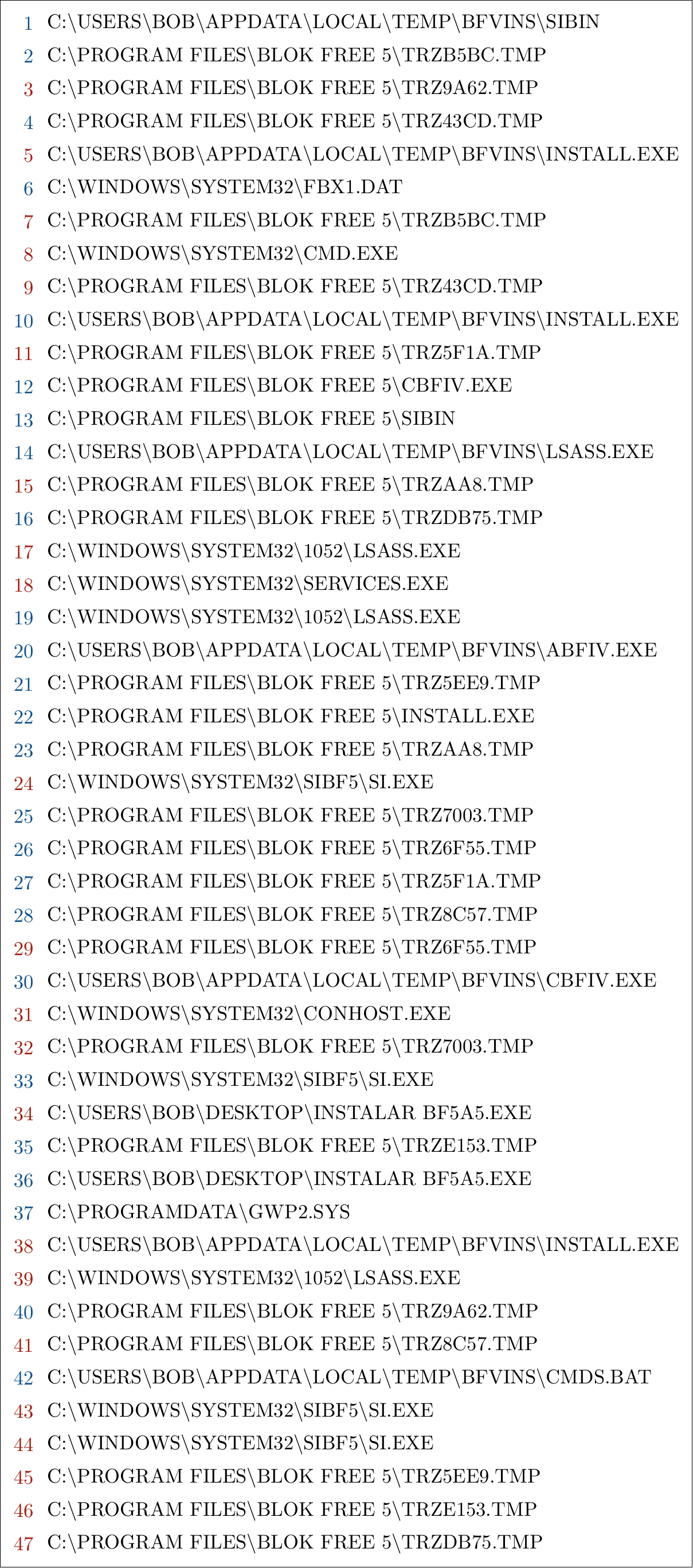}
    \caption[Legend for the first example.]{Fully qualified names of executables and processes in Figure~\ref{fig:idp_g1}. Generic username \texttt{"BOB"} was used for anonymization of the original user.}%
    \label{fig:g1l}
\end{figure}

\vfill
\newpage
\vfill

\begin{figure}[p]
    \includegraphics[width=\textwidth,height=0.95\textheight,keepaspectratio]{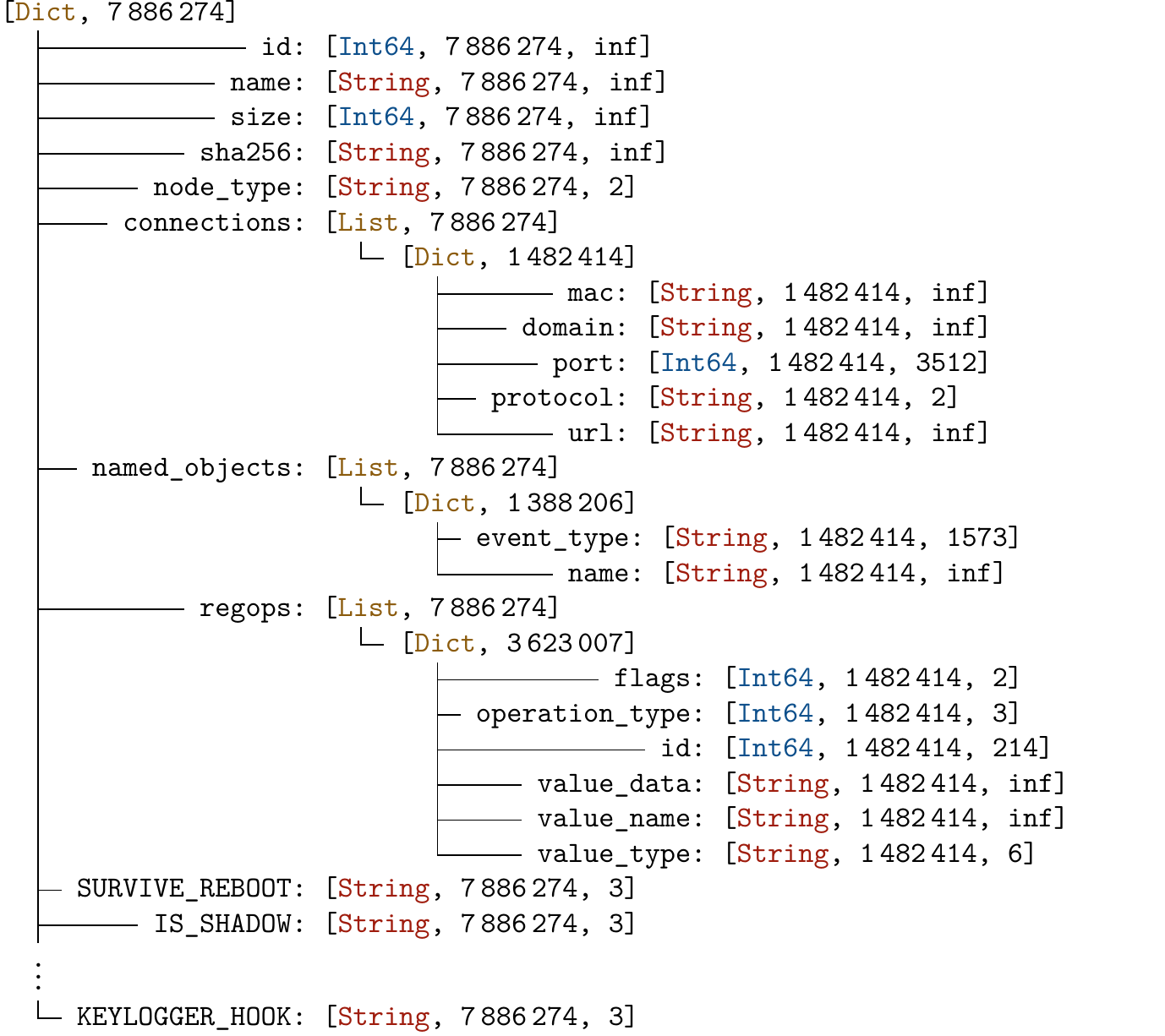}
    \caption[Schema of a vertex data description in IDP graphs.]{A (shortened) schema inferred from $ 7\,886\,274 $ vertices present in graphs in the dataset. See Figure~\ref{fig:deviceid_schema} for the exact interpretation of the format. From the schema we can deduce that missing data is not an issue for \gls{idp} graphs, since all top-level entries are updated same number of times. The keys describing individual characteristics are shortened for brevity.}%
    \label{fig:idp_schema}
\end{figure}

\vfill
\newpage
\vfill

\begin{table}[p]\small
    \caption[Examples of IDP graph characteristics.]{Examples of characteristics extracted from patents~\cite{patent_detection, patent_removal}. Note that some characteristics function as controls for the whole engine. We filtered these out.}%
    \label{tab:characteristics}
    \centering
    {\renewcommand\arraystretch{1.25}
        \begin{tabular}{cp{11.5cm}}
            \toprule
            name & description \\
            \midrule
            \emph{SURVIVE\_REBOOT} & The executable is configured to automatically restart. Malicious programs generally need to survive reboot in order to be effective at stealing information from the user. \\
            \emph{WINDOW\_NOT\_VISIBLE} & The executable does not display a window on the screen. This implies that the program is trying to prevent user from seeing its activity. \\
            \emph{PROCESS\_IS\_HIDDEN} & The process is hidden from the \texttt{Windows Task Manager}. A malicious program can use this technique to remain hidden from the user. \\
            \emph{SMALL\_IMAGE\_SIZE} & The size of the executable file image is very small. Malicious programs try to be stealthy, and one way to remain invisible is to minimize the impact on the underlying system. \\

            \emph{WRITES\_TO\_WINDIR} & The executable attempted to write to the \texttt{Windows} directory. Often, malicious programs install themselves in the \texttt{Windows} directory, as that directory contains many executables, and it is easy to remain unnoticed there. \\
            \emph{WRITES\_TO\_PGM\_FILES} & The executable attempted to write to the \texttt{Program Files} directory. Some malicious programs (particularly adware) install themselves in this directory. \\
            \emph{EXEC\_FROM\_CACHE} & The executable is executed from a cached area. \\
            \emph{EXEC\_FROM\_WINDIR} & The executable is executed from the \texttt{Windows} directory. \\
            \emph{EXEC\_FROM\_PGM\_FILES} & The executable is executed from the \texttt{Program Files} directory. \\
            \emph{IS\_SHADOW} & The executable has the same name as a legitimate executable. This is evidence of a common mechanism that trojans and other malicious code use to hide themselves on a computer. \\
            \emph{P2P\_CODE\_INJECTION} & The executable attempted to inject code into the address space of another process. This is generally evidence of malicious activity as the injected code could be the malicious payload or a rootkit trying to hide the real malicious process from detection. \\
            \emph{TURNS\_OFF\_FIREWALL} & The executable attempted to turn off the \texttt{Windows firewall}. \\
            \emph{HAS\_DOUBLE\_EXTENSION} & The file name of the executable has a double extension. \texttt{Windows} is configured by default to hide known file extensions and this might fool an unsuspecting user into opening a harmful executable. \\
            \emph{TERMINATE\_PROCESS} & The executable terminates another running process. Some malicious programs attempt to terminate security programs (such as anti-virus, anti-spyware) running on the machine in order to avoid detection. \\
            \emph{KEYLOGGER\_HOOK} & The executable attempted to install a keylogger by a legitimate mechanism. Malicious programs install keyloggers to capture keystrokes and steal logins, passwords and credit card numbers.\\

            \emph{MODIFIES\_HOSTS\_FILE} & The executable attempted to modify the \texttt{hosts} file. \\

            \emph{MODIFIES\_AUTOEXEC\_BAT} & The executable attempted to modify the \texttt{autoexec.bat} file. \\

            \emph{MODIFIES\_CONFIG\_SYS} & The executable attempted to modify the default set of drivers loaded at startup time. \\

            \emph{INSTALLED\_VIA\_IM} & The executable was installed by an instant messaging program. \\

            \emph{INSTALLED\_VIA\_EMAIL} & The executable was installed via an email reader. \\

            \emph{INSTALLED\_VIA\_BROWSER} & The executable was installed by a browser. \\

            \bottomrule
        \end{tabular}
    }
\end{table}

\clearpage

\ifprint\blankpage\fi%

\chapter{Modelling Internet communication (use case)}%
\label{ap:cisco}

\begin{table}[H]
    \centering
    \caption[Sizes of vertex sets in the \emph{Modelling Internet communication} use case.]{Sizes of vertex sets sorted by date, when bipartite graphs were observed. Here, $\mathcal{A}_d$ in the second column denotes all vertices assembled from all bipartite graphs. The rest of the columns specify sizes of `right' sets in bipartite graphs. $\mathcal{B}_c$ denotes clients, $\mathcal{B}_b$ binary files, $\mathcal{B}_{IP}$ \gls{ip} addresses, $\mathcal{B}_{ta}$ \gls{tls} issuer authorities, $\mathcal{B}_{th}$ \gls{tls} certificate hashes, $\mathcal{B}_{ti}$ \gls{tls} issue times, $\mathcal{B}_{we}$ WHOIS entry emails, $\mathcal{B}_{wn}$ WHOIS nameservers, $\mathcal{B}_{wr}$ WHOIS registrar names, $\mathcal{B}_{wc}$ WHOIS country, $\mathcal{B}_{wi}$ WHOIS registrar id, and finally, $\mathcal{B}_{wt}$ WHOIS timestamp.}%
    \label{tab:vertex_sizes}
    \renewcommand{\cellalign}{l}
    \resizebox{\textwidth}{!}{%
        \begin{tabular}{*{14}{c}}
            \toprule
            date & $|\mathcal{A}_d|$ & $|\mathcal{B}_c|$ & $|\mathcal{B}_b|$ & $|\mathcal{B}_{IP}|$& $|\mathcal{B}_{ta}|$& $|\mathcal{B}_{th}|$& $|\mathcal{B}_{ti}|$& $|\mathcal{B}_{we}|$& $|\mathcal{B}_{wn}|$& $|\mathcal{B}_{wr}|$& $|\mathcal{B}_{wc}|$& $|\mathcal{B}_{wi}|$ & $|\mathcal{B}_{wt}|$ \\
            \midrule
            \texttt{05-23} & $710\,167$ & $3\,419\,758$ & $245\,999$ & $2\,357\,441$ & $6\,217$ & $374\,531$ & $160\,343$ & $18\,537$ & $15\,972$ & $1\,111$ & $160$ & $959$ & $901$ \\
            \texttt{06-03} & $631\,828$ & $3\,151\,105$ & $232\,507$ & $2\,247\,940$ & $5\,631$ & $337\,558$ & $142\,944$ & $16\,442$ & $14\,803$ & $989$ & $157$ & $849$ & $901$ \\
            \texttt{06-10} & $644\,868$ & $3\,140\,231$ & $236\,052$ & $2\,268\,276$ & $5\,745$ & $348\,298$ & $148\,844$ & $16\,834$ & $14\,913$ & $969$ & $160$ & $833$ & $904$ \\
            \texttt{06-17} & $630\,532$ & $3\,114\,985$ & $242\,696$ & $2\,305\,323$ & $5\,618$ & $340\,934$ & $145\,032$ & $16\,808$ & $14\,973$ & $945$ & $161$ & $806$ & $904$ \\
            \texttt{06-26} & $648\,960$ & $3\,175\,500$ & $238\,535$ & $2\,316\,943$ & $5\,732$ & $350\,155$ & $149\,609$ & $17\,519$ & $15\,454$ & $954$ & $160$ & $805$ & $902$ \\
            \texttt{06-27} & $616\,958$ & $3\,106\,120$ & $240\,931$ & $2\,256\,445$ & $5\,484$ & $334\,996$ & $142\,453$ & $16\,468$ & $14\,942$ & $909$ & $158$ & $763$ & $901$ \\
            \texttt{07-01} & $613\,601$ & $3\,055\,840$ & $240\,909$ & $2\,277\,775$ & $5\,433$ & $332\,991$ & $141\,471$ & $16\,373$ & $14\,866$ & $894$ & $157$ & $743$ & $897$ \\
            \texttt{07-08} & $557\,253$ & $2\,894\,017$ & $224\,571$ & $2\,117\,016$ & $5\,077$ & $307\,541$ & $129\,249$ & $14\,319$ & $13\,756$ & $810$ & $160$ & $665$ & $890$ \\
            \texttt{07-15} & $608\,327$ & $2\,943\,900$ & $241\,631$ & $2\,222\,769$ & $5\,504$ & $330\,853$ & $140\,718$ & $16\,168$ & $14\,834$ & $861$ & $158$ & $717$ & $899$ \\
            \texttt{07-22} & $600\,814$ & $2\,929\,311$ & $239\,520$ & $2\,193\,113$ & $5\,423$ & $326\,592$ & $138\,604$ & $16\,214$ & $14\,764$ & $842$ & $156$ & $693$ & $898$ \\
            \texttt{07-29} & $588\,039$ & $2\,873\,682$ & $237\,436$ & $2\,162\,197$ & $5\,194$ & $320\,663$ & $135\,980$ & $15\,785$ & $14\,475$ & $806$ & $150$ & $656$ & $898$ \\
            \bottomrule
    \end{tabular}}
\end{table}

\begin{table}[H]
    \centering
    \caption[Numbers of edges in bipartite graphs in the \emph{Modelling Internet communication} use case.]{Numbers of edges in bipartite graphs sorted by date, when the graph was observed. The meaning of the abbreviations in subscripts is the same as in Table~\ref{tab:vertex_sizes}. Note that \gls{tls} issuer authority, \gls{tls} certificate hash and \gls{tls} issue time contain the same number of edges for each of the dates. This is not caused by a typo, but by the fact, that these three relation types, if obtainable, were obtained together.}%
    \label{tab:edge_sizes}
    \renewcommand{\cellalign}{l}
\resizebox{\textwidth}{!}{%
    \begin{tabular}{*{13}{c}}
    \toprule
    date & $|\mathcal{E}_c|$ & $|\mathcal{E}_b|$ & $|\mathcal{E}_{IP}|$& $|\mathcal{E}_{ta}|$& $|\mathcal{E}_{th}|$& $|\mathcal{E}_{ti}|$& $|\mathcal{E}_{we}|$& $|\mathcal{E}_{wn}|$& $|\mathcal{E}_{wr}|$& $|\mathcal{E}_{wc}|$& $|\mathcal{E}_{wi}|$ & $|\mathcal{E}_{wt}|$ \\
    \midrule
    \texttt{05-23} & $334\,320\,003$ & $4\,073\,050$ & $9\,314\,998$ & $465\,879$ & $465\,879$ & $465\,879$ & $44\,613$ & $122\,231$ & $42\,804$ & $39\,506$ & $46\,589$ & $46\,603$ \\
    \texttt{06-03} & $292\,659\,940$ & $3\,639\,635$ & $8\,356\,254$ & $419\,035$ & $419\,035$ & $419\,035$ & $38\,974$ & $108\,077$ & $37\,349$ & $34\,538$ & $40\,690$ & $40\,704$ \\
    \texttt{06-10} & $299\,532\,603$ & $3\,785\,134$ & $8\,902\,097$ & $430\,878$ & $430\,878$ & $430\,878$ & $39\,766$ & $110\,423$ & $38\,213$ & $35\,321$ & $41\,543$ & $41\,555$ \\
    \texttt{06-17} & $293\,466\,696$ & $3\,813\,715$ & $8\,803\,587$ & $420\,926$ & $420\,926$ & $420\,926$ & $39\,322$ & $109\,466$ & $37\,741$ & $34\,906$ & $41\,072$ & $41\,088$ \\
    \texttt{06-26} & $306\,114\,224$ & $3\,846\,992$ & $9\,141\,868$ & $431\,772$ & $431\,772$ & $431\,772$ & $41\,035$ & $114\,242$ & $39\,370$ & $36\,328$ & $42\,832$ & $42\,845$ \\
    \texttt{06-27} & $291\,065\,433$ & $3\,821\,056$ & $8\,740\,732$ & $412\,836$ & $412\,836$ & $412\,836$ & $38\,296$ & $107\,202$ & $36\,775$ & $33\,943$ & $39\,996$ & $40\,009$ \\
    \texttt{07-01} & $285\,920\,746$ & $3\,789\,277$ & $8\,649\,546$ & $410\,419$ & $410\,419$ & $410\,419$ & $37\,797$ & $106\,055$ & $36\,256$ & $33\,520$ & $39\,480$ & $39\,494$ \\
    \texttt{07-08} & $251\,616\,117$ & $3\,361\,849$ & $7\,917\,765$ & $376\,934$ & $376\,934$ & $376\,934$ & $33\,028$ & $936\,23$ & $317\,36$ & $292\,40$ & $345\,25$ & $345\,34$ \\
    \texttt{07-15} & $278\,227\,235$ & $3\,693\,630$ & $8\,869\,060$ & $406\,339$ & $406\,339$ & $406\,339$ & $37\,536$ & $105\,356$ & $36\,008$ & $33\,288$ & $39\,180$ & $39\,193$ \\
    \texttt{07-22} & $278\,975\,786$ & $3\,765\,391$ & $8\,627\,468$ & $400\,566$ & $400\,566$ & $400\,566$ & $37\,407$ & $104\,908$ & $35\,844$ & $33\,015$ & $39\,017$ & $39\,031$ \\
    \texttt{07-29} & $274\,507\,944$ & $3\,701\,055$ & $8\,475\,127$ & $392\,479$ & $392\,479$ & $392\,479$ & $36\,473$ & $102\,514$ & $35\,002$ & $32\,228$ & $38\,021$ & $38\,034$ \\
    \bottomrule
    \end{tabular}}
\end{table}

\begin{table}[htb]
    \centering
    \caption[Numbers of malicious domains in the \emph{Modelling Internet communication} use case.]{Numbers of malicious domains in each of the datasetes. In the first row, there is the number of domains that are both in the blacklist and in the observed dataset, and in the second row, we give the ratio of the number of blacklisted domains to the total number of domains in the dataset.}%
    \label{tab:mal_numbers}
    \renewcommand{\cellalign}{l}
\resizebox{\textwidth}{!}{%
    \begin{tabular}{c|*{13}{c}}
    \toprule
    & \texttt{05-23} &\texttt{06-03} &\texttt{06-10} &\texttt{06-17} &\texttt{06-24}&\texttt{06-26} &\texttt{06-27} &\texttt{07-01} &\texttt{07-08} &\texttt{07-15} &\texttt{07-22} & \texttt{07-29} \\
            \midrule
        $ \lvert L \cap \mathcal{A} \rvert  $ & $656$ & $ 578 $ & $ 535 $ & $ 557 $& $ 538 $ & $ 557 $ & $ 552 $ & $ 555 $& $ 512 $& $ 557 $& $ 577 $& $ 548 $\\
        $ \lvert L \cap \mathcal{A} \rvert / {\lvert \mathcal{A} \rvert} $ & $ 0.924 $\textperthousand & $0.915$\textperthousand & $ 0.830 $\textperthousand & $ 0.883 $\textperthousand & $ 0.865 $\textperthousand& $ 0.858 $\textperthousand& $ 0.895 $\textperthousand& $ 0.904 $\textperthousand& $ 0.919 $\textperthousand& $ 0.916 $\textperthousand& $ 0.960 $\textperthousand & $ 0.932 $\textperthousand \\
    \bottomrule
    \end{tabular}}
\end{table}

\begin{table}[H]
    \centering
    \caption[All relations used in the \emph{Modelling Internet communication} use case.]{All relations used in our experiments, with their cardinality type and several examples of specific relation pairs (edges in a bipartite graph). Each edge is specified as $ (u; v) $ to avoid confusion as commas may be used in $ u $ or  $ v $. \gls{m2m} stands for many-to-many and \gls{m2o} for many-to-one cardinality type. Longer names are for shortened using ellipsis (\ldots).}%
    \label{tab:relations}
    \renewcommand{\cellalign}{l}
    \resizebox{\textwidth}{!}{%
        \begin{tabular}{ccl}
            \toprule
            name (\emph{domain}-*) & card. type & examples \\
            \midrule
            *-\emph{client} & M2M & \makecell{(\texttt{tottenhamhotspur.com}; \texttt{S8g}) \\ (\texttt{loanstreet.com.my}; \texttt{2Pu3}) \\ (\texttt{healthlabtesting.com}; \texttt{2WLu})} \\
            \midrule
            *-\emph{binary} & M2M & \makecell{  (\texttt{kotonoha-jiten.com}; \texttt{15CBF8\ldots}) \\ (\texttt{wonderslim.com}; \texttt{B41781\ldots}) \\ (\texttt{pythonprogramming.net}; \texttt{CF6ACB\ldots})}   \\
            \midrule
            *-\emph{\gls{ip} address} & M2M & \makecell{  (\texttt{quickpayportal.com};  \texttt{208.78.141.18}) \\ (\texttt{jobwinner.ch};  \texttt{217.71.91.48}) \\ (\texttt{tottenhamhotspur.com};  \texttt{104.16.54.111})  }\\
            \midrule
            *-\emph{\gls{tls} issuer} & M2O & \makecell{ (\texttt{cratejoy.com};  \texttt{CN=Amazon, OU=Server CA 1B, O=Amazon, C=US}) \\ (\texttt{creative-serving.com};\\\quad  \texttt{CN=COMODO RSA Domain Validation Secure Server\ldots})\\ (\texttt{healthlabtesting.com}; \texttt{CN=Symantec Class 3 Secure Server CA\ldots})}\\
            \midrule
            *-\emph{\gls{tls} hash} & M2O & \makecell{ (\texttt{timeoutdubai.com}; \texttt{90c093\ldots}) \\ (\texttt{boomerang.com}; \texttt{e577e6\ldots}) \\ (\texttt{quickpayportal.com}; \texttt{85bdd8\ldots})}    \\
            \midrule
            *-\emph{\gls{tls} issue time} & M2O &  \makecell{(\texttt{jobwinner.ch}; \texttt{1496041693}) \\ (\texttt{healthlabtesting.com}; \texttt{1445558400}) \\ (\texttt{flatmates.com.au}; \texttt{1502150400})}   \\
            \midrule
            *-\emph{WHOIS email} & M2O & \makecell{ (\texttt{unstableunicorns.com}; \texttt{unstableunicorns.com@*sbyproxy.com})\\ (\texttt{albertlee.biz}; \texttt{abuse@godaddy.com})\\  (\texttt{crowneplazalondonthecity.com};\\ \quad \texttt{crowneplazalondonthecity.com@*sbyproxy.com })  } \\
            \midrule
            *-\emph{WHOIS nameserver} & M2O &  \makecell{(\texttt{grd779.com}; \texttt{ns2.hover.com}) \\ (\texttt{celeritascdn.com}; \texttt{lady.ns.cloudflare.com}) \\ (\texttt{smeresources.org}; \texttt{ns-495.awsdns-61.com})}   \\
            \midrule
            *-\emph{WHOIS registrar name} & M2O & \makecell{ (\texttt{unblocked.how}; \texttt{eNom, Inc.}) \\ (\texttt{rev-stripe.com}; \texttt{Amazon Registrar, Inc.}) \\ (\texttt{chisaintjosephhealth.org}; \texttt{Register.com, Inc.})} \\
            \midrule
            *-\emph{WHOIS country} & M2O & \makecell{ (\texttt{thefriscostl.com}; \texttt{CANADA}) \\ (\texttt{getwsone.com}; \texttt{UNITED STATES}) \\ (\texttt{notify.support}; \texttt{PANAMA})}\\
            \midrule
            *-\emph{WHOIS registrar id} & M2O & \makecell{ (\texttt{lo3trk.com}; \texttt{468}) \\ (\texttt{watchcrichd.org}; \texttt{472}) \\ (\texttt{bozsh.com}; \texttt{1479}) }\\
            \midrule
            *-\emph{WHOIS timestamp} & M2O & \makecell{ (\texttt{unblocked.how};  \texttt{15293}) \\ (\texttt{comicplanet.net}; \texttt{15159}) \\ (\texttt{unpublishedflight.com}; \texttt{15553})} \\
            \bottomrule
        \end{tabular}
    }
\end{table}

\newpage

\vfill

\renewcommand{\thesection}{D.\arabic{section}}
\renewcommand{\thefigure}{D.\arabic{figure} }
\renewcommand{\thetable}{D.\arabic{table} }

\section{Comparison to PTP}%
\label{ap:cisco_ptpcomp}

\begin{table}[H]
    \caption[AUPRC and AUROC metrics for three HMill models and the PTP algorithm.]{Values of the \gls{auprc} and the \gls{auroc} metrics for three \gls{hmill}-based models and the \gls{ptp} algorithm, rounded to 4 decimal places. Results on all three testing datasets are displayed and correspond to curves in Figures~\ref{fig:ptpcomp_pr} and~\ref{fig:ptpcomp_roc_log}. The greatest number in every column is written in bold.}%
    \label{tab:cisco_results_ptpcomp}
    \centering
    {\renewcommand\arraystretch{1.25}
        \begin{tabular}{c|cc|cc|cc}
            \toprule
            & \multicolumn{2}{c|}{\texttt{06-10}} & \multicolumn{2}{c|}{\texttt{06-27}} & \multicolumn{2}{c}{\texttt{07-22}} \\
              & \gls{auprc} & \gls{auroc}& \gls{auprc} & \gls{auroc}& \gls{auprc} & \gls{auroc} \\
            \midrule
            \gls{ptp} & $0.1561$ & $\mathbf{0.9621}$ & $0.1243$ & $0.9584$ & $0.1442$ & $0.9544$ \\
            $ m_B $ & $0.1242$ & $0.9579$ & $0.1561$ & $0.9590$ & $0.1544$ & $0.9545$ \\
            $ m_W $ & $\mathbf{0.2248}$ & $ 0.9560$ & $\mathbf{0.2402}$ & $0.9663$ & $\mathbf{0.2220}$ & $0.9574$ \\
            $ m_D $ & $0.1684$ & $0.9557$ & $0.1912$ & $\mathbf{0.9667}$ & $0.2047$ & $\mathbf{0.9580}$ \\
            \bottomrule
        \end{tabular}
}
\end{table}

\begin{figure}[H]
    \centering
    \begin{subfigure}[b]{0.329\textwidth}
        \centering
        \includegraphics[width=\textwidth]{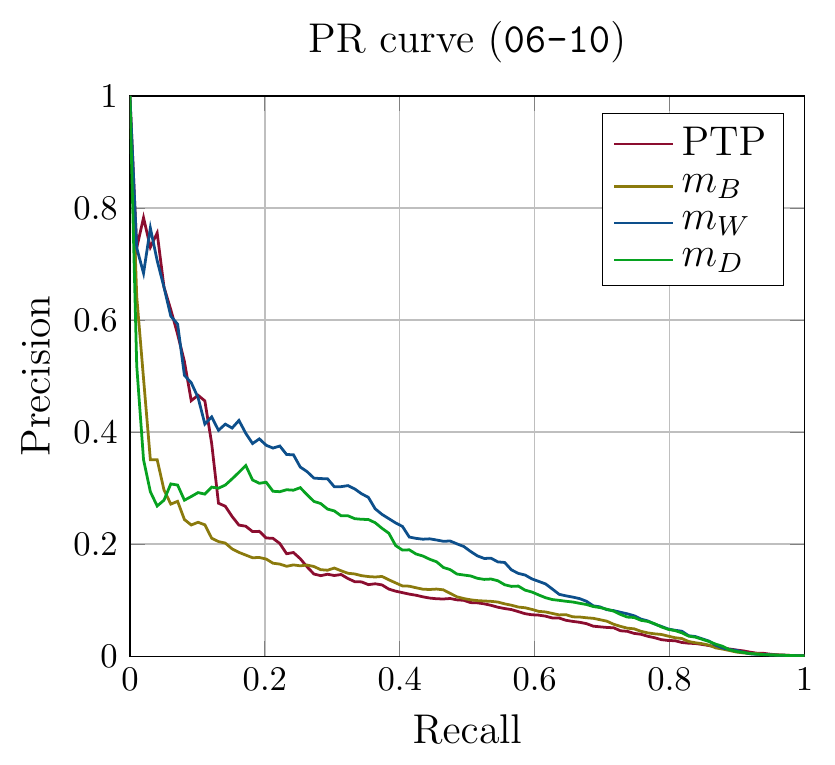}
    \end{subfigure}
    \hfill
    \begin{subfigure}[b]{0.329\textwidth}
        \centering
        \includegraphics[width=\textwidth]{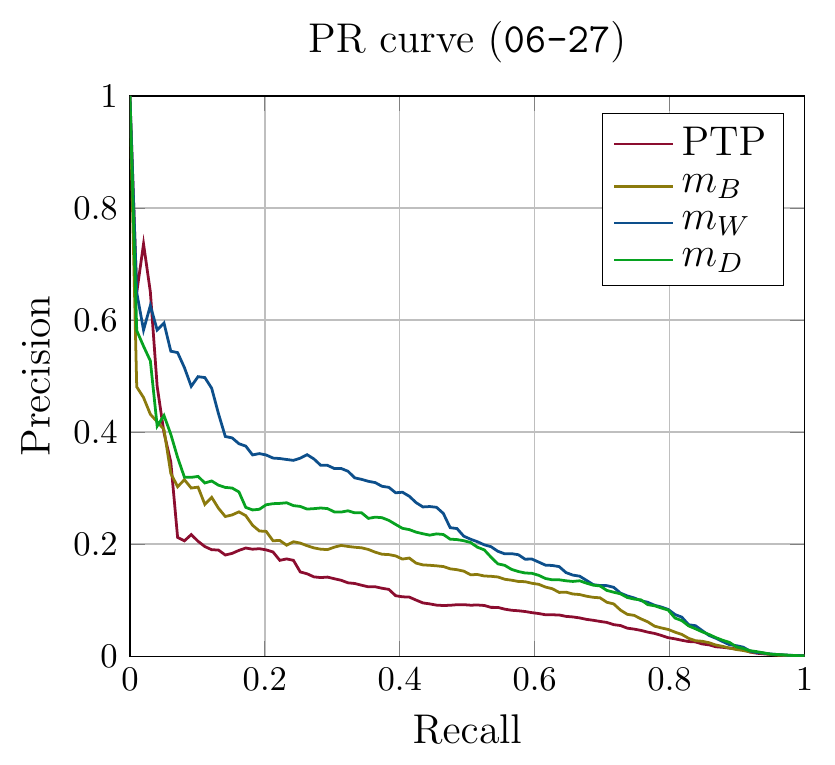}
    \end{subfigure}
    \hfill
    \begin{subfigure}[b]{0.329\textwidth}
        \centering
        \includegraphics[width=\textwidth]{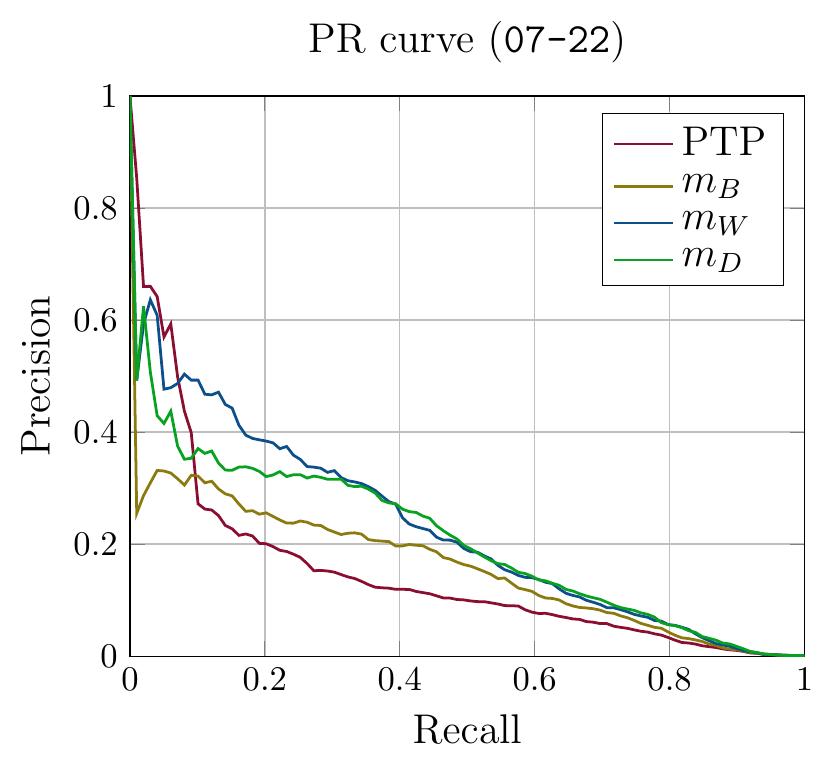}
    \end{subfigure}
    \caption[PR curves of three HMill models and the PTP algorithm.]{\gls{pr} curves comparing the performance of three \gls{hmill}-based models $ m_B $,  $ m_W $ and  $ m_D $ to the \gls{ptp} algorithm. Three figures are plotted, each corresponds to one of the testing datasets.}%
    \label{fig:ptpcomp_pr}
\end{figure}

\begin{figure}[H]
    \centering
    \begin{subfigure}[b]{0.329\textwidth}
        \centering
        \includegraphics[width=\textwidth]{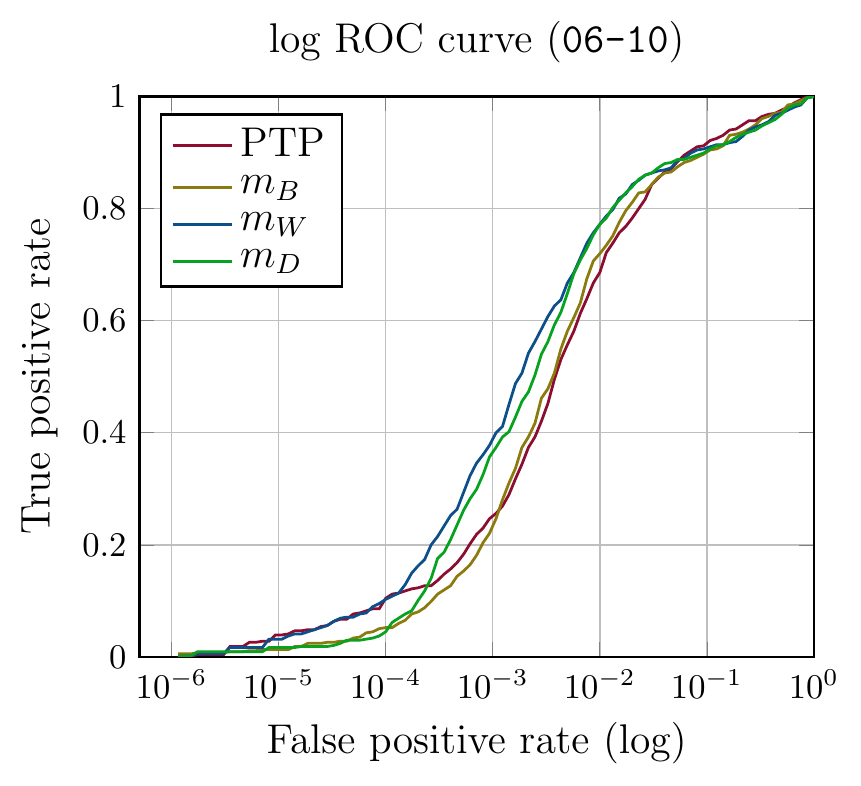}
    \end{subfigure}
    \hfill
    \begin{subfigure}[b]{0.329\textwidth}
        \centering
        \includegraphics[width=\textwidth]{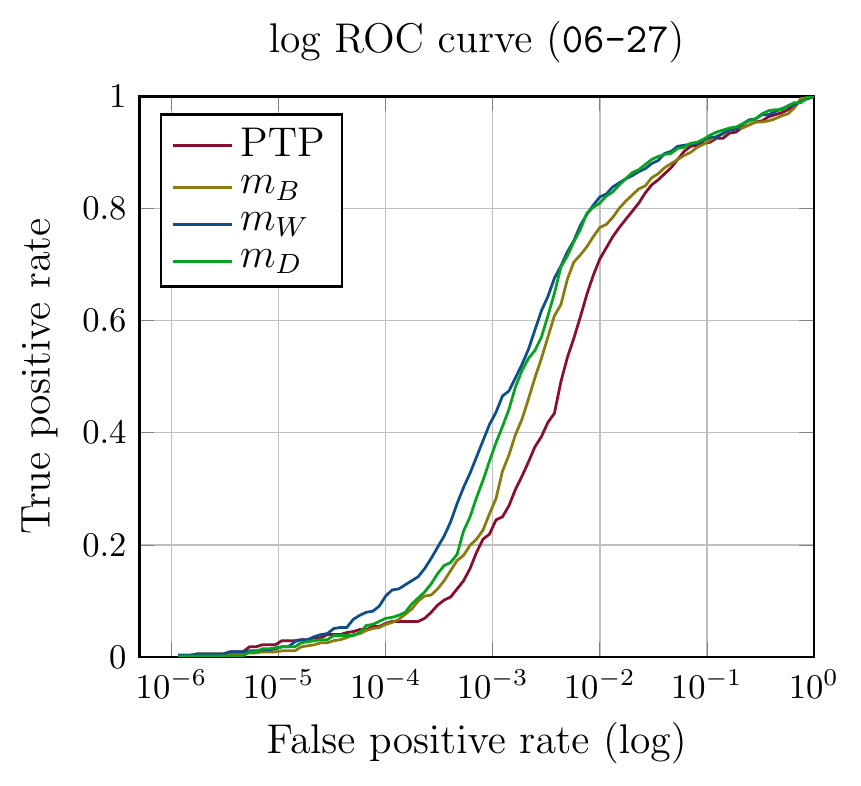}
    \end{subfigure}
    \hfill
    \begin{subfigure}[b]{0.329\textwidth}
        \centering
        \includegraphics[width=\textwidth]{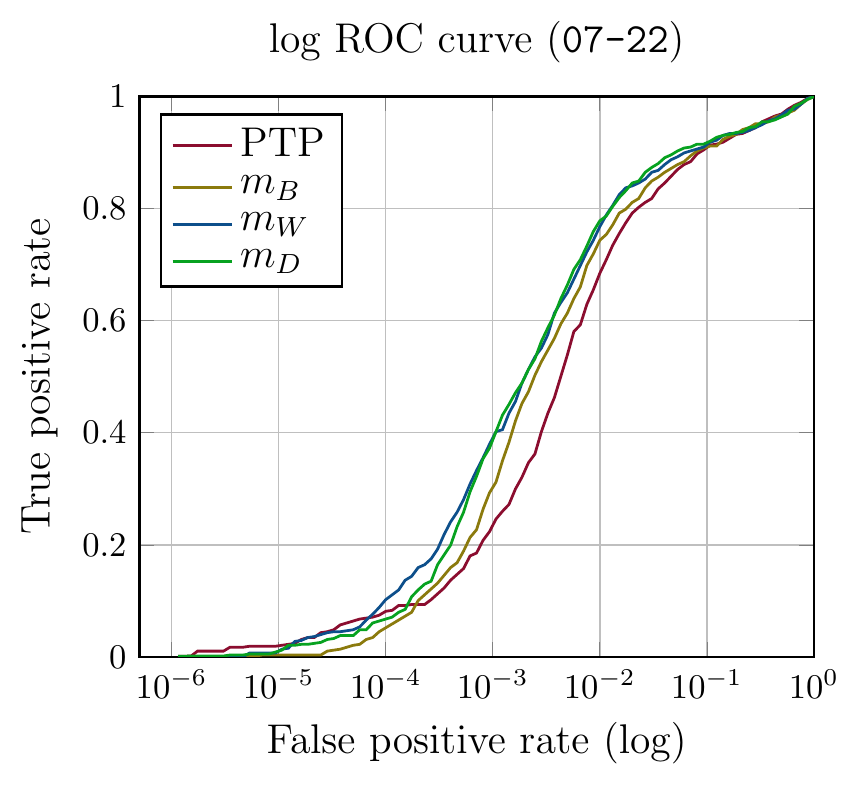}
    \end{subfigure}
    \caption[ROC curves of three HMill models and the PTP algorithm.]{\gls{roc} curves comparing the performance of three \gls{hmill}-based models $ m_B $,  $ m_W $ and  $ m_D $ to the \gls{ptp} algorithm. Three figures are plotted, each corresponds to one of the training datasets. The logarithmic scale is used for $ x $-axis.}%
    \label{fig:ptpcomp_roc_log}
\end{figure}

\vfill

\newpage

\vfill

\section{Additional relations}%
\label{ap:cisco_additional_relations}

\begin{table}[H]
    \caption[AUPRC and AUROC metrics for three HMill models and all eleven relations.]{Values of the \gls{auprc} and the \gls{auroc} metrics for three \gls{hmill}-based models, rounded to 4 decimal places. Here, all eleven relations are employed. Results on all three testing datasets are displayed and correspond to curves in Figures~\ref{fig:all_pr} and~\ref{fig:all_roc_log}. The greatest number in every column is written in bold.}%
    \label{tab:cisco_results_all}
    \centering
    {\renewcommand\arraystretch{1.25}
        \begin{tabular}{l|cc|cc|cc}
            \toprule
            & \multicolumn{2}{c|}{\texttt{06-10}} & \multicolumn{2}{c|}{\texttt{06-27}} & \multicolumn{2}{c}{\texttt{07-22}} \\
              & \gls{auprc} & \gls{auroc}& \gls{auprc} & \gls{auroc}& \gls{auprc} & \gls{auroc} \\
            \midrule
            $ m_b $ & $0.6419$ & $\textbf{0.9884}$ & $\textbf{0.6905}$ & $\textbf{0.9899}$ & $\textbf{0.6595}$ & $\textbf{0.9880}$ \\
            $ m_w $ & $\textbf{0.6422}$ & $0.9840$ & $ 0.6823$ & $0.9818$ & $0.6590$ & $0.9804$ \\
            $ m_d $ & $0.4498$ & $0.9780$ & $0.5031$ & $0.9831$ & $0.4719$ & $0.9796$ \\
            \bottomrule
        \end{tabular}
}
\end{table}

\begin{figure}[H]
    \centering
    \begin{subfigure}[b]{0.329\textwidth}
        \centering
        \includegraphics[width=\textwidth]{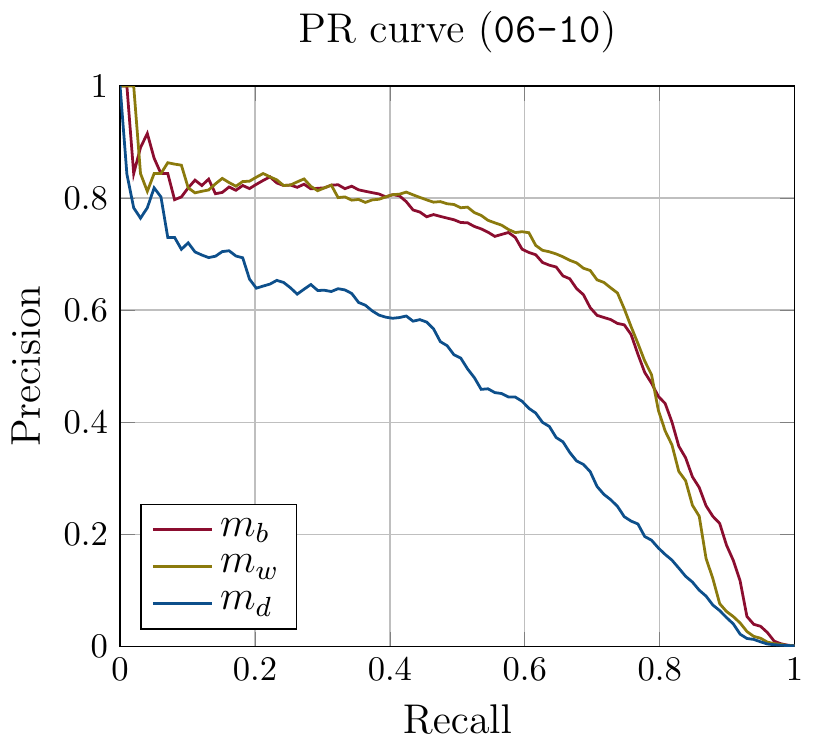}
    \end{subfigure}
    \hfill
    \begin{subfigure}[b]{0.329\textwidth}
        \centering
        \includegraphics[width=\textwidth]{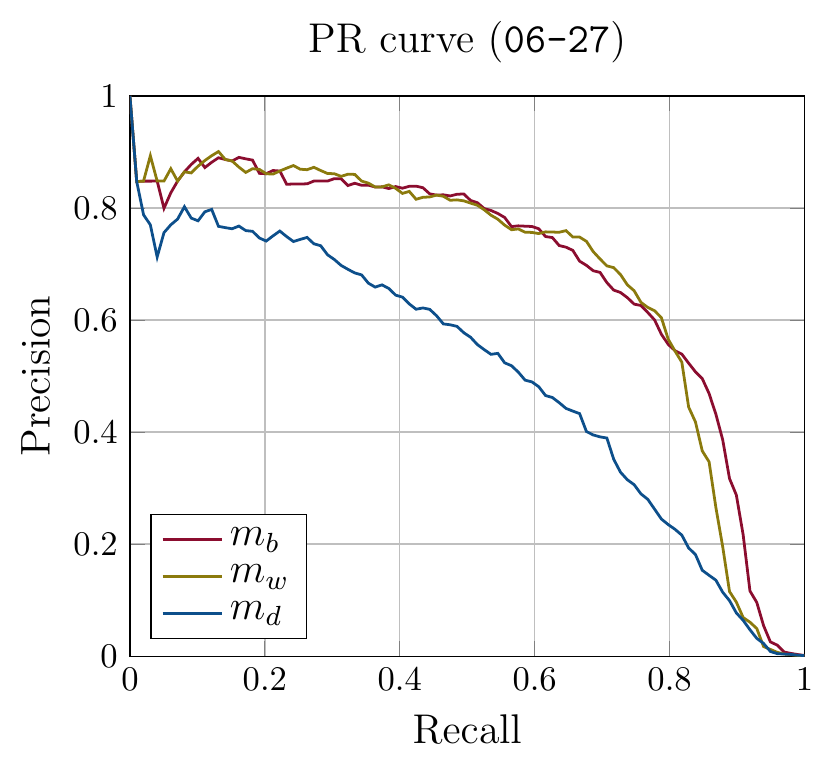}
    \end{subfigure}
    \hfill
    \begin{subfigure}[b]{0.329\textwidth}
        \centering
        \includegraphics[width=\textwidth]{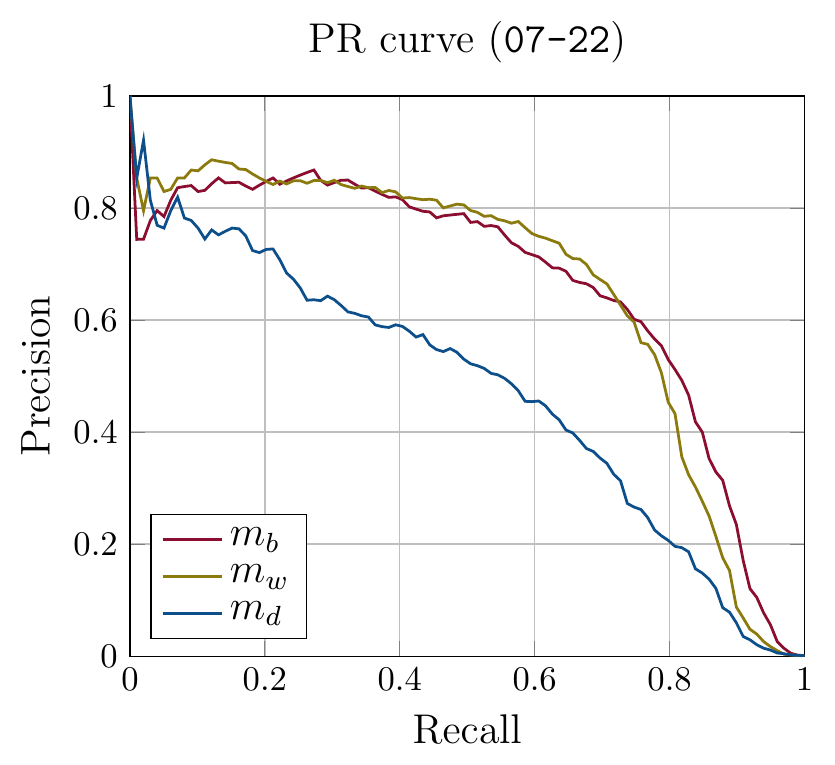}
    \end{subfigure}
    \caption[PR curves of three HMill models when all eleven relations are used.]{\gls{pr} curves comparing the performance of three \gls{hmill}-based models $ m_b $,  $ m_w $ and  $ m_d $ when all eleven relations are used. Three figures are plotted, each corresponds to one of the testing datasets.}%
    \label{fig:all_pr}
\end{figure}

\begin{figure}[H]
    \centering
    \begin{subfigure}[b]{0.329\textwidth}
        \centering
        \includegraphics[width=\textwidth]{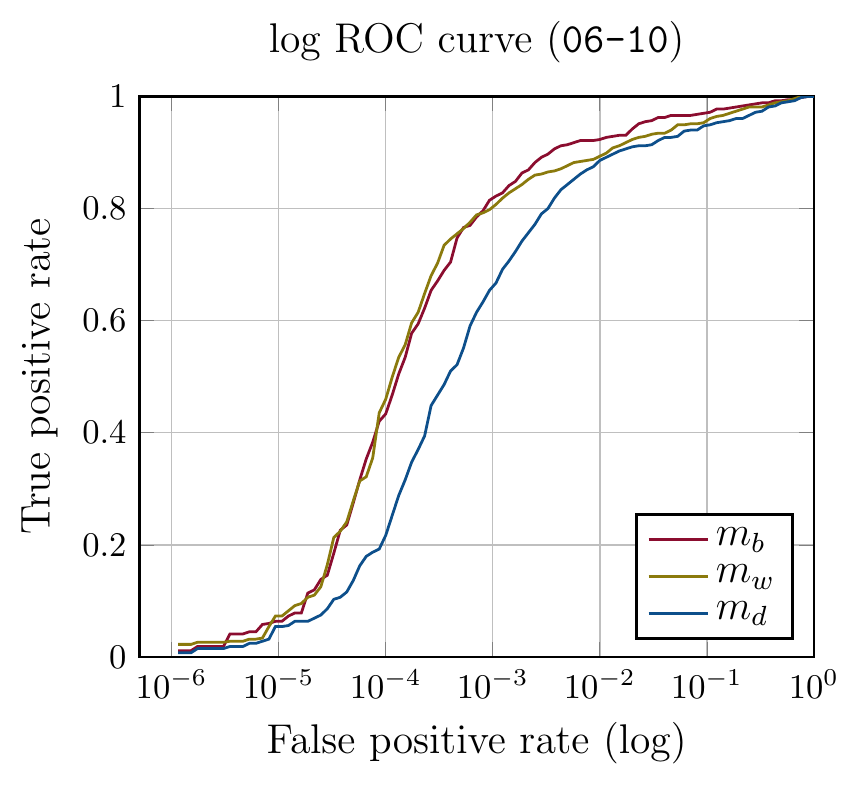}
    \end{subfigure}
    \hfill
    \begin{subfigure}[b]{0.329\textwidth}
        \centering
        \includegraphics[width=\textwidth]{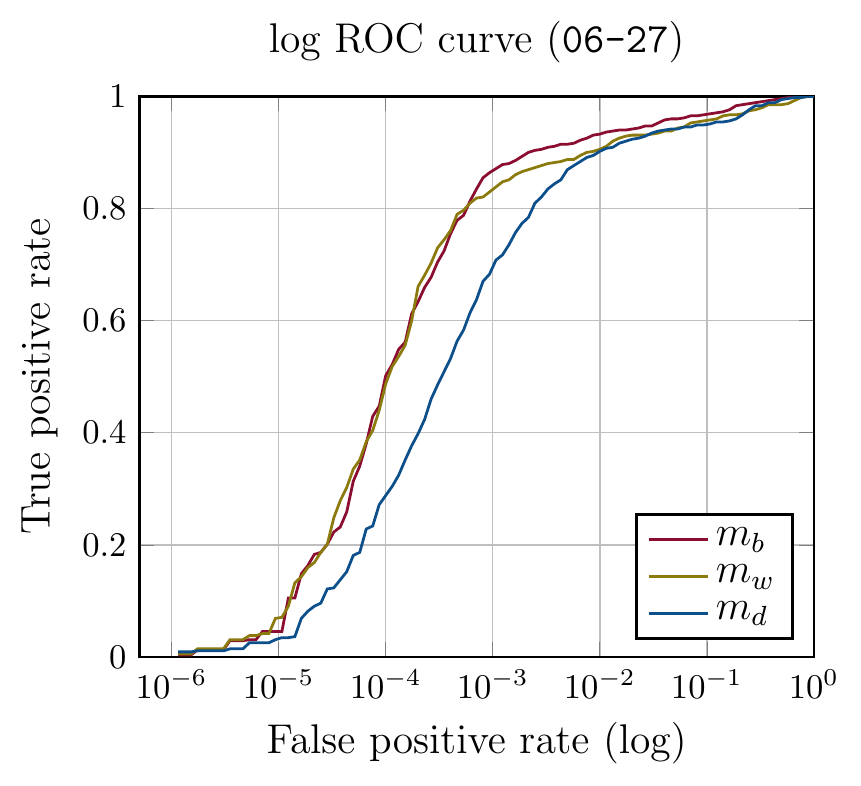}
    \end{subfigure}
    \hfill
    \begin{subfigure}[b]{0.329\textwidth}
        \centering
        \includegraphics[width=\textwidth]{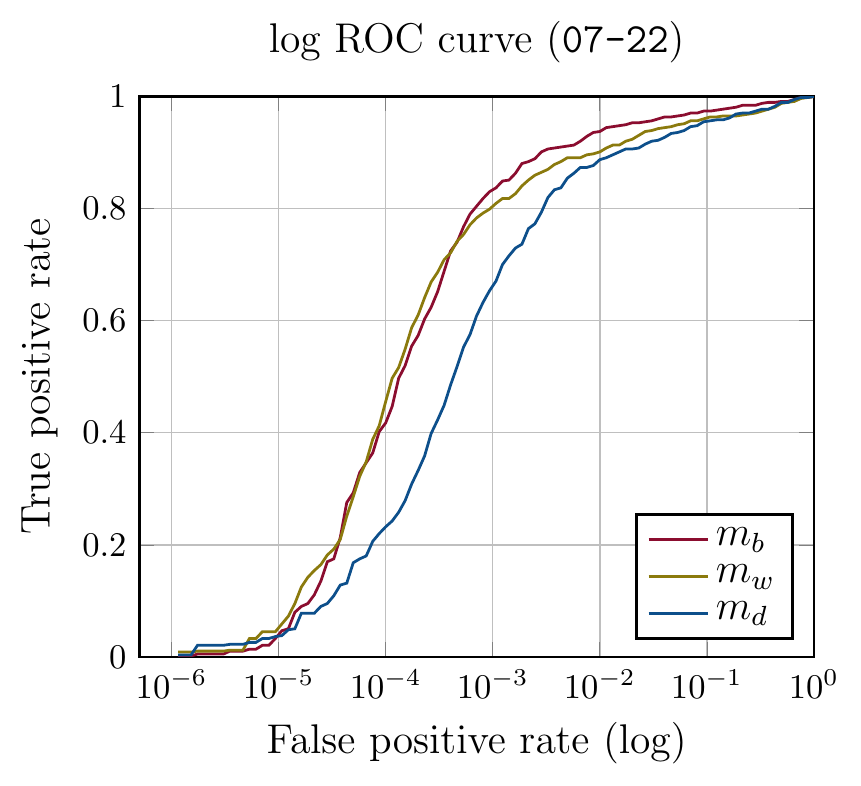}
    \end{subfigure}
    \caption[ROC curves of three HMill models when all eleven relations are used.]{\gls{roc} curves comparing the performance of three \gls{hmill}-based models $ m_b $,  $ m_w $ and  $ m_d $ when all eleven relations are used. Three figures are plotted, each corresponds to one of the training datasets. The logarithmic scale is used for $ x $-axis.}%
    \label{fig:all_roc_log}
\end{figure}

\vfill

\newpage

\vfill

\section{Fewer datasets trained on}%
\label{ap:cisco_lessgraphs}

\begin{table}[H]
    \caption[AUPRC and AUROC metrics for the `baseline' HMill model and a number of training dates.]{Values of the \gls{auprc} and the \gls{auroc} metrics rounded to 4 decimal places for the baseline \gls{hmill} model $ m_b $, in cases, where a different number of datasets is used. Specifically, in the first row, we used only the first dataset (date \texttt{05-23}), in the second row the first three of them and first five in the next row. In the last row, there is a performance of the model trained on all available training datasets, which is the identical setting to the one in Section~\ref{ssub:fewer_datasets}. Hence, the first row is identical to the first row in Table~\ref{tab:cisco_results_all}. Results on all three testing datasets are displayed and correspond to curves in Figures~\ref{fig:lessgraphs_pr} and~\ref{fig:lessgraphs_roc}. The greatest number in every column is written in bold.}%
    \label{tab:cisco_results_lessgraphs}
    \centering
        {\renewcommand\arraystretch{1.25}
            \begin{tabular}{l|cc|cc|cc}
                \toprule
            & \multicolumn{2}{c|}{\texttt{06-10}} & \multicolumn{2}{c|}{\texttt{06-27}} & \multicolumn{2}{c}{\texttt{07-22}} \\
              & \gls{auprc} & \gls{auroc}& \gls{auprc} & \gls{auroc}& \gls{auprc} & \gls{auroc} \\
            \midrule
                $ m_b $ ($1$ dataset) & $\textbf{0.6933}$ & $0.9739$ & $0.7005$ & $0.9796$ & $0.6708$ & $0.9750$ \\
                $ m_b $ ($3$ datasets) & $0.6834$ & $0.9825$ & $\textbf{0.7030}$ & $0.9843$ & $\textbf{0.6753}$ & $0.9810$ \\
                $ m_b $ ($5$ datasets) & $0.6381$ & $0.9762$ & $0.6760$ & $0.9839$ & $0.6363$ & $0.9774$ \\
                $ m_b $ ($8$ datasets) & $0.6419$ & $\textbf{0.9884}$ & $0.6905$ & $\textbf{0.9899}$ & $0.6595$ & $\textbf{0.9880}$ \\
                \bottomrule
            \end{tabular}
        }
\end{table}

\begin{figure}[H]
    \centering
    \begin{subfigure}[b]{0.329\textwidth}
        \centering
        \includegraphics[width=\textwidth]{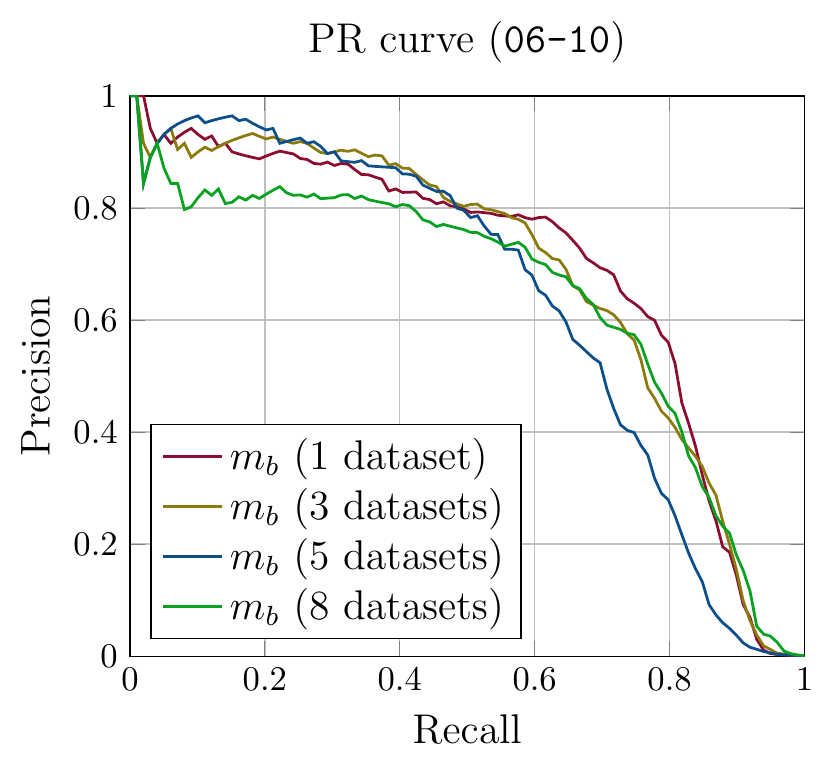}
    \end{subfigure}
    \hfill
    \begin{subfigure}[b]{0.329\textwidth}
        \centering
        \includegraphics[width=\textwidth]{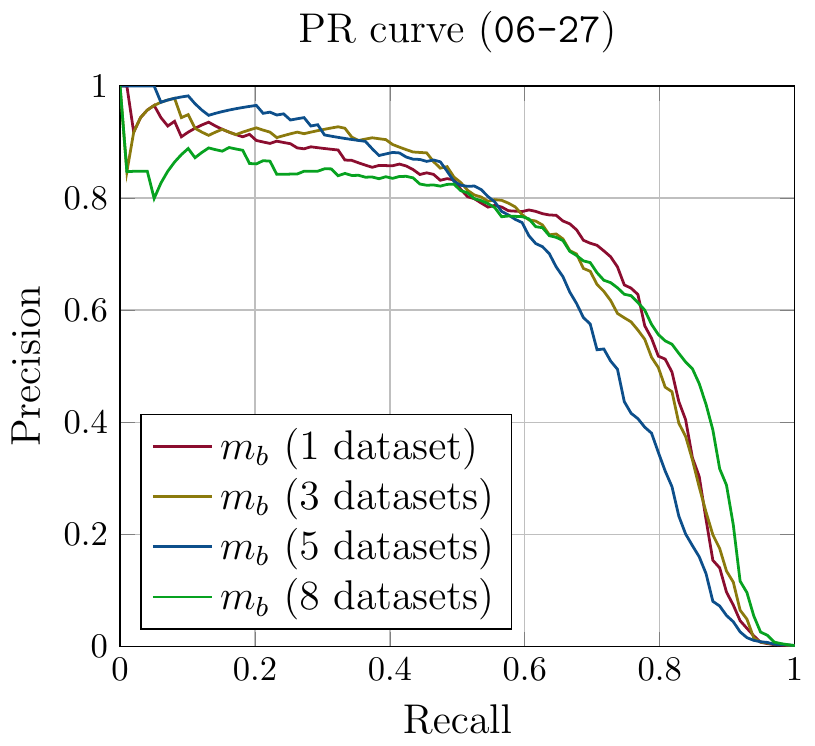}
    \end{subfigure}
    \hfill
    \begin{subfigure}[b]{0.329\textwidth}
        \centering
        \includegraphics[width=\textwidth]{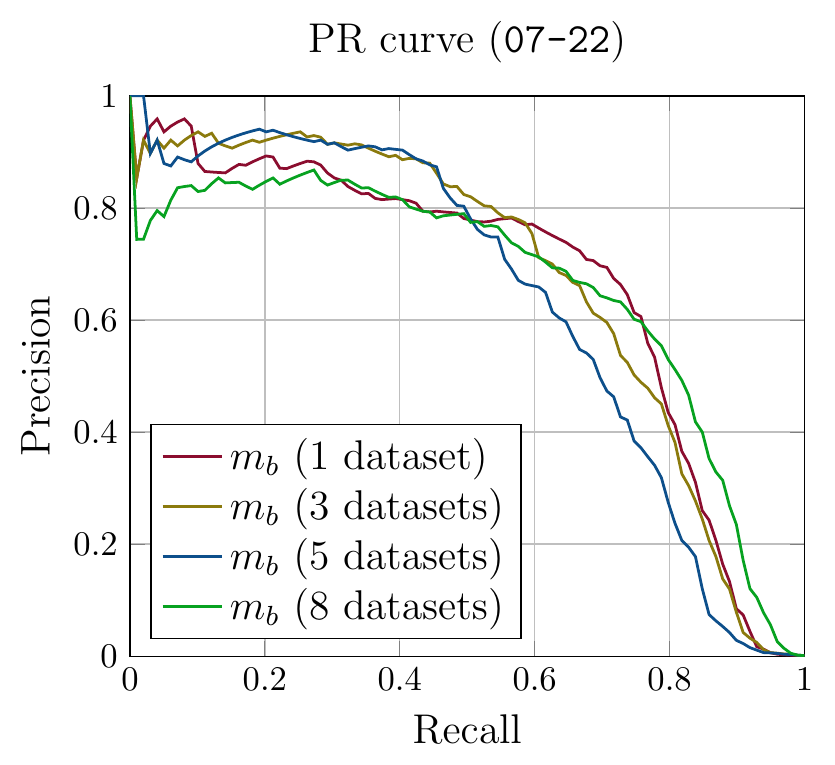}
    \end{subfigure}
    \caption[PR curves of the `baseline' HMill model when trained on a different number of datasets.]{\gls{pr} curves comparing the performance of the `baseline' \gls{hmill} model $ m_b $, when a different number of datasets is used for training. Three figures are plotted, each corresponds to one of the testing datasets.}%
    \label{fig:lessgraphs_pr}
\end{figure}

\begin{figure}[H]
    \centering
    \begin{subfigure}[b]{0.329\textwidth}
        \centering
        \includegraphics[width=\textwidth]{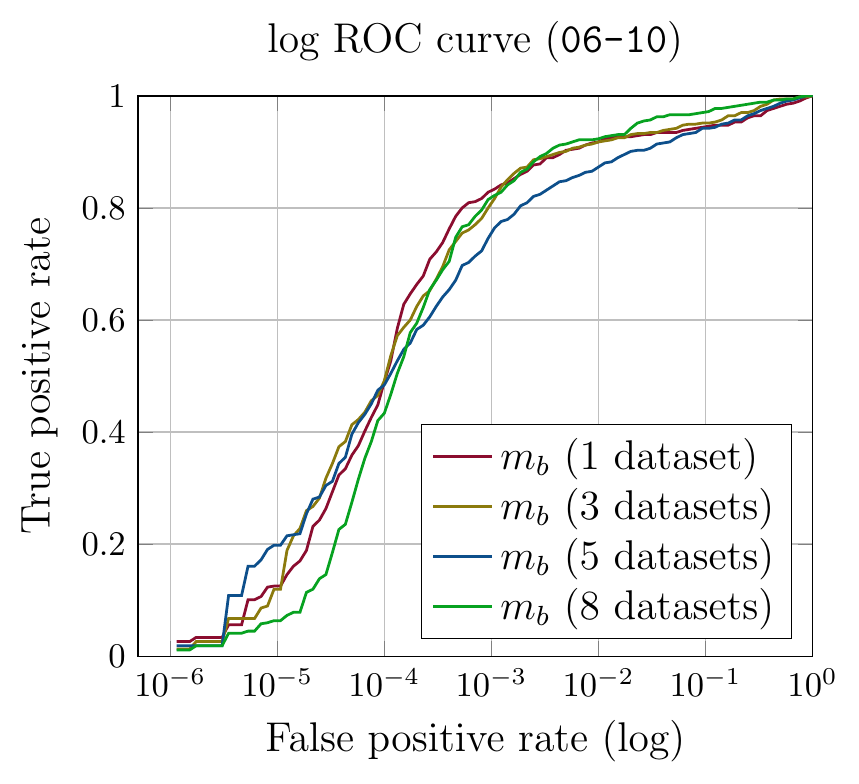}
    \end{subfigure}
    \hfill
    \begin{subfigure}[b]{0.329\textwidth}
        \centering
        \includegraphics[width=\textwidth]{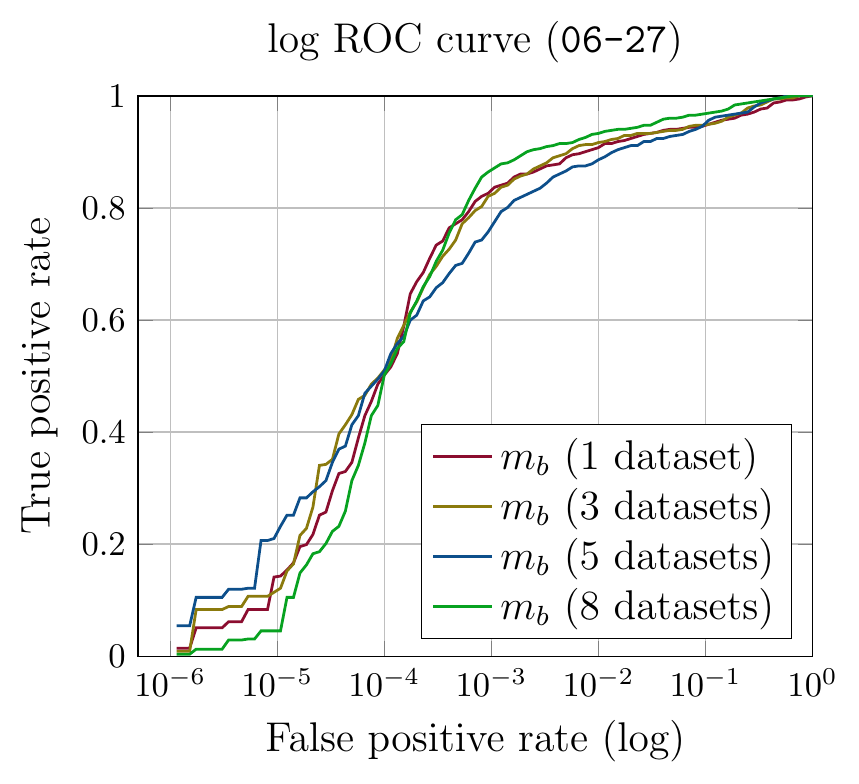}
    \end{subfigure}
    \hfill
    \begin{subfigure}[b]{0.329\textwidth}
        \centering
        \includegraphics[width=\textwidth]{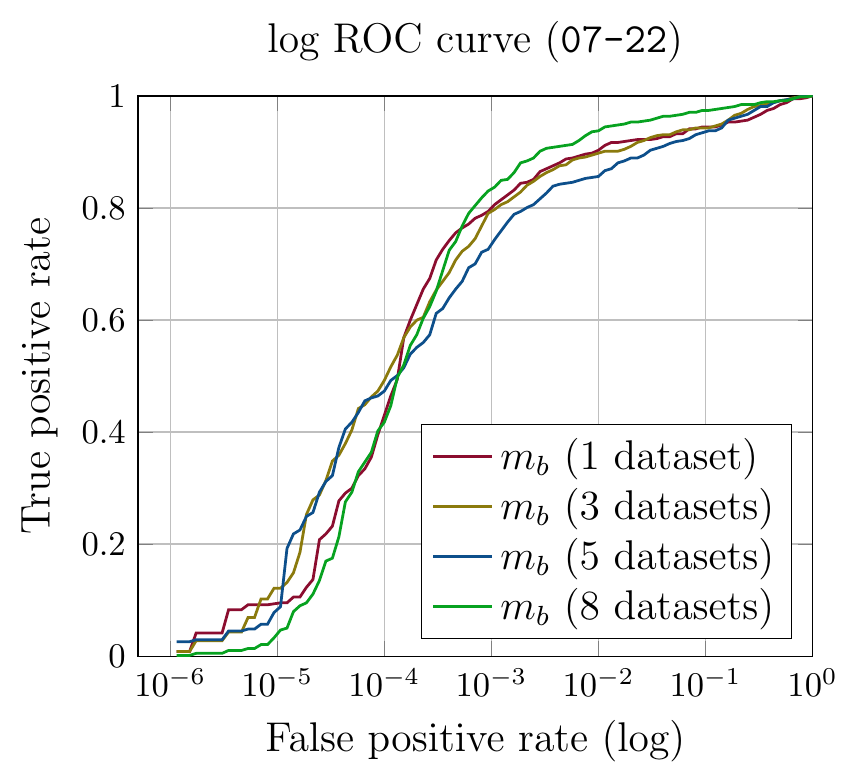}
    \end{subfigure}
    \caption[ROC curves of the `baseline' HMill model when trained on a different number of datasets.]{\gls{roc} curves comparing the performance of the `baseline' \gls{hmill} model $ m_b $, when a different number of datasets is used for training. Three figures are plotted, each corresponds to one of the training datasets. The logarithmic scale is used for $ x $-axis.}%
    \label{fig:lessgraphs_roc}
\end{figure}

\vfill

\newpage

\vfill

\section{Grill test}%
\label{ap:cisco_gtest}

\vfill

\begin{table}[H]
    \caption[AUPRC and AUROC metrics for the `baseline' HMill and Grill test.]{Values of the \gls{auprc} and the \gls{auroc} metrics rounded to 4 decimal places, for the baseline \gls{hmill} model $ m_b $ with and without Grill test. The first row is identical to the first row in Table~\ref{tab:cisco_results_all}. Results on all three testing datasets are displayed and correspond to curves in Figures~\ref{fig:gtest_pr} and~\ref{fig:gtest_roc_log}. The greatest number in every column is written in bold.}%
    \label{tab:cisco_results_gtest}
    \centering
    {\renewcommand\arraystretch{1.25}
        \begin{tabular}{l|cc|cc|cc}
            \toprule
            & \multicolumn{2}{c|}{\texttt{06-10}} & \multicolumn{2}{c|}{\texttt{06-27}} & \multicolumn{2}{c}{\texttt{07-22}} \\
              & \gls{auprc} & \gls{auroc}& \gls{auprc} & \gls{auroc}& \gls{auprc} & \gls{auroc} \\
            \midrule
            $ m_b $ & $\textbf{0.6419}$ & $\textbf{0.9884}$ & $\textbf{0.6905}$ & $\textbf{0.9899}$ & $\textbf{0.6595}$ & $\textbf{0.9880}$ \\
            $ m_b $ (Grill) & $0.4503$ & $0.9509$ & $0.4745$ & $0.9662$ & $0.4735$ & $0.9593$ \\
            \bottomrule
        \end{tabular}
}
\end{table}

\begin{figure}[H]
    \centering
    \begin{subfigure}[b]{0.329\textwidth}
        \centering
        \includegraphics[width=\textwidth]{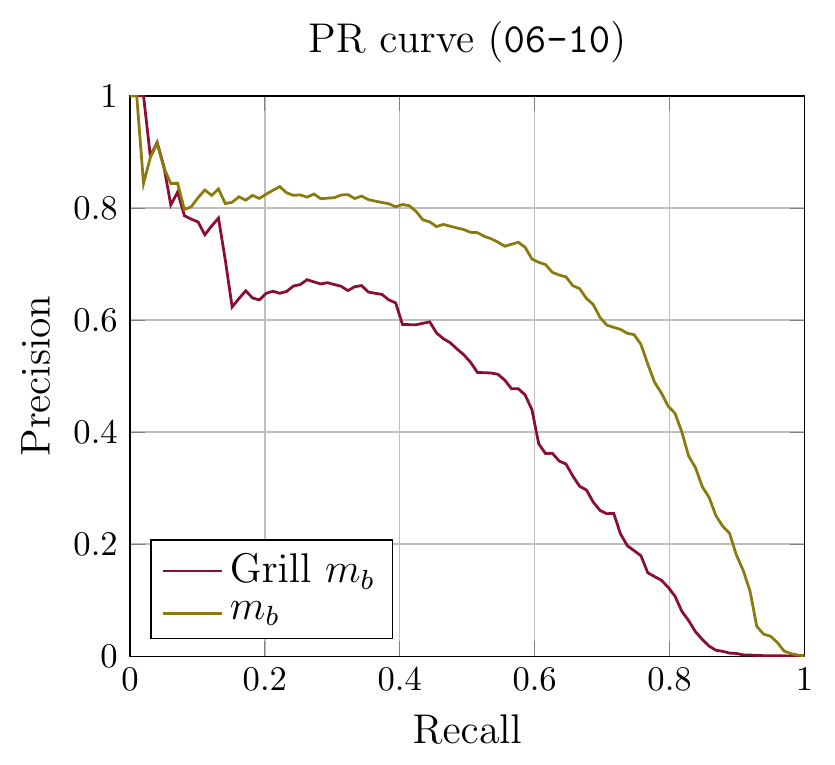}
    \end{subfigure}
    \hfill
    \begin{subfigure}[b]{0.329\textwidth}
        \centering
        \includegraphics[width=\textwidth]{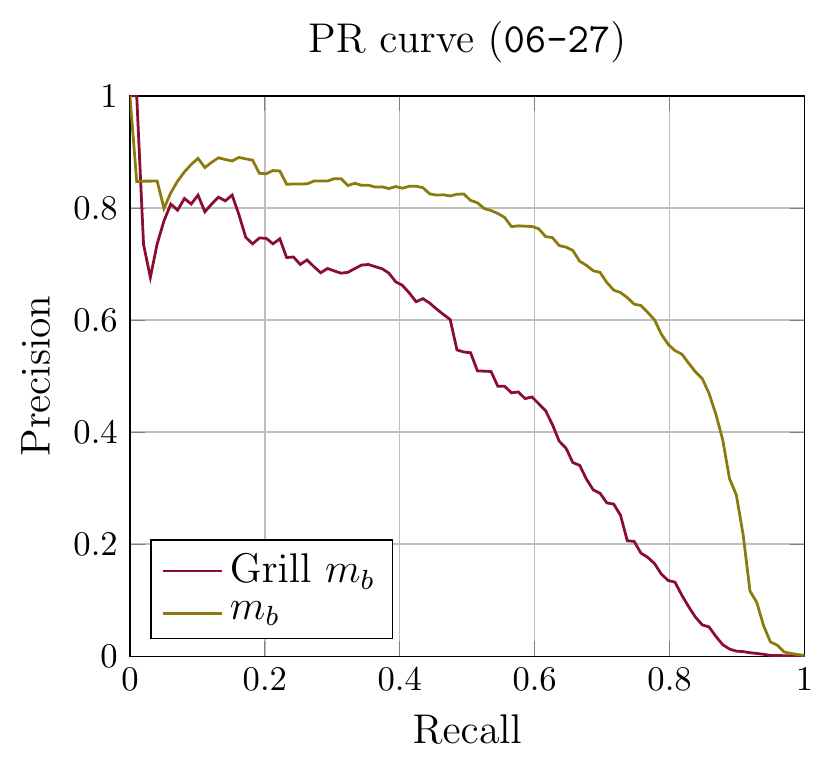}
    \end{subfigure}
    \hfill
    \begin{subfigure}[b]{0.329\textwidth}
        \centering
        \includegraphics[width=\textwidth]{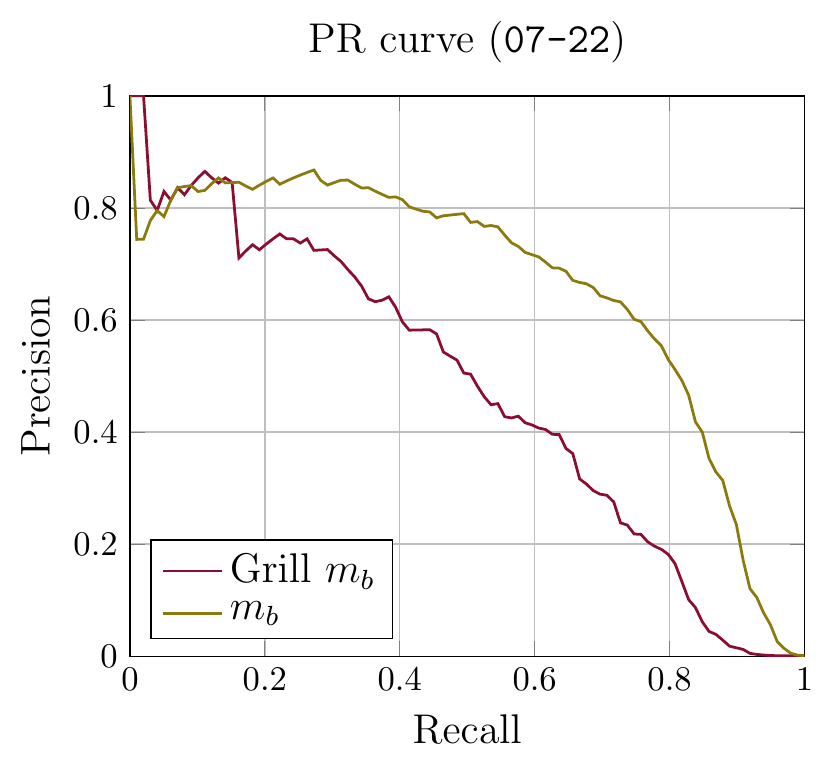}
    \end{subfigure}
    \caption[PR curves of the `baseline' HMill model with and without the Grill test.]{\gls{pr} curves comparing the performance of the `baseline' \gls{hmill} model $ m_b $, when the Grill test is (not) used. Three figures are plotted, each corresponds to one of the training datasets.}%
    \label{fig:gtest_pr}
\end{figure}

\begin{figure}[H]
    \centering
    \begin{subfigure}[b]{0.329\textwidth}
        \centering
        \includegraphics[width=\textwidth]{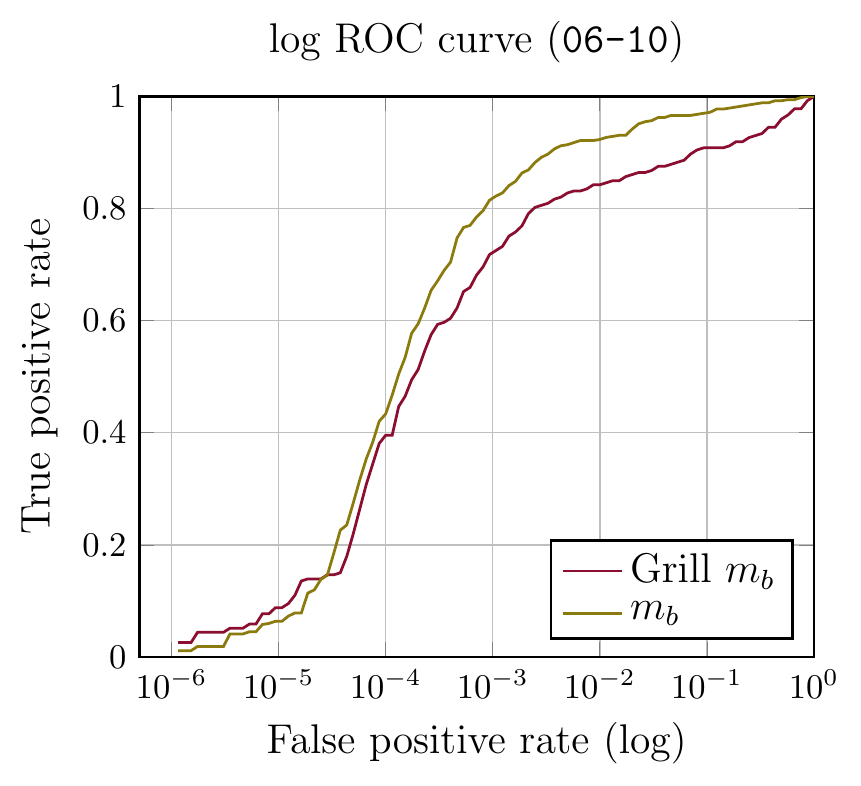}
    \end{subfigure}
    \hfill
    \begin{subfigure}[b]{0.329\textwidth}
        \centering
        \includegraphics[width=\textwidth]{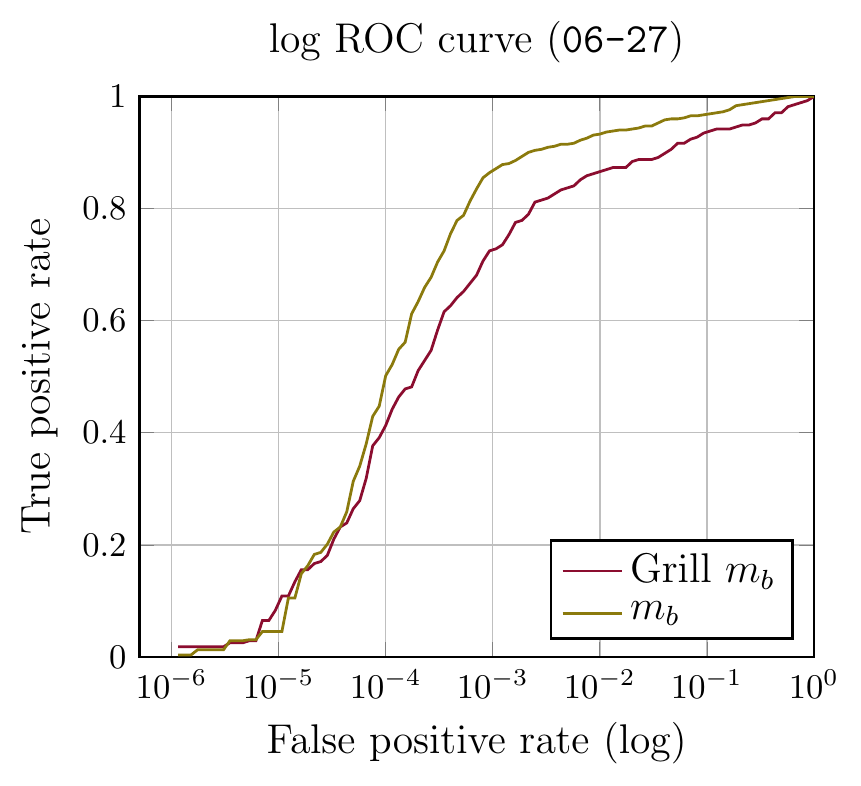}
    \end{subfigure}
    \hfill
    \begin{subfigure}[b]{0.329\textwidth}
        \centering
        \includegraphics[width=\textwidth]{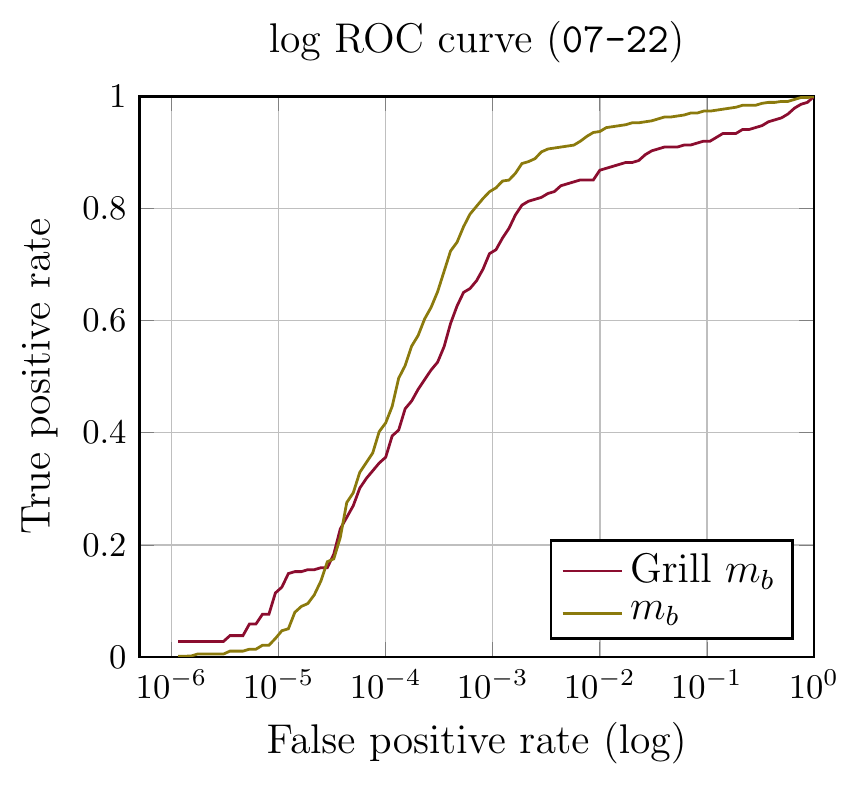}
    \end{subfigure}
    \caption[ROC curves of the `baseline' HMill model with and without the Grill test.]{\gls{roc} curves comparing the performance of the `baseline' \gls{hmill} model $ m_b $, when the Grill test is (not) used. Three figures are plotted, each corresponds to one of the training datasets. The logarithmic scale is used for $ x $-axis.}%
    \label{fig:gtest_roc_log}
\end{figure}

\vfill

\newpage

\newglossarystyle{cvutstyle}{%
\setglossarystyle{indexgroup}
	\renewcommand*{\glossentry}[2]{%
	\tabto{0.5cm}
	\emph{\glstarget{##1}{\glossentryname{##1}}}
	\tabto{3cm}\glossentrydesc{##1}
	\hfill [##2] \vspace{0.045in}\newline
}%
	\renewcommand*{\glsgroupheading}[1]{%
        {\color{cvutaccented}{\glsgetgrouptitle{##1}}\vspace{0.05in}\newline}%
	}
}

\glsnogroupskiptrue
\printglossary[style=cvutstyle]

\end{appendices}

\bibliographystyle{apalike}
\bibliography{thesis.bib}

\end{document}